\documentclass[12pt]{report}
\usepackage{utathesis}
\usepackage{soul}
\setuldepth{Berlin}
\usepackage{comment}
\usepackage{url}
\usepackage{algorithm}
 \usepackage{algpseudocode}
\usepackage{nth}
\usepackage{amssymb}
\usepackage{epsfig}
\usepackage{makecell}
\usepackage{hyperref}
\usepackage{wrapfig}
\usepackage{enumitem}
\usepackage{subfig}
\usepackage{color}
\usepackage{multirow}

\usepackage{graphicx}

\usepackage[utf8]{inputenc}
\usepackage[english]{babel}
\usepackage{amsmath}
\usepackage{amsthm}

\newenvironment{definition}[1][Definition]{\begin{trivlist}\item[\hskip \labelsep {\bfseries #1}]}{\end{trivlist}}

\usepackage{pgfplots}
\usepackage{csquotes}
\usepackage{makecell}

\usepackage{tikz}

\usepackage{xspace}
\usepackage[symbol]{footmisc}

\usepackage{amsmath}
\usepackage{amsfonts}
\usepackage{subfig}
\newcounter{Figcount}
\newcounter{tempFigure}

\usepackage[colorinlistoftodos]{todonotes}

\begin{document}
\graduationmonth{}
\graduationyear{}
\defensedate{October 2025}
\author{Hafsa Billah}
\committee{}{}{}{}{}

\title{Non-Traditional Domain-Independent Approaches To Analyzing and Identifying Situations Embedded in Videos}
\title{A Non-Traditional Domain-Independent framework For Analyzing and Identifying Situations Embedded in Videos}
\title{A Novel, Domain-Independent Framework For Analyzing and Identifying Situations Embedded in Videos}
\title{A General, Non-Traditional Approach To Analyzing and Identifying Situations Embedded in Videos}
\title{A General, Stream-Based Approach To Analyzing and Identifying Situations Embedded in Videos}
\title{A General, Stream-Based framework For Analyzing and Identifying Situations Embedded in Videos}
\title{A General, Non-Traditional Approach To Analyzing and Identifying Situations Embedded in Videos}
\title{Video Analysis for Situation identification: A General, Domain-Independent Stream-Based Approach}
\title{Video Analysis for Situation identification: An Efficient Approach wit Real-Time potential}

\title{VideoScoop: A Non-Traditional Domain-Independent Framework For Video Analysis}
\titlepage



\begin{abstract}
Due to the proliferation of cameras in handheld devices and the widespread use of CCTV, images and videos have become indispensable for capturing and disseminating information. Automated analysis for understanding image or video contents (e.g., objects, activities, background, situations of interest, etc.) is critical for many applications such as civic monitoring, surveillance (in general), monitoring activities in Assisted Living environments, and many more. Image and Video Analysis (IVA) research has been ongoing for several decades resulting in a number of techniques for algorithmically analyzing and understanding contents of images and videos.  

Image Analysis (IA) has advanced along several dimensions, including object recognition, image classification, and pose recognition, among others. Although Video Analysis (VA) subsumes image analysis (or analysis of individual video frames), it is far more complex than IA, as video contents change over time. Several decades of research in Video Content Extraction (VCE) has led to the development of advanced content extraction algorithms, such as object recognition, object tracking, etc. However, identifying interesting or complex activities (termed situations in this thesis) embedded and spanning multiple consecutive frames (e.g., a package being stolen from a premise, heavy traffic flow in an intersection, etc.) cannot be achieved by merely extracting the contents of each frame. To fully understand the video content, automated analysis of various, diverse situations is essential. Our focus is to identify situations from the extracted contents. For this purpose, the extracted contents need to be: i) represented using appropriate data models, ii) express situations on these models using an easy-to-use language (non-procedural queries or analysis expressions), and iii) develop algorithms for accurately and efficiently analyzing the contents to detect expressed situations.   However, to date, situation analysis has been accomplished, to a large extent, either manually or by using customized algorithms rather than using a \textbf{general-purpose, domain-independent} framework.

Broadly, two approaches can be identified for Video Situation Analysis (VSA): manual and custom algorithms. In manual analysis, a human watches a video and identifies interesting or specified situations in the video footage. Although this approach can be applied for forensic analysis (that is, after the fact analysis) or for real-time monitoring, it is impractical due to the significant human capital required. Further, manual situation analysis is subject to human fatigue, skill set disparities, and human errors.
In contrast, the custom solutions are designed for specific video types and/or situations. In addition, custom solutions are not fast enough for real-time situation analysis at this time. Furthermore, different custom algorithms are needed and has to be developed to identify various situations. Currently, these alternatives are typically applied after the fact to analyze situations of interest from an archived video. Even for forensic analysis, the custom solutions are not general-purpose. These algorithms cannot analyze all diverse videos (from different domains) or situations, as new algorithms or software packages need to be developed for each situation or video type (from different domains). Hence, a common limitation of the above alternatives is their inability to perform real-time and/or general-purpose situation analysis. 

In this thesis, we propose a general-purpose, novel framework that overcomes the limitations of current VSA alternatives. It is a general-purpose solution rather than a customized one, and is amenable to real-time video analysis capabilities. Our approach can also be used for forensic analysis of videos, which is the focus of this thesis, as it is the first step towards real-time video situation analysis. The proposed approach extracts contents (as much as possible) from video frames using state-of-the-art VCE technologies. These extracted contents are represented (modeled) using two alternative expressive data models -- relational model with extensions (termed R++) and graph models, including multiple graphs (or layers) per video. Once modeled appropriately, the extracted contents can be used as data streams (where the underlying model is the relational model). As a result, Continuous Query Processing (CQP) can be used on streams using the proposed Continuous Query Language for Video Analysis (CQL-VA) operators (specifically extended for video analysis). The use of non-procedural Structured Query Language (SQL) as the basis of CQL-VA provides a much easier way to express situations in terms of "what to identify" (intent) instead of "how to identify" (algorithm), shifting the burden of efficient query processing to the underlying query processor/optimizer. Unlike custom algorithms, the proposed approach for both representations supports \textbf{extraction of video contents only once}, and evaluation of as many queries or analysis expressions on that as needed, avoiding the development of new algorithms. \textit{A major advantage is that as VCE algorithms improve, they can be used without changing the situation analysis part. More accurately extracted information will lead to a more accurate situation analysis.}


Similarly, the graph models provide an alternative representation and facilitate the detection (or analysis) of certain situations that are difficult (or impossible) to achieve using the relational model, even with extensions. Here, existing graph algorithms (e.g., clustering) are utilized, and additional efficient algorithms have been developed to identify various embedded interesting situations. Together, these two models and associated analysis approaches, together, support the detection of a wide variety of situations embedded in the videos.

Finally, our approach is domain-independent, meaning that videos from various domains can be processed using the same framework. To accommodate domain independence in the proposed framework, primitive situations are identified and expressed as templates. These templates are used for identifying variants of primitive situations across domains by parameterizing them. Three contrasting domains -- Assisted Living (AL), Civic Monitoring (CM), and Surveillance (SL) -- have been used in this thesis to highlight domain independence. Proposed primitive operators or analysis algorithms are used to compose specific queries or analysis expressions to detect more complex and meaningful situations. Finally, extensive analysis has been performed on several interesting situations expressed using the framework for evaluating accuracy, efficiency, and robustness, utilizing datasets containing a variety of videos (of different lengths and types) from the above three domains.

\indent
\end{abstract}

\tableofcontents
\listoffigures
\listoftables
\addtocontents{toc}{\noindent\mbox{Chapter}\hfill\mbox{Page}}%


\chapter{Introduction}
\label{chap:introduction}


Images and videos are being used extensively nowadays because they are easy to capture and convey complex information quickly and effectively.
The availability of inexpensive mobile, handheld, and surveillance cameras has made this possible.\footnote{This situation is somewhat analogous to the availability of inexpensive sensors, which gave rise to the development of techniques, algorithms, and frameworks for sensor stream data processing for various kinds of monitoring (e.g., environment, water quality, infrastructure, etc.) Something similar is happening for images and videos! Moreover, video processing involves processing a stream of frames from a video.} An image captures a single moment (or snapshot) in time and contains spatial information about objects or action(s) happening at that time. A video is a sequence of images (or frames) captured at a fixed frame-per-second (fps) rate that depicts the changing information or video contents (e.g., objects and their location, action(s), etc.) change over time (or over a sequence of consecutive frames). In contrast to images, videos contain both spatial and temporal information about objects or action(s) happening over a period. In other words, videos contain \textbf{situation(s)} that are \textit{action(s) of interest present across multiple consecutive sequences of frames}. 

An example of an interesting situation in a Surveillance (SL) video is ``a delivered package missing from the place of delivery (before the owner collects it)'' or in a video recorded in an Assisted Living (AL) environment is ``a person not taking part in group activities (or is isolated)''. An example from the Civic Monitoring (CM) domain is to identify ``traffic congestion at intersections or parking lots, including variations by day and time'' (for adaptive tolling purposes). There are numerous other example situations in videos from different domains.

Automatically understanding the contents (or situations) of images or videos is essential for several applications, such as monitoring senior care facilities, traffic monitoring, surveillance of facilities (e.g., schools, shopping malls, grocery stores, etc.) Image and Video Analysis (IVA) research has been addressing the challenges associated with understanding image or video contents for several decades~\cite{Chellappa-NSF-report-2014}. Image Analysis (IA) is challenging because several environmental factors (e.g., lighting conditions, occlusion, motion, camera position, background, color of attire, use of glasses and hats, etc.) impact the image analysis. Many of these factors affect the correctness of the algorithms (e.g., object recognition) developed for IA. Over the past several decades, significant strides have been made in various IA tasks (e.g., object recognition, image classification, etc.) by designing algorithms, both traditional and neural network-based, that can interpret image contents correctly (to some extent) under different environmental conditions~\cite{survey/ibrahim2021survey, Chellappa-NSF-report-2014}. 

In contrast to IA, Video Analysis (VA) poses several additional challenges. Due to the evolving contents in videos, both spatial and temporal information in a video needs to be analyzed. Broadly, VA can be divided into two separate steps: (i) Video Content Extraction (VCE) and (ii) Video Situation Analysis (VSA). The VCE tasks or workflow process raw videos and extract meaningful spatial (e.g., objects and their location in the field of view) contents from each video frame or temporal contents (e.g., tracking objects~\footnote{Object tracking assigns a unique identifier to an object across a fixed sequence of frames (given as parameter)}) across several consecutive frame sequences. Different aspects of content extraction have also been researched over the past several decades. The advent of deep learning has addressed some of the near-term challenges (e.g., person identification, tracking occluded objects, etc.) in VCE to some extent, as reported in~\cite{Chellappa-NSF-report-2014}. The long-term challenges (e.g., extracting all different types of background, consistently generating object tracks, etc.) are still being researched according to~\cite{Chellappa-NSF-report-2014}.

On the other hand, VSA algorithms analyze meaningful situations by either processing raw videos or extracting video contents and analyzing them (termed a two-step approach in this thesis). The current VSA solutions can be described as: i) manual or ii) use of custom algorithms. In manual analysis, a human watches videos to identify specified or desired situations. This process is labor-intensive and subject to human fatigue (when video duration is longer), resulting in errors. The custom analysis algorithms~\cite{VIRAT-2013,videoQuerying/aref2003video,videoQuerying/zhang2017vsql,videoQuerying/bastani2020miris,videoQuerying/kang2017noscope,VideoQuerying/goswami2023active} are designed for specific situations and/or video types (domains).
They require new software or algorithms or significant retraining (for deep-learning based approaches~\cite{videoQuerying/bastani2020miris,videoQuerying/kang2017noscope}) for new situation types or videos belonging to different domains or applications, and sometimes need human intervention~\cite{VideoQuerying/goswami2023active}. However, the availability of CCTV cameras, dashcams, personal recording devices, and other camera types has increased the \textbf{volume and variety} of video data as well as  24/7 recording videos. Performing VSA on this large volume of videos manually, using a human-in-the-loop, is not practical. Also, the use of custom analysis algorithms cannot scale to new situation types or deal with the variety of videos generated in different domains. Finally, whether custom video analysis can be performed in real-time remains an open question due to the complexity of the algorithms and the current frame processing rate of these algorithms. However, in our approach, if the VCE can be done in real-time, there is hope that analysis can also be supported in real-time. 

In this thesis, we propose a two-step VSA approach, termed \textbf{VideoScoop}. The first step is VCE, which is a one-time extraction of video contents (using state-of-the-art packages). The second step (VSA) involves a separate analysis of the extracted contents to identify specific situations. This is our focus and where the contributions of this thesis lie. This is accomplished by specifying
situations using queries/analysis expressions and analyzing the extracted contents using a query-processing engine/analysis algorithms for analyzing the extracted contents from VCE. The separation of VCE and VSA provides several benefits. The VCE workflow can be updated as needed to reflect ongoing progress. It can utilize both traditional and deep-learning-based approaches, which are gaining momentum and showing significant promise for enhanced content extraction. Since VSA is separate from VCE, a large \textbf{volume and variety} of interesting situations can be specified on any video and processed using the operators/analysis algorithms developed. Multiple situation analyses can be done concurrently using a single pass on the extracted contents. New operators/analysis algorithms can be added in a modular manner. These operators, along with existing operators or the analysis algorithms, can be used to compose complex situations in a non-procedural manner or as a composition of analysis algorithms. All of the above features are difficult and challenging to incorporate into custom approaches.

Additionally, VSA can be performed as a forensic analysis on collected videos to understand their contents (or situations) and even to train LLMs~\cite{LLMSituationDetection/yuan2025empowering,LLMVideoUnderstanding/tang2025video} where appropriate. This type of analysis is helpful for criminal investigations, various security and surveillance applications, and pedagogical purposes. VSA can also be performed in real-time if VCE can provide real-time extraction of contents, which can be fed as a stream for VSA processing. Since the proposed approach is built on the stream processing paradigm, it can be extended to real-time video analysis by adapting algorithms (essentially one-pass algorithms) and techniques (such as scheduling and load shedding), and Quality of Service (QoS) and Quality of Experience (QoE) specifications to move towards real-time video analysis, which is very difficult or not possible using the traditional approaches. This capability is crucial for many applications where timely responses are critical, such as AL environments and real-time security monitoring to generate alerts in SL applications, etc. Recently, some approaches~\cite{videoQuerying/yadav2019vidcep} have adopted a two-step approach that separates VCE and VSA, built on the stream processing paradigm. However, these approaches cannot identify a variety of situations across different video types and are not general-purpose, compared to our proposed approach.

The following are some real-life scenarios that exemplify the advantages of our approach for both forensic and real-time video situation monitoring. Consider the event of Ms. Pelosi’s residence being breached~\footnote{\url{https://www.cbsnews.com/news/paul-pelosi-attack-break-in-caught-on-security-camera/}}. Even though the CCTV cameras were recording video footage 24/7, no one was watching those videos manually. Despite the availability of several custom algorithms developed for other situations, they could not be employed to perform forensic analysis on these videos after the incident had happened. Manual forensic analysis was performed \textbf{after the fact} to understand what happened. Using the proposed approach, a situation can be specified as ``someone lingering close to the front door or window, followed by a loud noise (breaking or trying to open)''. For another real-life example, consider the Las Vegas shooting in 2017~\footnote{\url{https://www.cbsnews.com/feature/las-vegas-shooting/?utm_source=chatgpt.com}}. An individual came through the \textit{same door several times in a short period, carrying big suitcases.} If this situation (\textit{"a person coming through the same door multiple times carrying big objects in a short interval"}) can be specified as a continuous query and computed, an alert can be generated in a timely manner using our proposed approach.

Given the limitations of the existing VSA alternatives, there is a clear need for an ``\textbf{out-of-the-box}'' solution to automate video situation analysis. We believe our proposed solution accomplishes that with low risk and high return. Low risk is due to our leveraging and extending mature and tested technologies, such as video content extraction and stream data processing for sensor data. The high return is due to its ability to move away from customized or manual solutions and provide an extensible path to real-time automation. However, many challenges need to be addressed and resolved along the way, as indicated in the general problem statement below. 

\section{Problem Statement} 

\textit{Develop an extensible, domain-independent framework for efficient, accurate, and robust video situation analysis to detect a wide range of interesting situations expressed as queries using a non-procedural query language or analysis expressions.}

For the two-step approach described earlier for the above general problem statement, several sub-problems need to be addressed to provide a complete framework. We have also chosen three diverse domains -- AL, SL, and CM -- to demonstrate the domain-independence of the proposed approach. The sub-problems that are addressed in this thesis are:

\begin{enumerate}
    \item \label{situations_and_framework}\textbf{Identification} of: (a) \textit{components of a two-step, domain-independent general-purpose VSA framework} to correctly identify situations from arbitrary videos, and (b) develop an approach to express interesting situations  (both primitive and composite) using the chosen domains for illustration.

    \item \label{vce_workflow} \textbf{VCE workflow:} Identify appropriate packages and algorithms to extract as much content as possible and create a workflow whose components can be easily updated as new algorithms and technologies emerge. 
    \item \label{representation_models} \textbf{Representation models for extracted contents:} Identify the representation models (such as relational or graphs) and their details that can capture extracted contents and support queries and/or analysis expressions on them. Chosen models need to be expressive and amenable to efficient processing.
    \item \label{cql_va} \textbf{Continuous Query Language for VA (CQL-VA):} Extend Continuous Query Language (CQL) seamlessly by adding new spatial and temporal operators needed for video situation analysis. Formally define the syntax as well as semantics of these operators, including backward compatibility with CQL/SQL. 
    \item \label{graph_analysis} \textbf{Graph algorithms for situation analysis:} Identify available algorithms for VSA for the graph model and their applicability to extracted video contents. Develop new algorithms that are needed for the situations identified in sub-problem~\ref{situations_and_framework}.
    \item \label{situation_expressions} \textbf{CQL-VA query evaluation:} Composing situations using operators developed above and efficient algorithms for computing them efficiently and accurately. 
    \item \label{implementation} \textbf{Implementation of VideoScoop Stream processing System:} Extend the current server developed for sensor stream data processing~\cite{phdThesis/Jiang05,msThesis/Kendai06,msThesis/Sonune03,msThesis/Gilani03} with new CQL-VA operators.
    As part of this, develop a web-based dashboard (termed \textbf{VideoScoop}) for specifying queries on videos graphically and analyzing the results interactively.
    
    
    \item \label{experimental_analysis} \textbf{Experimental analysis:} Experimentally validate the accuracy, efficiency, and robustness of situations detected using a wide range of diverse videos drawn from the domains of interest. Demonstrate robustness by mixing and matching situations to ensure that all occurrences are detected correctly. Videos of different lengths, frame rates, and complexity should be used for validation. 
\end{enumerate}

Chapters of the thesis correspond to the sub-problems indicated above. Related work is given in Chapter~\ref{chap:relatedwork}. Solutions to each of the sub-problems discussed above constitute the contributions of this thesis.

\section{Contributions and Thesis Road-map}

The contributions and thesis road-map are outlined below.

\begin{enumerate}


\item 
    (a) \textbf{A domain-independent VSA framework is introduced} in chapter~\ref{chap:problem-statement}. This approach extracts different types of video contents \textbf{once} by developing a workflow of available VCE algorithms of different categories. The extracted contents are represented using alternative representation models: the extended relational model (R++) or graphs. The rationale behind using two models is that one model is more expressive and that certain analyses are easier to perform in one model than in the other. Together, both models can analyze a wide variety of situations. Once appropriately represented, the extracted contents are streamed to a Continuous Query Processing (CQP) system or a Graph Analysis (GA) system, which can identify situation(s) of interest posed as queries/analysis expressions. The CQP system processes the queries using the CQL-VA operator extensions and the Stream Processing (SP) functionalities. The GA system processes the analysis expressions using the underlying analysis algorithms developed for each situation template. Finally, the video frames that contain the submitted situations are given as output to the user. In chapter~\ref{chap:problem-statement}, the challenges associated with developing each module of the framework and how solving these challenges can help in developing a domain-independent solution are discussed elaborately.

    (b) \textbf{Expressing interesting primitive and composite situations} using the domains  AL/SL/CM, the notion of situation templates is defined (discussed in chapter~\ref{chap:problem-statement}). Situation templates identify a generic primitive situation that can be parameterized, avoiding the need for multiple versions of the same situation across domains. Several parameterized situation templates are designed to express primitive situation variants across domains. The usefulness of these templates to develop a domain-independent VSA solution is discussed in more detail in chapter~\ref{chap:problem-statement}.

\item \textbf{A VCE workflow has been developed} to address sub-problem~\ref{vce_workflow} in chapter~\ref{chap:vce_workfolow}. Different categories of VCE algorithms were identified and assembled in the workflow to extract needed contents from a video for analysis.


\item \textbf{Alternative representation models}: In chapter~\ref{chap:Representation}, an extended relational model (termed R++) is proposed with new data types (sequences, vectors, arrables) to support various extracted content types. In addition, a graph model is also introduced. For flexibility in developing algorithms as well as efficiency, the use of three types of graph models is proposed, along with a discussion of their advantages, disadvantages, storage requirements, compression, etc.

\item \textbf{Several graph algorithms for video analysis (GA-VA)} for the analysis of alternative graph models are proposed in chapter~\ref{chap:graph-ana-videocontents} to address sub-problem~\ref{graph_analysis}. The advantages and disadvantages of these algorithms in terms of accuracy and efficiency are detailed in chapter~\ref{chap:graph-ana-videocontents}. Finally, the proposed algorithms are \textbf{experimentally evaluated} for accuracy, robustness, and efficiency (addresses sub-problem~\ref{experimental_analysis}). For this purpose, several datasets containing videos (with embedded situations of interest) of different lengths and categories (small/medium/large/mixed) from the three domains have been collected/generated, and extensive experimentation has been performed on these datasets for evaluating the proposed algorithms. 


\item For the R++ model, several new \textbf{CQL-VA} operators are  introduced in chapter~\ref{chap:CQL-VA} to address the sub-problem~\ref{cql_va}. 
New primitive operators for video analysis and extensions of existing operators for the correct and efficient processing of extracted video contents will be introduced in this chapter. The syntax, semantics, backward compatibility with the relational model, closure property, and complexity of these operators will be discussed elaborately in chapter~\ref{chap:CQL-VA}. Implementation of \textbf{CQL-VA} operators in an existing SP system ~\cite{phdThesis/Jiang05,msThesis/Kendai06,msThesis/Sonune03,msThesis/Gilani03} is also discussed in this chapter to address sub-problem~\ref{implementation}.

\item \textbf{Situations as CQL-VA queries and their evaluation} are discussed in  chapter~\ref{chap:situation-as-query} to address sub-problem~\ref{situation_expressions}. Alternative CQL-VA query or analysis expressions for situations, their advantage, and disadvantages will also be discussed. Algorithms are developed for new operators and extensions of existing operators. All algorithms will be assessed based on time complexity, accuracy, and response time. Finally, \textbf{extensive experimentation} will be performed to evaluate the accuracy, robustness, and efficiency of the queries to address sub-problem~\ref{experimental_analysis}.

\end{enumerate}

The rest of the thesis is organized as follows. In chapter~\ref{chap:relatedwork}, the literature related to this thesis is discussed along with its limitations. In chapter~\ref{chap:conclusion}, we draw the conclusion and discuss future work. This thesis has led to three \textbf{publications}~\cite{DEXA/BillahC24,ADBIS2023/BillahSC,PETRA2024/BillahSC24}.

\chapter{A Domain-Independent Framework for Video Analysis}
\label{chap:problem-statement}

In this chapter, we first define video situations and discuss different interesting situations for the three contrasting chosen domains -- Assisted Living (AL), Civic Monitoring (CM), and Surveillance (SL). Then, we discuss how domain-independence can be achieved by categorizing situation variants from these domains into parameterized situation templates. Finally, we introduce our proposed framework and discuss the challenges associated with developing its various components.

\section{Video Situation Definition}
A \textbf{video situation} comprises action(s) or events~\footnote{In this thesis, we do not differentiate between events and situations. There is considerable amount of work on event specification and processing~\cite{snoopibcont::raman,Cha+96:snoop-sem:tr,DKE/AdaikkalavanC06,msThesis/Krishnaprasad94,phdThesis/Adaikkalavan06}, but for our purpose that differentiation is not critical.} of interest occurring in a video across multiple consecutive sequences of frames. A situation may happen once or multiple times in a video. There can be multiple different situations occurring in a video concurrently or sequentially. As situations vary from simple (atomic) to complex, we categorize video situations as (i) primitive and (ii) composite situations.

A \textbf{primitive situation} (PS) is a single basic or primitive \textit{action} performed by one or more objects over multiple consecutive sequences of frames. In other words, a primitive situation is atomic and cannot be broken down into smaller situations. Some examples of primitive situations are: \textit{``A person or an object is \textbf{moving in a specific direction}''}, ``\textit{A car or vehicle is \textbf{turning left or right}}'', etc. Here, turning left, moving in a direction, etc., are primitive situations that occur across multiple consecutive sequences of frames. A list of primitive situations\footnote{Currently, we have not used pose vectors that can be extracted from videos for primitive situations. With their use, more primitive situations, such as "sitting on a chair" or "dropping an object" can be defined.} used throughout the thesis is shown in Table~\ref{tab:PrimitiveSituations}. These primitive situations are composed to detect composite or complex situations.

A composite or \textbf{complex situation} (CS) consists of multiple primitive actions occurring (mostly) sequentially or concurrently over multiple sequences of frames. In other words, complex situations are composed of multiple primitive or composite situations using operators such as SEQUENCE or FOLLOWED BY, AND, OR, NOT\footnote{NOT corresponds to a non-occurrence and has to be defined clearly and detected properly.} etc.~\cite{ChaMis91,ICDE/ChakravarthyKTB95}. For example, consider the complex situation ``\textit{A car \textbf{stopped} at an intersection with \textbf{heavy traffic flow}}''. This situation consists of two primitive situations occurring at the same time (or in a small number of consecutive frames). This situation can be composed as: ``\textit{A car has \textbf{stopped}}'' (primitive situation) \textbf{AND} ``\textit{ \textbf{heavy traffic flow} at an intersection}'' (primitive situation). Here, the \textbf{AND} operator is used to connect the above two primitive situations.
In other words, for this complex situation, two primitive situations — stopping and heavy traffic flow — need to be detected occurring at the same time (or in a small number of consecutive frames) in a video. 


Another example of a complex situation is \textit{"two people crossing each other"} can also be defined using the FOLLOWED BY (or SEQUENCE) on the two primitive situations ``\textit{Two people \textbf{are coming closer}}'' and ``\textit{Two people \textbf{are moving apart}}''\footnote{Note that for all the above situations, unless depth information is extracted from videos, it will not be possible to detect how far apart the objects are along the z-axis.}. The above illustrates how complex situations can be expressed using primitive and previously defined composite situations. 

In Table~\ref{tab:CompositeSituations}, several examples of composite situations and how they are composed using primitive situations are shown. In this table, the keywords \textit{FOLLOWED BY, AND, OR, NOT} are used to connect (or compose) the primitive situations. It is also possible to detect non-occurrence of a situation S2 (represented as Not(S1, S2, S3) \textbf{within an interval specified either by time or by the occurrence of other situations/events.} In the above specification, S1 and S3 can be other situations or temporal situations/events. For example, an object disappearing can be specified in terms placement of the object as the starting interval and t units of time as the end interval. End interval can also be specified as an event/situation, such as a door opening in front of which object was placed.

\section{Situation Templates and Domain-Independence}

A domain-independent solution should have the capability to identify primitive situations by developing generalized analysis algorithms/operators. Once appropriate analysis algorithms or operators are available to detect primitive situations, they can be composed further to identify composite or complex situations. To achieve this, the \textbf{primitive situations} from the situation variants from the three domains are grouped into parameterized templates. This approach can be easily extended to additional domains.

 \textit{A \textbf{Situation Template (ST)} is a generalized representation of a primitive situation, containing basic or atomic actions/events that are useful for specifying multiple actual primitive situations in a domain or from multiple domains}. 


We will demonstrate the importance of developing such templates below using simple example situations from Table~\ref{tab:PrimitiveSituations}. Consider the situations \textit{``A car moving towards/away from another car?"} (ST6(b) in Table~\ref{tab:PrimitiveSituations}) from the CM domain and \textit{``When was a person moving towards/away from another person?"} (ST6(a) in Table~\ref{tab:PrimitiveSituations}) from the SL domain. In both cases, different types of objects are responsible for the same situation, ``moving towards'' or `` moving away'', which require the same primitive computation (namely, the direction in which an object is moving with respect to another object). The above two situation variants can be expressed as a more generic situation template: \textit{An object moving in a specific direction} (shown as ST6 in Table~\ref{tab:PrimitiveSituations}). Here, the direction of moving (moving towards or away) and the object class label(s) are parameters for this template.

\begin{table}[!htb]
\footnotesize
\caption {\small{\textbf{Primitive situation templates and parameterized situation examples}. }}
\begin{tabular}{|p{.05\textwidth}|p{0.22\textwidth}| p{0.15\textwidth}|p{0.5\textwidth}|p{0.08\textwidth}|}
\hline
ID & Situation Template (ST) & Parameters & Example Primitive Situations & Domains \\ \hline

\multirow{3}{*}{ST1} & \multirow{4}{.22\textwidth}{\textbf{No. of object(s)} over/below a threshold in an interval} & \multirow{3}{.15\textwidth}{Class label(s), threshold, an interval} & (a) \textbf{Heavy traffic flow} in a traffic intersection. & CM  \\ \cline{4-4}
~ & ~ & ~ & (b) An interval during which parking lot was \textbf{empty/busiest}. & ~ \\ \cline{4-4}
~ &  ~ & ~ & (c) Number of people \textbf{entering} a building over a period. & ~ \\ \hline
\multirow{2}{*}{ST2} &
\multirow{2}{.22\textwidth}{\textbf{Occurrence/non-occurrence (in a specified interval)} of object(s)} & \multirow{2}{.15\textwidth}{Image(s), class label(s)} & (a) \textbf{Presence/absence} of specific person(s) using his/her image(s). & SL/AL \\ \cline{4-4}
~ & ~ & ~ & (b) Count the number of appearances (non-consecutive frame occurrences) of an object in a video. & ~ \\ \hline

ST3 & \multirow{2}{.22\textwidth}{\textbf{Duration} of object(s) presence} & Class label(s), duration & (a) A person \textbf{waiting} at a crosswalk/parking lot for t seconds. & CM \\ \cline{4-5}
 ~ & ~ & ~ & (b) A person \textbf{lingering} at a doorstep for t seconds & SL \\ \hline 
ST4 & Object(s) not \textbf{moving} for a duration & \multirow{2}{.15\textwidth}{Class label(s), duration} & (a) A car \textbf{stopped} (at an intersection) for t seconds &  CM \\ \cline{4-5}
~ & ~ & ~ & (b) An individual is \textbf{immobile}  (or standing) for t seconds & AL \\ \hline 
ST5 & Object(s) taking a \textbf{turn} & Class label(s), turning direction (left/right) & (a) A car/person taking \textbf{left/right} turn & CM/SL \\ \hline

ST6 & \multirow{2}{.25\textwidth}{Object(s) moving in a specific \textbf{direction}}) & \multirow{4}{.15\textwidth}{Class label(s), direction (towards/away)} & (a) When was a person moving \textbf{towards/away} from a check-post/another person?  & SL \\ \cline{4-4}
~ & ~ & ~ & (b) A car moving \textbf{towards/away} from another car in an intersection & CM \\ \hline


ST7 & \textbf{Same object(s)} present in two different videos & Class label(s) & (a) Has the \textbf{same person} entered and exited a shopping mall/building monitored using entry and exit cameras? & SL\\ \hline

ST8 & \multirow{3}{.20\textwidth}{\textbf{Group(s)} of objects of given size} & \multirow{3}{.15\textwidth}{Class label(s), size n} & (a) Which individuals are isolated (not part of a \textbf{group})? & AL \\ \cline{4-5}
~ & ~ & ~ & (b) The largest \textbf{group(s)} of individuals present in a premise & CM \\ \cline{4-4} 
~ & ~ & ~ & (c) \textbf{Group(s)} of n individuals (or size n) & ~ \\ \hline \hline 
 ST9 & Identify the $n^{th}$ occurrence of a specified situation/event & \multirow{2}{.15\textwidth}{$n$, parameters of the situation specified} & (a) Identify the first/last/$n^{th}$ occurrence of heavy traffic flow or empty/busy parking lot & CM \\ \cline{4-4}
 ~ & ~ & ~& (b) When was the first time two people were moving towards/away from each other? & SL \\ \hline
\end{tabular}
\label{tab:PrimitiveSituations}
\end{table}

By specifying situations from different domains with a parameterized template, developing an algorithm/operator for identifying ST6 (with different parameters) will be sufficient. Additionally, by developing such templates, computations designed for one or more domains can be applied to others with different parameters. This eliminates the need for developing domain- or situation-specific solutions and establishes the basis for a domain-independent solution.

Given the importance of situation templates in developing a domain-independent solution, we have categorized several interesting situations from the three domains (AL/SL/CM) into templates as shown in Table~\ref{tab:PrimitiveSituations} (column 2). For each ST, their parameters, one or more example situations from the three domains (AL/SL/CM) are shown. In our view, these situation templates, and the list of situations (both primitive and composite) shown in Table~\ref{tab:PrimitiveSituations} and~\ref{tab:CompositeSituations} are sufficient for laying out the foundation for a domain-independent framework.



\begin{table}[!htb]
\footnotesize
\caption{\textmd{\textbf{Complex situations using primitive situations}}}
\centering
\vspace{10pt}
\begin{tabular}{|p{0.03\textwidth}|p{0.20\textwidth}|p{0.6\textwidth}|p{0.1\textwidth}|}
\hline
\textbf{ID} & \textbf{Complex Situations} & \textbf{Primitive Situations Required} & \textbf{Domains} \\ \hline

CS1 & \multirow{2}{.2\textwidth}{A delivered object missing from the place of delivery} &   \makecell[l]{ An object is present for t seconds (ST2(a)) \\
\textit{FOLLOWED BY}
the same object is \textit{NOT} present (ST2(a)) \\ \textit{AND} the duration of non-presence of the object is \\ n seconds (ST3(a))} & SL \\ \hline

CS2 &  A car stopping in an intersection with heavy traffic flow & \makecell[l]{A car stopped in an intersection (ST4(a)) \\ \textit{AND} 
heavy traffic flow in an intersection (ST1(a))} & CM \\ \hline
CS3 & A car taking a U turn & \makecell[l]{A car taking a left turn (ST5(a)) \\
\textit{FOLLOWED BY}
the same car taking left turn again \\ (ST5(a))} & CM \\ \hline

CS4 & \makecell[l]{A person moving \\ towards  
a check post \\ and 
returning \\ without  crossing it} & \makecell[l]{A person moving towards a check post (ST6(a)) \\
\textit{FOLLOWED BY}
the \textbf{same} person moving away \\ from the check post (ST6(a))  \\
\textit{OR} the same person taking U turn (CS3) } & CM \\ \hline
CS5 & Two people crossing each other& \makecell[l]{Two people moving towards each other (ST6(a)) \\
\textit{FOLLOWED BY}
the \textbf{same} two people moving away \\
from each other (ST6(a))} & SL \\ \hline

CS6 & \multirow{2}{.2\textwidth}{A person is being picked up by a vehicle} & \makecell[l]{A person waiting at a parking lot for t seconds (ST3(a)) \\ 
\textit{FOLLOWED BY}
a car moving towards the person (ST6(a)) \\
\textit{FOLLOWED BY}
the same person \textit{NOT} present (ST2 (a)) \\ 
\textit{AND} the duration of non-presence is n seconds (ST3(a))} & CM \\ \hline


CS7 & Two individual(s)  \textbf{entering and exiting} a 
premise within 
t seconds
of each other & \makecell[l]{
A person present in the entry and exit videos (ST8 (a)) \\  
\textit{FOLLOWED BY} a different person present in the same \\ entry and  exit  videos (ST7(a)) 
after t seconds} & SL/CM \\ \hline

\end{tabular}
\label{tab:CompositeSituations}
\end{table}

\section{Proposed Video Situation Analysis Framework }

\begin{figure}
    \centering
    \includegraphics[width=1\linewidth,keepaspectratio=true]{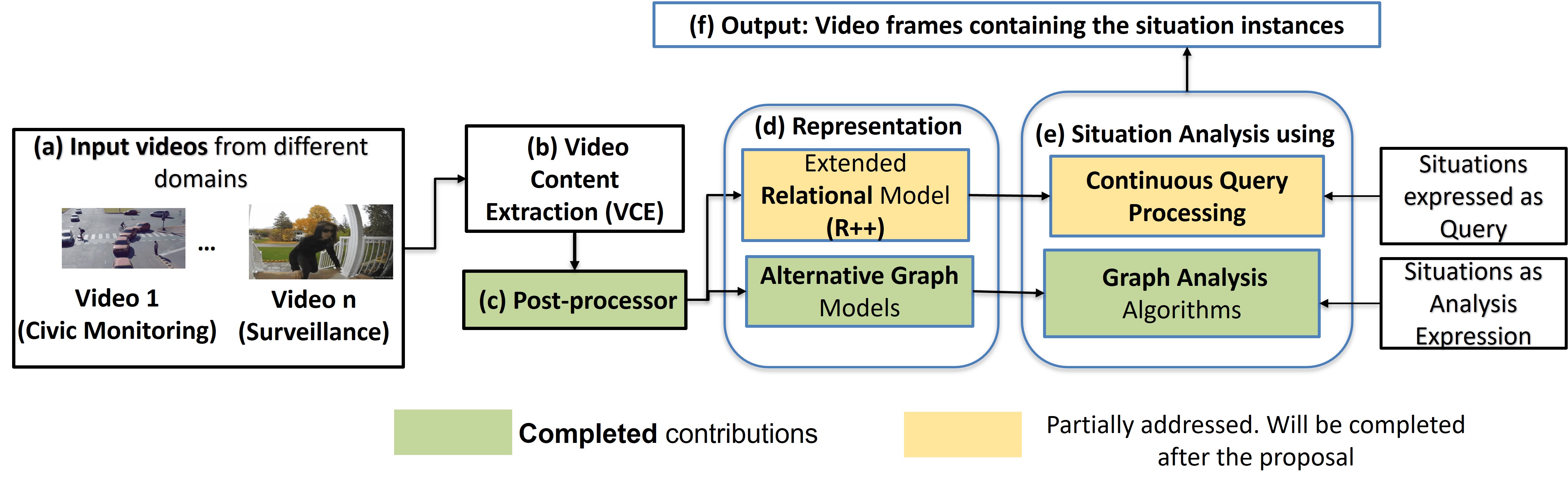}
    \caption{Components of proposed video situation analysis framework.}
    \label{fig:VA-Framework}
\end{figure}

In this thesis, a general-purpose Video Situation Analysis (VSA) framework (illustrated in Figure~\ref{fig:VA-Framework}) is proposed to facilitate the accurate, efficient, and robust detection of situations in given video(s). This framework takes as input a set of videos. The Video Content Extraction (VCE) module processes one video at a time, and extracts contents from the videos using a combination of different types of VCE algorithms to maximize the amount of extracted contents. The extracted contents are then represented using either the proposed extended relational model (R++) or alternate graph models. Once appropriately represented, these extracted contents are streamed to a Continuous Query Processing (CQP) system or a Graph Analysis (GA) system. The situations of interest are posed as queries using the proposed Continuous Query Language for Video Analysis (CQL-VA) operators or analysis expressions using the proposed Graph Algorithms for Video Analysis (GA-VA) and submitted to the proposed framework. The CQP system processes queries using the CQL-VA operators, an extension of Continuous Query Language (CQL), with new primitive operators and optimized extensions of existing CQL operators. The proposed Graph Algorithms for Video Analysis (GA-VA) are used to analyze the extracted video contents for identifying the given situation. Developing components of the proposed VSA solution involves several challenges. They are elaborated below.




\section{Video Content Extraction Challenges}
Several categories of VCE algorithms have been designed over the years for extracting different types of contents (e.g., object detection algorithms~\cite{objectRecognintion/yolov10} extract object bounding box location and class label, pose estimation algorithms extract object pose points~\cite{PoseEstimation/HRNeT}, etc.) The state-of-the-art (SOTA) is also different for each of these tasks. The SOTA algorithms from each VCE task must be assembled in a workflow to extract as much content as possible with the best accuracy. This is challenging because each VCE task expects different inputs (e.g., bounding box location of objects, video frames, etc.). Identifying how to propagate outputs extracted from a VCE algorithm to another algorithm of a different category is challenging, as the expected input for each algorithm type can be different. The above must be handled to extract the maximum possible information as accurately as possible. In this thesis, a workflow was developed addressing all of the above. A VCE workflow is elaborately discussed in Chapter~\ref{chap:vce_workfolow}.

\section{Challenges of Representing Extracted Video Contents}


\subsection{Need for Alternative Models}

The extracted video contents from the VCE workflow contain different data types (numeric, categorical, and multi-dimensional vectors of various types, etc.) The extracted contents can be represented in different ways. Currently, two widely used representation models in the literature are the relational model and the graph model. The relational model is known for representing data in a tabular format and has been widely used for several decades. Query optimization has been a strong point, relieving the end user from worrying about how to develop algorithms and optimize them. SQL query optimizer takes care of that, which will also be possible for our approach. It also supports a wide variety of Relational operators and is also being extended periodically as part of the SQL standard. The graph model, which is gaining popularity, is ideal for representing entities as nodes (or vertices) and the relationships among them as edges. Using node and edge labels, a temporal sequence can be modeled. They can also be modeled as individual graphs showing the evolution of objects and relationships. 

As we analyzed the templates developed  for video situation analysis (listed in Table~\ref{tab:PrimitiveSituations}), it became clear that some situations are better suited for the relational model and some for the graph model.  In other words, the relational model appears well-suited for analyzing certain categories of situations, while the graph model is better suited for others. Below, we explain why one model is better than another in certain cases with an example situation from Table~\ref{tab:PrimitiveSituations}.

\underline{\textit{Identify group(s) of different sizes in a video:}} This situation is useful in understanding the behavior of people in assisted living about their interaction with others or their isolation over a period of time. This situation requires clustering objects based on their location (extracted as bounding box coordinates, which is a rectangular area encompassing an object and containing the x and y coordinate positions of the bottom left and top right corners of the rectangle) across multiple video frames to identify the clusters/group(s) of different sizes. Applying traditional clustering algorithms such as K-means on a relational model (by representing each bounding box coordinate as separate columns) is extremely complicated, as clustering algorithms inherently require iterative recalculations (of cluster centroids, cluster assignments, etc.) until a convergence point is reached. Another alternative for performing clustering using a relational model is to use embedded Structured Query Language (SQL) within a host programming language such as C, Java, or Python. The data points (bounding box coordinates) need to be fetched using SQL queries, and the host language performs the computations for K-means. This incurs overhead for fetching the data and adds complexity to orchestrating the algorithm using the host language while maintaining database operations.

On the other hand, by modeling objects as nodes and their location (or bounding box) as node attributes, traditional clustering algorithms such as K-means can be easily applied to cluster the nodes. Further, expressing the situation as a graph analysis expression is easier than writing a query for a relational model.
Apart from these, in some situations, graph characteristic information (e.g., connected components, degree) can be leveraged for analysis.

In this thesis, we have explored alternative representation models to enable the efficient and accurate analysis of a wide variety of situations by utilizing the advantages of both relational and graph models. However, representing the extracted contents in each model poses several challenges, which we discuss below.



\subsection{Representing Extracted Contents with Relational Model}
The categorical and numerical data extracted by VCE can be modeled easily using a relational model (in its first normal form). However, modeling vectors of different sizes and types is not that straightforward. For example, the location of an object in each video frame is extracted as a bounding box. It would require four columns to represent the bounding box of an object using the traditional relational model. To perform further computation (e.g., the distance between bounding box centroids of two different objects) on these columns, a self-join must be performed, and eight columns from two relations need to be compared. This will become much more complicated when multidimensional vectors, such as feature vectors, need to be compared. 

On the other hand, the JSON array data type in SQL22, or the VARRAY data type in the Procedural Language extensions to the Structured Query Language (PL/SQL), can represent all the different vector types extracted by VCE. However, they do not support computations other than bounding box spatial relationships (e.g., overlap, left, right, etc.). User-defined procedures must be written to perform computations on other vector types (e.g., similarity comparison of feature vectors). Another way to represent these vectors is to use the Array DBMS~\cite{arrayDB/baumann2021array, ArrayDBMS/maier2013arrayql}, geospatial databases such as ArcMap~\cite{ArrayDBMS/ArcMap}, or Oracle spatial databases. Even though these databases support representing multi-dimensional vectors (or arrays), these systems are more suitable for spatial (or scientific) data, where each data point is a coordinate, and performing computations on specific array indexes (or geometric shapes) is essential. These models cannot represent the other extracted content types (numerical and categorical) and computations on different vector types (e.g., distance or spatial relationships between bounding boxes). User-defined functions must also be written for these systems to support computations on different vector types. \textit{This is similar to designing custom algorithms for each new situation.}

Another characteristic of videos is that the frames are generated sequentially. These frames and their contents must be processed in the sequence generated for specific situations (e.g., an object moving to the left). Column-oriented databases, such as AQuery~\cite{lerner2003aquery} or databases that support sequences as a data type and computations on them, such as Seq~\cite{seshadri1995seq} and SRQL~\cite{Sequence/SRQL}, were introduced to support computations that maintain order or operate on sequences. Even though the arrable data model proposed in AQuery can represent two-dimensional vectors and order-preserving computations, it cannot represent vectors of more than two dimensions, different types of vectors extracted by VCE, and computations on them. 
The arrable data model needs to be extended with different data types and support for multi-dimensional vectors so that different computations required for analyzing extracted video contents can be supported, and we can benefit from the column-oriented semantics of the arrable data model. 

Given the limitations of relational models and other data models built upon the relational model, we have proposed an extended relational model with new data types, along the column-oriented semantics of arrables, in Chapter~\ref {chap:Representation}.


\subsection{Representing Extracted Contents with Graph Model}
In videos, the same set of objects can be present in multiple consecutive sequences of video frames. The properties or extracted attributes of objects (e.g., bounding boxes) by VCE are likely to change from frame to frame. This presents a unique challenge for representation using the graph model: \textit{How to model objects and their attributes across frames?} 

Some recent works~\cite{videoQuerying/yadav2019vidcep} on graph modeling for extracted video contents model each object as a node and generate a Single Graph per Frame (SGF) of a video. The number of graphs generated equals the number of non-empty frames~\footnote{A non-empty frame is a frame where the VCE module extracts at least one object and corresponding attributes}. The other alternative in the literature is generating a Single Graph per Video (SGV) (e.g., Video Scene Graph Generation (SGG)~\cite{Scenegraph/ActionGenome,Scenegraph/VidSGG}, VEKG-TAG~\cite{VideoRepresentation/yadav2020knowledge}), where each object is represented as a vertex once, and all the non-empty frames are aggregated into one graph (containing a forest of graphs with large connected components). 

A third alternative, which has not been explored in the literature, is to generate Multiple Graphs per Video (MGV) for  representing several consecutive frames based on some property (e.g., objects are not split across graphs). This model provides a third alternative with several advantages. This approach is similar to multilayer networks (or MLNs) where a single complex graph is separated into layers with clearer semantics. This also enables parallel analysis, thereby improving efficiency.  If needed, analysis results from each graph (or layer) can be combined to perform global multi-graph or MLN analysis. This approach termed the decoupling-based approach~\cite{phdThesis/Santra20} has been shown to be efficient, scalable, and can preserve global analysis accuracy. However, if these graphs are generated carefully so that no two graphs have the same set of objects (or graphs are disjoint on objects), this allows for easier composition of results if the decoupling-based approach is applied for the analysis. The graph sizes (in terms of the number of nodes or frames) can also be balanced, so that each graph can take approximately the same amount of processing time, thereby reducing the overall processing time.

To the best of our knowledge, the above aspects of modeling and analyzing extracted contents have not been explored in the literature. Additionally, each of the above models has its advantages and disadvantages in terms of modeling clarity, storage requirement, and analysis accuracy and efficiency. Different aspects of modeling extracted video contents as graphs are elaborately discussed in chapter~\ref{chap:Representation}.

\section{Challenges of Analyzing the Extracted Video Contents} 

Once the extracted contents are represented appropriately, the next step is to design operators or analysis algorithms to perform those computations accurately, robustly, and efficiently. However, the existing relational model operators or graph analysis algorithms are insufficient for analyzing the situations listed in Table~\ref{tab:PrimitiveSituations} from the extracted contents. Below, we discuss the challenges associated with analyzing extracted contents using existing relational model operators and graph analysis algorithms.


\subsection{Relational Model Operators}
Some of the limitations of the traditional relational model and its operators for situation detection in videos are discussed below.

\underline{\textit{Comparison of vectors:}} 
For our purpose, multidimensional vectors of different types (e.g., feature vectors, bounding boxes) need to be compared in different ways. For example, to analyze the situation \textit{``Which individuals entered and exited a premise monitored using entry and exit cameras within t seconds of each other?"}, videos from two different cameras need to be joined to identify the same individuals entering and exiting. For this purpose, object feature vectors from two videos need to be matched as part of the join condition, which is not possible using existing relational model operators. A matching condition (with an appropriate similarity measure) is needed for comparing feature vectors, which can be used as part of the join condition. Similarly, a different primitive operator is needed to compare the spatial relationship (e.g., overlap, above, below, etc.) between bounding boxes. These operators should be designed in a manner that allows them to be expressed as part of select and join conditions, similar to relational operators. \textit{on a side note,  none of the existing custom algorithms support situations involving multiple cameras.}

\underline{\textit{Performing joins efficiently:}} Join algorithms have been optimized over four decades, mainly to reduce or optimize I/O. In any join algorithm, comparisons have to be made on each tuple from two relations. Video data is slightly different from traditional relational data and also from traditional stream data.  As mentioned earlier, consecutive frames in a video contain repetitive information (e.g., the same set of objects in multiple frames). Not all frames are necessary for many computations. For example, to analyze the above situation (identifying the same object in two video streams) it is not necessary to compare all the frames (or tuples generated from the frames) to identify the common objects from two different videos. Hence, new algorithms or optimizations of existing operators are needed to perform the joins accurately and efficiently. Again, there can be multiple alternatives for performing joins; each of these alternatives has its advantages and disadvantages in terms of accuracy and analysis efficiency. In Chapter~\ref {chap:CQL-VA}, we propose different flavors of join for video analysis and discuss their advantages and disadvantages.

\underline{\textit{Primitive Operators:}} Some of the situations listed in Table~\ref{tab:PrimitiveSituations}, require primitive computations (e.g., direction of movement, proximity of objects, etc.) Identifying a minimal number of primitive computations, efficient algorithms for these primitive computations, and expressing these operators as part of SQL or Continuous Query Language (CQL) constructs is challenging. 

Apart from the above, new temporal operators are also needed to identify situations that span multiple consecutive sequences of video frames. The operators designed should be backward compatible with the relational model and maintain the closure property, allowing them to be composed with existing relational model operators. In chapter~\ref{chap:CQL-VA}, the Continuous Query Language for video analysis (termed CQL-VA) is elaborated with new primitive operators to address many of the above challenges.

\subsection{Graph Analysis Algorithms}
As mentioned earlier, for certain types of analysis, the graph models are more suitable than the relational model. In this thesis, we have chosen two situations from Table~\ref{tab:PrimitiveSituations} to analyze using the alternative graph models. They are:
\begin{enumerate}
    \item ST8: Group(s) of objects of different sizes
      \item ST6: An object is coming closer or moving away from another object 
\end{enumerate}

The challenges of analyzing the above situations using graph models are discussed below.

\underline{\textit{ST8: Group(s) of objects of different size:}} To analyze this situation, objects need to be clustered using their bounding box centroid distance to identify group(s) of different sizes. However, the set of objects and their corresponding bounding boxes or locations change over time (after a certain number of frames). Hence, the group(s) size and the number of groups(s) change over time (or after some frames) in a video. Several alternative approaches have been proposed in the literature to compute group(s), which are discussed below, along with their limitations.

\textit{Alternative 1:} There are some existing approaches~\cite{GroupExtraction/dynamic-network,GroupExtraction/li2022evolutionary} in the literature where group(s) are extracted from dynamic networks (where nodes can be added or removed after some timestamp or t seconds). These algorithms cluster nodes after every t seconds and are designed for applications such as dynamic social networks or traffic networks, as mentioned in ~\cite{GroupExtraction/dynamic-network,GroupExtraction/li2022evolutionary}. If these approaches were to be applied to finding group(s) in video, clustering must be done after every t seconds (t changes depending upon the application). This is not a good choice in general, even for videos from the same domains, as objects do not enter or appear in the field of view (FOV) at a pre-determined frequency. Even if they did, that frequency is not known or easy to compute for any object. Objects can appear or leave at any point in time, and hence the group(s) size and membership can change.


\textit{Alternative 2:} Another existing alternative for finding group(s) is proposed in ~\cite{GroupExtraction/chen2017anchor}. Here, a Single Graph per Video is generated, and the edges are drawn based on extracted feature points similarity (score above a given threshold). This algorithm identifies each strongly connected component in the graph as a group in the video. In other words, this algorithm assigns a set of objects to one group throughout the entire video. However, this algorithm does not apply to videos where the group size and an object's group membership change at different points in time. Additionally, using a threshold for drawing edges is not a good choice, as it depends on this threshold to identify the group(s). 

In summary, neither of the existing approaches is appropriate to analyze groups from different types of videos. Another alternative solution to the above is to perform traditional clustering on every frame and identify group(s) or clusters of different sizes. Although this approach can correctly identify all the different group(s), performing traditional clustering on every video frame can be too expensive for analyzing long videos (of a few hours or even minutes). Hence, there is a need to develop accurate and efficient heuristic-based algorithms to avoid/minimize the number of times traditional clustering is performed. For this analysis, alternative graph models pose different challenges and have their advantages and disadvantages. 


\underline{\textit{ST6: Object(s) moving in a specific direction (moving closer/moving away):}} For this situation, it is important to identify the frames where the distance between objects bounding box centroids is decreasing (to identify if objects are coming close) monotonically or increasing (to identify whether objects are moving away) monotonically. Again, the easiest solution can be developed using model SGF or SGV by analyzing if the distance between each pair of objects (or nodes) is decreasing or increasing across all the frames. This should be done for all pairs of objects that appear together in at least one frame in the video. However, the bounding box information is an an approximate value estimated by the VCE algorithms. As a result, the computed distance between object pairs will fluctuate even in consecutive frames, even though in raw videos these fluctuations are ignored by humans. As a result, this approach may identify one instance of an object moving closer/far away as multiple instances. This will become a problem when multiple pairs of objects are moving closer/far away multiple times in a video. Additionally, this is a brute force approach and will not be efficient when the video grows large (containing thousands of frames) and the number of objects increases. 

There are some existing approaches developed based on supervised learning~\cite{twoObjectClose/Samaras2012} to detect this situation in the literature, which identify the existence of this situation in a given video. However, they are not designed to detect multiple occurrences of the situation.

For the above two situations, there is a need to develop new, accurate, and efficient algorithms for various situations using graph models. In chapter~\ref{chap:graph-ana-videocontents}, we have proposed alternative algorithm(s) for the different alternative graph models, and discussed their advantages and disadvantages for the above two situations.

\chapter{RELATED WORK}
\label{chap:relatedwork}

In this chapter, we discuss related work from three different areas that are relevant to this thesis: Video Content Extraction (VCE), Stream data  processing (both sensor and video), Event Processing, Video Content Analysis (VCA), and LLM-based Video Analysis systems. First, we summarize the important VCE systems from the literature, what contents they extract, and the state-of-the-art (SOTA) algorithms developed to solve them. The stream and event processing literature is vast and extensively researched. Hence, we have listed only the seminal works in this area and discussed their limitations from the perspective of video content analysis. Finally, we have discussed several video content analysis systems developed over the years. 
\section{Video Content Extraction (VCE) Systems}
\label{sec:VCE-relatedwork}
For our framework, pre-processing videos (termed VCE) is an important first step to extract meaningful contents before analyzing it. Note that the analysis is performed \textbf{using only the extracted content} and not the original video. \textbf{It is important to understand that errors in the content extraction process (which exist due to various reasons, such as algorithm limitations, occlusion, mis-identifying the same object as different objects, etc.) cannot be compensated during analysis}. One can only use the state-of-the-art algorithms and hope for the best. This affects the accuracy of the analysis. Analysis accuracy is dependent on the extracted content and not the original video\footnote{This is also true, to a large extent, for customized algorithms that work on the video directly without the extraction step!}.
This translates to performing accuracy comparisons on the VCE output if ground truth (GT) can be computed for that. And this GT is likely to be different from the actual video GT for most videos.

For general-purpose analysis, it is crucial to identify the contents that are needed and whether they can be extracted accurately by the current VCE algorithms. \textit{Pragmatically, the current state-of-the-art in content extraction and its accuracy puts a limit or upper bound on situations that can be detected}. Over the years, the Image and Video Analysis (IVA) community has divided the VCE tasks into several categories, each with a different purpose. Below, we elaborate on the purpose of each task, the SOTA~\footnote{SOTA for VCE changes vary rapidly -- every three to six months.} algorithms for each task, and what can and cannot be achieved by these algorithms. 

\subsection{Object Detection} 

Object detection is one of the fundamental and challenging tasks in VCE. It is the basis for many other content extraction tasks, such as object tracking, pose estimation, and action classification. Object detection extracts bounding box coordinates from an image or video frame and assigns a class label to the object, along with a confidence score, from a predefined set of classes. Current deep learning-based object detection algorithms can be broadly divided into two categories. They are Region Proposal Network (RPN) based algorithms (e.g., Faster R-CNN~\cite{ObjectDetection/ren2015faster}, Masked R-CNN~\cite{objectRecognintion/MRCNN}, etc.) and Single-shot detection methods (e.g., the family of YOLO (YOLOv7~\cite{objectRecognintion/wang2022yolov7}, YOLOv8~\cite{objectRecognintion/yolov8}, Yolov9~\cite{objectRecognintion/wang2025yolov9}, Yolov10~\cite{objectRecognintion/yolov10}, etc.) algorithms)

Before 2022, RPN-based algorithms were the preferred choice when accuracy was the primary goal, while single-shot methods were favored for performance-critical applications due to their speed. 
However, the release of YOLOv7 in 2022 marked a significant breakthrough, surpassing all RPN-based detectors in both speed and accuracy~\cite{objectRecognintion/wang2022yolov7}. Subsequently, YOLOv8-YOLOv10 has further improved upon YOLOv7, solidifying their position as SOTA object detectors across both categories. Recently, transformer-based object detection algorithms such as DETR~\cite{objectRecognintion/zhu2020deformable} have also become popular. Depending upon the application requirements, dataset used for training, algorithm type, and many other factors, multiple SOTA algorithms exist in object detection~\cite{Survey/ObjectDetection}.

\noindent\textit{\underline{Current Limitations:}} Even though the above object detection algorithms have advanced significantly, there are several challenges they still need to address. They are discussed below:
\begin{itemize}
    \item \textit{Background detection:} Current object detection algorithms cannot identify \textit{all of the background} information/objects in a frame. Some background objects, such as trees, staircases, etc., can be identified (if the videos have an appropriate Field of View (FOV)). 
    \item \textit{Object class labeling:} Correctly labeling objects is another problem in these algorithms. For example, they may end up labeling a person entering through a door as a door, or a backpack can be labeled as something else (e.g., a mouse), depending upon the FOV, lighting, etc. Identifying all different object classes correctly is an area of research still being investigated by the IVA community.
\end{itemize}

\subsection{Feature Extraction} Object detection algorithms extract features of the detected objects and assign labels based on them. However, the single-shot-based algorithms, such as the family of YOLO algorithms, do not output feature vectors (or feature maps)~\footnote{Feature maps are a collection of feature vectors} of each extracted object. Hence, algorithms that depend on object recognition need to use a separate algorithm to extract feature vectors (or feature maps). Before the advent of deep learning, histogram~\cite{objectRecognition/dalal2005histograms}, SIFT~\cite{lowe2004sift}, and SURF~\cite{SURF/bay2008speeded} based feature extractors were popular.
However, these feature extractors were not robust to different environmental conditions (e.g., lighting). Later, deep learning based feature extractors became popular because they generated more robust feature vectors than SIFT and SURF. One widely used and available feature extraction algorithm is Fast-ReId~\cite{ObjectDetection/ren2015faster}. \textit{Note that each deep learning algorithm extracts a different type of feature vector (or map), and the feature vector dimension also varies depending on the algorithm.}

\noindent\textit{\underline{Current Limitations:}} One of the biggest limitations of current feature extraction algorithms is generating discriminatory \textit{feature vectors for similar-looking objects or objects belonging to the same class}. For example, two individuals in a frame with the same colored cloth can generate similar feature vectors irrespective of the feature extraction algorithms used. It is also possible that objects having overlapping bounding boxes have very similar feature vectors (as they have common information). These problems are still under investigation by the IVA community.

\subsection{Object Tracking} Object tracking algorithms aim to track moving objects in a video over time (corresponding to a fixed number of frames). A threshold is imposed on how long an object can be tracked. Object tracking algorithms have improved significantly by exploiting deep learning approaches and using techniques for estimating object trajectories based on motion models such as optical flow~\cite{ObjectTracking/sundaram2010dense}, Kalman filters~\cite{ObjectTracking/li2010multiple}, etc. These algorithms can track multiple objects simultaneously over time (or a sequence of frames). Some of the current popular Multi-Object Tracking (MOT) algorithms are DeepSort~\cite{ObjectTracking/Wojke2018deep}, BotSort~\cite{objectReidentification/BotSort}, and transformer-based trackers~\cite{ObjectTracking/zhu2023cross}. 

\noindent\textit{\underline{Current Limitations:}} Even though the above algorithms significantly improved MOT, they still cannot consistently track objects in videos with complex backgrounds, moving cameras, etc. With these algorithms, two tracks (or object id) may be generated for the same object (specifically when detected as different objects by object detection, when objects enter or leave the FOV, or when objects occlude, etc.). Two objects may also be considered part of the same track (or object id) (because of the object's appearance, such as wearing the same colored clothes, etc.).

\subsection{Object Pose Estimation} Pose estimation algorithms are designed to extract the orientation (or pose points) of a specific object category (e.g., human, vehicle, animal, etc.). Some of the well-known and available human pose estimation algorithms are OpenPose~\cite{PoseEstimation/OpenPose}, LCRNet~\cite{PoseEstimation/LCRNet++}, ViTPose~\cite{PoseEstimation/xu2022vitpose}, and HRNet~\cite{PoseEstimation/HRNeT}.

\noindent\textit{\underline{Current Limitations:}} These algorithms cannot extract correct pose points when two objects are occluded.

\begin{table}[!htb]
    \centering
    \footnotesize
      \caption{Summary of existing VCE algorithms. FOV is Field of View}
    \begin{tabular}{|p{.15\linewidth}|p{.10\linewidth}|p{.22\linewidth}|p{.12\linewidth}|p{.30\linewidth}|}
    \hline
    \textbf{VCE Task} & \textbf{Category} & \textbf{Extracted Contents} & \textbf{SOTA} & \textbf{Limitations} \\ \hline
    Object Detection  & RPN and SSD & Object location, class, confidence score & YOLOv10
    ~\cite{objectRecognintion/yolov10}~\footnote{YOLOv10 is one of the SOTA algorithm among many} & Background extraction, Correct object class labeling \\ \hline
    Feature Extraction & ~ & Feature vector  & Fast-ReId~\cite{he2023fastreid} & Discriminative feature vectors for objects of different classes. \\ 
    \hline
    Object Tracking & Kalman-Filter & Unique identifier to each object &  BotSort~\cite{objectReidentification/BotSort} & Consistent object tracking when objects enter FOV\\ \hline

   Pose estimation & Human and Animal Pose & 2D/3D pixel points representing body parts & VitPose++
   ~\cite{PoseEstimationGeneric/xu2022vitpose+} & Correct point estimation in occlusion \\ \hline

   Action classification & Pose based & Action class to each object (person) & PA3D\cite{PoseEstimation/zhang20243d} & Limited number of action classes to a fixed number of objects. Applicable to videos with very clear scene partitions, such as movies. \\ \cline{2-4}
      ~ & Scene Graph & Action class in 'n' sequence of frames & ~\cite{Scenegraph/kim2024groupwise} & ~ \\ \cline{2-5} 
   ~ & Video-based & Action class to the whole video &  ~\cite{ActivityRecognition/ryali2023hiera} & Not applicable for long videos with multiple actions happening at the same time \\ \hline
 
    \end{tabular}

    \label{tab:VCE_Summary}
\end{table}

\subsection{Action Classification} Action classification algorithms extract actions (e.g., walking, talking, behind, etc.) performed by an object in an image or a particular video. Typically, these algorithms assign an action class to a video of a small duration (ranging from a few seconds to a minute). Several different action classification algorithms are available, such as two-stream networks~\cite{ActivityRecognition/simonyan2014two}, 3D spatiotemporal activity classification~\cite{ActivityRecognition/feichtenhofer2017spatiotemporal},
transformer-based action recognition~\cite{ActivityRecognition/ryali2023hiera}, etc. The above algorithms significantly improved accuracy, performance, and the number of action classes recognized. However, they cannot classify \textbf{each object's actions} separately and were rather designed to label an image or video with an action class. 
The pose-based action detection algorithms partially solved the problem by tracking the human poses in a sequence of images or video frames and assigning an action class to each tracked pose. Some of the available algorithms are Lcr-net~\cite{PoseEstimation/rogez2017lcr}, PA3D\cite{PoseEstimation/zhang20243d}, MMNeT~\cite{ActivityRecognitionPose/MMNET}, ~\cite{ActivityRecognitionPose/2020mmaction2, ActivityRecognitionPose/reilly2023just, ActivityRecognitionPose/duan2022revisiting}, etc. \textit{Even though these algorithms can assign action classes to each extracted object}, they cannot infer the relationship between objects or interaction between objects (e.g., a person sitting on a chair) or actions of a group of objects (e.g., the largest clusters/groups of individuals). These algorithms are also limited in assigning action classes by tracking poses for a short video duration. 
Another class of VCE tasks that can extract the activity relationship between objects in the form of subject-action-object~\cite{Scenegraph/ActionGenome}, where the subject and object are extracted objects. These algorithms model the video as a scene graph and predict the action classes across 'n' fixed frames (given as a threshold) between the objects based on the spatiotemporal feature extracted. Some well-known SGG models are VidSGG~\cite{Scenegraph/VidSGG}, ~\cite{Scenegraph/xu2017scene},~\cite{Scenegraph/li2024panoptic}, and ActionGenome~\cite{Scenegraph/ActionGenome}. These algorithms typically assume the set of objects is fixed (or does not change drastically) across the video frames, applicable for short videos, and videos where scenes are partitioned, such as movies.

\section{Stream Data Processing Systems}

\subsection{Sensor Stream Processing (SP) Systems} The sensor SP systems were developed to process streams of sensor data to identify events, such as monitoring accidents from car sensor data, making trading decisions based on changes in stock prices, monitoring infrastructures, environment/water quality monitoring 24/7, and others. These are required to perform continuous computation on an unbounded stream of data coming in real-time and meet certain Quality of Service (QoS) requirements. Several systems have been developed over the years to address the different aspects of stream processing. 

Aurora~\cite{SP/abadi2003aurora}, a sensor stream processing system, was developed for applications that require frequent triggers and can have imprecise data. Continuous queries (CQ) were processed in Aurora through a data flow diagram (containing boxes and arrows). The SQuAL query language used in Aurora has seven primitive operators (e.g., filter, aggregation, map, etc.) and different window semantics. Aurora also supports run-time query optimization, a two-level scheduling algorithm, and real-time QoS requirements (e.g., load-shedding). The successor of Aurora was Borealis~\cite{SP/abadi2005designBorealis}, which supported continuous query processing in a distributed manner. They supported dynamic revision of query results, dynamic query modification, and highly scalable optimization. 
Another well-known sensor SP system in that era was STREAM~\cite{SP/arasu2003stream}. STREAM had support for Continuous Query Language (CQL), an extension of Structured Query Language (SQL), with constructs for specifying streams, relations, and additional constructs for specifying windows. It also supports a chain scheduling strategy and load shedding. TelegraphCQ~\cite{SP/chandrasekaran2003telegraphcq} was developed to handle processing many queries over a large volume of data. 

Another sensor stream processing system \textbf{MavStream} ~\cite{phdThesis/Jiang05,msThesis/Gilani03,msThesis/Kendai06} addressed several aspects of SP holistically (modeling, scheduling, and load shedding, including implementation of MavStream) and from a QoS perspective. This system supports various CQ operators (e.g., join, group by, split) and window types (e.g., time-based, tuple-based, landmark, sliding window, etc.) When a CQ is submitted to the MavStream system along with a window specification and QoS requirements, the query processing engine of MavStream generates a query plan. The system has a scheduler that determines which operators to execute from the generated query plans. This system has support for a class of scheduling strategies (e.g., round robin, weighted round robin, path capacity scheduling, etc.). The MavStream system can estimate QoS metrics for query results using a queuing model (specifically, estimating tuple latency and memory requirements). The system also has the capability to monitor actual QoS at runtime and adjust scheduling and load-shedding strategies to support bursty sensor inputs. This is achieved through a runtime optimizer integrated into the system. 

Recently, several commercial systems such as Apache Kafka~\cite{thein2014apache}, Apache Flink~\cite{flinkarticle}, Apache Storm~\cite{SP/iqbal2015big}, and Twitter Heron~\cite{2015TwitterHS} support stream data processing but not in the same way the systems described above. 


\section{Event Processing System}
 In the 1980s, the Event-Condition-Action (ECA) paradigm in the active databases was introduced as a flexible mechanism to automate diverse DBMS functions—such as view materialization, situation monitoring, and others, enabling systems to raise alerts for events without manual intervention. During the 1990s, research in active databases expanded rapidly, focusing on integrating the ECA paradigm into object-oriented DBMSs. Numerous prototypes and event specification languages have been developed to formalize event semantics and enable the detection of complex events.  Sentinel~\cite{sentinel::sharma} was among the earliest active database systems to support ECA, which supported the event specification language Snoop~\cite{Cha+96:snoop-sem:tr,Vidya94}. Snoop was introduced to define both primitive and composite events using operators such as SEQ, AND, OR, etc. To restrict the number of composite events detected for an application, several parameter contexts were also proposed in Snoop (e.g., restricted, chronicle, continuous, and cumulative). 

After 2000, Complex Event Processing (CEP) systems evolved significantly due to the rise of sensor-based systems. Systems such as Aurora~\cite{Balakrishnan:Aurora}, Cougar~\cite{cougaar}, etc., integrated SP systems into an event-detection paradigm. Some commercial systems, such as StreamSQL~\cite{StreamSQL:URL} have also integrated SP and CEP systems and provided them as a standard query
language for stream and complex event processing.

Another important system developed at that time was the \textbf{MavEStream}~\cite{phdThesis/Jiang05,msThesis/Garg05} system. This system integrates the QoS-aware DSMS system MavStream with the event processing system Sentinel (also termed Local Event Detector or LED), developed based on the ECA rule paradigm. The events here are occurrences of interest, conditions are simple or complex queries, and actions are operations to be performed when the event is triggered. The MavEStream accepts Continuous Event Queries (CEQ), which allow for defining CQs without and with event specification. It also allows for specifying masks on the attributes of event operators to reduce the number of events generated. Once a CEQ is submitted, the system converts the output generated from a CQ into an event object. These event objects are passed to the LED system, which uses Event Detection Graphs (EDG's) to identify the composite events. Finally, an alert is raised by the system. The major advantage of this architecture is seamless separation of the SP and CEP modules, which allows optimization of individual components and addition of new functionalities in a modular manner more easily. In other words, SP and CEP systems can be used with full potential without overhead, and common computations can be performed by a CQ, and the CQ output can be propagated into different types of events of interest by defining a mask. In this thesis, the proposed operators for video situation analysis will be integrated into the MavEStream system.


\subsection{Video Stream Processing (VSP) Systems}
VSP systems gained popularity in the late 2000s because of the abundant video data available through different camera devices. These systems focus on processing and storing video frames and retrieving them in real-time. They were developed using the functionalities of the earlier sensor SP systems.
One popular real-time cloud-based VSP system is Amazon Kinesis~\cite{VSP/AmazonKinesis}. It includes content extraction features (e.g., object recognition) and efficient and secure storage and retrieval of video streams. 

Oracle Multimedia~\cite{ORACLE-Multimedia} also has the functionality to store and retrieve videos and images. Using a relational model, the ORVideo object type in Oracle Multimedia can represent different video characteristics, such as frame rate, frame width, video duration, etc. Stream-processing functionality for streaming video frames has also been incorporated into Oracle Multimedia. 

Videopipe~\cite{VSP/salehe2019videopipe} is another VSP system that allows streaming video feeds with content extraction features on edge devices (e.g., SMART IP camera devices, mobile devices, etc.) Their pipeline for streaming videos allows different VCE tasks on different devices for the same video to preprocess the video frames in near real-time. Another popular VSP system is SECP~\cite{VSP/zhang2019streaming}, which combines the power of SP systems and deep learning to extract video contents from edge devices in near real-time. They re-purposed the functionalities of Apache Storm~\cite{SP/iqbal2015big} and utilized the parallel processing power of GPUs to achieve the above.


In summary, the sensor SP systems have their merits in designing the semantics of CQL and developing several strategies (scheduling, load shedding, etc.) to satisfy QoS requirements in real-time. The most recent systems can also process many queries in a distributed manner. On the other hand, the VSP systems have addressed some of the challenges associated with near real-time video streaming and content extraction. \textit{Even with all the functionalities of these systems, they cannot process and analyze the video contents and identify situations automatically from them.} For video data, a whole set of new challenges needs to be addressed in the representation and analysis of the extracted video contents as elaborated in Chapter~\ref{chap:problem-statement}.

\section{Video Content Analysis Systems}
Video content analysis systems have been developed over a long time. A few of them have also exploited SP functionalities partially for situation detection. Instead of exhaustively reviewing video content analysis systems, we have categorized them into four categories: manual analysis, custom solutions, low-level content analysis systems, and graph analysis systems. We have discussed how different frameworks or solutions were developed in these categories over different periods and their limitations. We have also summarized the video content analysis systems on each category, their functionalities, and limitations in Table~\ref{tab:qvc-related-work-comparison}.

\subsection{Manual Analysis}

The history of manual video content analysis dates back to around the 1960s, when video footage or images were primarily examined by human operators for surveillance and security purposes. At that time, video lengths were short, and a smaller number and variety of videos were captured. Hence, manual analysis was feasible. Even though content extraction tasks have advanced significantly from the early 90s, manual analysis is still in practice today to some extent. Currently, the extracted contents (e.g., detected object classes) are visually analyzed by humans for forensic analysis.

\subsection{Custom Analysis Algorithms}
With the advent of IVA, numerous different custom algorithms became available for video content analysis. They are still being developed for various purposes. In this section, we have discussed the custom algorithms developed in three domains -- Surveillance (SL), Civic Monitoring (CM), and Assisted Living (AL). To best understand the progression of the variety of custom algorithms, we have divided them into periods of development (starting from the 90s) and discussed what has been developed in each period. Since numerous algorithms exist even in the above three domains, we have discussed only the important or popular algorithms.


\begin{table}[!htb]
\footnotesize
\caption {\textmd{Summary of existing Video Content Analysis literature.}}
\label{tab:qvc-related-work-comparison}

\centering
\begin{tabular}{|p{.1\textwidth}| p{.1\textwidth}|p{.15\textwidth}|p{.1\textwidth}|p{.25\textwidth}|p{.20\textwidth}|}
\hline
\textbf{Category} & \textbf{Domains} & \textbf{Systems}  & \textbf{Data Model} & \textbf{Supported} & \textbf{Not Supported}\\ 
\hline
 
Custom Solutions & SL/CM & VIRAT~\cite{VIRAT-2013}, VSAM~\cite{CustomAlgoSurveillance/collins2000system}, ~\cite{CustomAlgoSurveillance/freer1996automatic} & Video & Loading objects in vehicle, a person entering/exiting a building, etc., intruder detection & New algorithm required for new situation type \\ \cline{3-6}
 & ~ & NoScope \cite{videoQuerying/kang2017noscope}, MIRIS \cite{videoQuerying/bastani2020miris}, SVQ++~\cite{videoQuerying/chao2020svq}, THIA~\cite{videoQuerying/cao2021thia}, ~\cite{videoQuerying/nguyen2021traffic}, FrameQL~\cite{videoQuerying/kang2018blazeit}, SBU-Kinetics~\cite{twoObjectClose/Samaras2012} & Video & Searching for objects using feature vector/color/label, car taking a turn, two person approaching each other, etc. & Retraining required for new situations \\ \cline{2-5}

~  & AL & ~\cite{FallDetection/tohidypour2022deep,FallDetection/amsaprabhaa2023multimodal}, ~\cite{GaitMonitoring/romeo2023video, GaitMonitoring/ajay2018pervasive,SocialIsolation/prenkaj2023self} & Video & Fall detection, Gait monitoring, social isolation detection & ~ \\ \hline
Low-level Content Analysis Systems & SL/CM & QBIC~\cite{VideoQuerying/QBIC}, ~\cite{VideoQuerying/hjelsvold1994modelling}, CVQL~\cite{VideoQuerying/kuo2000content}, BilVideo \cite{videoQuerying/donderler2005bilvideo},  VDBMS \cite{videoQuerying/aref2003video} & Relational & Searching for objects based on color/shape, two objects meeting, distance decreasing between two objects, similarity of objects & Fixed event and feature database \\ \cline{3-6}
~ & ~ & LVDBMS \cite{aved2013scene,aved2014informatics} & Relational & Spatial relationship between objects (e.g., overlap, before, north, etc.) & Only boolean operators are supported \\ \cline{3-6}
~ & ~ & VideoStorm~\cite{videoQuerying/zhang2017live} & Relational & Counting cars by using information extracted from license plates. Query processing in a distributed manner & Only two types of queries can be answered \\ \hline 
Graph Analysis & SL/CM & VidCep \cite{videoQuerying/yadav2019vidcep, VideoRepresentation/yadav2020knowledge, videoQuerying/yadav2021vidwin} &	Knowledge Graph & Heavy traffic, a person riding a horse/bike, handshaking, punching, etc. & Complex queries involving join, finding group(s) of different sizes, etc. \\ \cline{2-5}
~ & Sports & Visual Query Answering \cite{vqa/xiong2019visual} &	Entity Attribute Graph & Background representation, searching for objects using the label, counting objects & ~ \\ \cline{2-5}
~ & CM & Equi-Vocal~\cite{videoQuerying/EQUI-VOCAL} & Scene graph & Spatial relationship (e.g., close, far, etc.) between objects & ~ \\ \hline

\end{tabular}

\end{table}



\underline{\textit{Surveillance (SL) and Civic Monitoring (CM) Systems:}}
During the \textbf{late 90s-early 2000s}, several custom algorithms were developed for surveillance and civic monitoring. One crucial framework developed at that time for battlefield monitoring was VSAM~\cite{CustomAlgoSurveillance/collins2000system}. They have developed several algorithms (or routines) to analyze human activities such as walking, running, etc., using the geo-locations of the objects, and modeled them as a scene. These scenes were sent to a central operator unit for a broader overview of further analysis (by humans). The system was designed for only analyzing humans and vehicles (cars) and four types of interaction and actions among them. Password~\cite{CustomAlgoSurveillance/ivanov1999video} is a framework for identifying pick-up and drop-off events from parking lot videos. The Password project was developed to analyze images for identifying potential anomalous activities (e.g., vandalism in a metro station, potential prowler in a parking lot, etc.).~\cite{CustomAlgoSurveillance/freer1996automatic} also developed a framework for detecting intruders. The major disadvantage of the above algorithms was that they were limited to one or two object classes and activities.

During the \textbf{late 2000s-mid 2010s}, several different custom surveillance and civic monitoring algorithms were developed by including more action classes. They also improved the robustness of the earlier algorithms. One crucial framework developed at that time was VIRAT~\cite{VIRAT-2013}. They have developed an L1 tracker for analyzing a person's activities. The actions and interactions were divided into three categories: person to self, person to object, and person to vehicle. They were able to analyze 17 different situations. They have also developed a dataset~\cite{VIRATDataSet} and laid the foundation of several video content analysis systems developed later. 


Since the \textbf{late 2010s-until now}, custom video surveillance algorithms have advanced rapidly because of deep-learning algorithms. Some of the critical custom video analytic systems developed for the surveillance domain using deep learning are NoScope \cite{videoQuerying/kang2017noscope}, MIRIS \cite{videoQuerying/bastani2020miris}, THIA \cite{videoQuerying/cao2021thia}, FrameQL \cite{videoQuerying/kang2018blazeit}, etc. In addition to identifying a wide range of objects, NoScope allows for searching a specific object using its image within minimal computation time (e.g., similar to SQL select). MIRIS \cite{videoQuerying/bastani2020miris} is designed to analyze situations related to traffic surveillance. It accelerates object recognition performance to enhance the time for query processing. THIA \cite{videoQuerying/cao2021thia} also accelerates the inference time while searching for an object in a video.\textit{The above systems are focused more on the optimization and performance of deep learning algorithms rather than situation analysis.}

During this era, algorithms have also been developed to analyze situations involving actions happening in the videos~\footnote{This is different from activity classification algorithms of VCE as they analyze the interaction between multiple objects and support event extraction to some extent}. SVQ++ \cite{videoQuerying/chao2020svq} answers fixed situations involving interaction (e.g., throw, catch) between two objects. TQN~\cite{videoQuerying/zhang2021temporal} can identify actions such as leap, side split, etc. Another customized framework for traffic monitoring was proposed by \cite{videoQuerying/nguyen2021traffic} for extracting the semantic relationship between a textual query (in the traffic surveillance domain) and a video. 
FrameQL~\cite{videoQuerying/kang2018blazeit} is a custom framework that can process aggregation queries such as count, selection, and projection on videos. Though users can express situations as SQL queries, there is no underlying query processing architecture to compute them. Instead, they employ specialized neural networks and write an algorithm (or software) to identify each situation individually.

\underline{\textit{Assisted Living (AL) Systems }}: The custom solutions in the AL domain are developed based on wearable sensors, Internet of Things (IoT) devices, etc., to classify a handful of actions performed in a controlled environment, such as Fall detection systems~\cite{FallDetection/tohidypour2022deep,FallDetection/amsaprabhaa2023multimodal}, daily life activity monitoring systems~\cite{ActionRecognition/Skeletal-Francisco}, gait analysis systems~\cite{GaitMonitoring/romeo2023video, GaitMonitoring/ajay2018pervasive}, social isolation monitoring~\cite{SocialIsolation/prenkaj2023self}, etc. These algorithms are also focused on classifying sensor data and improving the accuracy, performance, etc., of those classification algorithms rather than query processing or situation analysis.

The above systems have merits in analyzing specific situations and optimizing the performance of machine learning models. However, these systems cannot analyze a new situation type or cannot be applied to videos from a different domain. Either a new algorithm needs to be written, or these algorithms need to be retrained for each new situation class. They also cannot compose the situations detected by them into composite situations.

\subsection{Low-level Content Analysis Systems}
The low-level content analysis frameworks model the extracted video contents using a relational model and support querying on only specific content types using low-level spatial operators. Many of these frameworks also support SP functionalities to some extent.

During the \textbf{90s-2000s}, video content querying systems were developed for querying from a fixed database. One of the earliest systems in this era is the QBIC~\cite{VideoQuerying/QBIC} system. The QBIC system stores image contents (objects) in a fixed database along with several attributes such as color, texture, shape, etc. The video data is stored using representative frames for each shot in a video. The object features or shots (or frames) are matched and retrieved based on the submitted user query sketch. A data model was proposed by~\cite{VideoQuerying/hjelsvold1994modelling} for modeling different video components. They defined an Enhanced Entity Relationship (EER) diagram to model video stream, frame sequences, shots, and sequences of shots (as scenes). They supported five categories of content-based queries, such as retrieving videos related to a given scene.~\cite{VideoQuerying/kuo2000content} developed a content-based video querying language (CVQL). This system had the functionality to search for videos based on the temporal or spatial relationship among the video contents. It supported queries involving a specific object type, two spatial relationships of objects, and compound relationships between objects (e.g., meet, approach, decrease, etc.) Videos with those relationships are retrieved from a fixed database. Though the above systems developed query-processing architectures with content-matching capability to some extent, only simple relationships among the contents were considered. They are limited to retrieving only videos or images with specific contents from databases, not situations from a video.

 Unlike the solutions discussed above, several video content querying frameworks were developed to analyze the relationship of the contents within a video from the \textbf{early 2000s-2010}. One of the earliest solutions in this era was VDBMS~\cite{videoQuerying/aref2003video}. The system supported queries involving similarity matching of the bounding box location of two different videos. They supported joining videos based on frame timestamps and videos of different categories. Even though joining and other query optimizations are supported by this system, they rely on a fixed content database and do not support queries on all different types of extracted contents completely.

BilVideo~\cite{videoQuerying/donderler2005bilvideo} was another video querying system where users could pose queries using SQL-like query expressions. The system retrieves the query result from a stored video database (containing information about object properties such as feature vectors) using rule-based inferencing. The system includes logical, temporal, and spatial operators that recognize actions such as meeting, appearing in a place, and the trajectory of an object. However, maintaining a fixed feature database is infeasible since object feature vectors change from frame to frame and video to video. There was no operator or algorithm to compare or join using these feature vectors.~\cite{videoQuerying/csaykol2005database} designed a database model for analyzing surveillance videos. The system could process queries such as object entrance in a scene, crossing each other, etc. This system also maintained a predefined event database (with a fixed schema) for processing the queries. 


A modular, layered approach was taken to process live motion imagery data by LVDBMS~\cite{aved2014informatics,aved2013scene}. The live video query language (LVQL) incorporated in LVDBMS expresses situations with low-level abstraction (boolean operators). It can answer queries about whether a particular event happened (true or false). However, this system did not incorporate queries involving aggregation and joining. A cloud-based solution for video querying was designed in VideoStorm \cite{videoQuerying/zhang2017live}. The resource allocation and scheduling algorithms incorporated in VideoStorm architecture allow efficient execution of multiple queries in the cloud platform. Despite this, the system can only answer custom queries like counting cars and identifying license plates. 

The frameworks discussed above have tried to extend the relational model and facilitate query processing on extracted video contents. However, they use custom or low-level operators, do not consider all different extracted content types for modeling and analysis, and use fixed event or feature databases.

\subsection{Video Analysis Using Graphs}

Compared to the vast literature discussed earlier, the use of graphs for video analysis is fairly recent and limited. Recently, several solutions have been developed for video content analysis using graph models. VidCep~\cite{videoQuerying/yadav2019vidcep, VideoRepresentation/yadav2020knowledge}, and Peixi et al.~\cite{vqa/xiong2019visual} modeled extracted video contents using knowledge graphs. VidCep has support for detecting situations, such as handshaking, sitting on a chair, searching objects using their color or label, high traffic volume, etc. They represent each object (or object id) as a node and the relationships (e.g., distance) between them as an edge. This system generates a graph for each video frame and transforms it into an aggregated graph (termed VEKG-TAG). In the aggregated graph, the edges are drawn if the object pair has appeared in at least a frame together, and edge weights are a sequence of values where this pair of objects appears. Several spatial or temporal operators are designed by this system to identify the above situations. This system also supports four Complex Event Processing (CEP) operators: SEQUENCE, AND, OR, and EQ (for concurrent occurrence of events). Even though this system has many functionalities, there are several limitations of the system. First, generating graphs for each frame and then aggregating them is not an efficient approach, especially when large videos (length in hours) need to be processed. Second, the operators or algorithms designed for situation analysis perform some computation on every video frame and are applicable when the video length is small. Again, processing large videos will be a problem using this system. Finally, they do not have accurate, efficient, and robust operators or algorithms to identify both the primitive and complex situations addressed in this thesis. On the other hand, the system developed by Peixi et al.~\cite{vqa/xiong2019visual} answers only four simple situations (e.g., searching for a player wearing specific colored clothes, counting the number of players, etc.) related to the sports domain. 




Another system developed recently is Equi-Vocal~\cite{videoQuerying/EQUI-VOCAL}. They model the whole video as a graph and support only queries related to the spatial bounding box relationship between two objects in a frame. They have support for query optimization and improving the accuracy of the situation analysis through active learning, which requires a human to intervene to constantly guide their system for improving system accuracy. This incorporates humans in the loop to a large extent, and in this thesis, the goal is to minimize the involvement of humans.

Even though all of the recent systems have their merit, the major drawback of these systems is that they do not support a wide variety of situations, the algorithms are not efficient and robust for processing large videos, and some of them require human intervention.

Our approach to modeling video contents using graphs is quite different from the work in the literature. Our goal is to represent video contents in a general-purpose manner using multiple graph models (in addition to the R++ model) and use or develop generic graph analysis algorithms for dealing with complex situations composed from primitive situations shown in Table~\ref{tab:PrimitiveSituations}. We also discuss storage and performance tradeoffs between different graph models.

\subsection{Large Language Models for Video Analysis} Apart from the above systems, recently, the Large Language Models (LLMs) have become popular and support a wide range of Video Analysis (VA) tasks (termed also Video-LLMs) such as video captioning, video question answering, etc.~\cite{ICABCD2025/SantraC25}.


The Vision Language Models (VLMs) have also become very popular in recent years. These models are designed for specific VA tasks such as video summarization, situation analysis, and others. One such system was developed by Yuan~\cite{LLMSituationDetection/yuan2025empowering} \textit{et al.}. This system analyzes situations such as searching for objects using color or situations involving object spatial relationships in a video. The system is also capable of providing an overall description of the input image/video. Other important VLMs published in recent years are Video-LLama~\cite{LLM/zhang2023video} and Video-Glam~\cite{LLM/munasinghe2025videoglamm}.

These models have certain limitations. First, processing a large number and length of videos using these models is computationally expensive. Second, for certain situations such as ``\textit{finding group(s)} of objects" where objects need to be clustered, these models are not appropriate as they are mostly supervised or semi-supervised algorithms. Third, to address a new situation or set of situations, these models need to be retrained with a substantial amount of labeled data. Finally, these models are still in their infancy to be fully fleshed out for video situation analysis.

\chapter{Video Content Extraction (VCE) Workflow}
\label{chap:vce_workfolow}
This chapter outlines our approach to extract contents from videos (using appropriate algorithms/packages available in the public domain). We have developed a VCE workflow that extracts contents that can be modeled for our proposed video situation analysis (VSA). This workflow takes into account constraints of various extraction techniques, pipelines multiple extraction algorithms with proper input and output, and generates an output from which both representation models are generated using our post-processing algorithm. It is essential to note that VCE workflow and post-processing are applied \textbf{only once} to a video, and analysis can be performed as many times as needed using the generated models.

\section{General VCE Workflow Definition}
\noindent A video $\mathcal{V}$ can be viewed as a sequence  $<1,\ldots,\mathcal{F}>$ of frame ids (represented as $F^\mathcal{V}$) captured at a specific frames per second ($fps$) rate using a camera. Each video has the time (of generation), $fps$, frame resolution (width and height), etc., as metadata. Each camera has a camera or device id and may have a camera location (e.g., mounting location or area id, which is a unique identifier representing the geographical location where the camera is installed, etc.) as part of video metadata. The camera metadata may or may not be part of a video. Each video frame is considered a $w \times h$ dimensional array of pixel values, where $w$ is the width and $h$ is the height. All the frames of a video captured by a specific camera device (calibrated with a resolution) have the same width and height. Furthermore, for the purposes of this work, we assume that the camera is fixed and stationary, capturing the same field of view over time. This work supports videos from multiple stationary cameras (e.g., entry and exit cameras).

A VCE algorithm processes each video frame by analyzing pixel values and extracts contents from them~\cite{survey/ibrahim2021survey}. A VCE workflow comprises one or more VCE tasks or algorithms, which can extract objects, assign them an object id, and extract different properties of the objects (e.g., object location as a bounding box, class label, class label confidence score, etc.) in a frame.  
Different VCE workflows can be created depending on the object properties that need to be extracted. However, object detection and tracking are common to all of them. \textit{If no object is detected in a frame by the object detection algorithm, subsequent algorithms relying on object detection will fail to extract any information.} 


Frames in a video processed by the VCE workflow can be of two types: empty and non-empty. A frame is empty if the VCE workflow has \textit{not} extracted any object (irrespective of the existence of objects) from the frame. \textit{A frame is non-empty if at least one object has been extracted from the frame by a VCE algorithm.} The first non-empty frame does not necessarily need to be the first frame of the video. Empty frames may result due to errors induced by a VCE algorithm, where it fails to detect any object from a frame $f$.

A VCE workflow extracts $O_{f}$ set of objects (or object ids) from a non-empty frame $f \in F^\mathcal{V}$. The total number of objects in a frame is $|O_f|$. 
An object in a video is assigned an object id for a given number of frames (termed as track buffer or tracking threshold by some algorithms~\cite{objectReidentification/BotSort}). If an object reappears and the tracking threshold has expired, then the object is assigned a new object id by the VCE algorithm. In other words, \textit{depending upon the tracking threshold given, the same object can be assigned a different object id by a VCE algorithm}. 

An object is given the same object id $o_i$ in a sequence of frames by VCE workflow. In videos, objects can enter the field of view (or reappear) multiple times after a certain period. If the given tracking threshold is large enough (e.g., equal to the length of the video), they are typically assigned the same object id. Note that it is possible that an object can be assigned a different object id even if the tracking threshold is set as the entire video length (or a very large value) because of environmental conditions (e.g., an object being occluded), because of the errors from the feature extraction algorithm used by the tracking algorithm, etc. Hence, the frames in which an object id $o_i$ is detected need not be consecutive. For each non-empty frame $f$, a set of attributes is extracted by the VCE workflow for an object id $o_i$ (represented as $A_{o_i}^{f}$). These attributes can include the Feature Vector (FV), Class label (CL), Class label confidence (CLC) score, Bounding box ($BB$), and others. The attribute values are likely to vary from frame to frame. Hence, we will denote each attribute for an object id individually. For example, the class label, class label confidence score, feature vector, and bounding box attributes of object $o_i$ in frame $f$ are denoted as $CL_{o_i}^f$, $CLC_{o_i}^f$,$FV_{o_i}^f$, and $BB_{o_i}^f$ respectively. Note that the extracted attributes are not limited to these attributes, and there can be other extracted attributes. So, all the attributes of an object id $o_i$ in frame $f$ can be denoted as a set $\mathcal{A}_{o_i}^f$ consisting of $\{CL_{o_i}^f, CLC_{o_i}^f, FV_{o_i}^f$, $BB_{o_i}^f, ...\}$. All the attributes of an object id $o_i$ in the entire video can be denoted by ${\mathcal{A}_{o_i}^\mathcal{V}}$ and can be computed as 
$\bigcup_{f \in F^\mathcal{V}} ~\mathcal{A}_{o_i}^f$. 

In a video $\mathcal{V}$, there can be $UO^{\mathcal{V}}$ unique object id's, and a total of $OI^{\mathcal{V}}$ object id instances, where $UO^{\mathcal{V}} = |\bigcup_{f \in F^\mathcal{V}} O_{f}|$ and $OI^{\mathcal{V}} = \sum_{f \in F^\mathcal{V}}|O_f|$. On average, each frame can contain $OI^{\mathcal{V}}/\mathcal{F}$ number of objects. Typically, the number of object instances is much greater than the number of unique objects in a video.


\section{VCE Workflow Components}
\begin{figure}
    \centering
    \includegraphics[width=1\linewidth, keepaspectratio=true]{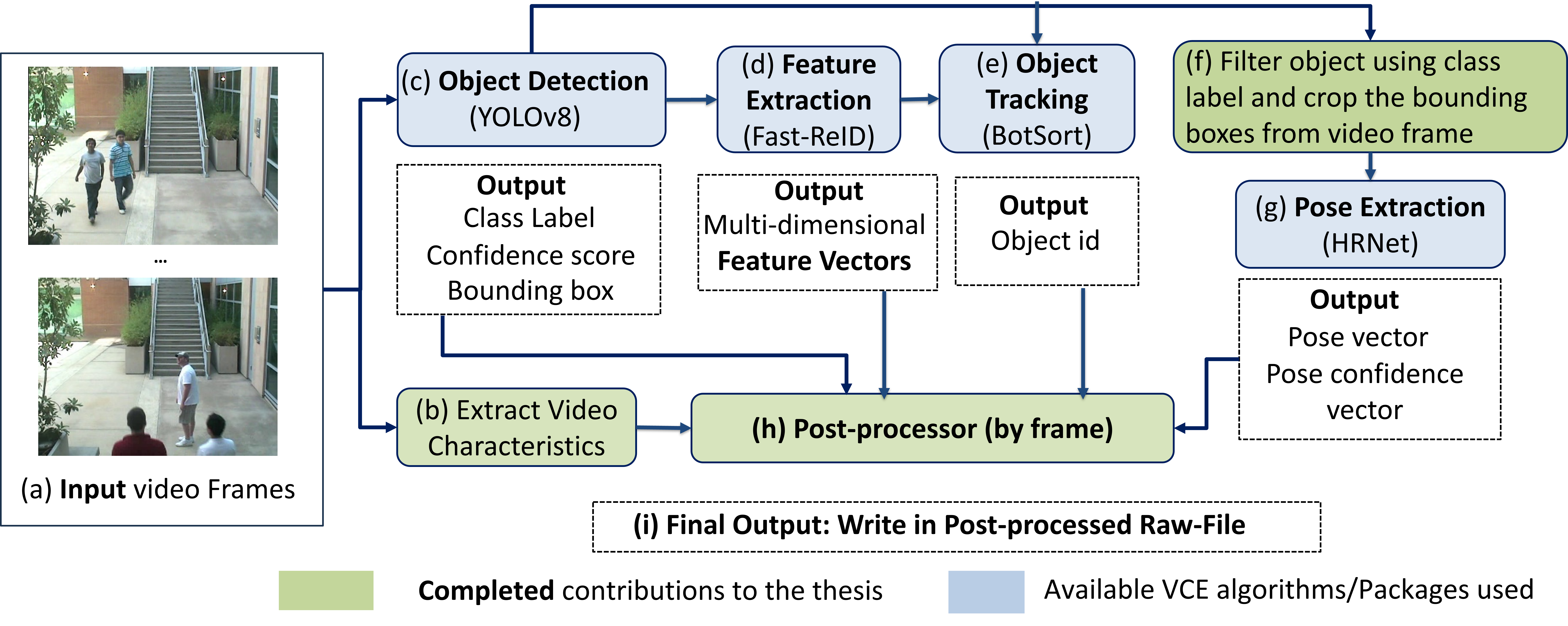}
    \caption{VCE workflow used for VideoScoop }
    \label{fig:vce-workflow}
\end{figure}

The proposed VCE workflow (shown in Figure~\ref{fig:vce-workflow}) takes as input a video and a set of parameters (e.g., tracking threshold, etc.) to be used by different VCE algorithms at various stages. In this work, the videos to be used are categorized into four classes: small (videos with length $< 5$ minutes), medium (videos with length $5-15$ minutes), large (videos with length $> 15$ minutes), and mixed. The mixed videos are generated by mixing the small videos. There are two types of mixed videos: mixed single and mixed random. The mixed single videos contain one known situation (multiple instances) from the small videos, concatenated with other videos from the medium and large categories. The purpose of generating mixed videos is to experimentally validate the robustness of the algorithms. The small, medium, large, mixed single, and mixed random video id's are prefixed with "S", "M", "L", "MS", and "MR".

The VCE workflow first extracts the \textbf{video characteristics}: video type from the video id, length (in seconds), total number of frames, fps, video generation date-time, frame width $w$, and frame height $h$ from video metadata. It writes this information to an output file as a header. 


The VCE module then processes frames from the video one at a time. A video frame is first passed to the object detection module (Figure~\ref{fig:vce-workflow}(c)). The object detection module (using the YOLOv8~\cite{objectRecognintion/yolov8} algorithm) extracts object location (as bounding box coordinates) and some attributes (class labels and their corresponding confidence scores). Note that any particular object detection algorithm can extract a fixed set of object classes, depending on what dataset they are trained on. For example, the YOLOv8 object detector used in this VCE workflow is trained on the COCO dataset~\cite{DataSet/COCO} and can extract 80 object classes. There exist other object extraction algorithms that can recognize more object classes. However, YOLOv8, trained on COCO, is widely used and sufficient for our purpose, as it contains all the common object types required to identify the situations of interest in this thesis. 

The extracted bounding box coordinates by the object detector and the input video frame are passed to a feature extractor module (Figure.~\ref{fig:vce-workflow}(d)). This module (using the Fast-ReID~\cite{he2023fastreid} algorithm) extracts the feature vector (or map) of the pixels inside each bounding box coordinate in the video frame. These feature vectors, bounding box coordinates, class labels, and confidence scores are passed to the object-tracking module (Figure~\ref{fig:vce-workflow}(e)). The object-tracking module (using the BotSort~\cite{objectReidentification/BotSort} algorithm) assigns a unique $object~id$ to each bounding box coordinate up to some number of frames using the extracted feature vectors and additional estimation algorithms~\cite{ObjectTracking/Wojke2018deep}.

These objects are filtered based on a given class label (in our case, it is person, as the situations of interest for this thesis involve only humans) before passing to the human pose estimator module, Figure~\ref{fig:vce-workflow}(g). The available pose estimation algorithms can extract pose points from an image or a given bounding box location of an image at a time. Hence, the extracted bounding box coordinates are used to crop the filtered objects from the input video frame for pose extraction. Finally, the cropped bounding box images of the filtered objects are passed to the pose extraction module, which outputs d pixel coordinates representing human body parts (e.g., head, eye, etc.) and confidence scores of each pose coordinate point. The current VCE workflow uses HRNet~\cite{PoseEstimation/HRNeT} as a pose extractor, which extracts 17 pose points. They are: nose, left eye, right eye, left ear, right ear, left shoulder, right shoulder, left elbow, right elbow, left wrist, right wrist, left hip, right hip, left knee, right knee, left ankle, and right ankle. Note that it is possible to replace the algorithms of each VCE task with the latest state-of-the-art (SOTA) algorithms as they become available. Finally, all the extracted contents at each stage for a frame are passed to a post-processor (Figure~\ref{fig:vce-workflow}(h)) The post-processor assigns a frame id, timestamp, and computes some additional meta information from the pre-processed output and writes it to the post-processed raw data file ($RDF^{\mathcal{V}}$).

\noindent \underline{\textit{Post-processed raw data file ($RDF^{\mathcal{V}}$) contents:}} The \textbf{post-processed} output of the VCE workflow contains the following extracted information. Here, vectors are represented by enclosing them with $[]$.

\begin{enumerate}
\item \textbf{Frame id ($fid$)}: A unique identifier of a frame assigned by the post-processor. 

\item \textbf{Object id ($oid$)}: A unique identifier assigned to each frame across 't' consecutive frames by the object tracking algorithm. 


\item \textbf{Object class label ($CL$)}:A categorical label assigned to each $oid$.

\item \textbf{Object class label confidence score ($CLC$):} A class confidence score (ranging between 0 and 1).

\item \textbf{Object bounding box ($[BB]$)}: A bounding box is a rectangular area representing the location of an object in a frame. The $[BB]$ coordinates are represented as a vector of four float values: $[x_{min}, y_{min}, x_{max}, y_{max}]$, where $x_{min}, y_{min}$ are the bottom left corner and $x_{max}, y_{max}$ are the top right corner of the bounding box within the frame. 

\item \textbf{Object feature vector ($[FV]$)}: A multi-dimensional vector of float values, where each element represents specific features (e.g., shapes, edges, etc.) of an object. Depending on the feature extraction algorithm $[FV]$, the dimension varies. For example, Fast-ReID extracts $1024 \times 1024 \times 3$ dimensional feature vectors, whereas SIFT~\cite{lowe2004sift} feature vectors have a size of $128$. 
\item \textbf{Pose Vector ($[PV]$)}: This is $d \times 2$ dimensional vector, where each vector element is x, y pixel coordinate (float values). Here, $d$ denotes the number of body pose points (e.g., ankle joint, torso, etc., of the human body) extracted by the pose extraction algorithm. Here also, depending upon the algorithm used $d$ varies. HRNet extracts $17 \times 2$ dimensional pose vectors. 

\item \textbf{Pose Confidence Vector ($[PCV]$)}: A pose confidence vector is a vector of length $d$, which contains confidence score values for each extracted pose point of an object.

\item \textbf{Timestamp ($ts$)}: A timestamp value in the format $<fid~ts:oid~ts>$ is assigned for each $fid, oid$ pair. The $fid~ts$ is an integer value starting from 1. The $fid~ts$ is increased for each frame (including the empty frames). The $oid~ts$ also starts from $1$ and is increased by 1 for each $oid$ in a frame. This allows us to assign unique timestamps to each object in a frame (which becomes a row or tuple in the relational model), which is helpful for applying stream processing (SP) functionalities (e.g., frame-based window) to the post-processed video data.


\end{enumerate}

\begin{figure}
    \centering
\includegraphics[width=1\linewidth]{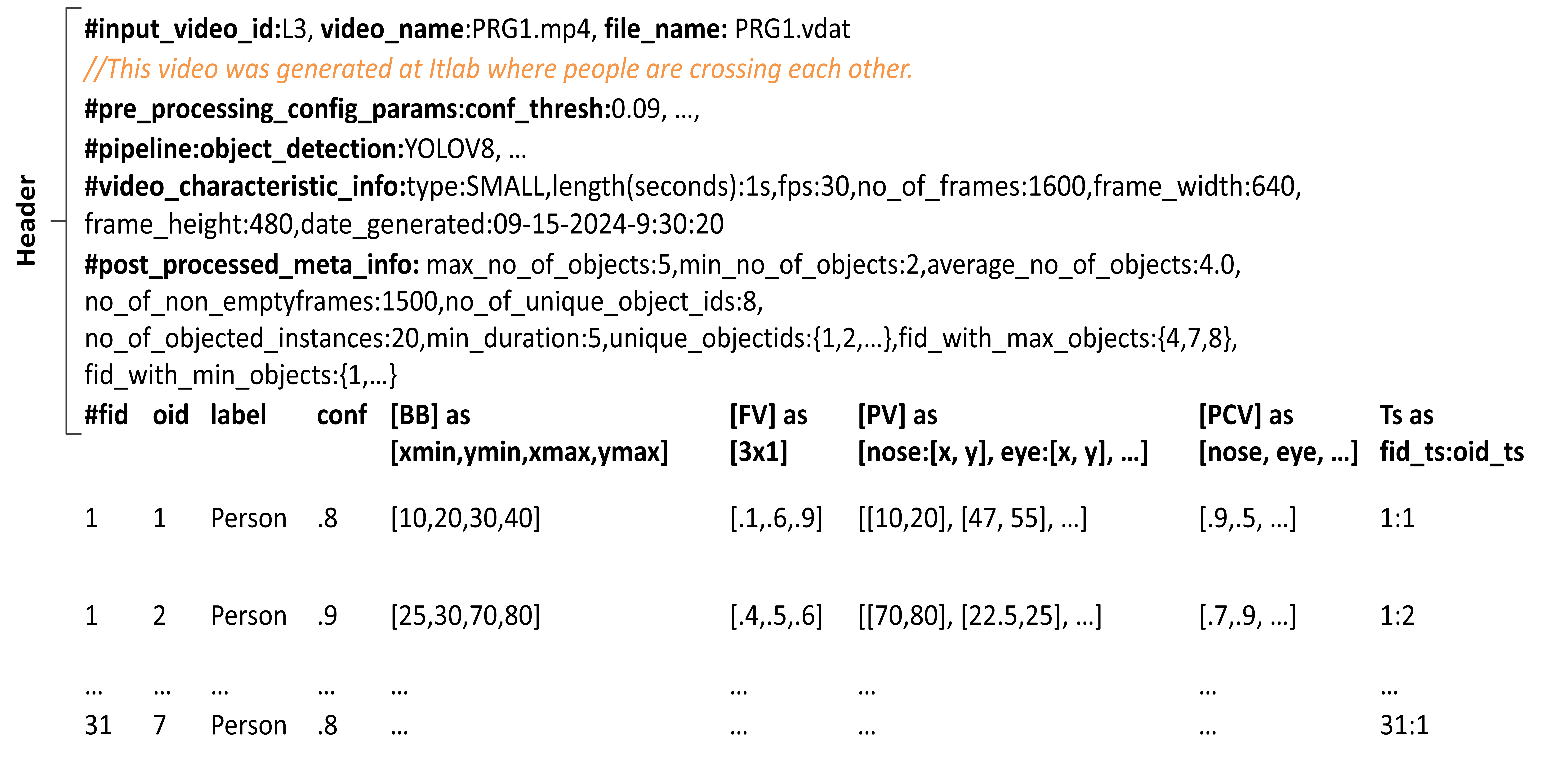}
    \caption{Sample post-processed output generated by a VCE workflow. The header contains various types of metadata gathered during VCE and post-processing.}
    \label{fig:VCE_raw_output}
\end{figure}

\noindent\underline{\textit{Meta-information computed by post-processor:}} The post-processor also computes some meta-information at the end of processing all the frames of the video. They are: the maximum, minimum, and average number of objects in a frame, the total number of non-empty frames, the set of unique object id's, the total number of unique object ids, the total number of object id instances, the minimum number of frames an object appeared in a video (min duration), and the sequence of frame id's having the maximum and minimum number of objects in the whole video.

These are aggregate information for the whole video and are useful as meta-information. Different meta-information is likely to be useful for different situation analysis.  Hence, these are extracted and stored for later use. The meta-information is also used for generating the representation models from the post-processed raw file. For example, we have used the number of non-empty frames and the number of unique object id's to generate one of the graph models described later in chapter~\ref{chap:Representation}. The $fps$ and the maximum number of object information are also used for analyzing the situation ``finding group(s) of different sizes'' The number of frames, the number of unique object id's, minimum duration of objects were used for analyzing the situation ``Two people coming closer to each other or moving away from each other''. The use of this meta-information for the above analysis is discussed in chapter~\ref{chap:graph-ana-videocontents}. In general, the meta-information, computed once, is used multiple times in situation analysis and reduces computation times of situations.



\noindent \underline{\textit{Post-processed file $RDF^{\mathcal{V}}$:}} An example of the sample post-processed file $RDF^{\mathcal{V}}$ generated by the VCE workflow is shown in Figure~\ref{fig:VCE_raw_output}.
This file contains multiple rows for the header, each depicting different pieces of information. The header line starts with 
the delimiter ``\#'', and comments (given by the user) start with the delimiter ``//''. As part of the header, the video characteristics, information about the algorithm used in the VCE workflow (starting with the keyword 'pipeline'), pre-processing configuration parameters (specified in a configuration file), and the computed meta-information during post-processing are written to the post-processed output file. After all the headers and comments are written, the data line starts. The data lines are tab-separated, and there is a column for each of the extracted attributes described above. There exists a row for each $fid, oid$ pair, along with the corresponding extracted attributes. In addition, the column header contains information about the data format (e.g., bounding box coordinates, feature vector dimension, pose vector categories, timestamp format, etc.) In Figure~\ref{fig:VCE_raw_output}, the VCE workflow extracted 2 objects (with $oid$ 1, 2) from $fid$ 1. There exist two rows for the $fid$ 1 as part of the data. For each $oid$, its label, bounding box, feature vector, etc., information is written. For timestamp, the format is $fid:oid$. Hence, two timestamps $1:1$ and $1:2$ are assigned to each row for $oid$ 1 and $oid$ 2 corresponding to fid 1. Similarly, for frame 31.  This timestamp is used for the frame-based window introduced for video processing.

Both data models — R++ and multiple graph models (SGF or Single Graph per Frame, SGV or Single Graph per Video, and MGV or Multiple Graphs per Video) used in this thesis are generated from the post-processed output file ($RDF^{\mathcal{V}}$).

\chapter{Modeling Alternatives for Extracted Video Contents}
\label{chap:Representation}

In this chapter, we discuss two data models that we think are appropriate and useful for representing extracted video contents. Both models are widely used for representing different kinds of data.  Relational or tabular representation is primarily used for modeling transactional data. Relational models used in DBMSs provide SQL (Structured Query Language) support for querying and analysis. SQL has undergone continuous revisions and optimization for over four decades, making it one of the most expressive and efficient languages for computations on the tabular data model. The focus of SQL optimization has been to reduce the amount of input/output (I/O) from secondary storage for the SQL operators. SQL has also been extended to CQL (Continuous Query Language) for processing stream data produced by sensors continuously, in a tabular or relational form. CQL provides additional functionality suited for sensor stream data. Unlike relational DBMSs, CQL also supports Quality of Service (QoS) requirements, which are crucial for processing stream data. 

On the other hand, graph models are particularly suited for modeling and analyzing applications with relationships between entities. Graph model analysis is also widely used, and many algorithms for aggregate and non-aggregate analyses exist.  Apart from algorithms for shortest paths, breadth-first and depth-first search, and graph characteristics such as degree, density, the analysis of graph representations has become important for social networks and other data that are heavy on relationships. Hence, analysis algorithms for clustering, community detection, and substructure detection have been developed for large graphs, and scalability has also been a focus.

For this problem, we have chosen the two models above because they offer different analytical approaches that may be useful in various situations, as exemplified in Chapter~\ref{chap:problem-statement}. We begin with a brief summary of graph types and their characteristics, followed by how they are used to represent extracted video contents in a manner that suits our purpose.

\section{Graph Types and Their Characteristics}
\label{sec:graph-types}

A \textbf{simple graph} (also referred to as single graph, network, or monoplex in the literature) is defined as (V, E) where V is a set of vertices or nodes and E is a set of edges connecting two \textit{distinct} vertices. E is a subset of V $\times$ V. The edges are typically unweighted, either directed or undirected and loops or multiple edges between nodes are not allowed. Typically, vertices have unique numbers, but the labels (can be more than one) of nodes and edges do not need to be unique. 
This simple graph model is adequate for many purposes and is widely used; however, it would be useful to be able to associate weights with nodes and edges to capture the semantics of real-world data sets, along with labels. Multiple edges and associated labels would allow one to represent multiple entity and relationship types, which leads to an \textbf{attributed graph}.

An \textbf{attributed graph} (also called a multigraph) is defined as (V, E, $\phi$) where V is a set of vertices or nodes, E is a set of edges connecting two distinct vertices, and $\phi$ is a function mapping of E to $\{ \{x, y\} \mid x, y \in E~ and~x~\neq~ y\}$. If the distinctness of nodes is removed, loops will also be allowed. The primary advantage of a multigraph or attributed graph from a modeling perspective is that it can capture multiple entity types and various relationships between these entity types. Multiple labels can be associated with nodes and edges.
With the attributed graph model, it is possible to include relevant information from the data description as labels, and hence it is more expressive as a model than a simple graph. Although an undirected edge is sufficient for most models, direction and weights can be associated with edges and weights can also be associated with nodes in addition to labels. 

This model is useful for capturing multiple relationships between the same type of objects in real-world data sets. For example, if one wants to represent a `friend' in Facebook as a relationship, and `connected' in LinkedIn as a relationship between the same objects, you need multiple edges with different labels representing their semantics. An attributed graph can model more complex information. Ability to represent multiple relationships with more clarity and analyze them individually or together leads to multilayer networks (or MLNs).

A \textbf{MultiLayer Network (or MLN)} is a \textit{network of simple graphs} (or forest.) In this model, every layer represents a \textit{distinct} relationship among entity instances of the same type with respect to a single (or combinations of) features. The sets of entities across layers, which may or may not be of the same type, can be related to each other too. Formally, a \textit{multilayer network}, $MLN (G, X)$, is defined by two sets of graphs: i) The set $G = \{G_1, G_2, \ldots, G_N\}$ contains simple graphs of N individual layers, where $G_i ~(V_i, E_i)$ is defined by a set of vertices, $V_i$ and a set of edges $E_i$. An edge $e(v,u) \in E_i$, which is a subset of $V_i$ $\times$ $V_i$, connects vertices $v$ and $u$, where $v,u\in V_i$ and ii) A set $X =\{X_{1,2}, X_{1,3}, \ldots, X_{N-1,N}\}$ consists of bipartite graphs (some of ehich may be empty). Each bipartite graph $X_{i,j} (V_i, V_j, L_{i,j})$ is defined by two sets of vertices $V_i$ and $V_j$, and a set of edges (also called links or interlayer edges) $L_{i,j}$, such that for every link $l(a,b) \in L_{i,j}$,  $a\in V_i$ and $b \in V_j$, where $V_i$ ($V_j$) is the vertex set of graph $G_i$ ($G_j$.) 

\textit{A MLN can be used to separate entities and corresponding relationships from an attributed multigraph into separate layers where each layer is a simple graph.} This provides more clarity in understanding and processing as well. Currently, MLN representations are being widely adopted for modeling complex data sets with multiple types of entities and multiple relationships between the same types of entities. They can also capture relationships between different types of entities.

\begin{figure}
    \centering
    \includegraphics[width=0.8\linewidth,keepaspectratio=true]{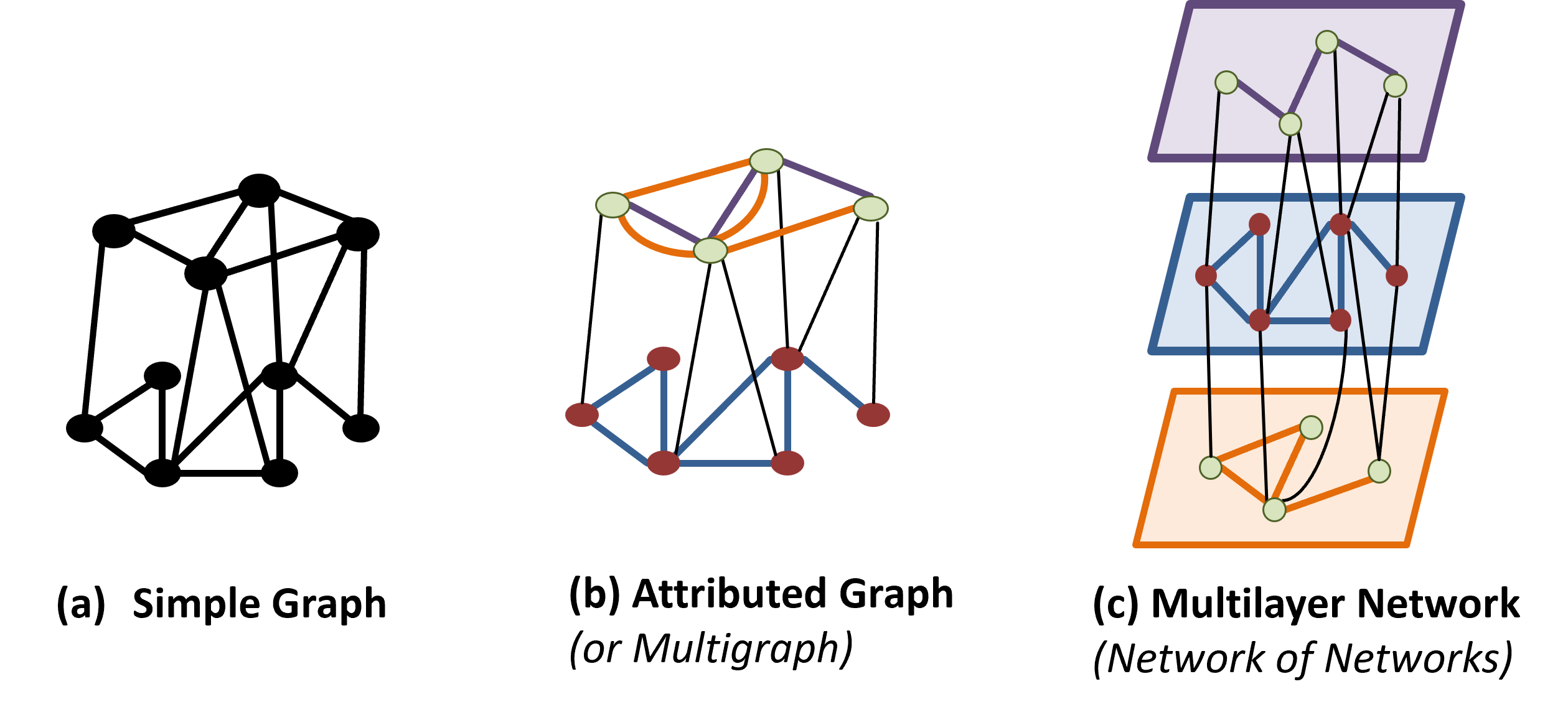}
    \caption{Types of Graph Models}    
    \label{fig:graph-types}
\end{figure}

Based on the types of relationships and entities, multilayer networks can be classified into various categories. Layers of a \textbf{homogeneous MLN (or HoMLN)} are used to model the diverse relationships that exist among the \textbf{same type of entity}, like movie actors who are linked based on co-acting (i.e., they act together in a movie) or have similar average rating. Thus, $V_i$'s have a common set of vertices (overlapping), and inter-layer edges are implicit, connecting the same nodes from two layers. Hence, inter-layer edge sets are empty as they are implicit. Relationships among \textbf{different types of entities} like researchers (connected by co-authorship), research papers (connected if published in the same conference), and year (related by pre-defined ranges/eras) are modeled through \textbf{heterogeneous MLN (or HeMLN).} The inter-layer edges represent the relationship across layers, such as writes, published-in, and active-in. In addition to being collaborators, researchers may be Facebook friends. Thus, to model multi-feature data that capture \textbf{multiple relationships within and across different types of entity sets}, a combination of homogeneous and heterogeneous MLNs is used, termed \textbf{Hybrid MLN (or HyMLN).} 

Figure \ref{fig:graph-types} shows different types of graph models that have been considered in this paper. Figure \ref{fig:graph-types} (a) shows a \textbf{simple graph} with a single type of nodes and edges, but without any label information. Figure \ref{fig:graph-types} (b) shows an \textbf{attributed graph (or multigraph)} that includes multiple node and edge types (illustrated using different colors). It also illustrates support for multiple edges between two nodes (multiple edges between light green-colored nodes). Finally, Figure \ref{fig:graph-types} (c) shows a \textbf{multilayer network} (specifically, HyMLN) where the attributed graph of Figure~\ref{fig:graph-types} (b) is separated into different layers such that each layer captures information about a single type of entity and relationship in the form of a simple graph/network. Due to the presence of three types of relationships (shown through orange, blue, and purple colored edges), three separate layers/simple graphs are generated. Also note that the first and third layer have the same node types but with different relationship types, making it a Hybrid MLN.

The above three graph types, as well as variants of MLNs, clearly provide a suite of alternatives for modeling any data set based on entity types and relationships, as well as objectives to be explored (in terms of labels retained). It also provides flexibility of choice as the same information can be represented as attributed graphs or MLNs. Simple graphs and MLNs also provide better clarity and separation of semantics, enhancing understanding of the dataset. Additionally, the availability of algorithms for a specific graph model will also play a crucial role in determining the choice of graph model. As an example, there are \textit{not many} algorithms available for attributed graphs in contrast to simple graphs. As there is considerable ongoing research in developing algorithms for multilayer networks \cite{MultiLayerSurveyKivelaABGMP13,aleta2019multilayer,CommFortunatoC09}, and the clarity of the model is better, MLNs are becoming popular and preferred for modeling complex datasets.

Multi-feature data comprises of multiple relationships among the same or different types of entities. Relationships among entities can be specified by explicitly (e.g., distance or spatial relationships between objects) or based on a similarity metric, depending on the type of feature, such as nominal, numeric, time, date, latitude-longitude values, text, audio, video, or image. For preserving the structure and semantics of complex data sets as well as their efficient analysis, multilayer networks (or MLNs) are a better choice than attributed graphs~\cite{ICCS/SantraBC17,BDA/SantraB17,MultiLayerSurveyKivelaABGMP13}.

\subsection{Available Algorithms for Different Graph Models}

Numerous algorithms have been developed for simple graphs. Many algorithms have been developed for shortest paths, spanning trees, Hamiltonian and Eulerian paths/cycles, and cliques on simple graphs. Breadth and depth-first approaches are used for many algorithms. There are multiple algorithms for aggregate analysis, such as community detection, centrality, and substructure analysis on simple graphs. On the other hand, very few algorithms exist for  attributed graphs (except for substructure discovery~\cite{KDD/HolderCD1994,book/PadmanabhanC2009,datamine/KuramochiK05, ICDM/YanH2002} for exact and inexact or similar substructures.) 
Hence, MLNs are preferred instead of attributed graphs. Instead of an attributed graph, an MLN can be generated and processed, as it provides many advantages. 
MLNs can be analyzed in many ways. Recently, a novel decoupling approach has been developed~\cite{phdThesis/Santra20} which is efficient, flexible, and scalable. We will use this approach for one graph type, where we generate multiple graphs for a video instead of a single graph. or forest.
Currently, algorithms are being developed for MLNs using the decoupling approach. As MLNs provide clarity of semantics and also preserve structure, it is a good alternative as it decomposes a large problem into smaller problems.

As indicated above, new algorithms are being developed to analyze MLNs efficiently without loss of structure and semantics. This approach -- termed the decoupling approach~\cite{ICCS/SantraBC17,ICDMW/SantraBC17,Arxiv/SantraKBC20,IC3K2023/PavelRSC23} -- also leverages existing algorithms available for simple graphs and composes aggregate analysis using them. In this approach, analysis results clearly preserve both structure and semantics, allowing for easy drill-down and understanding of the results.

\subsection{Summary}

In summary, our focus is on using simple and MLN graph types for modeling extracted video contents. Multiple relationships between the same objects can be represented using HoMLNs. Relationships between different entity types can also be represented using HeMLNs. We develop new analysis algorithms tailored for situations that utilize graph representation and use existing algorithms where available. These algorithms can also be composed to detect composite situations as needed. For more details on analysis algorithms for MLNs, see~\cite{ICCS/SantraBC17, ICDMW/SantraBC17,CICLing:Vu2019,DBLP:journals/dke/SantraKBC22,Arxiv/SantraKBC20,IC3K2023/PavelRSC23}. 

\section{Graph Representation of Extracted Video Contents}

As described in Chapter~\ref{chap:vce_workfolow}, a video $\mathcal{V}$ is a sequence $F^{\mathcal{V}}$ frames $<1,\ldots,\mathcal{F}>$ where $\mathcal{F}$ is the last frame. $\mathcal{F}$ or $|F^\mathcal{V}|$ indicates the total number of frames in video $\mathcal{V}$. Typically, some frames are empty, and most are non-empty. A VCE workflow extracts a set of objects (or object ids along with their class label, confidence, and other attributes) $O_f$, which can be described as $\{o_{id} ~|~ o_{id} \in f\}$ from each frame f. The total number of objects in a frame $f$ is $|O_f|$. This can vary from frame to frame. The object ids need not be consecutive in a frame. When referring to a specific object id $o_i$ in a non-empty frame $f$, we will use the notation $o_i^f$ to make it clear. For an empty frame $e$, $O_e$ is an empty set or $|O_e|$ is 0. An extracted object id $o_i$ is usually present in some number of frames, and these frames need not be consecutive, as an object $o_i$ can reappear after some period in a video and even multiple times. At the same time, the same object may be given a different object ID because of the object tracking threshold or due to a VCE quirk in the presence of occlusion, and other environmental factors. 
A VCE workflow extracts a number of attributes (such as feature vector, class label, bounding box, etc.) for each object id $o_i$  in frame f. We will denote each attribute for an object id individually. For example, we will denote the class label attribute of object $o_i$ in frame $f$ as $CL_{o_i}^f$ as its value varies from frame to frame. Similarly, $FV_{o_i}^f$ and $BB_{o_i}^f$ represent the feature vector and the bounding box attributes, respectively. So, all the attributes of an object id $o_i$ in frame $f$ can be denoted as a set $\mathcal{A}_{o_i}^f$ consisting of \{$CL_{o_i}^f$, $FV_{o_i}^f$, $BB_{o_i}^f, ...\}$. All the attributes of an object $o_i$ in the entire video can be denoted by  ${\mathcal{A}_{o_i}^\mathcal{V}}$ and can be computed as  
$\bigcup_{f \in F^\mathcal{V}} ~\mathcal{A}_{o_i}^f$.

Some global properties of video $\mathcal{V}$ are also useful and are extracted by the VCE workflow. The total number of unique object id's in a video $\mathcal{V}$ is denoted by $UO^\mathcal{V}$. Similarly, the total number of object instances extracted by the VCE is denoted by  $OI^\mathcal{V}$. Given $\mathcal{F}$ frames in video $\mathcal{V}$, average number of objects per frame is $OI^\mathcal{V}$/$\mathcal{F}$. 


The above extracted contents can be modeled as a graph by representing object id's as nodes/vertices, attributes of an object in a frame as node labels. When an object id occurs in multiple frames, all attributes from all frames are used as node labels, although the grouping and representation of attributes is left to implementation (or how we store the graph models). For example, each attribute type (e.g., CL, FV, or BB) from different frames can be grouped together as a vector and stored, or they can be stored separately by frame for efficient lookup using indexes. We will elaborate on this later. 

In a graph representation (regardless of the type), one or more relationships need to be identified to draw edges between nodes. This relationship may vary from one graph representation to another. It is also possible to have multiple relationships between the same object types in the graph. This will also be discussed further when we elaborate on graph models.

We have identified two primitive graph types: Single Graph per Frame (SGF), and Single Graph per Video (SGV), and a third one (Multiple Graphs per Video or MGV), which is a partitioned graph (termed layers) based on connected components. Either all connected components are kept in a single layer (SGV model) or partitioned in the MGV model. The SGF model is quite different from the other two, as the model is frame-based. Note that the SGF model cannot be interpreted as MGV as graphs are not disjoint from frame to frame.

We will be using simple and MLNs (for MGV) for our representation. In this thesis, we are using a single relationship between any two objects (mostly of type \textit{person}) and hence HoMLNs are sufficient for our purpose. MLNs are also sufficient even if we need to use multiple relationships between two objects. As most of our situations are about persons (objects with a specific label \textit{person}), we have not explored the use of HeMLNs. However, if we want to represent relationships between two distinct object types, such as a background object or a person, or two different types of background objects, HeMLNs can be used. Hence, the graph model types chosen for representation are sufficient to accommodate broader and more complex situations that are not considered in this thesis.

From the post-processed output of VCE, we generate our graph models that are suitable for analyzing various situations. Remember, as shown in Figure~\ref{fig:VCE_raw_output}, the post-processed VCE output raw data file (or $RDF^{\mathcal{V}}$) contains one row per object id detected in each frame sequentially, along with some meta information in the header.

For each graph type, two files -- Base Graph File (or $BGF^\mathcal{V}$) and Index Data File (or $IDF^{\mathcal{V}}$) -- are generated to separate information for easy lookup. $BGF^\mathcal{V}$ has the graph representation with node id's, some selected node labels (e.g., class label confidence score, frame id, etc.), edges, and edge labels. Depending on the model type, the node labels stored in the graph file differ. The other file ($IDF^{\mathcal{V}}$) contains all the node label information for each frame and is indexed on frame id and is used for looking up attributes of specific node label information in a frame (e.g., bounding box in a frame) using the index. Also, as we have indicated earlier, the relationship(s) of edges in SGF are based on that frame, whereas the relationship(s) in MGV and SGV are aggregate information. Graphs for empty frames do not contain any information and will not be stored in the graph files. We will elaborate on these file representations and the process of generating graphs later.

\subsection{Graph Models Used in VideoScoop}

As mentioned above, we have chosen three graph types for modeling and situation detection. Some of these are also used in the literature~\cite{VideoRepresentation/yadav2020knowledge,videoQuerying/EQUI-VOCAL,Scenegraph/ActionGenome} in slightly different ways than how we do. Some types are new to our approach. We use three graph types for modeling extracted video contents for different reasons elaborated along with the model description: i) one/single graph per frame or SGF, where a graph is generated for each frame in the video (a null graph is used for an empty frame), ii) one/single graph for the entire video or SGV in which we generate a forest of graphs or connected components. Each unique object id is represented as a node only once in the graph, along with all the attributes from all the frames in which the object id appears. A global relationship is used for edges between nodes, and iii) MGV, where a number of graphs (also can be thought of as MLN layers) are generated along the lines of SGV, but the number of nodes and frames used in each layer is based on some criteria. The number of MGV graphs (or layers) can also be tuned using various criteria, such as video length, total number of unique objects in the video, or total number of frames (corresponds to object instances).

Multiple graph alternatives offer flexibility in matching the situation being analyzed and performance requirements. One model may be more efficient than the other or provide parallel processing flexibility. In general, SGF is fine-grained and can be used to process a specific number of frames sequentially or to identify frames to be processed for a situation in a given duration in the video. SGV, on the other hand, is compact in terms of the number of nodes and edges and may be useful for aggregate analysis, either using a single connected graph or the whole forest, as we have done for the situation ``finding group(s) of different sizes''. Individual disconnected graphs can also be processed independently and in parallel if need be. MGV can be seen as an alternative that divides SGV based on connected components or groups frames of SGF based on object id. This model can be beneficial as it supports the balancing of nodes and/or frames using a specific criterion, and each graph (or layer) can be processed in parallel.  Results from individual graphs can also be combined as needed for detecting a situation, as is typically done for MLNs using the decoupling approach~\cite{phdThesis/Santra20}. Unlike SGF, time range-based processing is somewhat difficult in the alternative SGV and MGV representations. Furthermore, the number of graphs or layers can be controlled to minimize the total response time. Together, all three options provide a wider range of options than those currently proposed in the literature for detecting a range of situations. If we include HeMLNs, more situations can be expressed and computed without having to step outside the model. For example, if we want to identify a situation in which a person is seen together with a bridge or a checkpost, the HeMLN representation can be used. Below, we first define each model, followed by a discussion of their storage requirements, advantages, disadvantages, and compressibility.

\subsection{\textit{Single Graph Per Frame (SGF)}}

In this model, a video $\mathcal{V}$ with $\mathcal{F}$ frames generates a set of graphs $SGF^\mathcal{V}$, where 
$SGF^\mathcal{V} = \{SGF_1^\mathcal{V},\ldots,SGF_{\mathcal{F}}^\mathcal{V}\}$. Graph $SGF_f^{\mathcal{V}}$ for frame $f$ will have $|O_f|$ nodes in it, one for each object id in the set $O_f$. An empty frame will have no nodes in the graph. The labels associated with each node  in frame $f$ or  $o_i^f$ will be $\mathcal{A}_{o_i}^f$ which is \{$CL_{o_i}^f$, $FV_{o_i}^f$, $BB_{o_i}^f$, \ldots\}. In other words, the class label, feature vectors, bounding box, pose vectors, and others extracted by VCE (if used) for object id $o_i$ in frame $f$ will be stored as part of the vertex label. The frame id is used as the graph id in this model. Since there is a graph for each frame and objects appear in multiple frames, the same object id appears in many graphs.
As given earlier, the average number of nodes in a any $SGF^\mathcal{V}$ graph will be $OI^\mathcal{V}$/$\mathcal{F}$.

A relationship needs to be identified in order to draw the edges in this model. If we choose a relationship, such as spatial, using the bounding box (e.g., left, overlap, etc.), the edge will be directed. If we choose the distance relationship between any two objects, it will be undirected. 

\begin{figure}
    \centering
    \includegraphics[width=0.8\textwidth, keepaspectratio=True]{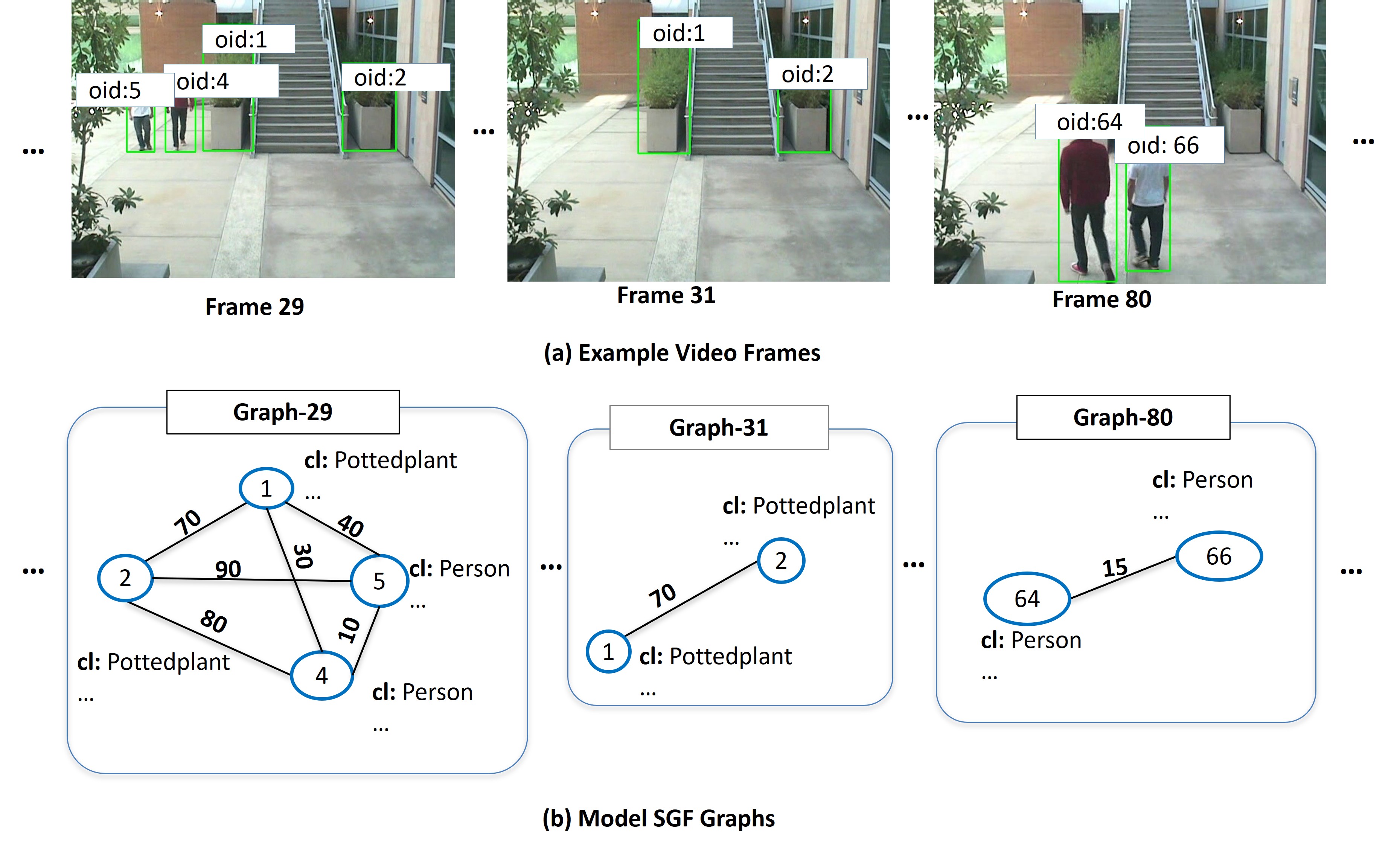}
   \caption{(a) Example video frames (b) Representation of each frame as a graph using model SGF. }
   \label{fig:Alternative-model-example}
\end{figure}

An example of this model using the distance relationship for the edges is shown in Figure~\ref{fig:Alternative-model-example} using example video frames. Here,
Figure~\ref{fig:Alternative-model-example}(a) shows three non-empty frames (frames 29,31,80) from a video in CAMNET~\cite{DataSet/Camnet} dataset. The extracted objects are shown by encompassing them with a green rectangular box, and their object ids are shown on top of the rectangle. In frame 29, four objects with object ids 1,2,4,5 are extracted, with class labels potted plant (for object ids 1,2) and person (for object ids 4,5). This frame is represented with graph-29 with four vertices and six edges in figure~\ref{fig:Alternative-model-example}(b). The bounding box centroid distances between each pair of objects in frame 29 are shown as edge labels. Similarly, frame 31 is represented with graph-31 (with two nodes for object ids 1 and 2), and frame 80 is represented as graph-80 (with two nodes for object ids 64 and 66). In this figure, only the class label ($o_{cl}$) is shown as a vertex label. Other extracted attributes, such as feature vector, bounding box, etc., are also part of the vertex label. Note that the same objects with object ids 1 and 2 are represented in both graphs 29, 31.

\subsection{\textit{Single Graph Per Video (SGV)}}

In this model, we create a connected graph or a forest of connected graphs (singleton nodes included) $SGV^\mathcal{V}$ for the entire video $\mathcal{V}$ with $F$ frames. Each unique object id in the extracted contents becomes a \textit{single} node in the graph/forest. Each node is associated with the set of attributes of that node from all the frames in which the object id appears in that video. That is, ${\mathcal{A}_{o_i}^{\mathcal{V}}}$ which is   $\bigcup_{f \in F^\mathcal{V}} ~\mathcal{A}_{o_i}^f$.
The total number of nodes in the graph $SGV^F$ is equal to the number of unique object ids $UO^\mathcal{V}$, which is equal to $|\bigcup_{f \in F^\mathcal{V}} ~O_{f}|$.

As each node represents all occurrences of that object in the video in this model, identifying a relationship to be drawn between objects is tricky. The same two objects need not be present in all the frames associated with the set of objects in those frames. The relationship used between objects needs to be applicable across frames, and the edge label must include that. Hence, we choose an \textit{aggregate relationship} between two objects across frames in which they appear. In this model, an edge represents the occurrence of two objects at least in a frame, and further, the frames in which the distance is minimum. We also keep the frames in which this distance is maximum for the same two objects. So, the edge label consists of $minimum ~distance, <f_i, ..., f_j>$ and $maximum ~distance, <f_k, ...,f_l>$, where $f_i, ..., f_j$ are the sequence of frame id's with minimum distance, and $f_k,\ldots,f_l$ are sequence of frame id's with maximum distance\footnote{In literature, SGV is used for representing fixed $n$ number of video frames together as a graph~\cite{videoQuerying/yadav2019vidcep}, and draws an edge between every pair of objects across $n$ frames (generates a complete graph always). The edge labels are typically represented as a sequence of distances or bounding box spatial relationships (e.g., overlap, left,etc.)~\cite{videoQuerying/yadav2019vidcep} and corresponding frame timestamps for all the $n$ frames. If two objects never appear together in a frame among the $n$ frames, the edge labels contain $X$ (do not care). This representation is complex, and storing all the sequences of edge labels for all the frames is not a storage-efficient choice.
 Another way of representing edge labels for this model in the literature is to assign a predicted label of action classes (e.g., drinking, sitting, etc.) using deep learning models~\cite{Scenegraph/VidSGG}.}.



 The above aggregated information can be useful for certain analyses, such as analyzing the distance between pairs of objects or the duration of stay of a pair of objects in a video. Some of this aggregated information is used for analysis in this thesis (discussed in chapter~\ref{chap:graph-ana-videocontents}).



 
\begin{figure}
    \centering
    \includegraphics[width=0.8\linewidth, keepaspectratio=True]{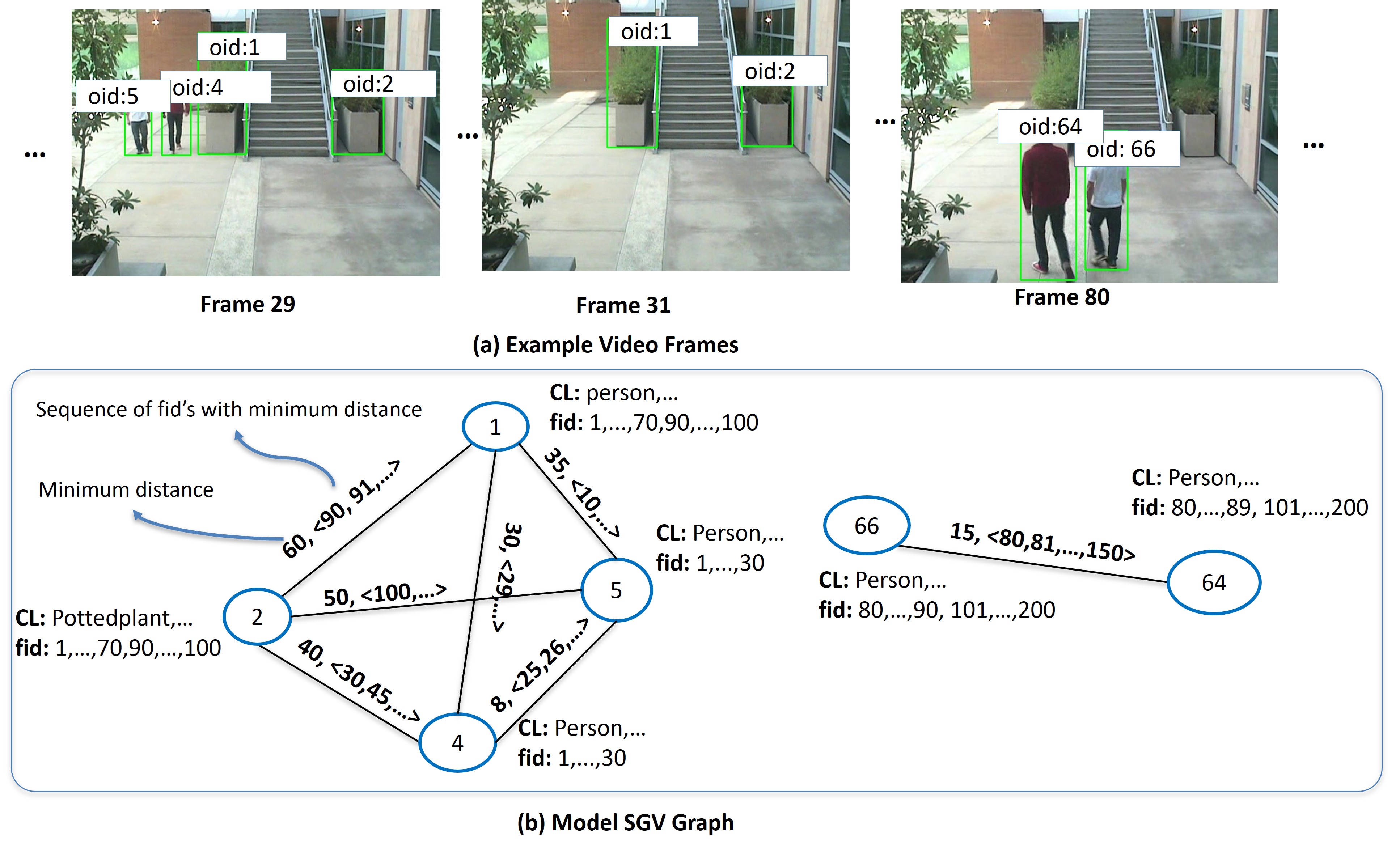}
    \caption{(a) Example video frames (b) Graph representation using model SGV. $CL$: object class label, and $fid$: frame id}
    
    \label{fig:SGV-example}
\end{figure}

An example of this model is shown in Figure~\ref{fig:SGV-example}. In Figure~\ref{fig:SGV-example}(a), three non-empty frames are shown (with object id at the top, and a bounding box as a green rectangular area) for a video. In Figure~\ref{fig:SGV-example}(b), an example graph representing all the video frames is shown. In this figure, the objects that appear in multiple frames (e.g., object ids 1 and 2, as shown in frames 29 and 31, respectively) are represented by a single vertex. The frame ids, and object class labels are shown as node attributes. This graph has two connected components: one with objects 1, 2, 4, and 5, and another with objects 64 and 66. Since objects 64 and 66 never appeared (or were detected) together with objects 1,2,4,5, there are no edges between these two sets of vertices. In this graph, only the minimum distance and the corresponding frame ids are shown as edge labels for simplicity.



\subsection{Multiple Graphs Per Video (MGV)}

This model represents a video $\mathcal{V}$ with $\mathcal{F}$ frames as $MGV^\mathcal{V} =\{MGV_1^\mathcal{V},\ldots,MGV_N^\mathcal{V}\}$ set of graphs where $1 \leq N \leq \mathcal{F}$. The graphs in each layer are the same as the graphs created for SGV, but a subset of them is in each layer. The number of graphs to be generated needs to be determined or computed based on the parameters of the video. For example, the number of unique object ids can be used to determine the number of (approximately) nodes using some threshold useful for parallel computation. Another way to generate the graphs is based on the number of frames. However, what is important is to balance the graphs (or layers) for computation to avoid a lopsided analysis of individual graphs. 

Each graph or layer generated will include one or more connected components. A connected component does not span more than one layer. In other words, \textit{a connected component will not be split between two graphs/layers.} This is important for \textbf{independent}  processing of layers in parallel. Since we are regrouping SGV into MGV, the total number of vertices, edges, and connected components in $MGV^\mathcal{V}$ is the same as the model SGV for the same chosen relationship. If graphs are disjoint among layers, the union (or some simple function, such as max or min) of the results computed will be the result for the entire graph.
 
\begin{figure}
    \centering
    \includegraphics[width=0.8\linewidth,keepaspectratio=True]{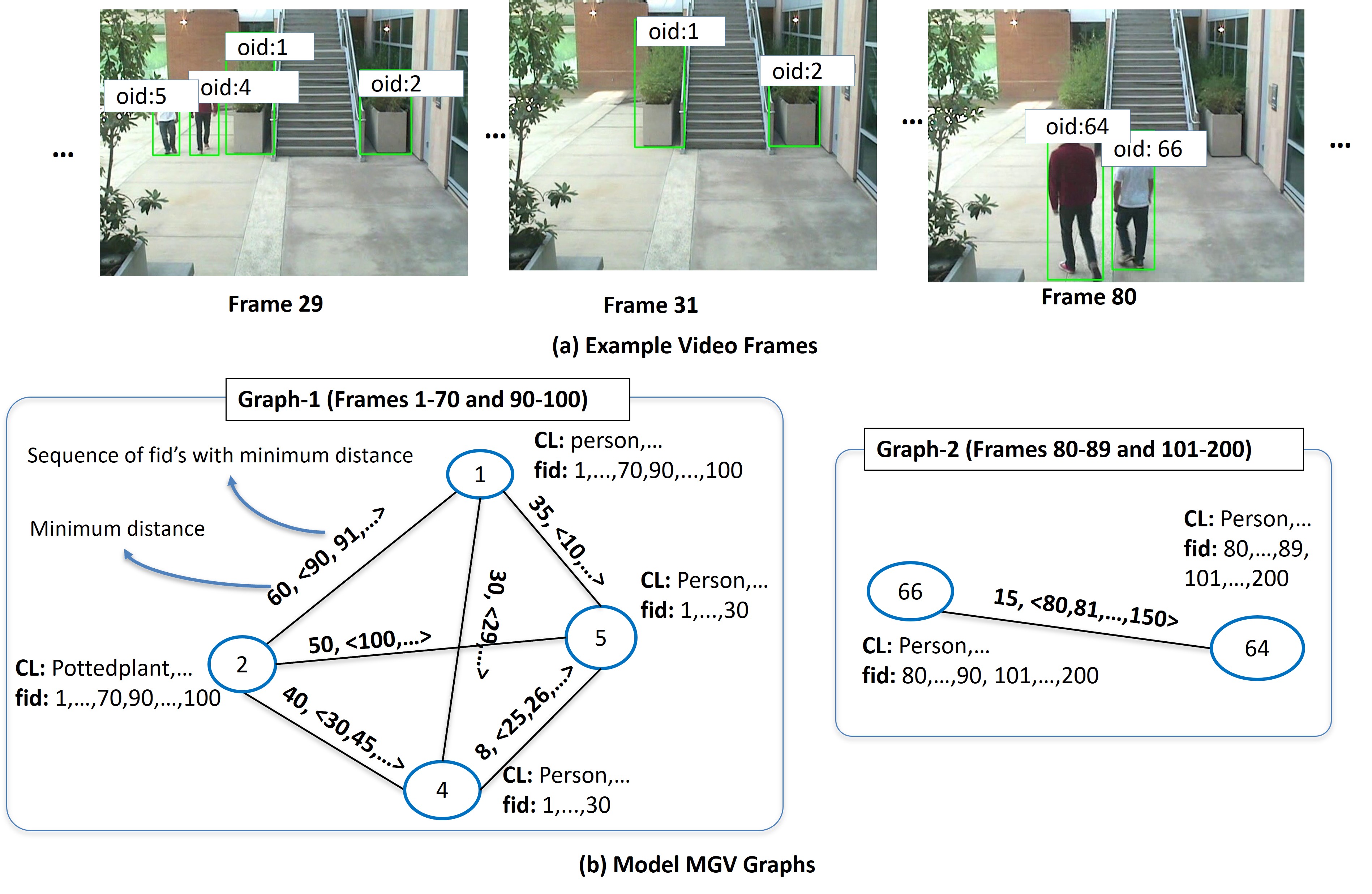}
    \caption{\small{(a) Example video frames, (b) Graph representation using model MGV. Here, $CL$: object class label and $fid$: frame id}} 
    
    \label{fig:MGV-example}
\end{figure}

Figure~\ref{fig:MGV-example}(b) shows an example of this model for the same video frames shown above for the other two models. In this figure, graph-1 is generated from non-empty frames 1-70, and 90-100, and graph-2 from non-empty frames 80-89 and 101-200. Note that here the two connected components of the model SGV (shown in Figure~\ref{fig:SGV-example}(b)) are represented with two different graphs, and the frames representing the graphs are not consecutive sequences.

\subsection{Discussion of Alternative Models}

All three models contain the same information generated by the VCE workflow in slightly different ways. The node labels are the \textit{same} in all three models. The edge labels are the same for models SGV and MGV. The difference comes mainly from the representation of the number of nodes and edges in different models. Hence, we briefly discuss those differences.

The total number of nodes and edges is the largest for SGF, as the number of nodes corresponds to the number of objects in instances $OI^\mathcal{V}$ for video $\mathcal{V}$, which is  $\sum_{f \in F^\mathcal{V}}|O_f|$. As each SGF graph is complete, the number of edges ($\sum_{f \in F^\mathcal{V}}\frac{|O_f|^2}{2}$). The size of the video in terms of the number of frames has a bearing on this. This representation may not be viable for large videos, as the situation processing time may increase non-linearly. At the same time, if situations need to be detected over a short time window, this representation can efficiently determine the number of frames needed.


The number of nodes in SGV and MGV graphs is the same and is equal to the number of unique object ids $UO^\mathcal{V}$. Total number of edges will be $(UO^\mathcal{V})^2/2$, if a complete graph. The number of edges is significantly less than the SGF graph for the same video. 




Below, we indicate the differences in representation in terms of the number of nodes and edges for four different categories of video type (small, medium, large, and mixed) in Table~\ref{tab:comparison-of-models-nodes-and-edges}. As discussed in chapter~\ref{chap:vce_workfolow}, the small videos are of length $< 5$ minutes, medium videos have length $5-15$ minutes, and large videos are of length $>15$ minutes. The mixed videos are of different lengths ($1$ min.-$2$ hours) generated by mixing the small, medium, and large videos in random order. The number of nodes and edges shown in Table~\ref{tab:comparison-of-models-nodes-and-edges} are from the actual graphs generated for models $SGF$ and $SGV$ on all these videos. This table shows that model SGF has approximately 47 times as many nodes on average for small videos as model SGV. Similarly, for medium, large, and mixed videos, SGF has, on average, 15, 58, and 32 times more nodes, respectively. The number of edges on average is also way more than SGV. Model MGV is not shown in this table, as the number of nodes and edges is the same as SGV.


\begin{table}[H]
    \scriptsize
    \centering
     \caption{Summary of the number of nodes and edges generated for different models}
    \label{tab:comparison-of-models-nodes-and-edges}
    \begin{tabular}{|c|c|c|c|c|}
        \hline
         Video Type & Avg. No. of Frames & Model Type & Avg. No. of Nodes & Avg. No. of edges \\ \hline
         Small (59 videos) & 1667 & SGF &3643 & 6601\\ \cline{3-5}
         ~ & ~ & SGV & 77 & 360 \\ \hline 
         
         Medium (13 videos) & 19073 & SGF & 61911 & 149069 \\ \cline{3-5} 
         ~ & ~ & SGV & 4032 & 13688\\ \hline
          Large (19 videos) & 42915 & SGF & 64463 & 136285 \\ \cline{3-5} 
    ~ & ~ & SGV & 1334 & 2498\\ \hline
         Mixed (18 videos) & 38206 & SGF & 71669 & 97505 \\ \cline{3-5} 
    ~ & ~ & SGV & 2188 & 8087 \\ \hline
    \end{tabular}
   
\end{table}

\noindent\underline{\textit{Analysis Efficiency:}} Analysis efficiency depends on the situation being detected and the model being used. Our aim is to generate all three models for a video and develop algorithms for all three models to come up with a heuristic on which model is best based on the type of computation or traversal involved. In general, SGF is too fine-grained and graphs are too small for parallel processing. SGV and MGV are better suited for parallel processing, depending on the size of the graph in each layer. The way MGV is formulated, each graph/layer is independent, and a minimal composition of results is needed, if at all. This may change when HeMLNs are also included. Based on the above analysis, SGF may not be appropriate for large videos or videos beyond a certain number of frames, unless time range-based situation detection is needed. For small videos, SGF may be viable.

To leverage the full benefit of parallel processing, the graphs generated for MGV need to be \textbf{balanced}, so that the processing time of each graph and of any single graph is not lopsided, thereby increasing the maximum response time and consequently, the overall response time. The graphs can be balanced either in terms of (i) the number of nodes or (ii) the number of frames in each graph. Depending upon the analysis performed, one of the heuristics may be beneficial for MGV graph generation. For example, analysis that requires individual frame processing (e.g., clustering objects to identify group(s) of different sizes), having approximately the same number of frames can be beneficial, which we have showcased later in chapter~\ref{chap:graph-ana-videocontents}. Now it is also possible, after balancing, that a graph can have one or more large connected components (if an object is present in every video frame, or if every consecutive frame has an object in common, or forms a chain). In all these cases, there will be a graph that dominates the processing time. It should be possible to merge graphs across layers to further balance this.


\noindent\underline{\textit{Compression:}} In the absence of compression, all models contain the same label information and hence should return the same accuracy for any situation detection. Response time is likely to vary from model to model.

If compression were to be used, node labels are the best target for that, as they form the bulk of the storage. Node label compression is difficult or not possible for SGV due to the model type. Compression can be used beneficially for the other two models.
Compression may be possible either in a lossless manner or in a lossy way. Lossless compression, in principle, should give the same accuracy as the non-compressed representation. On the other hand, lossy compression may compress better, but may result in reduced accuracy in detecting situations.

In general, SGF is difficult to compress in a lossless manner as labels are distributed across graphs. The other two representations are better and are more amenable to compression. SGF can be compressed in a lossy manner, taking into account the fact that videos have a frame rate and field of view that is not likely to change very quickly (especially for a static camera and subjects that are not fast-moving). So, it may be possible to keep only a few frames per second and drop the others. This may not affect the detection of many situations (e.g., counting the number of person objects in a video), but may affect other situations (e.g., disappearance of an object). 

For SGV and MGV, compression can be done on labels as they are grouped together for an object id. Some compression may be lossless, and others may not be lossless. For example, if the same class label appears in consecutive frames, they can be compressed without losing any information. On the other hand, compressing bounding boxes and feature vectors (using some threshold) may result in decreased accuracy and is difficult to predict without further research. A notion of payoff can be associated based on the percentage of storage reduction and the corresponding percentage of accuracy reduction. This is somewhat analogous to load shedding that has been researched for sensor data stream processing.  The nuisances of video processing need to be taken into account. This is part of our future research. 
We will not be addressing compression in this thesis, as our focus is on the framework and situation detection.

Storage reduction will be maximum for labels that are large, such as feature vectors, as compared to class labels. It may be possible to keep a few selected (based on a similarity threshold) and reduce storage significantly.

\subsection{Generating Graph Models from Extracted Video Contents}
As mentioned earlier, the graph models (SGF/SGV/MGV) introduced above are generated in \underline{\textit{one pass}} over the pre-processed data file $RDF^{\mathcal{V}}$ created by the VCE workflow. As described in chapter~\ref{chap:vce_workfolow}, file $RDF^{\mathcal{V}}$ for a video $\mathcal{V}$ (shown in Figure~\ref{fig:VCE_raw_output}) contains: (i) \textit{global video characteristics}  such as frame per second, video length, video type (small/medium/large/mixed), frame width, frame height, and video generation time as header, (ii)  computed meta information by post-processor (maximum, minimum, and average number of objects in a frame, frames with minimum and maximum number of objects, total number of unique object ids, total number of object id instances, minimum number of frames an object id appears (or minimum duration), and total number of non-empty frames in the whole video) as header, and (iii) a row for each $\langle frame~id, object~id \rangle$ pair with extracted attributes for the object id. A frame $f$ has $|O_f|$ rows in $RDF^{\mathcal{V}}$. In total $RDF^{\mathcal{V}}$ has $OI^{\mathcal{V}}+h$ number of rows, where $OI^{\mathcal{V}} =\sum_{f\in \mathcal{F}} |O_f|$, and $h$ is the number of header rows. For model generation, all rows of a frame $f$ are processed together. 


Two types of files are generated for each of the graph models from $RDF^\mathcal{V}$. A base graph file or $BGF^\mathcal{V}$ for model SGF and SGV or multiple $BGF^\mathcal{V}$ files for model MGV. A second data file indexed by frame id (termed $IDF^{\mathcal{V}}$), where index information is stored as a header. This separation keeps the base data graph file(s) compact and small with graph information. The indexed data files contain all the node label information for easy lookup as needed. Additionally, the node attributes contain a large amount of information of different types (e.g., multiple vector types). All of this information may not be necessary for every type of analysis. Instead of storing all these attributes, we have stored some minimal information (e.g., class label, confidence score, frame id) from the node/vertex labels in the graph file(s), which is required for most situation analyses in general. All the node label information (e.g., bounding box, feature vector, etc.) of a node in a frame is also stored as part of the $IDF^{\mathcal{V}}$, so that when frame-specific information for a node is required for an analysis, it can be fetched. Only one index data file $IDF^{\mathcal{V}}$ is created for all the models. The base data graph file(s) also store the post-processed meta-information and video characteristics information (fetched from $RDF^{\mathcal{V}}$ during graph generation) as a header. Additionally, some graph characteristic information (total number of connected components, largest connected component size, smallest connected component size, number of singleton nodes, maximum and smallest number of frames in a component, maximum, minimum, and average degree of a graph) for the graph(s) of model SGV and MGV is stored in the graph file header, which might be useful for some analyses.

Below, we provide an overview of how the models are generated and then stored in the two file types mentioned above. We then showcase our proposed algorithms for graph generation, along with the output files generated for each model.

 \noindent \underline{\textit{Model SGF}} generation is straightforward. For each frame $f$, after reading all $|O_f|$ rows from $RDF^{\mathcal{V}}$, the node labels of a graph $SGF_f^{\mathcal{V}}$ are written directly to the graph file $BGF^\mathcal{V}$ in the appropriate format, as no computation is required for generating nodes. The only computation needed for this model is to generate edge labels (e.g., bounding box centroid distance, bounding box spatial relationship, etc.) between each pair of objects in $O_f$. As soon as an edge label is computed, it is written to the graph file $BGF^\mathcal{V}$. For SGF generation, once a frame is processed, nothing needs to be stored in memory. We store only the class label and the class label confidence score from each graph's node labels in the graph file. The file representation is discussed in detail later for this model.

\noindent \underline{\textit{Model SGV}} generation requires aggregating all node labels across frames and computing aggregate edge label information to generate a graph $SGV^{\mathcal{V}}$, which involves additional computation. For all the \textit{new} object id instances in frame $f$ that have not appeared before in the video, \textit{new nodes} along with their labels need to be added to $SGV^{\mathcal{V}}$. For the object id instances that have appeared before frame $f$ (i.e., nodes that already exist in $SGV^{\mathcal{V}}$), some or all of their extracted attribute information (from $|O_f|$ rows in $RDF^{\mathcal{V}}$) is added as node labels in the graph file.

For this model, there are multiple choices for storing node labels in the base graph file $BGF^\mathcal{V}$. It is possible to group a node’s attributes together and store them as node labels (e.g., a sequence of bounding boxes and corresponding frame IDs). However, if an analysis such as clustering requires extracting all bounding boxes of the objects present in a frame, all vertices and their labels need to be scanned. In other words, multiple node labels would need to be accessed to extract all bounding boxes for objects appearing in a frame. \textit{This lookup is highly inefficient and complex.} Additionally, the number of objects in each frame can be beneficial for certain analyses (e.g., identifying groups of different sizes). Hence, we store the frame id and the number of objects in each frame as node labels in the base graph file $BGF^\mathcal{V}$. \textit{These are stored in descending order of the number of objects, which can again be advantageous for some analyses.} This way, when information such as the bounding box is required for a specific frame, it can be fetched from $IDF^{\mathcal{V}}$, and when only the frame id and number of objects are needed, they can be used from the graph file without a complex lookup.

In addition to the above, the class labels and corresponding confidence scores are stored in the base graph file $BGF^\mathcal{V}$ as part of the node labels. The class labels for a node can appear across a large number of frames (with some frames having the same label and others having different labels, each with different confidence scores). Rather than storing redundant information, the distinct class label categories, along with their highest confidence scores across all frames where the label is assigned, are stored in the graph file $BGF^\mathcal{V}$. In this way, if any analysis requires filtering certain nodes based on a label, it can be performed early on in this model.

For this model, the edge labels (e.g., bounding box centroid distance) for each pair of object instances in frame $f$ need to be computed, as they are used to update the global edge label. Using this information, the labels for \textbf{existing} edges are updated with aggregated information (e.g., minimum or maximum distance and the corresponding frame id sequences). For each \textbf{new} pair of object id instances in frame $f$, a new edge is added to $SGV^{\mathcal{V}}$. 

In videos, objects may reappear at any point in time and are typically assigned a unique object id for a limited number of frames (termed the tracking threshold, $th_{track}$). After $th_{track}$ frames, objects are assigned a new object id. The VCE pipeline must therefore assign new sets of object id's after each $th_{track}$ interval. Assuming this is the case, after every $th_{track}$ frames, the nodes and edges of $SGV^{\mathcal{V}}$ are written to a graph file $BGF^\mathcal{V}$, instead of storing all the nodes and edges in main memory, as their node and edge labels will no longer be updated. This reduces the number of nodes, edges, and corresponding labels stored in main memory, which is critical for processing very large videos (e.g., 24-hour-long videos) in general.

\noindent \underline{\textit{Model MGV}} generation is more complicated than SGV and SGF. As we have discussed above, model MGV can generate $N$ graphs for a video $\mathcal{V}$, which can be processed in a parallel manner during analysis. The graphs can be generated by balancing the number of frames or nodes, where each graph can have approximately the same average number of nodes or frames (denoted as $\mu$). However, if $N$ is chosen randomly, graphs of very small size (in terms of number of nodes or frames) can be created. Processing such small graphs in parallel may not improve efficiency for an analysis. Depending on the video type and length, the graph size should vary. For example, typically small videos have a very small number of nodes and frames (shown in Table~\ref{tab:comparison-of-models-nodes-and-edges}). For these videos, creating model MGV is not likely to be useful, as partitioning such a small number of frames/nodes into $N$ graphs and processing them in parallel will not significantly improve efficiency. For these videos, creating one graph (or model SGV) will be better for situation computations. Similarly, for medium and large videos, the number of graphs generated should vary depending on the number of nodes/frames. Hence, we need to specify the minimum graph size (termed $minGraphSize$) in terms of the number of frames or nodes, depending upon how we want to balance the graphs. This can be set as a system parameter. In this way, we can control the maximum number of graphs (termed $maxGraph$) that can be generated for MGV (an upper bound on how many graphs can be generated), as well as the average number of nodes/frames per graph. Depending on the video type, $N$ will vary ($N \leq maxGraph)$. 


Apart from the above, the MGV model generation should ensure that connected components are not spread across multiple graphs. To do this while processing a frame $f$, we need to determine which components among the existing graphs in $MGV^{\mathcal{V}}$ have common object id's/node ids with $O_f$. Here, there can be three cases: (i) one or more object id's from $O_f$ can have common node id's with multiple components within a graph $MGV_i^{\mathcal{V}} \in MGV^{\mathcal{V}}$, (ii) one or more object id's from $O_f$ can have common node id's with multiple connected components across different graphs in $MGV^{\mathcal{V}}$, and (iii) none of the object id's from $O_f$ have common node id's with any of the components of the existing graphs in $MGV^\mathcal{V}$. Case (i) can occur in a scenario, when one or more objects appear at different times and then re-appear together in the field of view after a \textit{very short duration}. In this case, all the components in a graph $MGV_i^{\mathcal{V}} \in MGV^{\mathcal{V}}$ that have a common node id's with $O_f$ should become one connected component. Case (ii) can occur when one or more objects appear at different times in a video and then reappear together in the field of view after a very \textit{long duration}; by that time, multiple graphs may have been generated. In this case, all the components from different graphs that have common node id's with $O_f$, should become one large connected component along with $O_f$, and \textit{be part of one graph}. Note that it is possible that cases (i) and (ii) both happen for some set of object ids in $O_f$, and in that case, components within a graph and then across graphs need to be merged. Case (iii) occurs when a completely new set of object id's $O_f$ appears in frame $f$. In this case, a new component should always be created. Depending on the graph size, we can put this component on any of the graphs. In our proposed algorithm, we place it either in the last graph if it can accommodate $|O_f|$ nodes. Otherwise, a new graph is created (if $N < maxGraph$) with the $O_f$ set of node.


It is also not possible to maintain exactly the same number of nodes or frames in each graph, and there should be a provision to ensure that each graph has some number of nodes more or less than the average number of nodes or frames allowed per graph. This can be handled by allowing each graph to have $\mu~\substack{+ \\ -}~\delta_{MGV_i^{\mathcal{V}}}$ number of nodes/frames, where $\delta_{MGV_i^{\mathcal{V}}}$ is the number of nodes/frames in the smallest connected component in a graph $MGV_i^{\mathcal{V}}$ (excluding singletons or components with one frame). Finally, all the node and edge label information can be written to the corresponding graph files after processing $th_{track}$ number of frames. However, to make the decision, whether a new graph should be created, the set of nodes/number of frames in the last graph should be kept track of, even after writing all the graph information in a file after $th_{track}$ number of frames. Information about the set of nodes/number of frames in other graphs is not necessary, as it is not needed to determine whether a new graph should be created.

It is also possible that the graphs generated by this model after processing all the $\mathcal{F}$ frames become imbalanced, where some graphs can contain a very small number of nodes/frames, in comparison to the rest of the graphs. In that case, the components of the graphs of $MGV^{\mathcal{V}}$ can be rearranged to further balance the graphs by making another pass on the graphs of $MGV^{\mathcal{V}}$. However, this is not handled by the proposed algorithm.



\subsection{Graph Model Generation Algorithms}
Taking all aspects of graph generation into account as discussed above, a graph generation algorithm is presented in Algorithm~\ref{alg:Graph-gen}. The table of notations used by this algorithm is shown in Table~\ref{tab:notations-graph-gen}. This algorithm takes as input the raw pre-processed data file $RDF^{\mathcal{V}}$, type of graph $model$ (SGF/SGV/MGV/$*$, $*$ to generate all models) to be generated, $edgeType$ (accepts different values depending upon model type), the tracking threshold $th_{track}$ used in the VCE pipeline, $balanceBy$ (node/frame), and $minGraphSize$ (in terms of number of nodes/frames). The $balanceBy, minGraphSize$ parameters are used for model MGV creation only (shown as underlined), and $th_{track}$ is used for MGV/SGV creation. The $edgeType$ parameter value is \textit{``distance''} (of bounding box centroids) or ``\textit{bounding box spatial relationship}'' for model SGF, and \textit{''min max distance''} for model SGV/MGV. Since the analyses in this thesis use bounding box centroid distances, only edge generation for $edgeType$ \textit{``distance''} (model SGF), and \textit{``min max distance''} (SGV/MGV) is shown in the algorithm. It outputs one or more $BGF^{\mathcal{V}}$ files, depending upon the model type, and the data file indexed on frame id $IDF^{\mathcal{V}}$.

\begin{table}[!htb]
\centering
\scriptsize
\caption{ \small Table of notations for proposed algorithms for identifying the situation ``Finding group(s) of different size''}
\label{tab:notations-graph-gen}
\begin{tabular}{|p{0.15\textwidth}|p{0.8\textwidth}|}
    \hline
    \textbf{Notation} & \textbf{Description} \\ 
    \hline
    $RDF^\mathcal{V}$ & Raw pre-processed file generated by VCE for a video $\mathcal{V}$ \\ 
    \hline
  $model$ & Type of graph model to generate: SGV/MGV/SGF/ \texttt{*} (all models). \\ \hline
    
   $edgeType$ & For SGF: ``\textit{distance}'' or ``\textit{bounding box spatial relationship}''.  \\
    ~ & For SGV/MGV: ``\textit{min max distance}''. \\ \hline
    
    $balanceBy$ & Balances model MGV generation by ``node'' or ``frame''. \\ \hline
    $minGraphSize$ & Minimum number of nodes or frames per graph in MGV.  \\ \hline
    $maxGraph$ & Maximum number of graphs MGV can generate. \\ \hline
    $\mu$ & Average number of nodes/frames allowed per graph in model MGV. \\ \hline
    $th_{track}$ & Tracking threshold. Number of frames for which VCE assigns a unique object id in a video. \\ \hline

    $H_f$ & A dictionary constructed from $|O_f|$ rows of $RDF^{\mathcal{V}}$. The key is object id. The values are stored in the format $<BB_{o_i}^f ~centroid,CL_{o_i}^f, CLC_{o_i}^f>$ for an object id $o_i$. \\ \hline
    

     $\delta_{MGV_i^{\mathcal{V}}}$ & Smallest connected component size (excluding singletons)/component with the least number of frames in a graph 
    $MGV_i^{\mathcal{V}}$ in $MGV^{\mathcal{V}}$ \\ \hline
    
    $\delta_{MGV^{\mathcal{V}}}$ & A list storing $<\delta_{MGV_1^{\mathcal{V}}}, \ldots, \delta_{MGV_N^{\mathcal{V}}}>$ values for all the graphs in $MGV^{\mathcal{V}}$. Maintains one-to-one correspondence with the graphs in $MGV^{\mathcal{V}}$. \\ \hline

    $C_{MGV_{i}^{\mathcal{V}}}$ & $\{\, C | ~C$ is the set of node id's or object id's in a connected component of graph~$MGV_i^{\mathcal{V}}$. For example, $C_{MGV_i^{\mathcal{V}}}$ can contain $\{\{o_1,o_2\},\{o3,o4,o5\}\}$, where $\{o_1,o_2\}$, $\{o3,o4,o5\}$ are two different components of graph $MGV_{i}^{\mathcal{V}}$. \\ \hline

    $C_{MGV^{\mathcal{V}}}$ & List of all component sets across graphs, i.e., $C_{MGV^{\mathcal{V}}} = < C_{MGV_1^{\mathcal{V}}}, \ldots, C_{MGV_N^{\mathcal{V}}} >$, maintaining one-to-one correspondence with the graphs in $MGV^{\mathcal{V}}$.\\ \hline

    $intersectingComp$ &  A dictionary with key as graph id, and value as the set of node id's in a connected component of the graph intersecting with $O_f$. Generated while processing a frame and used for shifting components while generating model MGV. \\ \hline
    
\end{tabular}

\label{tab:graph-gen-notation}
\end{table}

Algorithm~\ref{alg:Graph-gen} computes $maxGraph$ and $\mu$ if the given $model =~''MGV''$ in lines 2-4. It computes $maxGraph = \mathcal{F}/minGraphSize$, and $\mu = \mathcal{F}/maxGraph$ (average number of frames per graph) when $balanceBy =~''frame''$. When $balanceBy =~''node''$, it computes $maxGraph = UO^{\mathcal{V}}/minGraphSize$, and  $\mu = UO^{\mathcal{V}}/maxGraph$ (average number of nodes per graph). In line 3, Algorithm~\ref{alg:Graph-gen}) writes out all the computed meta information and video characteristic information from $RDF^{\mathcal{V}}$ header to the graph file(s) header. It then starts reading a line from file $RDF^{\mathcal{V}}$ (line 6). It reads all the lines having frame id $f$, and stores their information in $H_f$ (lines 10-14 in Algorithm~\ref{alg:Graph-gen}). For each object instance $o_i^f$ in frame $f$ (line 11), the algorithm extracts its attribute information $A_{o_i}^f$ and appends it to $IDF^{\mathcal{V}}$ in an appropriate format. From $A_{o_i}^f$, the bounding box $BB_{o_i}^f$, the class label $CL_{o_i}^f$, and the confidence score $CLC_{o_i}^f$ are extracted. The $BB_{o_i}^f$ centroid, $CL_{o_i}^f$, $CLC_{o_i}^f$ values are inserted into $H_f$, where the key is $o_i^f$ and the value is $<BB_{o_i}^f~centroid, CL_{o_i}^f, CLC_{o_i}^f >$ (line 12). Only these attributes are stored in $H_f$ because bounding box centroids are needed for edge computation, while class labels and confidence scores are stored as node labels in graph files. After processing all rows for frame $f$, an index on frame id $f$ is created for the $IDF^{\mathcal{V}}$, and the index information is stored in a dictionary called $Index$ (key: frame id, value: the location of the record corresponding to frame $f$ in file $IDF^{\mathcal{V}}$) in line 15. $H_f$ is then used for the generation of all models, and lines 8-15 are common across all model types. Once all the $\mathcal{F}$ frames are processed, the index information stored in $Index$ is inserted into the $IDF^{\mathcal{V}}$ header.

\noindent \underline{\textit{Model SGF Generation}} is shown in lines 16-20 in Algorithm~\ref{alg:Graph-gen}. This algorithm computes the number of nodes and edges from $H_f$ and writes them to the graph file, along with the node-label information stored in $H_f$ (line 17). If the given $edgeType$ is distance, then it computes the edge label using Euclidean distance between bounding box centroids (unit is pixels) for each pair of object id instances in $H_f$, and writes out the edge label information in the graph file. The algorithm only handles $distance$ for $edgeType$ currently.

\begin{figure}
    \centering
    \includegraphics[width=1\linewidth,keepaspectratio=true]{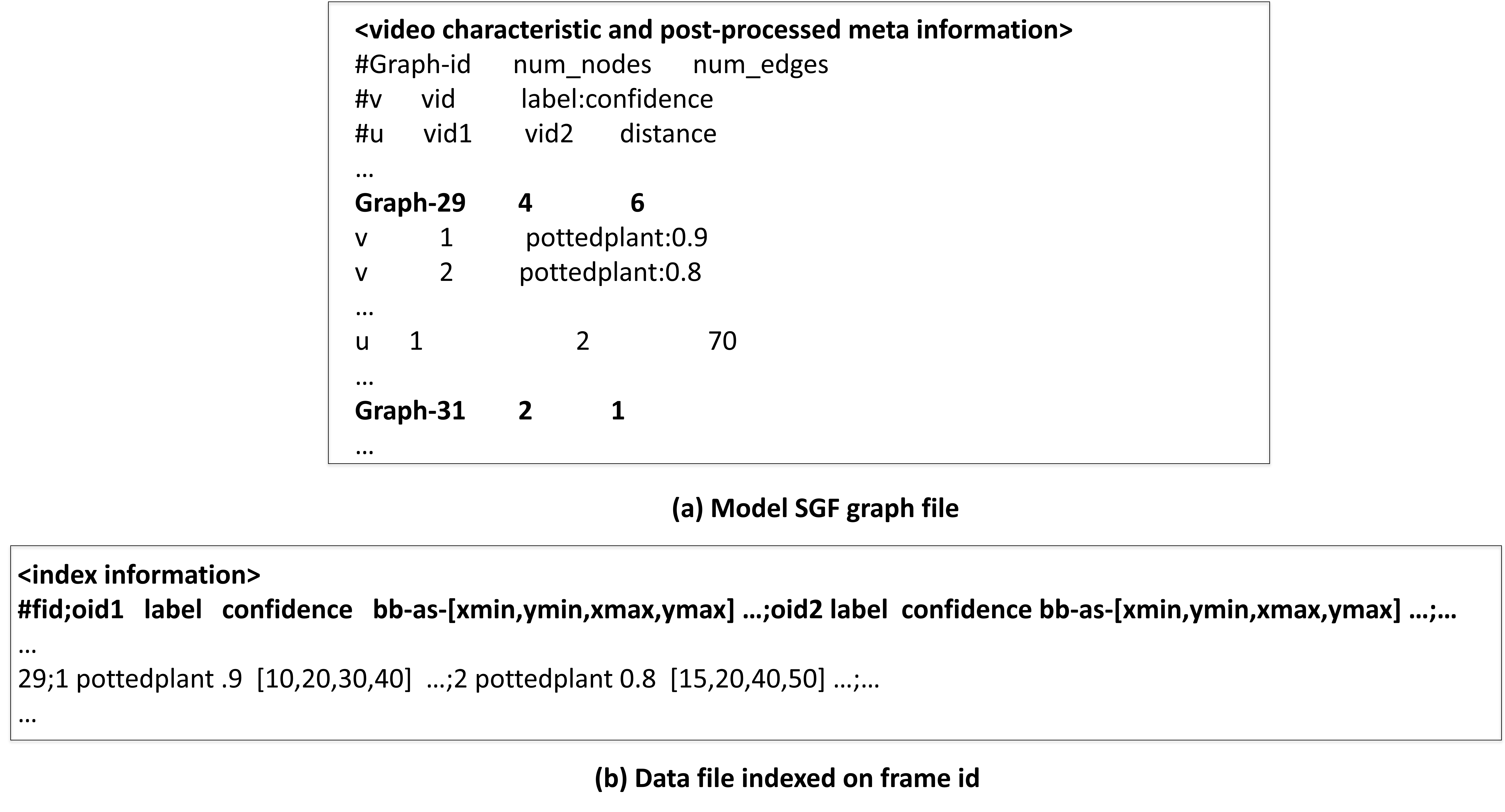}
    \caption{File representation of model SGF graphs and corresponding data file indexed on frame id}
    \label{fig:graph-file-format-SGF}
\end{figure}

The base graph file $BGF^{\mathcal{V}}$ contains a line for each node and edge in a graph, as well as a line for storing the graph id, the number of nodes, and the number of edges.
An example of the \textbf{output graph file} (tab separated) generated by Algorithm~\ref{alg:Graph-gen} for model SGF is shown in Figure~\ref{fig:graph-file-format-SGF}(a). The header stores video characteristics and metadata extracted from the pre-processed file (line 3 of Algorithm~\ref{alg:Graph-gen}). Here, the node lines are prefixed with ``v'' and edge lines are prefixed with ``u'' (for undirected edges). 
A vertex line lists the vertex id, class label, and confidence score, where the class label and confidence score are separated using the delimiter ``:''.  


\begin{algorithm}[H]
\scriptsize
\caption{\small Generate Graph Model}
\label{alg:Graph-gen}
\begin{algorithmic}[1]
\renewcommand{\algorithmicrequire}{\textbf{Input:}}
\renewcommand{\algorithmicensure}{\textbf{Output:}}
\Require $RDF^{\mathcal{V}}$, $model$, $edgeType$, $th_{track}$, \underline{$balanceBy$}, \underline{$minGraphSize$} 
\Ensure A graph file/$N$ graph files, A data file indexed on frame id $IDF^{\mathcal{V}}$
    \State $SGV^{\mathcal{V}} \leftarrow \emptyset, MGV^{\mathcal{V}} \leftarrow \emptyset, C_{MGV^{\mathcal{V}}} \leftarrow [], f \leftarrow 1, N \leftarrow 1, Index \leftarrow \emptyset, \delta_{MGV^{\mathcal{V}}} \leftarrow []$ 
    \If{$model =~''MGV''$}
        \State Compute $maxGraph$ and $\mu$ (avg. no of nodes or frames per graph) based on $balanceBy$ 
    \EndIf
    \State Extract the meta information from $DF^\mathcal{V}$ header and write to graph file header
    \State $line \leftarrow$ Read a line from $DF^\mathcal{V}$
    \While{$line \neq EOF$} 
        \State $H_f \leftarrow \emptyset$
        \State $f \leftarrow $ Extract frame id $f$ from $line$
        \While{$line$ contains frame $f$ information} \Comment{Collect all the rows corresponding to frame $f$}
            \State Extract the $A_{o_i}^f$ attributes of object id instance $o_i^f$ and append them in the data file
            \State $H_f[o_i^f] \leftarrow $ $<BB_{o_i}^f$ centroid, $ CL_{oi}^f, CLC_{oi}^f>$
            \State $line \leftarrow$ Read a line from $RDF^{\mathcal{V}}$
        \EndWhile
        \State Create index for frame $f$ for $IDF^{\mathcal{V}}$ and store indexes into $Index$
        \If {\textbf{$model = ~''SGF''$} } 
            \State Write the no. of nodes, no. of edges computed from $H_f$, and node labels from $H_f$ in the graph file
            \If {$edgeType =~''distance''$ }
                \State For each object id pair in $H_f$, compute the bounding box centroid distances and write in the graph file
            \EndIf
        \ElsIf{$model = ~''SGV''$} 
            \State $SGV^{\mathcal{V}} \leftarrow$ UpdateNodesAndEdges($SGV^{\mathcal{V}}, f, H_f, edgeType$)
            \If{$th_{track}$ number of frames has been processed}
                \State Write the nodes and edges in $SGV^{\mathcal{V}}$ in the graph file
                \State $SGV^{\mathcal{V}} \leftarrow \emptyset$
            \EndIf
        \ElsIf{$model = ~''MGV''$}
            \State $MGV^{\mathcal{V}} , C_{MGV^{\mathcal{V}}}, N, \delta_{MGV^{\mathcal{V}}} \leftarrow$ CreateMGV($MGV^{\mathcal{V}},  C_{MGV^{\mathcal{V}}}, f, H_f, N, \mu,  \delta_{MGV^{\mathcal{V}}} , edgeType, balanceBy, maxGraph$)
            \If{$th_{track}$ number of frames has been processed}
                \State Write all the $N$ graphs in $MGV^{\mathcal{V}}$ in $N$ $BGF^{\mathcal{V}}$ files.
            \EndIf
        \EndIf
\EndWhile

\State Write the remaining graph(s) in $MGV^{\mathcal{V}}$ or the graph $SGV^{\mathcal{V}}$ and insert graph characteristics in the header of $BGF^{\mathcal{V}}$ files
\State Insert the $Index$ in the header of $IDF^{\mathcal{V}}$
\end{algorithmic}
\end{algorithm}

For this model, there can be $\sum_{f \in \mathcal{F^V}} (|O_f|+\frac{|O_f|^2}{2})+\mathcal{F}+h$ rows in the graph file, where $h$ is the number of header lines, $\sum_{f \in \mathcal{F^V}} (|O_f|+\frac{|O_f|^2}{2})$ is number of lines for edges and vertices, and $\mathcal{F}$ is the number of lines to store graph id, the number of nodes, and the number of edges. 

An example $IDF^{\mathcal{V}}$ file generated by Algorithm~\ref{alg:Graph-gen} is shown in  Figure~\ref{fig:graph-file-format-SGF}(b). It contains a row for each frame id, along with all corresponding object id's and the extracted attributes for each object id within a frame. The elements of a frame (object ids and their extracted attributes) are separated using the delimiter ";". The attributes of each object id are separated using a tab. This file has a total of $\mathcal{F}$ lines. Any analysis performed on SGF requires both the graph and $IDF^{\mathcal{V}}$ file.

\noindent \underline{\textit{Model SGV Generation}} is shown in lines 21-26 in Algorithm~\ref{alg:Graph-gen}. Since this model generates only one graph, $SGV^{\mathcal{V}}$ is used for adding nodes and edges. Algorithm~\ref{alg:Graph-gen} either creates new nodes and edges or updates the nodes and edge labels of existing nodes and edges to graph $SGV^{\mathcal{V}}$ using the \textbf{function UpdateNodesAndEdges called in line 22}. The UpdateNodesAndEdges function is also used for model MGV creation later and shown in Algorithm~\ref{alg:update-node-edges}. 

UpdateNodesAndEdges function takes as input a graph $G$, frame id $f$, a dictionary $H_f$, and $edgeType$. It iterates over each object id instance $o_i^f$ stored in $H_f$ (as key), retrieves the class label $CL_{o_i}^f$, and class label confidence score $CLC_{o_i}^f$ from $H_f$ for $o_i^f$. It also computes the number of objects $n$ present in $f$ (which is the number of keys stored in $H_f$). It adds a node in graph $G$ if \textit{there is no existing node} for object id $o_i$ and adds $<CL_{o_i}^f, CLC_{o_i}^f>$, and $<f,n>$ as node labels for object id $o_i$ (line 6). 

If the input graph \textit{$G$ has a node for object $o_i$}, then it inserts frame id $f$, the number of objects in $f$ in \textbf{the descending order of the number of objects} in the node label (line 8). For this, a priority queue is used. This information will be written to the graph file later. It also updates the existing class label confidence score by taking the maximum of the existing confidence score stored in node label for a class and $CLC_{o_i}^f$ or adds new $CL_{o_i}^f$, and $CLC_{o_i}^f$ as label if $CL_{o_i}^f$ is encountered first for $o_i^f$ in frame $f$ (line 9).


\begin{algorithm}[!ht]
\scriptsize
\caption{\small Update Nodes and Edge Labels of Model SGV/MGV}
\label{alg:update-node-edges}
\begin{algorithmic}[1]
\renewcommand{\algorithmicrequire}{\textbf{Input:}}
\renewcommand{\algorithmicensure}{\textbf{Output:}}
\Require A graph $G$, frame id $f$, Dictionary with frame $f$ information $H_f$, $edgeType$
\Ensure Updated graph $G$ with nodes and edge labels
\Function{UpdateNodesAndEdges}{$G, f, H_f, edgeType$}
    \For{each object id instance $o_i^f$ in $H_f$}

        \State Retrieve $CL_{o_i}^f$, and $CLC_{o_i}^f$ from $H_f$
        \State $n \leftarrow $ Compute no. of objects in frame $f$ from $H_f$
        \If  {$G$ do not have a node with object id $o_i$} \Comment{Create new node}
            \State Add a node for object id $o_i$ in G and store $<CL_{o_i}^f,CLC_{o_i}^f>$ and $<f,n>$ as node labels
        \Else \Comment{Update node labels}
            \State Add $<f,n>$ in \textbf{descending order of number of objects} in node label to the sequence of frames in node labels
            \State Update with max of \text{existing confidence score} and $CL_{o_i}^f$/add new $CL_{o_i}^f$, and $CLC_{o_i}^f$ in node label
        \EndIf    

        \If{$edgeType =~''min~ and~ max~distance~''$}
            \For{each object id instance $o_k^f$ in $H_f$}
                \State $d \leftarrow$ Compute Euclidean distance between $o_i^f$, and $o_k^f$'s bounding box centroids stored in $H_f$
                
                \If{$G$ have an edge for $(o_i,o_k)$}  \Comment{Update existing edge label}
                    \State Update edge labels with new minimum/maximum distance and corresponding frame id sequences
                \Else  \Comment{Create a new edge}
                    \State Add an edge for $(o_i, o_k)$ with label $<d,f>,<d,f>$
                \EndIf
            \EndFor
        \EndIf
    \EndFor
\State \textbf{Return} $G$
\EndFunction
\end{algorithmic}
\end{algorithm}

For generating/updating edges labels Algorithm~\ref {alg:update-node-edges}, first computes the Euclidean distance $d$ between bounding box centroids (stored in $H_f$) of a pair of object id instances $o_i^f, o_k^f$ in line 13. If there is an already existing edge for object pair $(o_i,o_k)$ it extracts the edge labels $minimum~distance, <f_i,..., f_j>$ and $maximum~distance, <f_k,...,f_l>$, adds frame id $f$ to the edge label if the minimum/maximum distance is the same as $d$. If $d$ is less than $minimum~distance$ in the current edge label, then the new edge label contains $d, <f>$ and $maximum~distance, <f_k,...,f_l>$, where $d$ becomes the new minimum. Similarly, if $d >$ maximum distance, then the new edge label contains $minimum~distance, <f_i,...,f_j>$ and $d, <f>$, where $d$ becomes new maximum. If there are no edges for $(o_i,o_k)$ in graph $G$, then it adds an edge with label $d,<f>,d,<f>$, where $d$ is both the minimum and maximum distance. After updating the nodes and edges, it returns the updated graph $G$ to Algorithm~\ref{alg:Graph-gen}. 

As mentioned earlier, for this model, the nodes and edges in $SGV^{\mathcal{V}}$ need to be stored in main memory until $th_{track}$ number of frames, as objects will be assigned a different object id if they reappear after $th_{track}$ frames. After $th_{track}$ frames, the node labels and edge labels will not be updated for objects that appeared before $th_{track}$ frames. Hence, Algorithm~\ref{alg:Graph-gen} writes the nodes and edges of graph $SGV^{\mathcal{V}}$ in a graph file $BGF^{\mathcal{V}}$ after processing $th_{track}$ frames. Apart from the above, the algorithm calculates the \textbf{graph characteristic} information (total number of connected components, largest connected component size, smallest connected component size, number of singleton nodes, maximum and smallest number of frames in a component, maximum, minimum, and average degree of the graph) when new nodes and edges are added to graph $SGV^{\mathcal{V}}$ (not shown in the algorithm to keep it simple), which can be useful for analysis later. They are inserted into the graph file $BGF^{\mathcal{V}}$ header at the end of processing all the $\mathcal{F}$ frames. 

\begin{figure}
    \centering
    \includegraphics[width=0.6\linewidth,keepaspectratio=true]{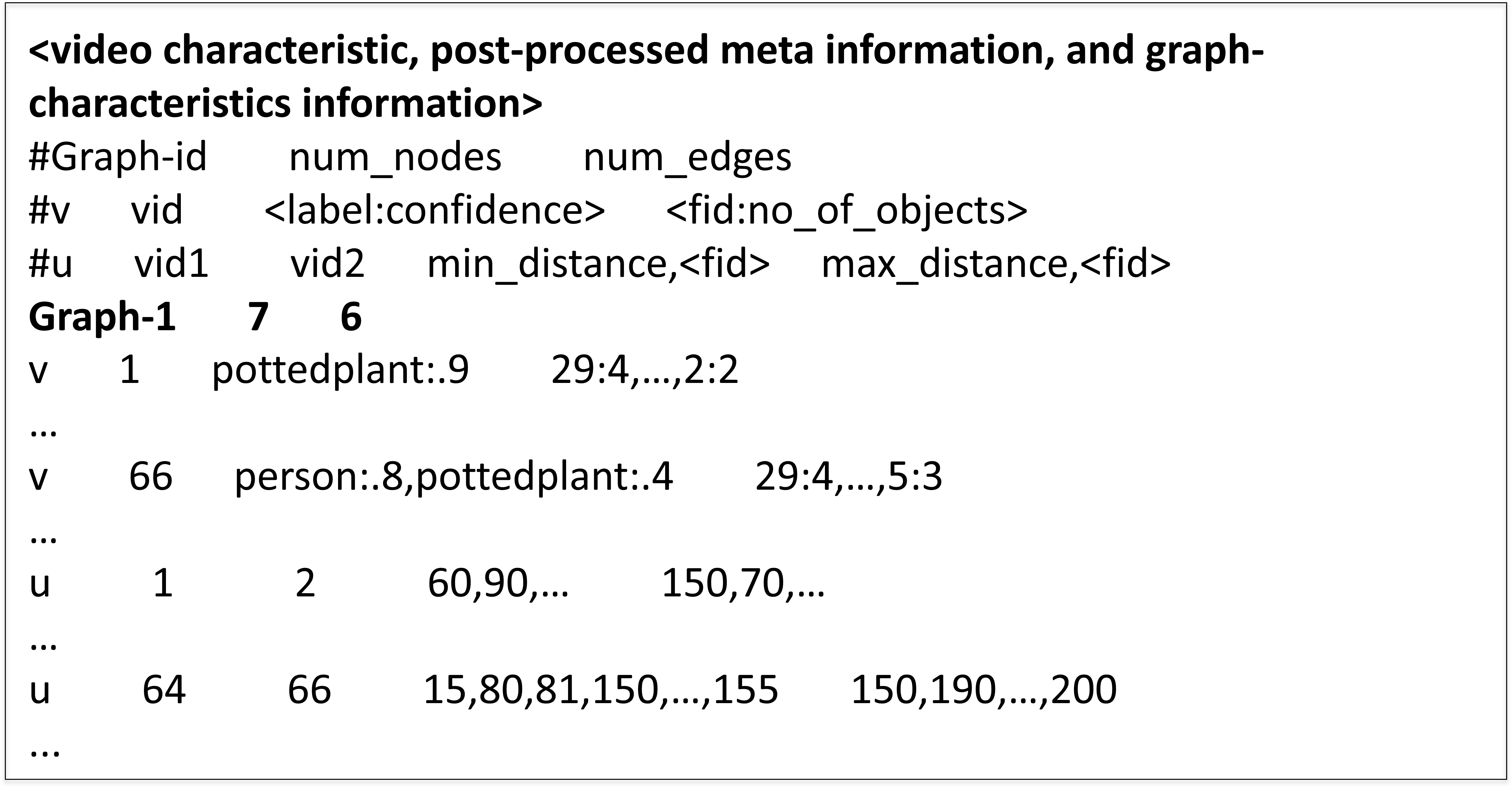}
    \caption{File representation of model SGV}
    \label{fig:graph-file-format-SGV}
\end{figure}

An example of an \textbf{base graph file} generated by Algorithm~\ref{alg:Graph-gen} for model $SGV$ is shown in Figure~\ref{fig:graph-file-format-SGV}. This file contains a line for each vertex (prefixed with 'v') and edge pair (prefixed with 'u'). It is a tab-separated file. The class label and confidence score are separated by the delimiter ":" in the vertex label. Similarly, the frame id and the number of objects in that frame id are separated by the delimiter ":". In this Figure, a single graph is generated with seven nodes and six edges. Vertex 66 is assigned two different labels (person and potted plant) at two different time points (or sequences of non-empty frames) in the video. For the person label, the highest confidence score is $0.8$ across all frames. Similarly, for the label potted plant, the highest confidence score is $.8$ across all the frames. Vertex 66 appeared in frame 29 (having 4 objects),..., frame 5 (having 3 objects). There are other frames in between where vertex 66 appeared, and they are stored in descending order of the number of objects. In this representation, multiple edge labels are separated by tabs. It separates each element of an edge label using ",". For example, in Figure~\ref{fig:graph-file-format-SGV}, there is an edge between object id 64 and 66. The first edge label represents their minimum distance, and frames with minimum distance are $80,81,150,\ldots,155$. Similarly, the second label represents that this pair of objects had a maximum distance of 150, and frames with maximum distance are $190,\ldots,200$. The indexed data file generated is the same as shown in Figure~\ref{fig:graph-file-format-SGF}(b).

In total, the graph file for this model can contain $UO^{\mathcal{V}}+\frac{(UO^{\mathcal{V}})^2}{2}+h$ lines, where $h$ is the number of header lines, $UO^{\mathcal{V}}$ and $\frac{(UO^{\mathcal{V}})^2}{2}$ are total number of lines for vertices and edges.

\noindent\underline{\textit{Model MGV Generation}} is shown in lines 27-32 in Algorithm~\ref{alg:Graph-gen}. Algorithm~\ref{alg:Graph-gen} calls the \textbf{function Create\_MGV} in line 28 for generating $MGV^{\mathcal{V}}$set of graphs for model MGV.

While generating this model, the algorithm allows each graph $MGV_i^{\mathcal{V}} \in MGV^{\mathcal{V}}$ to deviate slightly from $\mu$ (average no. of frames/nodes per graph) using the parameter $\delta_{MGV_i^{\mathcal{V}}}$, which is the smallest connected component size excluding components of size 1 (when $balanceBy =~''node''$)  or the component with the smallest number of frames (when $balanceBy =~''frame''$). Each graph can have the number of nodes or frames within the range $\mu-\delta_{MGV_i^{\mathcal{V}}}$ and $\mu+\delta_{MGV_i^{\mathcal{V}}}$ as described above. In this algorithm, $N$ is used to keep track of the total number of graphs created for model MGV, where $N \leq maxGraph$. Here $MGV_N^{\mathcal{V}}$ denotes the $N^{th}$ (or last) graph created for model MGV. All the $\delta$ values of $N$ graphs are stored in a list called $\delta_{MGV^{\mathcal{V}}}$ (see Table~\ref{tab:notations-graph-gen} for details), maintaining one-to-one correspondence with the set of graphs in $MGV^{\mathcal{V}}$.

As mentioned above, for creating MGV, the components of all the existing graphs having common object id's (or node id's) with $O_f$ need to be identified while processing a frame. For this, we need to know which object id's or nodes are part of a connected component in a graph, and we keep track of the node id's belonging to a component for each graph separately. For this, we keep track of the set of object id's forming a connected component in a graph $MGV_i^{\mathcal{V}}$. Here, $C_{MGV_i^{\mathcal{V}}}$ is a collection of sets where each item is the set of node id's forming a connected component. For example, $C_{MGV_i^{\mathcal{V}}}$ can contain $\{\{o_1,o_2\},\{o_3,o_4,o_5\}\}$, where $\{o_1,o_2\}$, and $\{o_3,o_4,o_5\}$ are part of two connected components in graph ${MGV_i^{\mathcal{V}}}$. This set is updated whenever nodes are added to a graph. 
In this algorithm, $C_{MGV^{\mathcal{V}}}$ is a list that contains all the sets of components across graphs. In other words, $C_{MGV^{\mathcal{V}}}$ is a list containing $C_{MGV_1^{\mathcal{V}}},\ldots, C_{MGV_N^{\mathcal{V}}}$. This list is used to move components between graphs. Note that we do not need to actually merge components in a graph; we just need to find out which graphs we need to add an $O_f$ set of nodes. Once the edges are drawn, the connected component of each graph will be automatically built. The reason for keeping track of object id's belonging to the connected components in each graph using $C_{MGV_i^{\mathcal{V}}}$ is to avoid performing a connected component analysis for each frame $f$. Finally, \textit{we will use the terminology connected component when referring to $C_{MGV_i^{\mathcal{V}}}$ even though it contains set of object id's for simplicity.}

\begin{algorithm}[H]
\scriptsize
\caption{\small Generating Model MGV}
\label{alg:create-mgv}
\begin{algorithmic}[1]
\renewcommand{\algorithmicrequire}{\textbf{Input:}}
\renewcommand{\algorithmicensure}{\textbf{Output:}}
\Require $MGV^{\mathcal{V}}, C_{MGV^{\mathcal{V}}},H_f$, $N$, $\mu$, $\delta_{MGV^{\mathcal{V}}}$, $edgeType, balanceBy, maxGraph$

\Ensure $MGV^{\mathcal{V}}, C_{MGV^{\mathcal{V}}}, N, \delta_{MGV^{\mathcal{V}}}$ 
\Function{CreateMGV}{$MGV^{\mathcal{V}}, f, H_f, N, \mu, \delta_{MGV^{\mathcal{V}}}, edgeType, balanceBy, maxGraph$}
    \State $k \leftarrow 0, createNewGraph \leftarrow False, intersectingComp\leftarrow \emptyset $
    \State $O_f \leftarrow $ Extract set of object id's in frame $f$ from the keys in $H_f$
    \For{each $C_{MGV_i^{\mathcal{V}}}$ in $ C_{MGV^{\mathcal{V}}}$}  \Comment{\textbf{Case (i): merge components and find intersecting components}}
        \State $C \leftarrow $ A set with object id's in different components of $C_{MGV_i^{\mathcal{V}}}$ having common object id's with $O_f$ 
        
        \If{$C \neq \emptyset$ } 
            \State $intersectingComp[i] \leftarrow C$   
            \State Update $C_{MGV_i^{\mathcal{V}}}$ by merging all the components having object id in $C$
            \State $k \leftarrow \arg\max_i |intersectingComp[i]|$ \Comment{Graph id with largest intersecting components} 
        \EndIf 
        
    \EndFor
    \If{$intersectingComp\neq \emptyset $} \Comment{\textbf{Case (ii): shift components across graphs}} 
        \State Extract $MGV_k^{\mathcal{V}}$ from $MGV^{\mathcal{V}}$\Comment{Graph with the largest connected component intersecting with $O_f$}
        \For {each key $i$ in $intersectingComp$ where $i \neq k$} 
            \State Extract graph $MGV_i^{\mathcal{V}}$ from $MGV^{\mathcal{V}}$
            \State $C \leftarrow  intersectingComp[i]$ \Comment{Component of graph $MGV_i^{\mathcal{V}}$ that intersects with $O_f$}
            \State Move the component $C$ from graph $MGV_i^{\mathcal{V}}$ to $MGV_k^{\mathcal{V}}$
            \State Remove $C$ from $C_{MGV_i^{\mathcal{V}}}$; Update $\delta_{MGV_i^{\mathcal{V}}}$ 
        \EndFor    
        \State $MGV_k^{\mathcal{V}} \leftarrow$ UpdateNodesAndEdges($MGV_k^{\mathcal{V}}, f, H_f, edgeType$) 
        
        \State Update $C_{MGV_k^{\mathcal{V}}}$ with union of all the shifted components and $O_f$ and $\delta_{MGV_k^{\mathcal{V}}}$
        \State \textbf{Return} $MGV^{\mathcal{V}}, C_{MGV^{\mathcal{V}}}, N, \delta_{MGV^{\mathcal{V}}}$ 
    \Else \Comment{Case (iii): \textbf{New component} with $O_f$ set of nodes created}
        \State Extract $MGV_N^{\mathcal{V}}$ from $MGV^{\mathcal{V}}$ and $\delta_{MGV_N^{\mathcal{V}}}$ from the list $\delta_{MGV^{\mathcal{V}}}$
       \If {$(N < maxGrph)$ and ($balanceBy =~''node''$ and no. of nodes in $MGV_N^{\mathcal{V}}+|O_{f}|$$ > \mu +\delta_{MGV_N^{\mathcal{V}}}$)}
            \State $createNewGraph \leftarrow True$
       \ElsIf{$(N < maxGrph)$ and ($balanceBy =~''frame''$ and  no. of frames in $MGV_N^{\mathcal{V}} > \mu +\delta_{MGV_N^{\mathcal{V}}}$) }
        \State $createNewGraph \leftarrow True$
       \EndIf

         \If {$createNewGraph = True$}
            \State $N \leftarrow N+1, MGV_N^{\mathcal{V}} \leftarrow \emptyset$  
            \State $MGV_N^{\mathcal{V}} \leftarrow $ UpdateNodesAndEdges($MGV_N^{\mathcal{V}}, f, H_f, edgeType$) \Comment{A new graph is created. }
            \State  $C_{MGV_N^{\mathcal{V}}} \leftarrow \{O_f\}$; Add graph $MGV_N^{\mathcal{V}}$ to $MGV^{\mathcal{V}}$ and add $C_{MGV_N^{\mathcal{V}}}$ to $C_{MGV^{\mathcal{V}}}$ 
        \Else
            \State $MGV_N^{\mathcal{V}} \leftarrow $ UpdateNodesAndEdges($MGV_N^{\mathcal{V}},  f, H_f, edgeType$) 
            
            \State $C_{MGV_N^{\mathcal{V}}} \leftarrow  C_{MGV_N^{\mathcal{V}}} \bigcup~\{O_f\}$
         \EndIf
        \State Update $\delta_{MGV_N^{\mathcal{V}}}$ in list $\delta_{MGV^{\mathcal{V}}}$ 
    \EndIf
\State \textbf{Return} $MGV^{\mathcal{V}}, C_{MGV^{\mathcal{V}}},N, \delta_{MGV^{\mathcal{V}}}$    
\EndFunction
\end{algorithmic}
\end{algorithm}

The \underline{\textit{CreateMGV function}} shown in Algorithm~\ref{alg:create-mgv}, takes as input the set of graph $MGV^{\mathcal{V}}$created until frame $f$ is processed, list of connected components of all graphs $C_{MGV^{\mathcal{V}}}$,$f,~H_f$, $N$, $\mu$, $\delta_{MGV^{\mathcal{V}}}$, $edgeType, balanceBy,$ and $maxGraph$ as parameter.
Algorithm~\ref{alg:create-mgv} handles case (i), where different components of a graph can have common nodes with $O_f$ by iterating over the list $C_{MGV^{\mathcal{V}}}$ (lines 4-11). For each set of components $C_{MGV_i^{\mathcal{V}}}$ for graph ${MGV_i^{\mathcal{V}}}$, it finds out the set of object id's from different components that have a common node id or object id with $O_f$ and stores them in $C$ in line 5. It also builds a dictionary $intersectingComp$ while iterating over the elements of $C_{MGV^{\mathcal{V}}}$ (line 7). The key to this dictionary is the graph id, and the value is the connected component $C$ that has common object id's or node id's with $O_f$. This dictionary holds for each graph which components have overlapping object id with $O_f$. It is later used for shifting components from one graph to another. It also determines the graph id $k$ that has the largest component having a common node id with $O_f$ (line 9). This will also be used later to shift components between graphs. We also update $C_{MGV_i^{\mathcal{V}}}$ with $C$ (represents a merged component).

In lines 12-22, Algorithm~\ref{alg:create-mgv} handles case (ii) for creating MGV, where multiple connected components across graphs in $MGV^{\mathcal{V}}$can have common node id's (or object id's) with $O_f$. Together, all of the intersecting components will form a larger connected component. By placing them in a single graph in this algorithm, we ensure that \textit{the connected components are not broken across multiple graphs}. To achieve this, we utilize the dictionary $intersectingComp$ constructed earlier in lines 4-11. Each intersecting component from a graph $MGV_i^{\mathcal{V}}$ is shifted to graph $MGV_k^{\mathcal{V}}$ (contains the largest component intersecting with $O_f$) in lines 14-19. If we choose $MGV_k^{\mathcal{V}}$ in such a manner, we will \textit{have to shift the minimal number of nodes and edges in total}. We also remove the component that has been shifted from $C_{MGV_i^{\mathcal{V}}}$ in line 18.
After all the components are shifted across graphs, $MGV_k^{\mathcal{V}}$ is updated with new nodes and edges from $O_f$ (by calling the function UpdateNodesAndEdges in line 20). This step makes all the shifted components, along with $O_f$, a large connected component, because of drawing edges from the new object id's present in $O_f$. It also updates the set $C_{MGV_k^{\mathcal{V}}}$ with the union of all the object id's in the shifted components and $O_f$, and recomputes $\delta_{MGV_i^{\mathcal{V}}}$ in line 21.

In lines 23-39, Algorithm~\ref{alg:create-mgv} handles case (iii), where $O_f$ \textit{does not have any object id's common with any of the graphs created till now}. In this case, the dictionary $intersectingComp$ will be empty. Now, for this, the algorithm checks if the nodes generated from $O_f$ can be accommodated in the last graph created (graph $MGV_N^{\mathcal{V}}$). For this, it uses the $balanceBy$ parameter, $maxGraph$, and $\delta_{MGV_N^{\mathcal{V}}}$. If $N < maxGraph$ and $balanceBy =~''node''$ (line 24), then the algorithm checks if adding all the $|O_f|$ nodes would cause the total number of nodes in $MGV_N^{\mathcal{V}}$ (last created graph) to exceed $\mu+\delta_{MGV_N^{^\mathcal{V}}}$. If $N < maxGraph$ and $balanceBy =~''frame''$ (line 26), then the algorithm checks if adding all the $|O_f|$ nodes from frame $f$ would cause the total number of frames in $MGV_N^{\mathcal{V}}$ to exceed $\mu+\delta_{MGV_N^{^\mathcal{V}}}$(line 65). For the above two conditions, it sets the value of $createNewGraph$ to True, indicating that a new graph should be created when either of the above conditions is satisfied. 

If $createNewGraph$ is true, the algorithm increments the value of $N$, creates a new graph $MGV_{N}^{\mathcal{V}}$ with $|O_f|$ nodes using the function UpdateNodesAndEdges in line 31. It also initializes the set $C_{MGV_N^{\mathcal{V}}}$ with $O_f$, which means there is only one component with $|O_f|$ nodes in the graph $MGV_{N}^{\mathcal{V}}$ and adds it to the list $C_{MGV^{\mathcal{V}}}$ (line 31). If $createNewGraph$ is False, meaning $|O_f|$ nodes can be added to the last created graph $MGV_N^{\mathcal{V}}$, and it adds $|O_f|$ nodes by calling the function UpdateNodesAndEdges in line 35. It also adds the set $O_f$ to $C_{MGV_N^{\mathcal{V}}}$. \textit{Note, whether we create a new graph/add nodes to the last graph, a new component with $|O_f|$ nodes will always be created for case (iii)}.

The CreateMGV function returns $MGV^{\mathcal{V}}, C_{MGV^{\mathcal{V}}}, N, \delta_{MGV^{\mathcal{V}}}$ to Algorithm~\ref{alg:Graph-gen}. In Algorithm~\ref{alg:Graph-gen}, once $th_{track}$ number of frames are processed for model MGV (line 27 in Algorithm~\ref{alg:Graph-gen}), it writes all the graphs into $N$ different base graph files and keeps only the number of frames and set of node id's the graph was generated from, which is useful for making decisions about creating new graphs while processing subsequent frames (not shown in the algorithm to keep it simple). 


For this model, we generate $N$ $BGF^{\mathcal{V}}$ graph files with the same representation as model SGV. In each graph file, the graph characteristics information for that graph is stored as a header. It also stores in the header the maximum and minimum number of objects in a frame and the frame id's with the maximum and minimum number of objects among all the $F_i$ frames, from which a graph $MGV_i^{\mathcal{V}}$ is generated. Some of these have been used for analysis later in Chapter~\ref{chap:graph-ana-videocontents}.

\subsection{Complexity Analysis of Algorithm~\ref{alg:Graph-gen}}

\noindent \underline{\textit{I/O Complexity:}} Algorithm~\ref{alg:Graph-gen} takes $\mathcal{O}(OI^{\mathcal{V}}+h)$ I/O time to read from file $RDF^{\mathcal{V}}$, which is $\mathcal{O}(OI^{\mathcal{V}})$ in simplified form. It takes $\mathcal{O}(F)$ I/O time to write $IDF^{\mathcal{V}}$ file for generating any model. Here $OI^{\mathcal{V}}$ is the number of object id instances in video $\mathcal{V}$ or the number of data lines in $RDF^{\mathcal{V}}$, and $h$ is the number of header lines in $RDF^{\mathcal{V}}$. This step is common to all models.

To \textit{generate model SGF}, the algorithm takes an additional $\sum_{f \in \mathcal{F^V}} (|O_f|+\frac{|O_f|^2}{2})+\mathcal{F}$ I/O's to write in the graph file. The I/O complexity for generating all the $\mathcal{F}$ graphs and the indexed data file for this model is $\mathcal{O}(OI^{\mathcal{V}}+ \sum_{f \in \mathcal{F^V}} (|O_f|+\frac{|O_f|^2}{2})+2\mathcal{F})$. For this model, on average, \textit{each graph has} $\frac{OI^{\mathcal{V}}}{\mathcal{F}}$ nodes, and $\frac{(OI^{\mathcal{V}})^2}{2\mathcal{F}^2}$ edges. Therefore, the overall I/O complexity for generating all $\mathcal{F}$ graphs can be simplified to $\mathcal{O}(OI^{\mathcal{V}}+\frac{(OI^{\mathcal{V}})^2}{2\mathcal{F}}+\mathcal{F})$, which is $\mathcal{O}(OI^{\mathcal{V}}+(OI^{\mathcal{V}})^2+\mathcal{F})$ in simplified form.


For \textit{model SGV}, the algorithm writes $UO^{\mathcal{V}}$ nodes and $\frac{UO^{\mathcal{V}^2}}{2}$ edges (a complete graph will be generated in the worst-case scenario) to the graph file. 
The I/O complexity for generating both the graph and indexed data file for this model is $\mathcal{O}(OI^{\mathcal{V}}+UO^{\mathcal{V}}+\frac{UO^{\mathcal{V}^2}}{2}+F)$, which simplifies to $\mathcal{O}(OI^{\mathcal{V}}+(UO^{\mathcal{V}})^2+F)$. The I/O time for models SGV and MGV is the same. 


If the I/O times of models SGF and SGV/MGV are compared, what makes a difference is the I/O time for writing edges, as the file read time and indexed data file generation time are the same for both. Model SGV/MGV will have very few edges than model SGF ($(UO^{\mathcal{V}})^2 <<(OI^{\mathcal{V}})^2$), and hence SGV generation will take less I/O time than SGF generation.

\noindent \underline{\textit{CPU Time Complexity:}} For model SGF, the main computation in a frame $f$ is computing edge labels (bounding box centroid distances) between all object pairs in $O_f$, which takes $\mathcal{O} (|O_f|^2)$ time. The CPU time complexity for processing all the $\mathcal{F}$ frames is $\mathcal{O}(\sum_{f \in \mathcal{F^V}}|O_f|^2)$. If each graph has $\frac{OI^{\mathcal{V}}}{\mathcal{F}}$ nodes and $\frac{(OI^{\mathcal{V}})^2}{2\mathcal{F}^2}$ edges on average, then the CPU time complexity for SGF generation will be $\mathcal{O}(\frac{OI^{\mathcal{V}^2}}{\mathcal{F}}) \approx \mathcal{O}(OI^{\mathcal{V}^2})$.

For model SGV generation, frame id's where an object id appeared are added in the node label in descending order of the number of objects (for line 8 in Algorithm~\ref{alg:update-node-edges}). Lets assume, each object id $o_i \in UO^{\mathcal{V}}$ appears in $\frac{\mathcal{F}}{UO^{\mathcal{V}}}$ frames on average. The complexity for generating the node labels in a sorted manner (using a priority queue) for an object id $o_i$ is $\mathcal{O} (\frac{\mathcal{F}}{UO^{\mathcal{V}}} \times \log~\frac{\mathcal{F}}{UO^{\mathcal{V}}})$. CPU time complexity for \textit{node generation} of all the $UO^{\mathcal{V}}$ object id's is $\mathcal{O}(\mathcal{F} \times \log~\frac{\mathcal{F}}{UO^{\mathcal{V}}})$. Another major computation is computing the distance between each pair of objects in frame $f$ (line 13 in Algorithm~\ref{alg:update-node-edges}). This time will be the same as the model $SGF$ edge generation time for all the $\mathcal{F}$ frames, which is $\mathcal{O}(OI^{\mathcal{V}^2})$. The CPU time complexity for model SGV is $\mathcal{O}(OI^{\mathcal{V}^2}+ \mathcal{F} \times \log \frac{\mathcal{F}}{UO^{\mathcal{V}}})$. 


For model MGV, the edge and node generation time is the same as model SGV. Additionally, another major computation for this algorithm comes from performing a \textbf{set intersection in each frame} to determine which components to merge across and within graphs or which components to shift while processing each frame. In the worst case, this will take $\mathcal{O}(UO^{\mathcal{V}})$ for performing set intersection in a frame, when each object id $o_i \in UO^{\mathcal{V}}$ appears in all the $\mathcal{F}$ frames. It will take $\mathcal{O}(\mathcal{F} \times UO^{\mathcal{V}})$ time to perform set intersection in all the $\mathcal{F}$ frames. The CPU time complexity for generating this model in the worst case is $\mathcal{O}(OI^{\mathcal{V}^2}+ \mathcal{F} \times \log \frac{\mathcal{F}}{UO^{\mathcal{V}}}+\mathcal{F} \times UO^{\mathcal{V}})$, which is $\mathcal{O}(OI^{\mathcal{V}^2}+\mathcal{F} \times UO^{\mathcal{V}})$ in simplified form.




The CPU time complexity for model MGV is higher than that of model SGV because of the additional time required for set intersections in each frame. Model SGV CPU complexity is higher than model SGF because of the additional time required to generate node labels in a sorted manner. However, in general, for all the models, the dominant CPU time complexity comes from the distance computation time for each pair of objects in each frame.

\section{R++ Model}
\label{sec:R++}
The VCE module discussed above generates three different categories of information: traditional numerical, multidimensional vectors, and categorical. Some multi-dimensional vectors can have an underlying inherent order (e.g., pose vectors are always generated in an order). The situations addressed in this thesis have certain requirements. One of them is that the extracted contents cannot be processed arbitrarily. The order-preserving property of arrable data model AQuery~\cite{lerner2003aquery} allowed representing ordered arrays. However, the arrays (or vectors) in the arrable data model allow only two-dimensional vectors, whereas the extracted contents can have multidimensional vectors (e.g., pose vectors are two-dimensional, feature vectors can be three-dimensional, etc.) 

Inspired by the concept of the arrable data model, the R++ model supports four different data types. They are basic type (T), vector (V), Sequence ($<S>$), and arrables. The basic type consists of traditional relational model data types. Apart from this, an enumerated data type is incorporated to represent the categorical attributes extracted by VCE. A vector is a collection of numerical, enumerated, or vector elements. All elements of a vector attribute [V] have the same dimension. A sequence $<S>$ is an ordered collection of numerical, categorical, or vector elements. The order can be inherent, or a sequence can be generated by ordering on attributes. Unlike vectors, sequence attribute values can have different dimensions if the value is a vector.

\begin{definition}
    An R++ relation $R$ consists of attributes $A_1, A_2,...,A_n$. An attribute $A_i$ can be of basic type, vector, or sequence. 
\end{definition}
\begin{figure}
\centering
\includegraphics[keepaspectratio=true, width=1\linewidth]{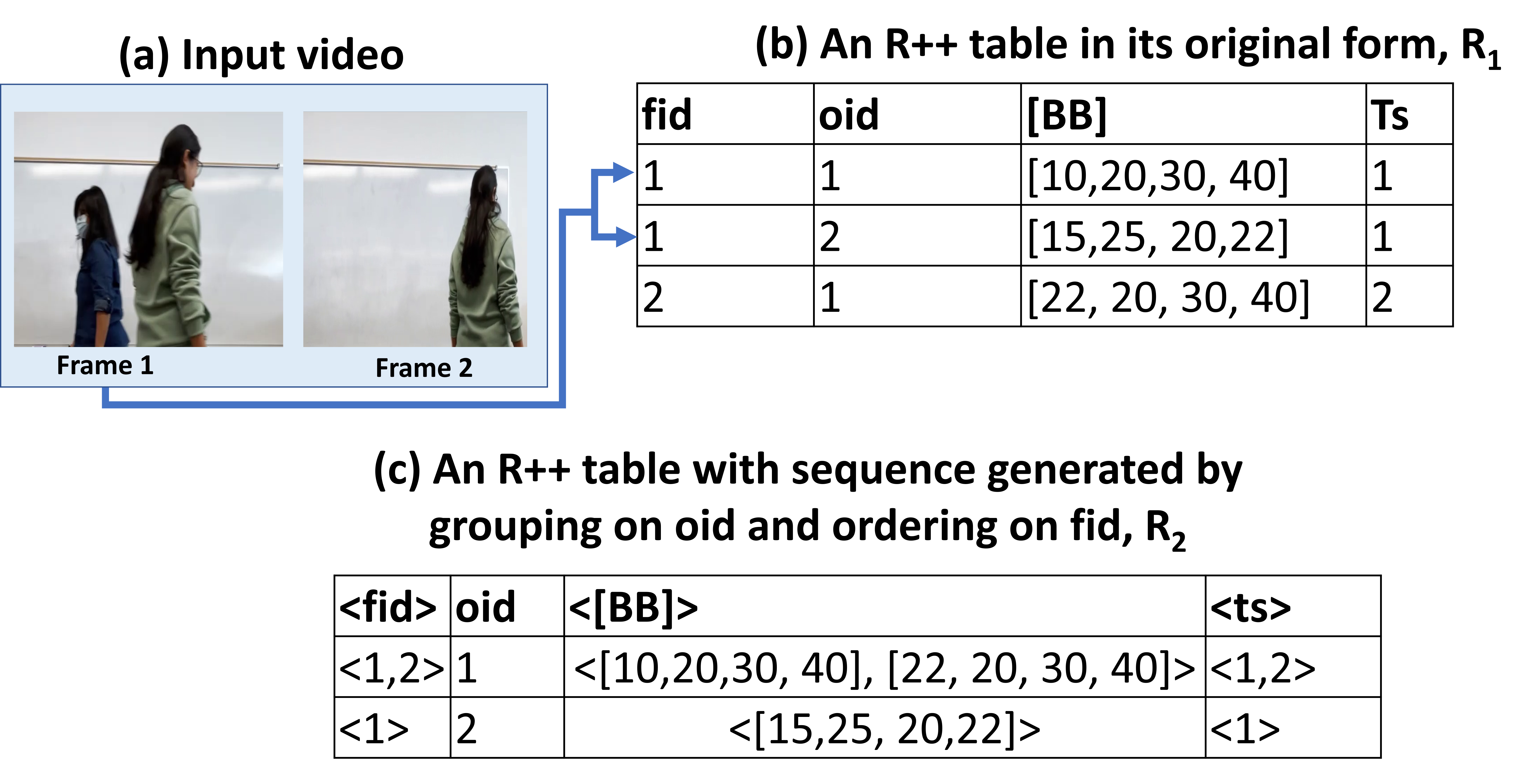}
\caption{\footnotesize{{\textbf{Video Content Representation using R++ model}. (a) An input video with 2 frames captured at 1 frame per second, (b) An R++ table representation of the video. Here $f_{id}$, $o_{id}$ and $ts$ are scalar attribute types and $[BB]$ is vector attribute. (c) An R++ table with Sequence attribute, generated by grouping on $o_{id}$ and ordering on $f_{id}$. }}}
\label{fig:R++Model}
\end{figure}

An example of an R++ table is shown in Fig.~\ref{fig:R++Model}(b).
For ease of explanation, only four different VCE content attributes are shown here. A tuple is generated for each object in a frame in the original R++ table shown in Fig.~\ref{fig:R++Model}(b). The attributes $f_{id}$, $o_{id}$, and $ts$ are numerical types whereas [$BB$] is vector type. The size of the [$BB$] vectors is four here. However, for pose vectors, this size can vary across tuples of a particular column, as the size of pose vectors varies depending upon the object type (or class). In Figure~\ref{fig:R++Model}(b), frame 1 contains two objects with $o_{id}$ 1 and 2. There are two tuples for frame 1 (one for each object). For supporting time-based windows, each frame is associated with an actual timestamp (shown as an integer for convenience). Examples of R++ table with sequence attributes are shown in Fig.~\ref{fig:R++Model}(c). Here, $R_2$ is generated by grouping on oid and ordering on timestamp. All the other attributes except the grouping attribute are converted into sequences here. This allows us to represent all the tuples belonging to a group (in this case oid) using a sequence across the entire relation. This can be important if one is interested in answering queries related to an object. For example, if one wants to compute the direction of an object, one can just use the first and last element of the [$BB$], instead of iterating over all the rows. Similarly, grouping and ordering other attributes is also possible.

\chapter{Situation Detection Using Graph Models}
\label{chap:graph-ana-videocontents}
In this chapter, we demonstrate the applicability of different graph models for detecting situations. 
We have selected two representative situations from Table~\ref{tab:PrimitiveSituations} that we believe the graph model is better suited for. We develop algorithms for their detection and compare the accuracy and performance on different graph models. Algorithms are parameterized as they detect multiple variations of the same situation. For example, the same algorithm is used for detecting group(s) of the largest size, above a size, or within a given range.  The two situations chosen are: i) finding groups/clusters of objects of different sizes based on object type, such as persons, and ii) whether two object types are moving towards/away from each other.

\section{Identifying Group(s) of Objects of Different Sizes}
\label{sec:group-computation}

\noindent \textit{\underline{Definition of group in a video:}} A group in a video is the set of objects or objects of a specific type (or class label $CL$ such as person) in closest proximity (or clusters of objects) formed over a sequence of frames.
An object id or a set of object id's can belong to different groups or clusters at different points in time. The size of a group in a video can range between 1 and $M$, where $M$ is the maximum number of objects in the video. The size of the largest group in a video is $\leq M$. Given the definition of a group in a video, the situation in question can be formulated as follows.   

``\textit{Given a set of graphs $SGF^{\mathcal{V}},MGV^{\mathcal{V}}$ or a graph $SGV^{\mathcal{V}}$ generated from a video $\mathcal{V}$ with $\mathcal{F}$ frames, find out all the groups or clusters of objects of given class label $CL$ and a given size $s$ or above a given size $s$ or within a range $[s, p]$ or the largest groups or clusters, where $1 \leq s \leq p \leq M$}".

In the above formulation, finding the largest cluster is most challenging, as we do not know the actual largest cluster size (the lower bound) beforehand, and the algorithm must determine the largest cluster size. For the other different sizes, $s$ serves as the lower bound and is given. The proposed algorithms and experimental results presented in this section are explained in the context of finding the largest cluster, although they can handle all the cases mentioned above. 

Before explaining the proposed algorithms, we will explain some preliminaries about traditional clustering algorithms, a baseline approach for addressing the situation in question, and its limitations. We will then propose our algorithms for various graph models, which improve upon the baseline approach in terms of performance. The accuracy of the algorithms is computed with respect to the baseline approach. Finally, we will also discuss the experimental results.

\subsection{Clustering Objects using K-means and Elbow }

As mentioned in chapter~\ref {chap:problem-statement}, the objects in \textit{a video frame} can be clustered using a traditional clustering algorithm (e.g., K-means) using their bounding box centroid distances, for a given $K$. $K$ denotes the number of groups (or clusters) in a frame. K-means clustering yields clusters of bounding box centroids and centroid points of $K$ clusters.
Traditionally, K is determined by the Elbow method by plotting the Sum of Squared Error (SSE)~\footnote{SSE of a clustering is calculated by taking the average sum of squared error from each data point (in our case, the bounding box centroid) and the assigned clustered centroid} for different values of K (2 to a chosen maximum, which will be total number of objects in a frame in our case). The value of K at the elbow point in the plot shows that increasing K beyond this point will not significantly reduce the SSE, and therefore, a better cluster cannot be obtained. In other words, the elbow point is \textbf{the optimal number of clusters} or $K$ for that dataset. To apply the Elbow method for our purpose, the Elbow point needs to be determined automatically from the plot. This is typically done by computing the point (or value of K) where the value of the second derivative of the elbow curve is maximum. For this, we need at least three points. In other words, \textit{to determine $K$ using the Elbow method, at least three objects or bounding box centroids are required in a frame.} Hence, in this thesis, for all the algorithms developed, \textbf{frames with two objects are dropped}. 


\subsection{Challenges and Key Observations}

\noindent \underline{\textit{A Baseline Approach:}} The situation in question can be addressed by performing K-means and Elbow on each video frame $f$ (after filtering the vertices with the given label $CL$). After clustering in frame $f$, clusters of specific size $s$, the largest cluster (found until frame $f$), or clusters within a range can be filtered out. Depending on the graph model type used, the order in which the frames are processed differs for this approach. For all the models, the output will be the same for this approach. However, response time will vary depending on which graph type is given as input.

The primary issue with the above baseline approach is that performing K-means and the Elbow method on all the video frames is computationally expensive. K-means itself requires a certain number of iterations to determine the centroids of a cluster. Apart from this, Elbow runs K-means $|O_f|$ number of times to determine $K$ for a frame $f$ with $O_f$ set of object id's. Repeating the above for all the $\mathcal{F}$ frames in a video, even after setting the number of iterations for K-means to a small value, will be very inefficient, particularly when processing large videos (length in hours). Therefore, the main challenge is to develop a solution that can \textit{``avoid performing K-means and Elbow on all the video frames and still obtain the same results as the baseline''}. To design such a solution, we need to understand how group(s) change in a video.


\noindent\underline{\textit{How do group(s) change in videos}}? In videos, objects enter or leave the field of view gradually, and groups are formed over a sequence of frames. These group(s) (or clusters of objects) \textit{can remain the same for some number of frames}. For those frames, the \textbf{cluster centroids remain approximately the same, and $K$ is also the same}. After some period (or sequence frames), these group(s) change. As a result, the number of clusters or group(s) $K$, and their centroids change. Group(s) can change in two cases: (i) when new objects enter or some objects leave the field of view of the video, or (ii) when the objects remain the same for some number of consecutive frame sequences, but group membership of objects changes. For case (i), both $K$ and cluster centroids change, and for case (ii), only the cluster centroids change; the number of clusters $K$ can change or remain same. 


In frames where the \textbf{group(s) do not change}, K-means and Elbow can be avoided by using the $K$ and the centroid from the previous frame for clustering (termed \textbf{simple clustering}). This approach can avoid the significant computational costs incurred by performing K-means (with multiple iterations) and Elbow (for different values of K), while still obtaining the same clusters, as the centroids remain approximately the same. However, \textbf{when the group(s) change in a frame}, K-means and Elbow need to be applied to determine the new cluster centroids and K. \textit{The difficulty lies in finding out in \textbf{which frame(s) the group(s) or clusters change} in a video.} Appropriate heuristics can be developed to capture at which frame(s) group(s) or clusters change. For the above, the assumption is that the frames are processed in a sequential manner.

\noindent\underline{\textit{How can we drop frames?:}} Another important aspect of videos is that the frames that have number of objects less than the largest cluster size or number of objects less than the lower bound of the given size parameters, clustering is not required in those frames as they cannot contain the largest cluster or clusters of desired sizes. This criterion is particularly useful for identifying the largest cluster. 
If we knew the actual largest cluster size, we could drop the largest number of frames, which is the most optimal way. However, we cannot know the actual largest cluster size until the entire video or all the frames have been processed. Hence, as we perform clustering on a frame $f$, we can drop it if it has $<s$ objects, where $s$ represents the largest cluster size found until frame $f$ (a local maximum). $s$ can be updated with the new largest cluster after performing clustering on frame $s$, and eventually $s$ will have the maximum cluster size. We can drop a large number of frames in this way, even if it is not the optimal way of doing it. The earlier we find the largest cluster size, the more frames we will drop. The order in which the frames are processed may have some advantages in terms of the number of frames we drop for certain video types, which we will discuss later below. This criterion can be applied when $s$ is given as well. 


Based on the above observations, we propose two parameterized algorithms for the model SGF: (i) Group computation using Heuristics (GC\_Heuristic shown in Algorithm~\ref{alg:histogram-of-objects}), and (ii) Histogram of Objects (HO) algorithm (shown in Algorithm~\ref{alg:histogram-of-objects}), and one algorithm for the models SGV and MGV: Vertex Traversal (VT) algorithm (shown in Algorithm~\ref{alg:SGV-algo-group-computation}). Apart from the above, GC\_Heuristic and VT algorithms are parameterized in such a manner that baselines for model SGF and SGV can be generated for accuracy and performance comparison. The GC\_Heuristic algorithm processes the graphs of the model SGF sequentially and applies appropriate heuristics for capturing cases (i) and (ii) of group change to avoid K-means and Elbow in a frame as well as drops graphs with $<s$ objects. Here $s$  $s$ can be the largest cluster size found until frame $f$ (or the whole video) or can be the lower bound of given size parameters. The HO algorithm groups the graph of model SGF in some manner, and drops graphs with $<s$ objects. The same principle is used by VT algorithm to drop graphs with $<s$ objects in a different manner for the model SGV/MGV. Since model $MGV$ graphs can be processed in parallel, a composition function for combining results from all the graphs is proposed in Algorithm~\ref{alg:MGV-composition-group-computation}. The table of notations used by these algorithms is shown in Table~\ref{tab:Group-computation-algo-notation}.

\subsection{Group Computation Using a Heuristic-based Approach Using Model SGF}

We are proposing a \underline{\textit{one-pass}} heuristic-based algorithm, termed GC\_Heuristic (shown in Algorithm~\ref{alg:heuristic-algo}), for the model SGF, addressing all the aforementioned aspects of group change. Below, we will first provide an overview of the algorithm's various components, along with introducing alternative heuristics for identifying group changes, and then explain it in detail, line by line. 

This algorithm processes one graph $SGF_f^{\mathcal{V}}$ in $SGF^{\mathcal{V}}$ set of graphs generated by model $SGF$ at a time (or sequentially processes the graphs in $SGF^{\mathcal{V}}$). It performs K-means or Elbow on a graph \textit{when it detects that group(s) have changed using a heuristic}. For subsequent graphs it clusters object id's by carrying forward the centroids and $K$ from the previous graph until the heuristic condition (introduced later) is satisfied again.

This algorithm also drops graphs having $<s$ objects, where $s$ contains the largest cluster size found until graph $SGF_f^{\mathcal{V}}$ is processed (or frame $f$), or a lower bound of given size parameters. However, if a large number of frames/graphs are dropped before processing graph $SGF_f^{\mathcal{V}}$,
then the previously computed cluster centroids and $K$ may no longer be correct as \textit{group(s) may have changed in between frames/graphs}. K-means and Elbow need to be done on the graph $SGF_f^{\mathcal{V}}$ after dropping a certain number of graphs or frames. This algorithm addresses this by maintaining a threshold (termed $th_{drop}$). K-means and Elbow are done on the graph $SGF_f^{\mathcal{V}}$ if $th_{drop}$ number of graphs have been dropped for having fewer than $s$ objects. 

However, $th_{drop}$ cannot be set as an arbitrary number of graphs, as determining this for each video type will be difficult. If $th_{drop}$ is s\textit{et too small, K-means and Elbow will be run too often, leading to unnecessary computations even when the groups have not truly changed.} If it is too large, the algorithm may proceed with simple clustering using an incorrect $K$ and cluster $centroids$, resulting in incorrect clustering. In videos, the frequency at which group(s) change also depends on how fast/slow objects are moving. Since this movement is observed relative to the frame rate ($fps$), this can be used to determine $th_{drop}$ (or how many frames should be considered a big drop for a certain video type). $th_{drop}$ is defined as a factor $\alpha$ of $fps$ in this algorithm and $th_{drop}$ values can be $1/2~fps$, $1~fps$, etc. In this way, the $th_{drop}$ need not be changed for every video type. \textit{Note that for all of these,  we exclude frames with $\leq 2$ objects}.

\noindent \underline{\textit{Alternative heuristics for identifying group change:}} We are proposing two threshold-based heuristics: (i) Jaccard Dissimilarity (JD) heuristic (for capturing case (i) above for group change), and (ii) Cluster Quality (CQ) heuristic for capturing case (ii) of group change explained above. We also propose a heuristic (independent of any threshold) based on cluster size (termed the \textbf{Cluster Size or CS heuristic}) to determine when K-means and Elbow method should be applied in a frame or graph, which can \textbf{capture both cases (i) and (ii) above for group change}. All three heuristics, along with their advantages and disadvantages, are discussed below. 

\noindent \underline{\textit{JD heuristic:}} As mentioned above, cluster centroids and $K$ can change when objects enter/leave a video in case (i). In other words, if the object id's in two consecutive frames $f-1$ and $f$ (or graphs generated from those frames) are significantly dissimilar, $K$ and $centroids$ can change. Dissimilarity of object id's in two graphs $SGF_f^{\mathcal{V}}$ and $SGF^{\mathcal{V}}_{f-1}$ can be measured using JD metric, where $ JD(SGF^{\mathcal{V}}_f, SGF^{\mathcal{V}}_{f-1}) = 1- \frac{|O_f \bigcap O_{f-1}|}{|O_f \bigcup O_{f-1}|}$, where $O_f$, $O_{f-1}$ are the set of vertices of graphs $SGF_f^{\mathcal{V}}$ and $SGF^{\mathcal{V}}_{f-1}$  respectively. $JD$ score ranges between 0 and 1. $JD (SGF^{\mathcal{V}}_f, SGF^{\mathcal{V}}_{f-1}) =1$ means the two graphs $SGF_f^{\mathcal{V}}$, $SGF^{\mathcal{V}}_{f-1}$ have completely disjoint set of vertices (or no common object id's in frames $f$ and $f-1$), and $JD (SGF^{\mathcal{V}}_f, SGF^{\mathcal{V}}_{f-1}) =0$ means the two graphs $SGF_f^{\mathcal{V}}$, $SGF^{\mathcal{V}}_{f-1}$ have same set of vertices (or all the object id's in frames $f$ and $f-1$ are same).

To measure how different the object ids are in two consecutive frames, we compare their JD score against a threshold $th_{JD}$ (range 0-1). If $JD(SGF^{\mathcal{V}}_f, SGF^{\mathcal{V}}_{f-1}) \geq th_{JD}$, the difference between the two graphs is considered significant, and K-means and Elbow are applied on $SGF_f^{\mathcal{V}}$. Otherwise, the graphs are deemed sufficiently similar, and simple clustering is carried out using the $K$ and centroids from $SGF^{\mathcal{V}}_{f-1}$. 


\noindent \underline{\textit{CQ heuristic:}} In videos, objects can also change group membership even if the set of objects remains the same (i.e., they do not leave the field of view) as mentioned above for case (ii). In this case, the cluster's centroid will change when objects move to another group, and $K$ will remain the same. Hence, if the \textit{simple clustering} explained above is used, the cluster quality will drop significantly in graph $SGF_f^{\mathcal{V}}$ from the last graph (or frame) where K-means and Elbow were performed. 

Let's assume K-means and Elbow is used for clustering in graph $SGF_j^{\mathcal{V}}$ (representing frame $j$), and the $centroids$ and $K$ values from graph $SGF_j^{\mathcal{V}}$ are carried forward until graph $SGF_f^{\mathcal{V}}$ ($j < f$). 
The clusters of bounding box centroids obtained from $SGF_j^{\mathcal{V}}$ and $SGF_f^{\mathcal{V}}$ are $C^{bb}_j$ and $C^{bb}_f$ respectively (see Table~\ref{tab:Group-computation-algo-notation} for examples). After performing a simple clustering on $SGF_f^{\mathcal{V}}$, if the quality of the clusters in $C^{bb}_f$ obtained from graph $SGF_f^{\mathcal{V}}$, drops from the quality of the clusters in $C^{bb}_j$ obtained from $SGF_j^{\mathcal{V}}$ \textbf{significantly}, then we can perform K-means and Elbow in $SGF_f^{\mathcal{V}}$. The relative drop in cluster quality can be computed as $CQ_{drop} = \frac{CQ(C^{bb}_j)-CQ(C^{bb}_f)}{CQ(C^{bb}_f)}$, where $CQ(\cdot)$ represents their cluster quality scores, and $CQ(C^{bb}_j) < CQ(C^{bb}_f)$. The cluster quality is traditionally measured using the Silhouette score, which requires the clustered points and their corresponding centroids. Silhouette score ranges from -1 to 1, where -1 means all the object id's are put in the wrong cluster and 1 means objects are well-clustered. We have used the Silhouette score for $CQ$ as well, and scaled the Silhouette scores to range from 0-2, so that $CQ_{drop}$ value can range from 0-1. If the $CQ_{drop}$ is more than a given threshold termed $th_{CQ}$ (range 0-1), then it means $centroids$ have been changed significantly in frame $f$, and group(s) have changed. K-means and Elbow need to be performed on the graph $SGF_f^{\mathcal{V}}$. Otherwise, simple clustering can be carried forward.

\noindent \underline{\textit{Limitations of the threshold-based heuristics:}} The problem with the above threshold-based heuristics is that these thresholds can change depending upon the video, and need to be adjusted for each video. Given the limitations of VCE algorithms, which do not guarantee detection in consecutive video frames or consistent tracking across them, the threshold-based approach will not yield high accuracy for all types of video. Determining such a threshold value is also challenging. Apart from this, depending on how the thresholds are set, both heuristics can end up performing K-means and Elbow on all video frames, with \textit{added time for computing the heuristic formulas} for $JD$ and $CQ$, which we are trying to avoid.

Below, we introduce a simpler heuristic that is independent of any threshold and can be applied to any type of video.

\noindent \underline{\textit{$CS$ heuristic:}} In videos, if the group/cluster centroids or $K$ changes in a frame $f$ \textit{the size of at least one group or cluster changes (increases/decreases) among all the groups/clusters formed in frame $f$}. If clustering in frame $f$ is done using $K$ and centroids carried from a previous frame (simple clustering), objects can be assigned incorrectly to a cluster. As a result, \textit{the cluster size can become very large due to the use of incorrect cluster centroids}. This criterion can be used to determine whether K-means and Elbow need to be performed on a frame $f$.

Let's assume K-means and Elbow is used for clustering in graph $SGF_j^{\mathcal{V}}$ (representing frame $j$), and the centroids and $K$ values from graph $SGF_j^{\mathcal{V}}$ are carried forward until graph $SGF_f^{\mathcal{V}}$ ($j < f$). The largest cluster size in graph $SGF_j^{\mathcal{V}}$ (obtained by doing K-means and Elbow) is $s_j$, and in graph $SGF_f^{\mathcal{V}}$ (obtained by simple clustering) is $s_f$, respectively.  If the largest cluster obtained from $SGF_f^{\mathcal{V}}$ after performing simple clustering is larger than the largest cluster obtained from graph $SGF_j^{\mathcal{V}}$ (where K-means and Elbow was last performed), then it means graph $SGF_f^{\mathcal{V}}$ vertices or object id's were clustered by carrying forward wrong $centroids$ and $K$. \textit{This heuristic will be true if $s_f > s_j$, and the algorithm will perform K-means and elbow on graph $SGF_f^{\mathcal{V}}$} to obtain new $centroids$ and $K$. In this way, we can avoid \textit{carrying forward incorrect centroids and K value}. This heuristic is applicable to all the different size parameters discussed above and is not dependent on any thresholds, unlike the alternatives mentioned above. 

For the CS heuristic, it is also possible that, from frame j to f, cluster sizes decrease (e.g., if objects leave the field of view). In such cases, this heuristic will fail, allowing the algorithm to perform simple clustering. \textit{Since we are dropping frames with $<s$ objects, K-means and Elbow will be performed after dropping the $th_{drop}$ number of frames, and the correct $K$ and $centroids$ will be obtained eventually}.

\begin{table}[!htb]
\centering
\scriptsize
\caption{\small Table of notations for proposed algorithm for identifying groups of different sizes}
\label{tab:Group-computation-algo-notation}
\begin{tabular}{|p{0.1\textwidth}|p{0.85\textwidth}|}
\hline
\textbf{Notation} & \textbf{Description and Example} \\ \hline
$SGF^{\mathcal{V}}$ & Set of input graphs generated by model SGF. \\ \hline
$SGV^{\mathcal{V}}$ & Input graphs generated by model SGV. \\ \hline
$MGV^{\mathcal{V}}$ & Set of input graphs generated by model MGV. \\ \hline
$BGF^{\mathcal{V}}$ & Base graph file(s) generated for model SGF/SGV/MGV \\ \hline
$IDF^{\mathcal{V}}$ & A data file indexed on frame id. Used to fetch bounding boxes for each object id in a frame. \\ \hline
$M^{\mathcal{V}}$ & Meta information associated with video $\mathcal{V}$ (stored in the graph file header). \\ \hline
$s, p$ & Size parameters specified as part of analysis. If $p = s$, clusters of size $s$ are filtered; if $p =*$, and $s$ is given, then clusters with size $>s$ are filtered. If $p=max$, the largest cluster size is computed (\textit{$s$ contains the largest cluster size} at the end of processing a frame or the whole video. \\ \hline
$CL$ & Given class label to filter objects of specific type (e.g., person). An optional input parameter. \\ \hline
$O_f$ & Set of objects in frame $f$. \\ \hline
$BB_f$ & Set of bounding boxes in frame $f$ for each $o_i \in O_f$. These are fetched from $ID^{\mathcal{V}}$ during an analysis.\\ \hline
$BB_{centroid}^f$ & Set of bounding box centroids computed from $BB_f$ for each $o_i \in O_f$. It has one-to-one correspondence with $O_f$. 
\\ \hline
$K$ & Number of clusters. \\ \hline
$C_f^{bb}$ & Set of clustered bounding box centroids obtained from a clustering algorithm. $C_f^{bb}$ can contain $\{\{bb_1,bb_2\}, \{bb_3,bb_4,bb_5\}\}$, where $\{bb_1, bb_2\}$ and $\{bb_3,bb_4,bb_5\}\}$ belongs to two different clusters, and $bb_i$ is bounding box centroid of object id $o_i$ in $O_f$. \\ \hline
$C_f$ & Clusters of object ids extracted from $C_f^{bb}$. Example: $\{\{o_1,o_2\}, \{o_3,o_4,o_5\}\}$. \\ \hline
$OC$ & Output clusters generated by an algorithm. Stores $cluster~size, cluster~of~object~id's, <f_i,\ldots,f_j>$, where $f_i,\ldots,f_j$ are the frame sequence, where the cluster was formed. For example, $OC$ can contain $2,<\{o_1,o_2\},<1,2,3>,<\{o_2,o_3\},<5>$. Here, it contains two clusters of size 2. The first cluster $\{o_1,o_2\}$  was formed in frame sequence $1,2,3$. The second cluster $\{o_2,o_3\}$ and was formed in frame 5.\\ \hline
$Histogram$ & A dictionary containing a histogram of objects (used by algorithm $HO$), where the key is the number of objects ($n$) in a frame $f$, and the value is a sequence of frame ids with $n$ number of objects. \\ \hline
$DoF$ & A dictionary of frames for a vertex $v$, which has key as frame id, value as number of objects in the frame. Used by algorithm VT to store vertex labels. \\ \hline
\end{tabular}
\end{table}

Based on all of the above aspects, the proposed GC\_Heuristic algorithm is shown in Algorithm~\ref{alg:heuristic-algo}. This algorithm takes as input $SGF^{\mathcal{V}}$ set of graphs generated by the model SGF (stored in a single base graph file $BGF^{\mathcal{V}}$), the indexed (on frame id) data file $IDF^{\mathcal{V}}$, video characteristic information $\mathcal{M}^{\mathcal{V}}$ (stored in $BGF^{\mathcal{V}}$ header), a parameter $generateBaseline$ (value can be True or False), class label $CL$ (optional parameter), size parameters s and p, and a factor $\alpha$. It also takes as input a parameter $heuristic$ to choose which heuristic to apply, along with the corresponding thresholds. $heuristic$ parameter accepts the value $~''JD''/''CQ''/''CS''$. If $heuristic =~'' JD~''$, then $th_{JD}$ should be given as input; if $heuristic =~'' CQ~''$, then $th_{CQ}$ should be given as input; and if $heuristic =~'' CS~''$, we do not need to input any threshold. The $generateBaseline$ parameter is used to generate the baseline for model SGF when it is set to $True$. The algorithm sets $doKMeans$ to $True$ when it decides to perform K-means and Elbow on a graph. It stores the clusters obtained at the last frame K-means and Elbow were done using $C_j^{bb}$, where $j$ denotes the frame id/graph id. 


Algorithm~\ref{alg:heuristic-algo} initializes the value of $s$ to 0 (serves as a lower bound for filtering clusters based on size) if the computation is for finding the largest cluster ($p=max$) in line 3. For other computations, it assumes $s$ will be given. In line 4, it extracts the $fps$ information from video meta information $\mathcal{M}^{\mathcal{V}}$, and computes $th_{drop}$ using given $\alpha$. Then it makes \underline{\textit{one pass}} on the graphs of $SGF^{\mathcal{V}}$ (stored in a base graph-file). While processing a graph $SGF_f^{\mathcal{V}}$ generated from frame $f$, it filters out the vertices with the given class label $CL$ in line 7. $O_f'$ contains the filtered set of vertex id's or object id's. It performs K-means and Elbow on the first graph, which has more than two nodes in line 9. In line 9, if the $generateBaseline$ parameter is set to true, it sets the value of $doKMeans$ to true. This means the algorithm will perform K-means and Elbow on all graphs with more than two nodes, and the $heuristics$ and dropping condition will not be applied. If $generateBaseline$ is false, then the algorithm applies a heuristic and drops graphs with $<s$ objects (lines 10-31).

In lines 10-11 of Algorithm~\ref{alg:heuristic-algo}, it drops graph $SGF_f^{\mathcal{V}}$, if number of vertices $|O_f'| < s$ and $|O_f'| > 2$ . It also counts the number of graphs that have been dropped for this condition using $D$ until K-means and the Elbow Method are applied to a graph $SGF_f^{\mathcal{V}}$. When a graph $SGF_f^{\mathcal{V}}$ has $\geq s$ vertices (checked in line 13), it first extracts the bounding box centroids $BB_f$ for all the filtered vertices in $SGF_f^{\mathcal{V}}$, from file $IDF^{\mathcal{V}}$ using the index on frame $f$, and computes the bounding box centroids $BB_{centroid}^f$ in lines 14-15. Here $BB_{centroid}^f$ is a list containing the centroids of all the object id's in $O_f'$. It then checks if the number of dropped frames, $D \geq th_{drop}$, and sets $doKMeans$ to true (lines 16-18).

\begin{algorithm}[H]
\scriptsize
\caption{\small GC\_Heuristic algorithm for finding group(s) using model SGF}
\label{alg:heuristic-algo}
\begin{algorithmic}[1]
\renewcommand{\algorithmicrequire}{\textbf{Input:}} \renewcommand{\algorithmicensure}{\textbf{Output:}}
\Require $SGF^{\mathcal{V}}$, $IDF^{\mathcal{V}}$, ${\mathcal{M^V}}$, $generateBaseline$, $\underline{CL}$, $s$, $p$, factor of fps $\alpha$, $heuristic$, \underline{$th_{JD}$}, \underline{$th_{CQ}$}
\Ensure Output clusters $OC$
\State $OC \leftarrow \emptyset$,  $K\leftarrow 1$, $centroids \leftarrow \emptyset, D \leftarrow 0, s_f \leftarrow 0, s_j \leftarrow 0, doKmeans \leftarrow False$
\State $C_j^{\mathcal{V}} \leftarrow \emptyset$ \Comment{The clusters obtained from last frame K-means and Elbow were done}
\State Set $s \leftarrow 0$ if $p=max$, otherwise use given $s$
\State $th_{drop} \leftarrow \alpha \times fps$ (extracted from $M^{\mathcal{V}}$) \Comment{Compute the threshold for dropping frames}
\For{each graph $SGF^{\mathcal{V}}_{f}$ in $SGF^{\mathcal{V}}$ (from the graph file) } 
    \State $doKmeans \leftarrow False$
    \State $O_f' \leftarrow $ Extract set of vertices in graph $G^{\mathcal{V}}_{f}$ with class label $CL$ (from the graph file) 
    \If {first graph with $|O_f'| > 2)$ or $generateBaseline = True$} 
        \State $doKMeans \leftarrow True$
    \ElsIf{$2 < |O_f'| < s$ } \Comment{\textbf{Drop the graphs}} 
        \State $D \leftarrow D + 1$ \Comment{Compute number of graphs dropped} 
        \State \textbf{Continue}
    \ElsIf{$|O_f'|>2$ \text{ and } $|O_f'|\geq s$} 
   \Comment{No. of vertices is more than the largest cluster size/given size $s$}
        \State Fetch the line for frame $f$ from $IDF^{\mathcal{V}}$ and Extract $BB_f$ for each object id in $O_f'$ 
        \State $BB_{centroid}^f \leftarrow$ Compute the bounding box centroids from $BB_f$
        \If{$(D \geq th_{drop})$ or ($heuristic =~''JD''$ and ($JD(SGF_f^{\mathcal{V}}, SGF_{f-1}^{\mathcal{V}})$  $> th_{JD})$)}
            \State $doKMeans \leftarrow True$
        \Else \Comment{Perform \textbf{ simple clustering} using $K$ and $centroids$ from previous graph}
            \State $ C_{f}^{bb} \leftarrow$ Cluster $BB_{centroid}^f$ using $centroids$ and $K$ 
            \State $centroids \leftarrow$ Recompute centroids for each cluster in $C_{f}^{bb}$
           
            \If{$~(heuristic=~''CS'')$} \Comment{Check \textbf{cluster size heuristic condition}}
                 \State $s_f \leftarrow $ largest cluster size in $C_{f}^{bb}$; 
                 $s_j \leftarrow $ largest cluster size in $C_{j}^{bb}$ 
                \If{$s_f > s_j$}
                    \State $doKMeans \leftarrow True$
                \EndIf
            \ElsIf{$heuristic=~''CQ''$} \Comment{Check cluster quality heuristic condition}
                \If{$CQ(C_f^{bb}) < CQ(C_j^{bb})$ and $CQ_{drop}(C_f^{bb}, C_{j}^{bb}) \geq th_{CQ}$}
                    \State $doKMeans \leftarrow True$
                \EndIf
            \EndIf
            
        \EndIf
        \If{$doKMeans =True$} \Comment{\textbf{Perform K-means and elbow}}
            \State $K \leftarrow$ Elbow($BB_{centroid}^f$)
            \State $C_{f}^{bb}, centroids \leftarrow$ K-means($K$, $BB_{centroid}^f$)      
            \State $D \leftarrow 0$
            \State $C_j^{\mathcal{V}} \leftarrow C_f^{\mathcal{V}}$
            \Comment{Update the clusters obtained from the last frame, K-means was done}
        \EndIf
        \State $C_{f} \leftarrow$ Determine the cluster membership for each $oid$ from the bounding box centroids in $C_{f}^{bb}$ 
        \State $OC, s \leftarrow$ \Call{Update\_Output\_Clusters}{$OC$, $f$, $s$, $p$, $C_{f}$}  
    \EndIf
\EndFor
\end{algorithmic}
\end{algorithm}

\begin{algorithm}[!htb]
\scriptsize
\caption{\small Update Output Clusters}
\label{alg:update-output-cluster}
\begin{algorithmic}[1]
\renewcommand{\algorithmicrequire}{\textbf{Input:}} 
\renewcommand{\algorithmicensure}{\textbf{Output:}}
\Require $OC, f, s, p, C_{f}$
\Ensure $OC$, $s$
\Function{Update\_Output\_Clusters}{$OC, f, s, p, C_{f}$}
    \For{each cluster $c ~in~ C_{f}$}
        \If{$p = max$ \textbf{and} $|c| > s$} \Comment{A larger cluster has been found}
            \State $s \leftarrow |c|$, $OC \leftarrow \emptyset$
            \State Insert $c$ and frame id $f$ in $OC$
        \ElsIf{$p = max$ \textbf{and} $|c| = s$} \Comment{New cluster of size $s$ is found/existing largest clusters are in frame $f$}
            \State Insert $c$ and frame id $f$ in $OC$/Add $f$ to the existing frame ids of $c$
        \Else
            \State $OC$ Insert/Update clusters with size $>s$ or within range $[s,p]$ or size $=s$ from $C_{f}$ and frame id $f$
        \EndIf
    \EndFor
    \State \Return $OC$, $s$
\EndFunction

\end{algorithmic}
\end{algorithm}


 In line 17, it also checks for the $JD$ heuristic condition \textit{if $heuristic =~''JD''$}. If the dropping condition or the heuristic condition is not satisfied in line 17, it clusters the bounding box centroids in $BB_{centroid}^f$ using the centroids and $K$ obtained from the previously processed graph, and computes cluster centroids in lines 19-20. The clustered bounding box centroids are stored in $C_{f}^{bb}$. 

After performing simple clustering, if the given heuristic is $CS$, it evaluates the heuristic conditions in lines 21-25. Here, it extracts the largest cluster size $s_f, s_j$ from the clusters of $C_{f}^{bb}$, $C_{j}^{bb}$ respectively. If $s_f > s_j$ \textbf{($CS$ heuristic condition) in line 23}, then it sets $doKMeans$ to true. The condition for the $CQ$ heuristic is evaluated in line 27. Here, it computes the cluster qualities of $C_{f}^{bb}$, and $C_{j}^{bb}$, and check if cluster quality of $C_{f}^{bb}$ has dropped from cluster quality of $C_{j}^{bb}$ above given $th_{CQ}$, using the $CQ_{drop}$ formula introduced earlier (line 27). If the heuristic condition is satisfied, then it sets the value of $doKeameans$ to $True$. 


Algorithm~\ref{alg:heuristic-algo} performs K-means and Elbow when $doKMeans = True$ in lines 32-37, and resets the value of $D$ to 0 in line 35, so that after performing K-means and Elbow on a graph, $D$ can be newly calculated. It also updates the value of $C_{j}^{bb}$ with new clusters obtained after performing K-means and Elbow on graph $SGF_{f}^{bb}$. 

Whether we perform simple clustering or K-means, the clustered bounding box centroids will be returned as $C_{f}^{bb}$ in lines 19 or 34. The cluster membership of each object id corresponding to the bounding box centroids in $C_{f}^{bb}$ is extracted and stored in $C_f$ (line 38). Finally, clusters of different sizes from $C_{f}$ are filtered out based on the parameters $s$ and $p$, and the final output clusters $OC$ are updated by calling the function Update\_Output\_Clusters shown in Algorithm~\ref{alg:update-output-cluster} in line 39. 

The \textbf{Update\_Output\_Clusters} algorithm checks if a larger cluster is found in frame $f$ if the computation is for the largest cluster (line 3 and line 6). If it finds a larger cluster, it removes all the elements of $OC$ and updates it with the new largest cluster (line 4). In this case, the largest cluster size increases, and \textit{$s$ is updated with the new largest cluster size}. If the algorithm finds a new largest cluster (without the largest cluster size being changed), it adds the clusters and their corresponding frame id in $OC$ (lines 6-7). For other computations, where $s$ is given, the algorithm filters clusters of size s, $>s$, or range $[s,p]$ and updates $OC$ accordingly (line 9). It returns updated output cluster $OC$, and the largest cluster size/lower bound of given size $s$. \textit{Note $s$ is only updated when the computation is for finding the largest clusters.}

\noindent \underline{\textit{Correctness:}} The heuristic-based algorithm, because of approximating which frames to perform K-means and Elbow, will not achieve 100\% accuracy for all different kinds of videos. Accuracy will only be 100\% if it finds all the frames with correct clusters of given size $s$ or $>s$ or within range $[s,p]$ or the largest cluster. This is likely to occur when the group formation (or size) remains unchanged across frames.

\noindent \underline{\textit{Complexity Analysis of Algorithm~\ref{alg:heuristic-algo}:}} Here, we will analyze the CPU time complexity and I/O complexity of the algorithm for both baseline generation and when heuristics are applied. We will assume that all the $\mathcal{F}$ frames of the video contain more than two objects, and the algorithm will process $\mathcal{F}_s$ frames in total with $>s$ objects for simplicity ($\mathcal{F}_s \leq \mathcal{F}$). A video $\mathcal{V}$ has $OI^{\mathcal{V}}$ total object id instances, and each graph in model SGF can have $OI^{\mathcal{V}}/F$ average number of nodes. We will also assume the $CS$ heuristic is applied for the heuristic condition.

\noindent \underline{\textit{CPU Time Complexity:}} The CPU time complexity of performing \textit{K-means clustering using Lloyd's algorithm} in a graph $SGF_f^{\mathcal{V}}$ is 
$\mathcal{O}(K \times |O_f| \times I \times d \times  2)$, where $K$ = number of clusters, $|O_{f}|$ = number of vertices in graph $SGF_f^{\mathcal{V}}$, $I$ = number of iterations for K-means, $d$ is the cost of computing distance, and $2$ is vector dimension (bounding box centroids are two dimensional). In the worst case, $K=|O_f|$, meaning that each object id/vertex in the graph will be assigned to an individual cluster. This can be simplified as $\mathcal{O}(|O_{f}|^2)$. The Elbow method applies the K-means algorithm $|O_f|$ times in a graph to determine $K$. Time complexity for performing Elbow method in a graph is $\mathcal{O}(|O_f|^3)$. 

The \textit{CPU time complexity of performing K-means and Elbow} in a graph $SGF_f^{\mathcal{V}}$ is $\mathcal{O}(|O_f|^2+|O_f|^3) \approx \mathcal{O}(|O_f|^3)$. Here, the Elbow method's time will be dominant. For this algorithm, the most computationally expensive steps are the K-means and the Elbow method, and \textbf{the number of graphs/frames K-means and Elbow done will dominate CPU time complexity}.

\noindent \underline{\textit{CPU time complexity for generating baseline}} using Algorithm~\ref{alg:heuristic-algo} will be $\mathcal{O}(\sum_{f \in \mathcal{F^V}}|O_f|^3)$, as it will perform K-means and Elbow on all the $\mathcal{F}$ graphs of model SGF. If each graph in model SGF has $OI^{\mathcal{V}}/\mathcal{F}$ average number of nodes, then the baseline approach complexity will be $\mathcal{O}({\mathcal{F} \times(OI^{\mathcal{V}}/\mathcal{F})^3})$

\noindent \underline{\textit{CPU time complexity for applying the heuristic-based approach}} using Algorithm~\ref{alg:heuristic-algo}: For this, we need to know the CPU time complexity of simple clustering in a graph $SGF_f^{\mathcal{V}}$, which is $\mathcal{O}(K \times |O_{f}| \times d \times 2)$. In worst-case ($K = |O_{f}|$), and this can be simplified as $\mathcal{O}(|O_f|^2)$. In the worst case, all graphs will have more than $ s$ objects (no graphs will be dropped), and the heuristic will be satisfied in every frame. The complexity for evaluating $CS$ heuristic at every graph is nominal and can be ignored. In this case, K-means and Elbow will be applied to all the $\mathcal{F}$ graphs generated by the model SGF, and complexity will be the same as the baseline.

In the average case, Algorithm~\ref{alg:heuristic-algo} will analyze $\mathcal{F}_s$ graphs. K-means and Elbow will be applied on half of the graphs, and \textit{simple clustering} on the other half (when the heuristic is not satisfied). Hence, the average case complexity of this algorithm will be $\mathcal{O}({\frac{F_s}{2}((OI^{\mathcal{V}}/F)^3+ (OI^{\mathcal{V}}/F)^2)})$, for $OI^{\mathcal{V}}/F$ average number of vertices per graph. This can be simplified as $\mathcal{O}({\frac{F_s}{2}(OI^{\mathcal{V}}/F)^3})$ as the number of graphs where K-means and Elbow are applied will dominate the time complexity.

\noindent \underline{\textit{I/O complexity:}} This algorithm iterates over all the $\mathcal{F}$ graphs generated by model SGF, which means it reads all the lines in the base graph file $BGF^{\mathcal{V}}$. A graph file for model SGF will contain $\sum_{f \in \mathcal{F^V}} (|O_f|+\frac{|O_f|^2}{2})+\mathcal{F}+h$ rows in the graph file, where $h$ is the number of header lines, $\sum_{f \in \mathcal{F^V}} (|O_f|+\frac{|O_f|^2}{2})$ is number of edge and vertex lines in the graph file, and $\mathcal{F}$ is the number of lines to store graph id, the number of nodes, and the number of edges lines. In this algorithm, a clustering (simple/K-means) will always be applied on a frame, which will require fetching bounding box centroids for all the $\mathcal{F}$ frames in the worst case. Hence, the I/O complexity for this algorithm is the same for both baseline generation and when the heuristic is applied. The I/O complexity for this algorithm is $\mathcal{O} (\sum_{f\in \mathcal{F^V}} (|O_f|+|O_f|^2)+2\mathcal{F})$, which is $\mathcal{O}(OI^{\mathcal{V}}+ (OI^{\mathcal{V}})^2/\mathcal{F}+\mathcal{F})$ for $OI^{\mathcal{V}}/\mathcal{F}$ average number of nodes per graph. The average case I/O complexity for this algorithm will be $\mathcal{O}(OI^{\mathcal{V}}+ (OI^{\mathcal{V}})^2/\mathcal{F}+\mathcal{F}_s)$ to fetch $\mathcal{F}_s$ lines will be fetched from $IDF^{\mathcal{V}}$.





\subsection{Histogram of Objects (HO) algorithm for model SGF}

As mentioned earlier, to find groups or clusters of different sizes, K-means and Elbow need to be used for clustering on only the graphs with more than $s$ vertices, where $s$ is the lower bound of the given size or contains the largest cluster size found until frame $f$ or the whole video after processing all the frames of the video. Then, 100\% accuracy (when compared with the baseline) can be achieved. 

In videos, typically, a very small number of frames contain the maximum number of objects ($M$) or a large number of objects (unless it is a video of a crowded area, sports video, etc.) This can be beneficial for \textbf{finding the largest cluster problem}. Here, if K-means and Elbow are performed on the graphs in the descending order of the number of objects or vertices, then the actual largest cluster size can be found earlier by processing a very small number of graphs, and a large number of graphs or frames can be dropped. If the actual largest cluster size in the video is closer to $M$, then processing graphs in the descending order of the number of vertices will perform clustering on less number of frames than sequentially performing clustering on each graph in $SGF_f^{\mathcal{V}} \in SGF^{\mathcal{V}}$. For finding clusters of size $>s$, or given size $s$ or within a range $[s,p]$, no matter what order the graphs are processed, we will end up doing K-means and Elbow on the same number of frames, as we have discussed earlier. The closer given size $s$ is to $M$ in a video, the larger the number of frames that will be dropped. Based on the above, Algorithm~\ref{alg:histogram-of-objects} is proposed.


\begin{algorithm}[!htb]
\scriptsize

\caption{\small Histogram of Objects (HO) for finding group(s) using Model SGF}
\label{alg:histogram-of-objects}
\begin{algorithmic}[1]
\renewcommand{\algorithmicrequire}{\textbf{Input:}} \renewcommand{\algorithmicensure}{\textbf{Output:}}
\Require $SGF^{\mathcal{V}}$, $IDF^{\mathcal{V}}$, ${\mathcal{M^V}}$, $\underline{CL}$, $s$, $p$
\Ensure Output clusters $OC$
\State $OC \leftarrow \emptyset$,  $K\leftarrow 1, Histogram \leftarrow \emptyset$
\State Set $s \leftarrow 0$ if $p=max$, otherwise use given $s$
\State $M \leftarrow $ Extract maximum number of objects from $M^{\mathcal{V}}$
\For{each graph $SGF_f^{\mathcal{V}}$ in $ SGF^{\mathcal{V}}$ (from the base graph file)}
    \State $n \leftarrow |O_f|$ \Comment{No. of nodes in graph $SGF_f^{\mathcal{V}}$}  
    \If{$n > 2$ and $n \geq s$}  \Comment{Build the histogram}
        \State $Histogram[n] \leftarrow $ Add frame id/graph id $f$ 
    \EndIf

\EndFor
\While{True}   
    \If{$M < s$}
        \State \textbf{break}
    \EndIf
    \State $F_{M} \leftarrow Histogram[M]$ \Comment{List of frame ids with $M$ objects}
    \For{each frame $f$ in $ F_M$}
 
       \State Fetch the line for frame $f$ from $IDF^{\mathcal{V}}$ and
       Extract $O_f$ with class label $CL$ and corresponding $BB_f$ 
          \State $BB_{centroid}^f \leftarrow$ Compute the bounding box centroids from $BB_f$
        \State $K \leftarrow$ Elbow($BB_{centroid}^f$)
        \State $C_{bb}^f$, centroids $\leftarrow$ K-means($K$, $BB_f$) 
        \State $C_f \leftarrow$ Determine the cluster membership of each corresponding $oid$ of $C_{bb}^f$
        \State $OC, s \leftarrow$ \Call{Update\_Output\_Clusters}{$OC$, $f$, $s$, $p$, $C_f$} \Comment{Filter clusters of different size or largest cluster}
    \EndFor
    
    \State $M \leftarrow M-1$
\EndWhile
\end{algorithmic}
\end{algorithm}

Algorithm~\ref{alg:histogram-of-objects} takes as input $SGF^{\mathcal{V}}$ set of graphs generated by the model SGF (stored in a single graph file $BGF^{\mathcal{V}}$), the indexed (on frame id) data file $IDF^{\mathcal{V}}$, video characteristic information $\mathcal{M}^{\mathcal{V}}$ (stored in base graph file header), class label $CL$ (optional parameter), and size parameters s and p. It initializes the value of $s$ to 0 (serves as a lower bound for filtering clusters based on size) if the computation is for finding the largest cluster ($p=max$) in line 2. For other computations, it assumes $s$ will be given. The algorithm first extracts the maximum number of objects, $M$, from the video metadata, ${\mathcal{M^V}}$, in line 3. Then it performs \underline{\textit{two passes}} on the graphs in $SGF^{\mathcal{V}}$. In the first pass (lines 4-9), the algorithm iterates over all the graphs in $SGF^{\mathcal{V}}$, and builds a dictionary $Histogram$ (where the key is the number of vertices in a graph, and the value is the list of graph ids/frame ids having the same number of vertices) based on the number of objects. In other words, it groups graphs that have the same number of vertices or objects and drops the graphs that have fewer than $s$ vertices (line 6). 


Once the histogram is built, in the second pass (lines 10-22), the algorithm iterates over the keys of the dictionary $Histogram$ in descending order of the number of objects using $M$. Note that, after the histogram is built, the graph file is not needed anymore. It extracts the list of graph ids/frame ids with $M$ objects, $F_M$ from $Histogram$ in line 14. It then iterates over the frame id's in $F_M$ one by one (line 15). On each frame $f$ in $F_{M}$, this algorithm first extracts the set of object id's $O_f$ with class label $CL$, and corresponding bounding boxes $BB_f$ from the indexed data file $IDF^{\mathcal{V}}$ using the frame id index (line 16). It then performs K-means and Elbow using the bounding box centroids $BB_{centroid}^f$ (lines 18-19), filters out clusters of different sizes/maximum clusters along with the maximum cluster size by calling the function Update\_Output\_Clusters (line 21). After processing all the frames in $F_M$, it decrements the value of $M$. It \textbf{terminates} if the value of $M$ is less than $s$ (line 11-13), which means there are no graphs/frames left with more than $s$ vertices or objects.

\noindent \underline{\textit{Correctness:}}
This algorithm always performs K-means and elbow on all the frames with $>s$ objects. Since a graph or frame with $<s$ objects cannot contain a cluster of given size or size $>s$ or within range $[s,p]$ or the largest cluster, this algorithm will always be 100\% correct (if compared with baseline approach).



\underline{\textit{CPU Time Complexity of Algorithm~\ref{alg:histogram-of-objects} :}} The complexity of building $Histogram$ in first pass is $\mathcal{O}(\mathcal{F})$, as it will iterate over all the $\mathcal{F}$ graphs. In the worst case, the largest cluster size will be one, or given size $s=1$. This algorithm will perform K-means and Elbow on all the $\mathcal{F}$ video frames in the second pass in worst-case, and CPU time complexity will be $\mathcal{O}(\mathcal{F}+\sum_{f \in \mathcal{F^V}}|O_f|^3)$, where $\mathcal{O} (\sum_{f \in \mathcal{F^V}}|O_f|^3)$ is CPU time complexity for performing K-means and Elbow on all the $\mathcal{F}$ frames. If each frame/graph contains $OI^{\mathcal{V}}/\mathcal{F}$ objects on average, then the worst-case complexity will be $\mathcal{O}(\mathcal{F}+\mathcal{F} \times (OI^{\mathcal{V}}/\mathcal{F})^3)$. In the average case, there will be $\mathcal{F}_s$ frames with more than $>s$ objects, where $\mathcal{F}_s < \mathcal{F}$, and it will perform K-means and Elbow on them. The complexity will be  $\mathcal{O}(\mathcal{F}+\mathcal{F}_s \times (OI^{\mathcal{V}}/\mathcal{F})^3)$.

\noindent \underline{\textit{I/O Time Complexity of Algorithm~\ref{alg:histogram-of-objects}:}} This algorithm iterates over all the $\sum_{f \in \mathcal{F^V}} (|O_f|+\frac{|O_f|^2}{2})+\mathcal{F}+h$ lines for all the vertices and edges in the graph file in the first pass. In the second pass, for the worst-case scenario, it will fetch the bounding box information of $\mathcal{F}$ frames from $IDF^{\mathcal{V}}$. The I/O complexity will be $\mathcal{O} (\sum_{f\in \mathcal{F^V}} (|O_f|+|O_f|^2)+ 2\mathcal{F})$. This can be simplified as $\mathcal{O}(OI^{\mathcal{V}}+ (OI^{\mathcal{V}})^2/\mathcal{F}+\mathcal{F})$ for $OI^{\mathcal{V}}/\mathcal{F}$ average number of objects per frame. In the average case, $\mathcal{F}_s$ frames will be fetched from $IDF^{\mathcal{V}}$. I/O complexity will be $\mathcal{O}(OI^{\mathcal{V}}+ (OI^{\mathcal{V}})^2/\mathcal{F}+\mathcal{F}_s)$ in average case.

\subsection{Vertex Traversal (VT) Algorithm for Model SGV and MGV}

As described in chapter~\ref{chap:Representation}, model SGV generates a single graph or a graph with multiple connected components $SGV^{\mathcal{V}}$, and model MGV generates $MGV^{\mathcal{V}}$ set of graphs from a video, where $MGV^{\mathcal{V}}=\{MGV_1^{\mathcal{V}}, \ldots, MGV_N^{\mathcal{V}}\}$ by distributing the components in a balanced manner (either by frames or nodes). Hence, the algorithm developed for the SGV model can be applied to the MGV model for $N$ graphs in parallel, and partial results obtained from the $N$ graphs need to be combined for a final global output. In this section, we propose an algorithm that processes the model SGV graph, and later we explain how the model MGV can use the same algorithm along with a composition algorithm for this model.   

 As mentioned above, in videos, typically, very few frames contain the maximum number of objects. If we can perform K-means and Elbow on these frames, then a lower bound of the largest cluster size can be obtained early on. The same principle used for algorithm \textbf{HO} can be applied to develop an algorithm for model SGV in a different manner, without resorting to building a histogram. Model SGV has certain advantages for this analysis. First, for this model, the graph file header has the sequence of frame id's with the maximum number of objects ($M$) among all the $\mathcal{F}$ frames of the video. If we perform K-means and Elbow on these frames, which will be very few in general, we can get a lower bound for identifying the largest cluster. 
 Second, for each vertex, the vertex label stores all the frame id's in which the vertex appears, along with the corresponding number of objects in each frame. These frame id's are stored in descending order of the number of objects. Because of this, all the frame id's of a vertex need not be scanned, and we can stop scanning the labels of a vertex $v$ if any frame contains $<s$ objects, where $s$ contains the largest cluster size until some number of frames are processed/a lower bound of the given sizes. Additionally, the same frame id can appear in multiple vertex labels, and we need to keep track of which frames K-means and Elbow were applied. For this analysis, traversing the edges is not required. 
 
 For model MGV, the maximum number of objects in a frame ($M_i$) is computed from all the frames used to generate the graph. This is a local maximum and $M_i\leq M$. Hence, the graph file header for graph $MGV_i^{\mathcal{V}}$ contains the frame id's with $M_i$ objects. If this algorithm is performed on model MGV, each graph will perform K-means and Elbow on the frames with $M_i$ objects to get a lower bound on the largest cluster size. This may lead to performing K-means and Elbow on more frames than the model SGV. However, because this model processes the $N$ graphs (or graph files) in parallel, this will not have any significant effect on efficiency in general.

Based on the above, a group computation algorithm (termed Vertex Traversal or VT) is proposed, which is shown as Algorithm~\ref{alg:SGV-algo-group-computation}. This algorithm first extracts the frame id's with the maximum number of objects from the header information of the input graph file, and performs K-means and Elbow on all of these frames, and obtains a lower bound on the largest cluster size $s$ (local maximum). Then it traverses the frame ids in the vertex labels stored in descending order of the number of objects. It stops scanning the vertex labels when the number of objects in a frame is $<s$. Once all the frames from which the input graph is generated are analyzed, this algorithm terminates. The proposed algorithm is also capable of generating the baseline for the model SGV, based on a parameter.

Since we will be using this algorithm for both SGV and MGV models, certain notations are employed to make it applicable to both models. Algorithm~\ref{alg:SGV-algo-group-computation} takes as input a graph $G$ which can be a graph $SGV^{\mathcal{V}}$ or a graph $MGV_i^{\mathcal{V}}$ in the set of graphs generated for $MGV$, an indexed (on frame id) data file $IDF^{\mathcal{V}}$, a parameter $generateBaseline$ (value is $True/False$), video meta information ${\mathcal{M^V}}$ (stored in graph file header), an optional parameter $CL$ to filter objects of given type, and size parameters $s$ and $p$. The input graph $G$ is stored in a base graph file $BGF^{\mathcal{V}}$. For this algorithm, we denote $\mathcal{F}_G$ as the number of frames $G$ was generated from ($\mathcal{F}_G = \mathcal{F}$ when the input graph is from the model SGV), and $\mathcal{F}_G^M$ as the set of frame id's having the maximum number of objects in a frame among the $\mathcal{F}_G$ frames. If model SGV is input, this will contain the frame id's that have $M$ objects among all the $\mathcal{F}$ video frames. 


\begin{algorithm}[H]
\scriptsize

\caption{\small Vertex Traversal (VT) Algorithm}
\label{alg:SGV-algo-group-computation}
\begin{algorithmic}[1]
\renewcommand{\algorithmicrequire}{\textbf{Input:}} \renewcommand{\algorithmicensure}{\textbf{Output:}}
\Require A graph $G$ ($SGV^{\mathcal{V}}$ or $MGV_i^{\mathcal{V}}$), $IDF^{\mathcal{V}}$, $generateBaseline$, ${\mathcal{M^V}}$, $\underline{CL}$, $s$, $p$
\Ensure Output Clusters $OC_i$
\Function{VT}{$G$, $IDF^{\mathcal{V}}$, $generateBaseline$, ${\mathcal{M^V}}$, $\underline{CL}$, $s$, $p$}
\State $OC \leftarrow \emptyset$; $\mathcal{F}_G \leftarrow$ Number of frames graph $G$ is generated from
\State $\mathcal{F}_{Kmeans} \leftarrow \emptyset$ \Comment{Set of frames where K-means is performed}
\If {$generateBaseline = False$}
    \State $\mathcal{F}_G^M \leftarrow$ Extract Set of frames with the maximum number of objects (from base graph file header)
    
            \For{each frame $f $ in $\mathcal{F}_G^M$} 
               \State Fetch the line for frame $f$ from $IDF^{\mathcal{V}}$ and
       Extract $O_f$ with class label $CL$ and corresponding $BB_f$ 
              \State $BB_{centroid}^f \leftarrow$ Compute the bounding box centroid from $BB_f$
               \State $K \leftarrow$ Elbow($BB_{centroid}$)
                \State $C_{bb}$, centroids $\leftarrow$ K-means($K$, $BB_{centroid}$) 
                \State $C_f \leftarrow$ Determine the cluster membership of each $oid$ from $C_{bb}$
                \State $OC, s \leftarrow$ \Call{Update\_Output\_Clusters}{$OC$, $f$, $s$, $p$, $C_f$}
                \State Add $f$ to $\mathcal{F}_{Kmeans}$
            \EndFor
\EndIf
\For{each vertex for object id $o_i$ in $G$} \Comment{Traverse the vertex list}
    \State $DoF \leftarrow \emptyset$
    \If{$\mathcal{F}_G = |\mathcal{F}_{Kmeans}|$} \Comment{K-means and Elbow \textbf{is performed} on all the frames in $G$}
        \State \textbf{break}
    \EndIf
    \If{$o_i$ has class label $CL$}
    \State Build $DoF$ with key frame id, value no. of objects from vertex label of $o_i$
    \For{each frame $f$ in $DoF$}
        \State $n \leftarrow$ $DoF[f]$ \Comment{Extract No. of objects in frame $f$}
        \If{$n < s$ and $generateBaseline =False$}              
            \State \textbf{break}
        \EndIf
        
        \If { $f$ not in $\mathcal{F}_{Kmeans}$} \Comment{Perform K-means and Elbow on frame $f$ }
             \State Fetch the line for frame $f$ from $IDF^{\mathcal{V}}$ using index and
       Extract $O_f$ and  $BB_f$ with label $CL$ 
            \State $BB_{centroid}^f \leftarrow$ Compute the bounding box centroid for each object id in $O_f$ from $BB_f$

            \State $K \leftarrow$ Elbow($BB_{centroid}$)
            \State $C_{bb}$, Centroids $\leftarrow$ K-means($K$, $BB_{centroid}$) 
            \State $C_f \leftarrow$ Determine the cluster membership of each corresponding $oid$ of $C_{bb}$
            \State $OC_i, s \leftarrow$ \Call{Update\_Output\_Clusters}{$OC_i$, $f$, $s$, $p$, $C_f$}
            \State Add $f$ to $\mathcal{F}_{Kmeans}$
        \EndIf
    \EndFor
\EndIf
\EndFor

\State \Return $OC$
\EndFunction
\end{algorithmic}
\end{algorithm}

The $generateBaseline$ parameter is used to generate a baseline for model SGV/MGV. If it is set to true, the algorithm will perform K-means and Elbow on all the $\mathcal{F}$ frames with $\geq 2$ objects by traversing the vertex labels only. We generate the baseline using this model to compare the performance of model SGV and MGV with $N$ graphs (for different values of $N$). This algorithm will output clusters $OC$, which denote the clusters obtained from the graph $G$. 


Algorithm~\ref{alg:SGV-algo-group-computation} maintains a \textbf{set} $\mathcal{F}_{Kmeans}$, which stores all the frame id's where K-means and Elbow is applied to avoid redoing K-means and Elbow on the frames where K-means and Elbow have already been applied. While traversing the vertices, the algorithm also builds a \textbf{dictionary of frames ($DoF$) from the vertex label of a vertex or object id $o_i$, which has key frame id and value no. of objects in that frame}. The keys are stored in descending order of the number of objects (as it is stored in this order in the vertex label). This dictionary is created for each vertex when it is traversed.

Algorithm~\ref{alg:SGV-algo-group-computation} extracts $\mathcal{F}_G$ in line 1, and  $\mathcal{F}_G^{M}$ from the graph file header in line 5. It then performs K-means and Elbow on all the frames in $\mathcal{F}_{G}^M$. For each frame $f$ in $\mathcal{F}_{G}^M$, it fetches the set of object id's with label $CL$, and corresponding bounding boxes of the objects from $IDF^{\mathcal{V}}$ (line 7), computes their centroids (line 8), performs K-means and Elbow (lines 9-10), and filters largest cluster/different sizes by calling function Update\_Output\_Clusters (line 12). It also adds the frames in $\mathcal{F}_{G}^M$ to $\mathcal{F}_{Kmeans}$, so that if we avoid performing K-means in these frames later, when they appear in vertex labels. For baseline generation, steps 4-15 are not executed.

The algorithm then traverses each vertex for an object id $o_i$ of graph $G$. It first extracts the set of frames where $o_i$ appears and the number of objects in each frame from the vertex label, and then builds $DoF$ for object id's $o_i$ from these (see line 22 in Algorithm~\ref{alg:SGV-algo-group-computation}). Then it iterates over each frame $f$ in $DoF$ and checks if K-means and Elbow have already been applied on frame $f$ using $\mathcal{F}_{Kmeans}$ (line 28). If not, then it performs K-means and Elbow on frame $f$, and filters out clusters of different sizes or the largest cluster size in lines 29-34. It stops scanning $DoF$, when the number of objects in frame $f$ is less than $s$ (line 25), and starts traversing the next vertex. The algorithm terminates if K-means and Elbow are performed in all the $\mathcal{F}_i$ frames (line 18). Finally, it returns the output clusters $OC$ for the graph $G$.

\begin{algorithm}[H]
\scriptsize
\caption{\scriptsize Compose Clustering Results }
\label{alg:MGV-composition-group-computation}
\begin{algorithmic}[1]
\renewcommand{\algorithmicrequire}{\textbf{Input:}} 
\renewcommand{\algorithmicensure}{\textbf{Output:}}

\Require Output clusters from $N$ graphs $OC_{1,N}$, $p$
\Ensure Composed output clusters $OC$
\Function{Compose\_Clustering\_Results}{$OC_{1,N}$, $p$}
\State $OC \leftarrow \emptyset, s \leftarrow 0$
\For{each $OC_i$ in $OC_{1,N}$}
    \If{$p = max$ and $OC \neq \emptyset$}
        \State $l \leftarrow$ Largest cluster size in $OC_i$
        \If{$l > s$}  \Comment{$OC_i$ contains a larger cluster than $OC$}
            \State $OC \leftarrow OC_i,\; s \leftarrow l$
        \ElsIf{$l = s$} 
            \State Add clusters of $OC_i$ to $OC$
        \EndIf
    \Else 
        \State Add clusters of $OC_i$ to $OC$
    \EndIf
\EndFor
\State \Return $OC$
\EndFunction
\end{algorithmic}
\end{algorithm}

\noindent \underline{\textit{Composition Algorithm for Model MGV:}} The VT algorithm shown in Algorithm~\ref{alg:SGV-algo-group-computation} can be applied on the $\{MGV_1^{\mathcal{V}},\ldots,MGV_N^{\mathcal{V}}\}\}$ set of graphs by calling the function VT $N$ times. From this, we will obtain a list of partial output clusters, $OC_{1,N} =<OC_1,\ldots,OC_N>$. These partial outputs are composed using the composition function proposed in Algorithm~\ref{alg:MGV-composition-group-computation}. If the given computation is for finding clusters of given size $s$ or $> s$ or within range $[s,p]$, then composition is easier, as all the partial outputs need to be merged into one output. Here, clusters are already filtered during the analysis phase because $s$ is given. For the largest cluster computation, each graph may output a different largest cluster, and the largest cluster size in the two partial outputs $OC_i$ and $OC_j$ in $OC_{1,N}$ may vary. The final output $OC$ must therefore contain the globally largest cluster.

The composition function shown in Algorithm~\ref{alg:MGV-composition-group-computation} takes as input a list $OC_{1,N}$ and the parameter $p$. From $p$, we can know if the computation is for finding the largest cluster (if $ p=max$). This algorithm iterates over each output cluster $OC_i$ at a time. For the largest cluster computation (lines 4-11), it checks, 
at each iteration, if $OC_i$ has a larger cluster than those already in $OC$, it replaces $OC$, and updates the current maximum cluster size $s$ (lines 6-8). If $OC_i$ contains clusters with size $s$ then it is added to $OC$; otherwise, $OC_i$ is discarded. For other computations, all the outputs are merged into $OC$ one by one (line 12).

\noindent \underline{\textit{Correctness:}} This algorithm will be 100\% correct (if compared to baseline approach) always, as it drops frames only with $<s$ objects.

\noindent \underline{\textit{CPU Time Complexity:}}  Algorithm~\ref{alg:SGV-algo-group-computation} has two major components to analyze for CPU time complexity: the time taken for traversing the vertex labels, and the number of frames K-means and Elbow is done. 

For \underline{\textit{Model SGV}} Algorithm~\ref{alg:SGV-algo-group-computation} will traverse $UO^{\mathcal{V}}$ (number of vertices in model SGV), where $UO^{\mathcal{V}}$ is the number of unique object id instances in video $\mathcal{V}$. In the worst case, each object id $o_i$ will appear in all the $\mathcal{F}$ frames, and it will take $\mathcal{O(F} \times UO^{\mathcal{V}})$ time to traverse all the vertex labels. It will perform K-means and Elbow on all the frames in the worst-case. The worst-case time complexity will be $\mathcal{O} (\mathcal{F} \times UO^{\mathcal{V}}+\sum_{f \in F^V}|O_f|^3)$, where $\mathcal{O}(|O_f|^3)$ is the CPU time complexity of K-means and Elbow in a frame $f$. In the average case, each unique object id can appear in $\mathcal{F}/UO^{\mathcal{V}}$ frames on average. The CPU time complexity for traversing all the vertex labels will be $\mathcal{O}(UO^{\mathcal{V}} \times (\mathcal{F}/UO^{\mathcal{V}}))$, which is equal to $\mathcal{O} (\mathcal{F})$. If there are $OI^{\mathcal{V}}/\mathcal{F}$ objects in a video frame on average, and $F_s$ frames have more than $s$ objects ($\mathcal{F}_s < \mathcal{F}$), then the average case time complexity will be $\mathcal{O} (\mathcal{F}+ \mathcal{F}_s \times (OI^{\mathcal{V}}/\mathcal{F})^3)$. Note that if the baseline is generated using  Algorithm~\ref{alg:SGV-algo-group-computation}, the complexity will be the same as the worst case complexity.

For \underline{\textit{Model MGV}} all $N$ graphs are processed in parallel, and the graph that takes the maximum amount of time to process will dominate the time complexity. In other words, the maximum time taken among the $N$ graphs is considered as the response time for this model. In the worst case, this model will generate one graph; the time complexity will be the same as model SGV. Let us assume, each graph has $\mathcal{F}/N$ frames on average, and among them $\mathcal{F}_s/N$ frames have $>s$ objects. Then, on average, each graph will take the same amount of time, and the average case complexity will be $\mathcal{O} (\frac{\mathcal{F}_s}{N}+ \frac{\mathcal{F}_s}{N} \times (OI^{\mathcal{V}}/\mathcal{F})^3)$. On average, $MGV$ can take $N$ time smaller CPU time than $SGV$ for the same algorithm.

\noindent \underline{\textit{I/O Complexity:}} Algorithm~\ref{alg:SGV-algo-group-computation} reads each vertex of the graph file for model $SGV$, and it will read a total of $UO^{\mathcal{V}}$ lines (all the vertex lines) from the graph file. In the worst case, it will perform K-means and Elbow on all the $\mathcal{F}$ frames, and it will fetch $\mathcal{F}$ lines from $IDF^{\mathcal{V}}$. In worst case I/O complexity will be $\mathcal{O}(UO^{\mathcal{V}}+\mathcal{F})$. Model MGV will have the same I/O complexity as $SGV$ in the worst case.

In the average case, it will perform K-means and Elbow on $\mathcal{F}_s$ frames for model SGV, and fetch $\mathcal{F}_s$ frames information from $IDF^{\mathcal{V}}$. In average case, the I/O complexity for model MGV will be $\mathcal{O}(UO^{\mathcal{V}}+\frac{\mathcal{F}_s}{N})$ for fetching $\mathcal{F}_s/N$ frame information $IDF^{\mathcal{V}}$.
 

\subsection{Summary}
We have proposed two algorithms for model SGF, and one algorithm for models SGV and MGV, and showcased that all the models can be used for the same analysis. Except for the GC\_Heuristic algorithm, all the algorithms will achieve 100\% correctness (if compared with baseline). For all of these algorithms, the number of frames K-means and Elbow dominate the time complexity. However, depending upon the model type, in which order we process frames or how we traverse the graphs also affects the performance. The summary of the accuracy and time complexity of all the algorithms for different models is shown in Table~\ref{tab:complexity-comparison-group-computation}. The heuristic-based algorithm (GC\_Heuristic) for the SGF model is not guaranteed to be 100\% correct, since its success depends on how objects enter or exit the field of view. Without prior knowledge of the video, and given the enormous diversity of video types and the potential errors introduced by VCE algorithms, defining a heuristic (based on approximation) that achieves (near 100\%) accuracy is very challenging. However, the advantage of this algorithm is that it is a single pass and is suitable for eventual real-time applications.

\begin{table}[ht]
\centering
\scriptsize

\caption{\small Comparison of group computation algorithms' accuracy and time complexity (average case). Here, $(OI^{\mathcal{V}}/\mathcal{F})$ is the average no. of objects, $\mathcal{F}_s: $ no. of frames with $>s$ objects, and $\mathcal{F}_s < \mathcal{F}$. }
\label{tab:complexity-comparison-group-computation}
\begin{tabular}{|c|c|c|c|c|}
\hline
\textbf{Model} & \textbf{Algorithm}       & \textbf{CPU Time Complexity} &   \textbf{I/O Time Complexity}        & \textbf{Accuracy} \\ \hline
SGF                 & GC\_Heuristic          &      $\mathcal{O}({\frac{\mathcal{F}_s} {2} \times (OI^{\mathcal{V}}/\mathcal{F})^3})$ & $\mathcal{O}(OI^{\mathcal{V}}+(OI^{\mathcal{V}^2}/\mathcal{F})+\mathcal{F}_s)$ & 0\%-100\%    \\ \hline
SGF                 & HO          &     $\mathcal{O}({\mathcal{F}+\mathcal{F}_s \times (OI^{\mathcal{V}}/\mathcal{F})^3})$ & $\mathcal{O}(OI^{\mathcal{V}}+(OI^{\mathcal{V}^2}/\mathcal{F})+\mathcal{F}_s)$   & 100\%  \\ \hline    
SGV                 & VT          &   $\mathcal{O}({\mathcal{F}+\mathcal{F}_s \times (OI^{\mathcal{V}}/\mathcal{F})^3})$ & $UO^{\mathcal{V}}+\mathcal{F}_s$& 100\%          \\ \hline
MGV                 & VT          &    $\mathcal{O} (\frac{\mathcal{F}_s}{N}+ \frac{\mathcal{F}_s}{N} (OI^{\mathcal{V}}/\mathcal{F})^3)$ &  $UO^{\mathcal{V}}+\frac{\mathcal{F}_s}{N}$ & 100\%         \\ \hline

\end{tabular}
\end{table}
Algorithm HO for model SGF is 100\% accurate, as graphs are dropped only based on the number of objects. Its main drawback is that it requires two passes, one for building the histogram and another pass over the histogram. Building the histogram can hamper efficiency if the video size is very large. Between the two algorithms for model SGF, GC\_Heuristic is more efficient in terms of CPU time complexity, though it sacrifices accuracy. For both of these algorithms for model SGF, the CPU time complexity is dominated by clustering (K-means with Elbow), which incurs a cubic cost. In the worst case, the GC\_Heuristic algorithm will take more I/O time than HO, as it will need to fetch the information of all the $\mathcal{F}$ frames to perform a clustering (either simple/K-means) on. In the average case, both algorithms' I/O complexity will be the same.

The VT algorithm for models SGV and MGV is also 100\% accurate (when compared to baseline). Algorithm VT is more efficient if model MGV (with N graphs) is used than model SGV with one graph on average. Note, depending upon $N$, the characteristics of the graphs in model MGV, this algorithm may not be able to optimize efficiency after a certain point. This depends upon how the connected components are distributed across graphs. It is possible that there can be a very large connected component in a graph with a large number of frames (closer to $\mathcal{F}_s$). In those cases, the performance may become closer to $SGV$. Without breaking the connected components across the graphs, performance may not be optimized after a certain point for MGV!

If HO (model SGF) and VT (model SGV) are compared, the VT algorithm does not require extra time for building histograms, as frames are grouped in some order in the vertex label. However, vertex scanning time takes some time for this algorithm. In videos, typically objects appear for a very small number of frames, and the number of objects in a video is much less than the number of frames. Hence, scanning the vertex labels for all the labels for VT will take less time than scanning the frames and building histograms for HO, even though the average case CPU time complexity is the same for both algorithms. This we have showcased, later in the experiments.

 If all the algorithms with 100\% accuracy across the three models are compared, model MGV will be the most efficient on average, and model MGV should be preferred. If performance is the priority, then GC\_Heuristic should be preferred.



\subsection{Experimental Evaluation}
In this section, we will elaborate on the experimental analysis for the situation in question. Before that, certain quirks of experimental evaluation for situation analysis in general is discussed below. 

The current state-of-the-art VCE algorithms have certain limitations, as mentioned several times in this thesis. Because of this, there is a discrepancy between what we perceive as a situation instance in a video (or actual ground truth) and what can be considered as a situation from extracted outputs from VCE algorithms. Additionally, manually labeling videos (specifically for medium and large-sized videos), which can contain multiple instances of the situation of interest, is challenging. Apart from this, the existing systems (or custom algorithms) for addressing some of the situations in this thesis are not open-source, or their datasets are not publicly available. Hence, setting up a \textbf{baseline against which the algorithms can be compared is a challenging task}. For some situations, it is possible to develop an exhaustive solution and compare the proposed algorithms' accuracy and performance against it. However, it might not be possible to develop an exhaustive solution for all different kinds of situations (which we have showcased later for a different situation).

The analysis algorithms/operators developed in this thesis are evaluated based on accuracy, efficiency, and robustness. They are described below:

\noindent \underline{\textit{Accuracy:}} Depending upon the situation, the accuracy metric definition can differ. For this situation, the True positives (TP) are defined as the number of frames present in both $baseline$ and $OC$, False positives (FP) are the number of frames in $OC$ but not present in $baseline$, and False negatives (FN) are the number of frames present in $baseline$ but not in $OC$. The accuracy is computed using the F1-score using the following formula. 

\[
\text{F1-score} = 2 \times \frac{\text{Precision} \times \text{Recall}}{\text{Precision} + \text{Recall}}, \quad 
\text{where Precision} = \frac{TP}{TP+FP}, \; 
\text{Recall} = \frac{TP}{TP+FN}
\]

\noindent \underline{\textit{Robustness:}} 
 In videos, it is possible that multiple instances of the same situation occur at different times (or after a certain number of frames). It is also true that numerous different types of situations can occur in a video. An operator or algorithm is highly robust if the same accuracy is obtained \textit{with and without the presence of other situations} in the video. For this validation, all the queries or algorithms are evaluated on several Mixed videos. These videos are generated in a manner that contains multiple different situations at various points in time, as well as instances of the same situation at different points in time. This establishes that the proposed algorithms and operators can accurately detect a situation, even in the presence of other arbitrary situations.

\noindent\underline{\textit{Efficiency and Scalability:}}
\noindent Efficiency is the time taken for the same computation using different operators or algorithms. Scalability measures the performance of the operators or algorithms when the video size increases. The goal is to develop algorithms or operators that are more efficient than the baseline algorithm (if it can be developed), yet still obtain the correct result. The efficiency analysis is carried out by measuring and comparing the response time of each algorithm or operator with respect to the baseline algorithm (if available).

\subsection{Dataset}
Selecting the appropriate dataset is crucial for evaluating the accuracy, robustness, and efficiency of the queries/analysis algorithms developed for the various situations addressed in this thesis. For this, it is essential to ensure that the dataset contains multiple instances of both known and arbitrary situations and is sufficiently large (in hours).

Over the years, several large-scale action recognition datasets have been created, including UCF-101~\cite{DataSet/ucf101}, HMDB-51~\cite{DataSet/HMDB}, and ActivityNet~\cite{DataSet/Activitynet}. These contain short video clips (1–10 seconds) of simple actions (e.g., walking, sitting, picking up objects), but they do not capture the complex situations of interest in this thesis (e.g., group formation, interactions between individuals over time). The UT-Interaction~\cite{DataSet/UT-Interaction-Data} dataset, although smaller in size, includes complex scenarios and multiple situations occurring simultaneously, making it particularly valuable for evaluation for this thesis.

Other datasets have been designed for specific domains: CAMNET~\cite{DataSet/Camnet} (CCTV tracking), Toyota Smart Home~\cite{DataSet/Toyota_Smarthome} (elderly daily activities), DETRAC~\cite{Dataset/CVIU_UA-DETRAC} and Urban-Tracker~\cite{DataSet/jodoin2014urban} (traffic monitoring), and VIRAT~\cite{VIRATDataSet} (parking lot activities). Compared to action recognition datasets, these datasets feature longer videos, more objects, and diverse situations, often with multiple simultaneous events, making them useful for evaluating the accuracy, robustness, and scalability of the approaches in this thesis. We have carefully chosen videos that contain situations of interest (multiple of them) from these datasets.

 However, the available datasets discussed above do not encompass all the relevant situations of interest in this thesis. Hence, the MavVid~\cite{DataSet/MavVid} dataset was prepared in ITLAB, containing videos (e.g., two people entering and exiting within n minutes of each other, two people crossing, etc.) of situations of interest addressed in this thesis. Additionally, for some situations specifically relevant to the Assisted Living domain (e.g., staying isolated, etc.), relevant videos were downloaded from YouTube, as it was difficult to create videos relevant to those situations. 

\begin{table}[!htb]
\scriptsize
    \caption{Dataset description. SL is surveillance, CM is civic monitoring, and AL is assisted living.}
    \centering
    \begin{tabular}{|p{.10\textwidth}|p{.10\textwidth}|p{.20\textwidth}|p{.10\textwidth}|p{.15\textwidth}|p{.1\textwidth}|}
    \hline
         Video Category & Length & Dataset & No. of Videos  & Average No. of frames &  Domain  \\ \hline
         Small &  $< 5$ min. & MavVid & 26 & 1348 & AL/SL \\ \cline{3-6}
         ~ & ~ & UT-Interaction & 23 &  1895  & SL/CM \\ \cline{3-6}
        ~ & ~ & DETRAC & 5 & 1071 &  CM \\ \cline{3-6}
        ~ & ~ & Youtube & 2 & 4267 &  AL \\ \cline{3-6}
    ~ & ~ & Urban-Tracker & 3 & 2210 &  CM \\ \hline
    \multicolumn{2}{|c|}{\textbf{Total small videos}} & ~ & 59 & 1667 & ~ \\ \hline
       Medium & 5 min.-15 min. & MavVid & 1 & 41800 & SL/CM \\ \cline{3-6}
        ~ & ~ & Youtube & 5 & 19030  & AL \\ \cline{3-6}
    ~ & ~ & Urban-Tracker & 1 & 4542  & CM \\ \cline{3-6}
    ~ & ~ & VIRAT & 6 & 17742 & CM \\ \hline
    \multicolumn{2}{|c|}{\textbf{Total medium videos}} & ~ & 13 & 19073  & ~  \\ \hline
     Large & 15 min.-1hour & MavVid & 1 & 75780  & SL/CM \\ \cline{3-6}
        ~ & ~ &Toyota SmartHome & 7 & 30460  & AL \\ \cline{3-6}
    ~ & ~ & CAMNET & 10 &  48488  & CM \\ \cline{3-6}
    ~ & ~ & VIRAT & 1 & 29589  & CM \\ \hline
    \multicolumn{3}{|c|}{\textbf{Total large videos}} & 19 & 42915 & ~\\ \hline
    Mixed-Single & $< 5$ min. & MavVidR & 4 & 2278  & SL/CM \\ \cline{3-6}
    ~ &  15min.-30min. & MavVidR & 1 & 7385 &  SL/CM/AL \\ \cline{2-6}
        ~ &  1 hour. & MavVidR & 3 & 35912 & SL/CM/AL \\ \hline
    Mixed-Random & 5-15 min. & MavVidR & 4 & 13216 & SL/CM/AL \\ \cline{2-6}
~ & 15-30 min. &  MavVidR & 3 & 44562 & SL/CM/AL \\ \cline{2-6}
~ & 1 hour  & MavVidR & 3 & 87931 & SL/CM/AL \\ \hline
\multicolumn{3}{|c|}{\textbf{Total mixed videos}} & 18 & 38206 & ~ \\ \hline
\multicolumn{3}{|c|}{\textbf{Total videos}} & 109 & ~ & ~ \\ \hline
    \end{tabular}
    \label{tab:dataset}
\end{table}

A summary of the datasets used in this thesis is shown in Table~\ref{tab:dataset}. The videos in the above datasets are categorized as small, medium, and large. The small videos are 1 to 5 minutes long and contain only one/two known situation instances (from the situations being addressed in this thesis) per video. The medium videos are 5 to 15 minutes long and contain multiple situations and boundary cases for the situations addressed in this thesis. The large videos have a length of 15 minutes to 1 hour. There is a scarcity of datasets containing videos of more than 30 minutes. Even the CAMNET dataset does not contain videos longer than 30 minutes. Hence, we have concatenated the videos (having the same background) from the CAMNET dataset to generate two 1-hour videos in the large category. The small and medium videos are used to evaluate the accuracy of the operators/analysis algorithms for identifying individual situations, when single and multiple instances of the same situations are present. The videos in the large dataset evaluate accuracy when multiple instances of the same and different situation types are present, as well as efficiency when the video size increases.

As mentioned earlier, a robust situation analysis system should be able to correctly identify all the situations in a video, even in the presence of arbitrary situations. For this purpose, we have created a complex mixed dataset category (named MavVidR~\cite{DataSet/MavVid_merged}). This dataset contains two different types of videos, mixed-single and mixed-random. The mixed-single videos are generated by concatenating arbitrary videos (or video fragments) from large and medium videos, with the videos from the small category containing known situations. The mixed-random dataset was generated by randomly selecting a number of videos from small, medium, and large categories and then concatenating them.

\subsection{Experiment Results}
\noindent \underline{\textit{Baseline generation:}} For the situation ``Finding group(s) of different size'', $baseline$ generated for the two models SGF and SGV separately. 
For model SGF, $baseline$ is generated by performing K-means and Elbow Analysis on each graph with more than 2 vertices, filtering out clusters of size s or above, within the range [s,p], or the largest clusters using Algorithm~\ref{alg:heuristic-algo}. For model SGV, $baseline$ is generated by traversing each vertex of the graph and then applying K-means and Elbow to each frame in the vertex label that has more than 2 objects using Algorithm~\ref{alg:SGV-algo-group-computation}. Although the above will generate the same output for both models, they are generated individually to fairly compare the response times of the algorithms within each model.


\noindent \underline{\textit{Experiment set up:}} For all the analysis in this thesis, the videos are pre-processed using YOLOv8~\cite{objectRecognintion/yolov8}, Bot-sort~\cite{objectReidentification/BotSort}, and HRNet~\cite{PoseEstimation/HRNeT} implementation in Python on an NVIDIA Quadro RTX 5000 GPU with 10 GB of memory. The graph-analysis algorithms are implemented using Python, and experiments were conducted on a machine equipped with 2 Intel Xeon processors (48 cores, 756 GB of main memory).

 We have experimented with all the videos in Table~\ref{tab:dataset} to find the largest cluster, as this is the most complex task. We have also experimented with different values of size $s$ for models SGV and SGF. Finally, experiments are conducted to determine the largest cluster on model MGV (with varying values of $N$) for large videos, allowing for an analysis of efficiency. All experiments are conducted on the given class label  ``person''. Since the algorithms drop graphs or frames with $\leq$2 objects, they generated output for \textbf{38 small videos (out of 59) and 14 large videos (out of 19)}. The rest of the videos in the small and large categories have fewer than two objects on average in a frame. The algorithms generated outputs for all the medium and mixed videos.
Below, the experimental results for each model are discussed individually, and then the algorithms across models are compared based on accuracy and efficiency.


\subsection{Finding out the largest group(s) or clusters of individuals}
\noindent \underline{\textit{Model SGF Algorithms:}} Two algorithms are proposed for this model: GC\_Heuristic and HO. The GC\_Heuristic algorithm can employ either threshold-based heuristics ($JD$ or $CQ$) or cluster size-based heuristic $CS$. Experiments were conducted for each of the heuristics individually. For $JD$ heuristic experiments, threshold values of 0.5, 0.4, 0.3, and 0.2 were used. Similarly, the $CQ$ heuristic experiments were conducted on threshold values ranging from 0.15 to 0.5.

\begin{figure}[!ht]
    \centering
    \includegraphics[width=\linewidth, keepaspectratio=true]{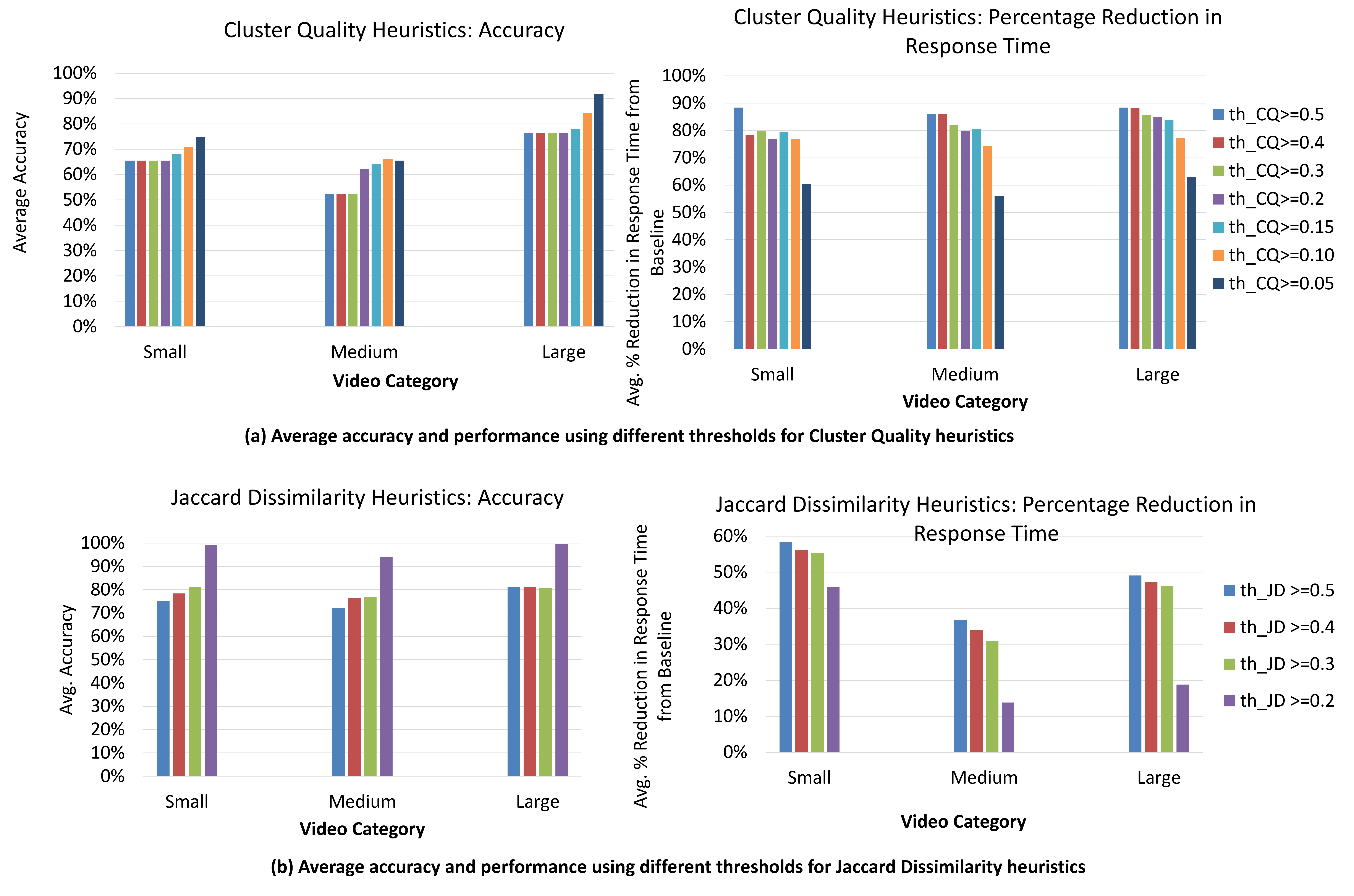}
    \caption{Average accuracy and performance on different categories of videos for threshold-based heuristics}
    \label{fig:gc-heuristic-threshold}
\end{figure}

The accuracy and percentage reduction from baseline for these heuristics on all small, medium, and large videos are shown in Figure~\ref{fig:gc-heuristic-threshold}. As we decrease the threshold value for $CQ$ or $JD$ heuristic, the number of frames K-means and Elbow increases, as the heuristic is satisfied in a larger number of frames. The more frames K-means and Elbow are used, the accuracy generally increases for both heuristics. For the $CQ$ heuristic, even when relaxing the threshold to a very small value of 0.05, the average accuracy remains below 80\% for small and medium videos, and for large videos, the highest accuracy obtained is 92\%. It achieves significantly better performance than the baseline (57\%-60\% reduction in response time at the smallest threshold value of 0.05). On the other hand, for $JD$ heuristic for the smallest threshold value (0.2), the accuracy of small, medium, and large videos is 99\%, 94\%, and \textbf{100\%} respectively. In comparison to $CQ$ heuristic, this heuristic achieves better accuracy. However, response time is reduced by 14\% for medium videos (whereas $CQ$ reduces it by 57\%) and by 19\% for large videos (whereas $CQ$ achieves 63\%) in comparison to the baseline. After adjusting the threshold to a very small value, accuracy improves to 100\% overall. However, performance does not improve drastically, and the improvement is significantly lower than that of the $CQ$ heuristic. In summary, \textit{it can be concluded that the $JD$ heuristic is good for accuracy and the $CQ$ heuristic is good for performance}. Overall, \textbf{neither can achieve good accuracy and performance together.}

\begin{figure}[h]
    \centering
    \includegraphics[width=1\linewidth, keepaspectratio=True]{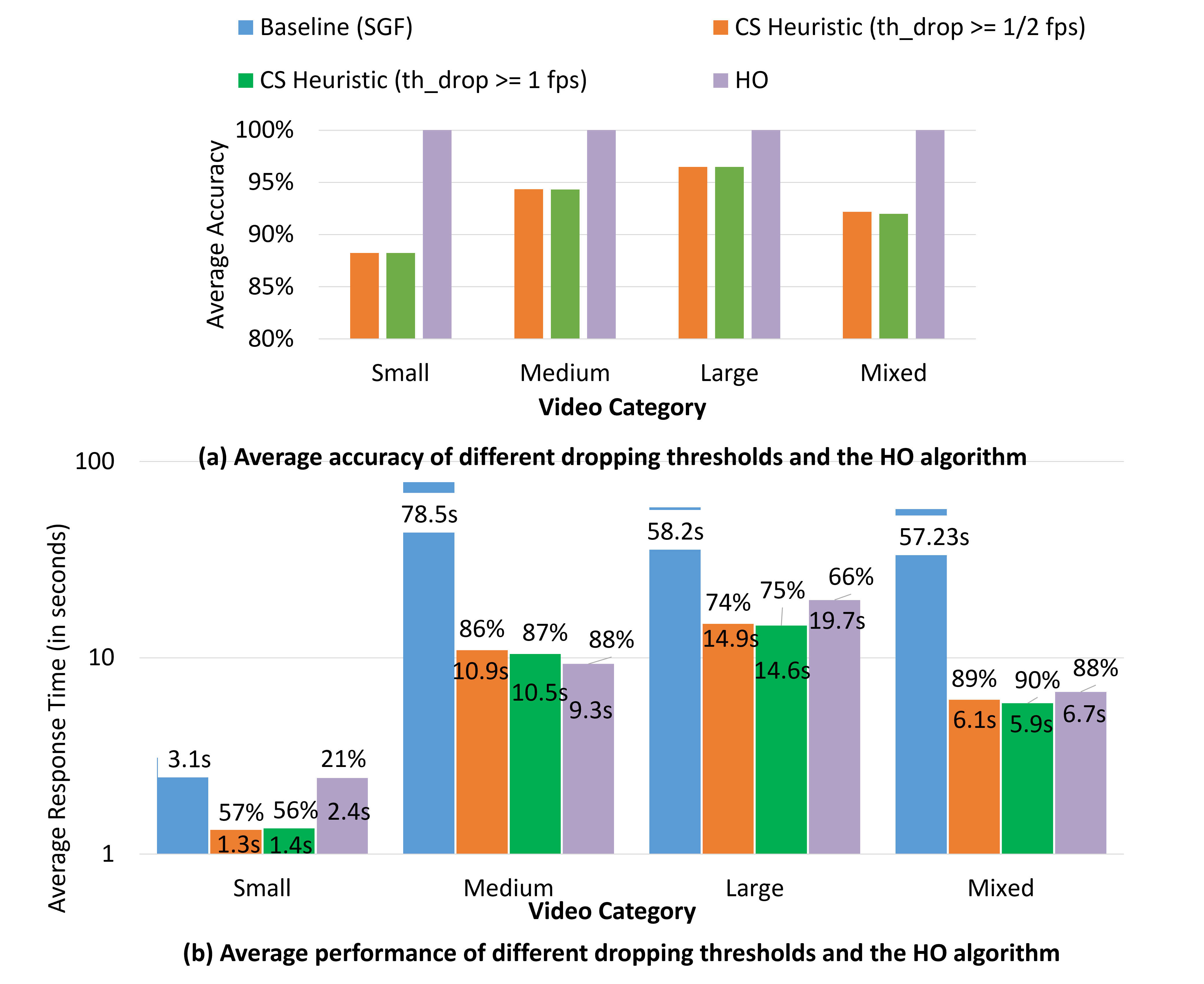}
    \caption{Average accuracy and performance on different categories of videos for GC\_Heuristic (with CS heuristic) and HO algorithm. The percentage reduction in response time from baseline is displayed at the top of the bar. Response time is plotted on a log scale.}
    \label{fig:gc-heuristic-Cs-Ho}
\end{figure}

The experiment results for the CS heuristic with different $th_{drop}$ (1 fps and 0.5 fps) values are shown in Figure.~\ref{fig:gc-heuristic-Cs-Ho}. In Figure~\ref {fig:gc-heuristic-Cs-Ho}, for all different categories of videos, the CS heuristic achieves above 88\% accuracy consistently for both $th_{drop} \geq 1/2~fps$ and $th_{drop} \geq 1 ~fps$. Similarly, the percentage reduction in response time from baseline ranges from  $56\%-90\%$ for different categories of videos. In general, there is no significant difference in performance across videos when $th_{drop}$ is changed from $1/2 ~fps$ to $1~fps$, and the range of percentage reduction in response time from baseline is from $56\%-90\%$. The reason is that in the majority of the videos, frames with fewer than $s$ objects appear for 1/2 second to 1 second. Additionally, the response time depends on which frame and how many objects were in a frame when K-means and Elbow were performed. Although there are slight differences in response time (in milliseconds) for both thresholds, averaging the response times for all videos in a specific category yields similar results. To better understand the performance, the number of frames for K-means and Elbow is performed on average for each video type is shown in Table~\ref{tab:Algo-K-means}. It can be seen from this table that both thresholds perform K-means and Elbow significantly fewer frames than the baseline for all video categories. However, on average, when $th_{drop} \geq 1 fps$ is used, it performs K-means and Elbow in a smaller number of frames than $th_{drop} \geq 1/2 fps$ (see Table~\ref{tab:Algo-K-means}, columns 4,5) for medium, large, and mixed videos. For small videos, both the thresholds perform K-means and Elbow on the same number of frames. The reason is that the majority of the small videos have the same set of objects, and the group(s) size does not change in general. As a result, even after changing the $th_{drop}$, it ends up performing K-means and Elbow on the same set of frames in the majority of the small videos. In general, both thresholds \textit{$th_{drop} \geq 1 fps$ should be used, as it achieves better performance} than $th_{drop} \geq 1/2 fps$ with the same accuracy.

Note that, for this heuristic, performance does not drop drastically for certain video categories (medium and large videos) as compared to the threshold-based heuristics. For example, for the medium videos, the CS heuristic with $th_{drop} \geq 1~fps$ achieves 94\% accuracy, and the percentage reduction in response time from baseline is 87\%. On the other hand, for the $JD$ heuristic with the smallest threshold value (0.2), it achieves 94\% accuracy with a percentage reduction in response time from baseline of 14\%, and for the $CQ$ heuristic with a threshold of 0.05, it achieves 66\% accuracy with a percentage reduction in response time from baseline of 56\%. This also occurs with large videos. On the other hand, for small videos, the $CQ$ heuristic with a threshold of 0.05 achieves a slightly better percentage reduction in response time from baseline (60\%) than the CS heuristics (56\%). However, accuracy is 75\% if the $CQ$ heuristic is used, whereas the CS heuristic achieves an accuracy of 88\%. For CS heuristics with 88\% accuracy, this slight performance drop is acceptable. \textit{In general, the CS heuristic is better than other threshold-based heuristics for various video types, achieving a \textbf{balance between accuracy and performance}.}

\begin{table}[H]
\centering
\scriptsize
\caption{Average number of frames K-means is performed for different video types.}
\label{tab:Algo-K-means}
\begin{tabular}{|p{.1\textwidth}|p{.05\textwidth}|p{.12\textwidth}|p{.08\textwidth}|p{.15\textwidth}|p{.15\textwidth}|p{.1\textwidth}|p{.1\textwidth}|}
\hline

\textbf{Video Type} & Avg. \# of Frames & Max cluster size (\# of videos) & \multicolumn{5}{c|}{\textbf{Avg. \# of Frames K-Means was Performed}} \\ \cline{4-8}
 ~ & ~ & & \textbf{Baseline} & \textbf{GC\_Heuristic (SGF) with CS ($th_{drop}\geq$ 1/2 fps)} & \textbf{GC\_Heuristic (SGF) with CS ($th_{drop}\geq$ 1 fps)} & \textbf{HO (SGF)} & \textbf{VT (SGV)}\\
\hline
Small (38 videos)  & 1923  & 2-3 (32 videos), $>3$ (6 videos) & 258  & 22  & 22  & 205 & 208 \\
\hline
Medium (13 videos) & 19073 &  3 (1 video), $>3$ (12 videos) & 6065 & 324 & 282 & 592 & 815 \\
\hline
Large  (14 videos) & 45372 & 2-3 (11 videos), $>3$ (3 videos) & 4334 & 403 & 375 & 1487 & 1487 \\
\hline
Mixed (18 videos) & 38026 & 2-3 (9 videos), $>3$ (9 videos) & 3934 & 162 & 142 & 426 & 428\\
\hline
\end{tabular}
\end{table}


In Figure~\ref{fig:gc-heuristic-Cs-Ho}, the accuracy and percentage reduction in response time from baseline of algorithm HO are also shown. This algorithm achieves 100\% accuracy for all the videos. The percentage reduction in response time is 21\% in small videos. The reason is that the \textbf{maximum cluster size is 2/3 in 32 out of 38 small videos}, and it ultimately performs K-means and Elbow on all the frames in those 32 videos. It is also evident from Table~\ref{tab:Algo-K-means}, as it can be seen that $baseline$ and algorithm HO perform K-means and Elbow approximately on the same number of frames for small videos. This trend is also evident in large videos, as it has the second-lowest percentage reduction in response time from baseline, and the maximum cluster size is 2/3 in 11 out of 14 large videos. The algorithm performs K-means and elbow in 65\% fewer frames than $baseline$, as shown in Table~\ref{tab:Algo-K-means}, row 3, column 6. For medium and mixed videos, the percentage reduction in response time from baseline is significantly higher than for small and large videos, as the maximum cluster size ranges between 2 and 18 in these videos. It also performs K-means and Elbow around 90\% fewer frames than $baseline$ (as shown in Table~\ref{tab:Algo-K-means}, row 2,3, column 6) in medium and mixed videos. It is evident that, depending on the maximum cluster size, the number of dropped frames increases/decreases.

\begin{figure}
    \centering
    \includegraphics[width=1\linewidth]{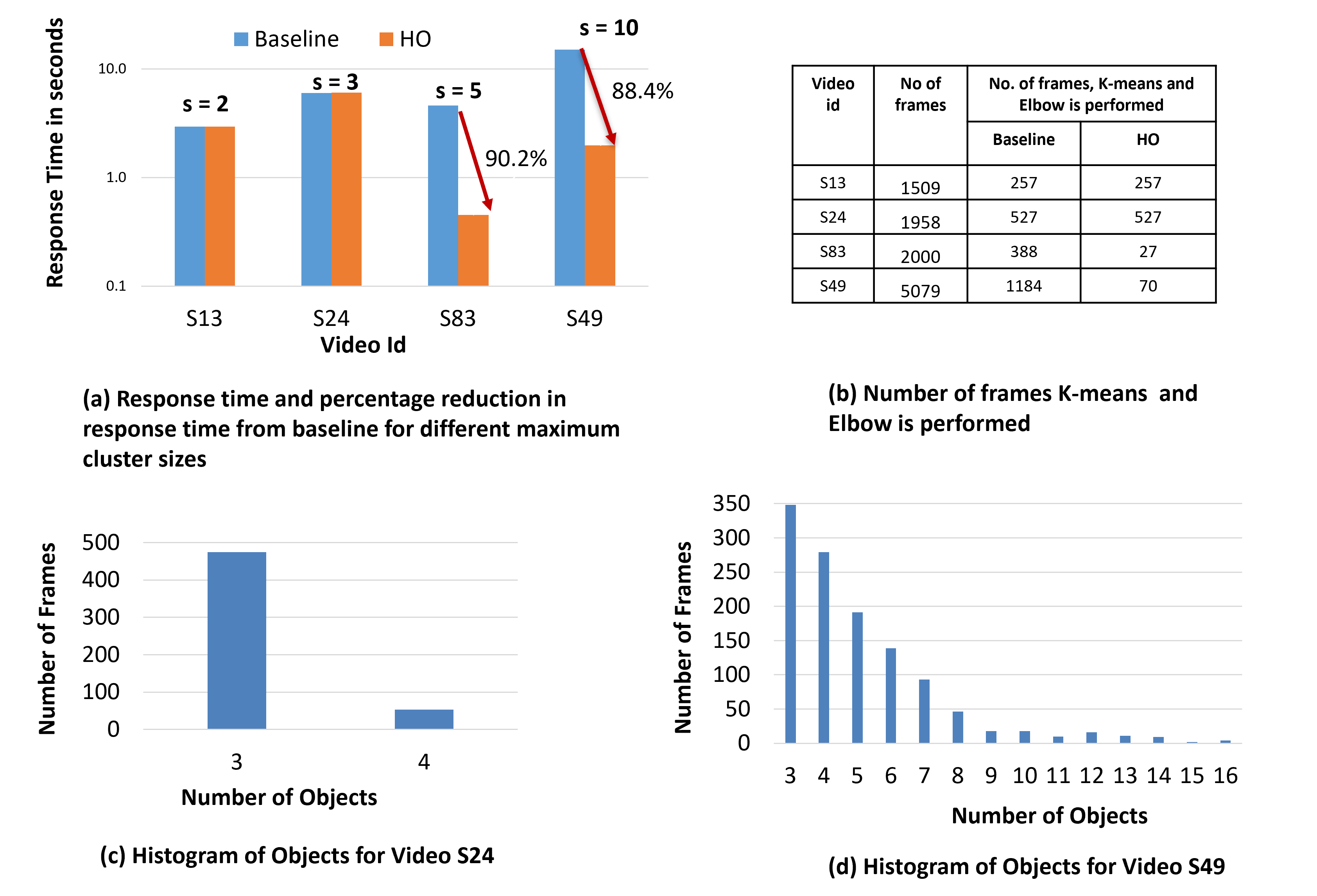}
    \caption{Response time (plotted in log scale) of sample small videos and the histogram of objects of two videos. Here $s$ is the maximum cluster size.}
    \label{fig:Ho-video-char}
\end{figure}

In Figure~\ref{fig:Ho-video-char}(a), the response time of four different small videos with different maximum cluster sizes is shown for algorithm HO. For videos S13 and S24, the maximum cluster size is 2 and 3, respectively. Algorithm HO performs K-means and Elbow on the same number of frames as $baseline$ (as shown in Figure~\ref{fig:Ho-video-char} (b), row 2,3) in these two videos. In Figure~\ref{fig:Ho-video-char}(c), the histogram for video S24 is shown. As can be seen, it has all 527 frames with either 3/4 objects, and the maximum cluster size is 3. Hence, HO and $baseline$ perform K-means and Elbow in all 527 frames. All the 32 videos that have the same response time as $baseline$ have the same type of histogram as S24. For videos S83 and S49, the maximum cluster size is 5 and 10, respectively, and they are dropping more frames in comparison to S13 and S24, as shown in Figure~\ref{fig:Ho-video-char}(b), rows 3,4. The histogram of video S49 is shown in Figure~\ref{fig:Ho-video-char}(d), where it can be seen that very few frames have more than 10 objects. Among the 38 small videos, six have histograms similar to S49. As a result, the percentage reduction in response time from baseline is around 80\% in those videos. It proves our initial statement that the larger the maximum cluster size in the video, the better the performance of the algorithm HO.

If the GC\_Heuristic algorithm (with CS heuristic, $th_{drop} \geq 1 fps$) and the HO algorithm are compared, it is observed that GC\_Heuristic outperforms HO at the expense of a slight loss in accuracy. Across small, medium, large, and mixed videos, the GC\_Heuristic algorithm performs K-means and Elbow in significantly fewer frames than the $baseline$ and HO algorithms. Hence, it can be concluded that among the model SGF algorithms, if accuracy is a priority, the HO algorithm should be chosen; if performance is a priority, the GC\_Heuristic algorithm with CS Heuristic ($th_{drop} \geq 1 fps$) should be chosen.

\begin{figure}
    \centering
    \includegraphics[width=1\linewidth]{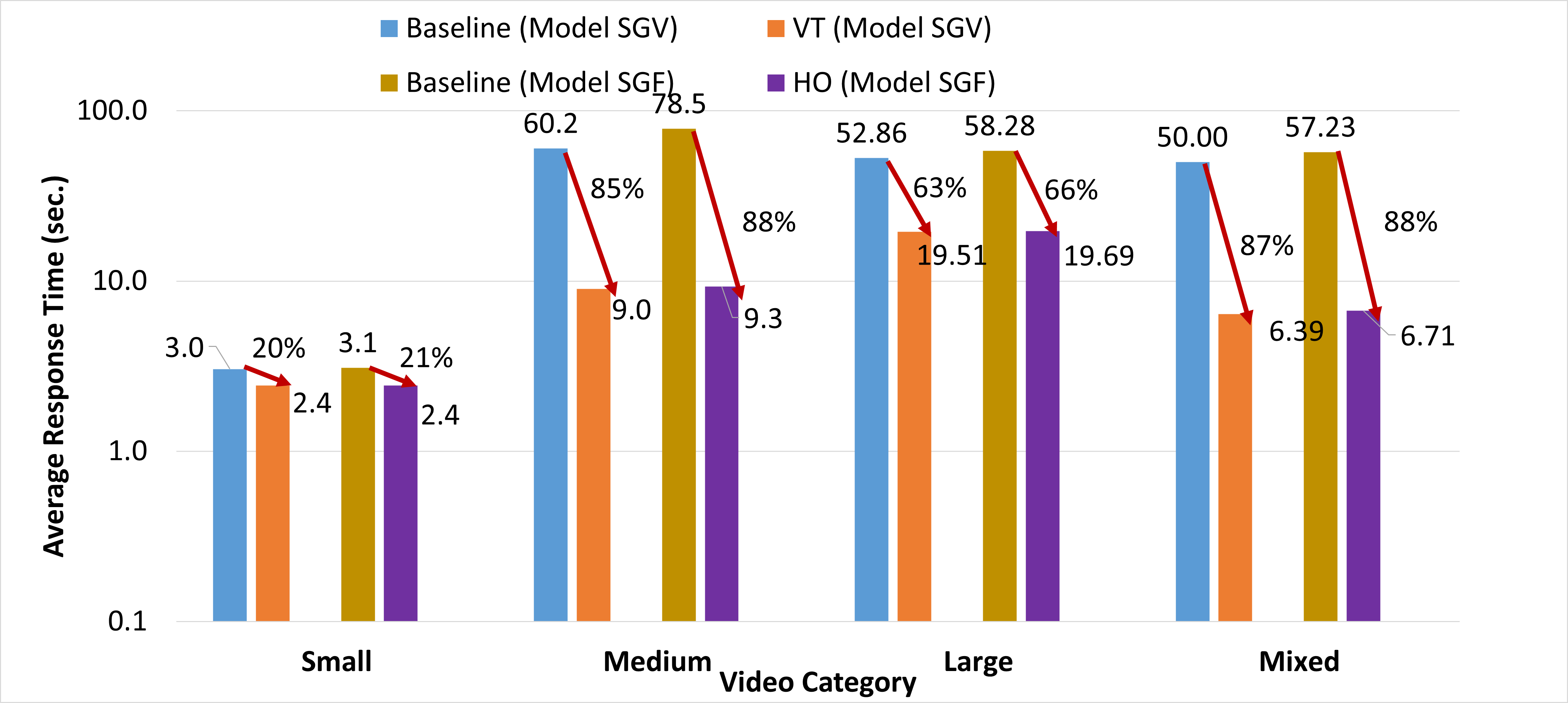}
    \caption{Response time (plotted in log scale) for VT algorithm for model SGV and HO algorithm for model SGF}
    \label{fig:SGV_VT}
\end{figure}

\noindent \underline{\textit{Vertex Traversal Algorithm using Model SGV}:} For this algorithm, we have experimented with all the small, medium, large, and mixed videos using the graphs generated by model SGV (number of graphs = 1). For all the videos, this algorithm obtained 100\% accuracy. The response time of the VT algorithm and the performance reduction percentage from baseline are shown in Figure~\ref{fig:SGV_VT}. Since this algorithm is designed based on the same principle as algorithm HO (for model SGF), it behaves similarly to algorithm HO. This means that for the same 32 small videos (out of 38), it has the same response time as $baseline$, and 11 out of 14 large videos have the same response time as $baseline$ because they have a maximum cluster size of 2/3. This algorithm also performs better for medium and mixed videos due to having a larger cluster size in those videos. Algorithm VT (for model SGV) has a better response time than algorithm HO (for model SGF) in general for medium, large, and mixed videos, as shown in Figure~\ref{fig:SGV_VT}. The reason is that it does not require extra time for building a histogram, as the frames are grouped in some order.

 For some videos (specifically medium videos), the VT algorithm performs K-means and Elbow in a larger number of frames than HO (see row 2 in Table~\ref{tab:Algo-K-means}). In those videos, the VT algorithm finds out the correct maximum cluster size after traversing a certain number of nodes. On the other hand, the HO algorithm finds the correct max cluster size earlier because it processes the graphs in decreasing order of the number of objects. However, in those videos, HO spends a significant amount of time building the histogram by grouping graphs. As a result, even after performing K-means and Elbow in more frames, the VT algorithm performs better than HO in medium videos.
\begin{figure}
    \centering
    \includegraphics[width=1\linewidth]{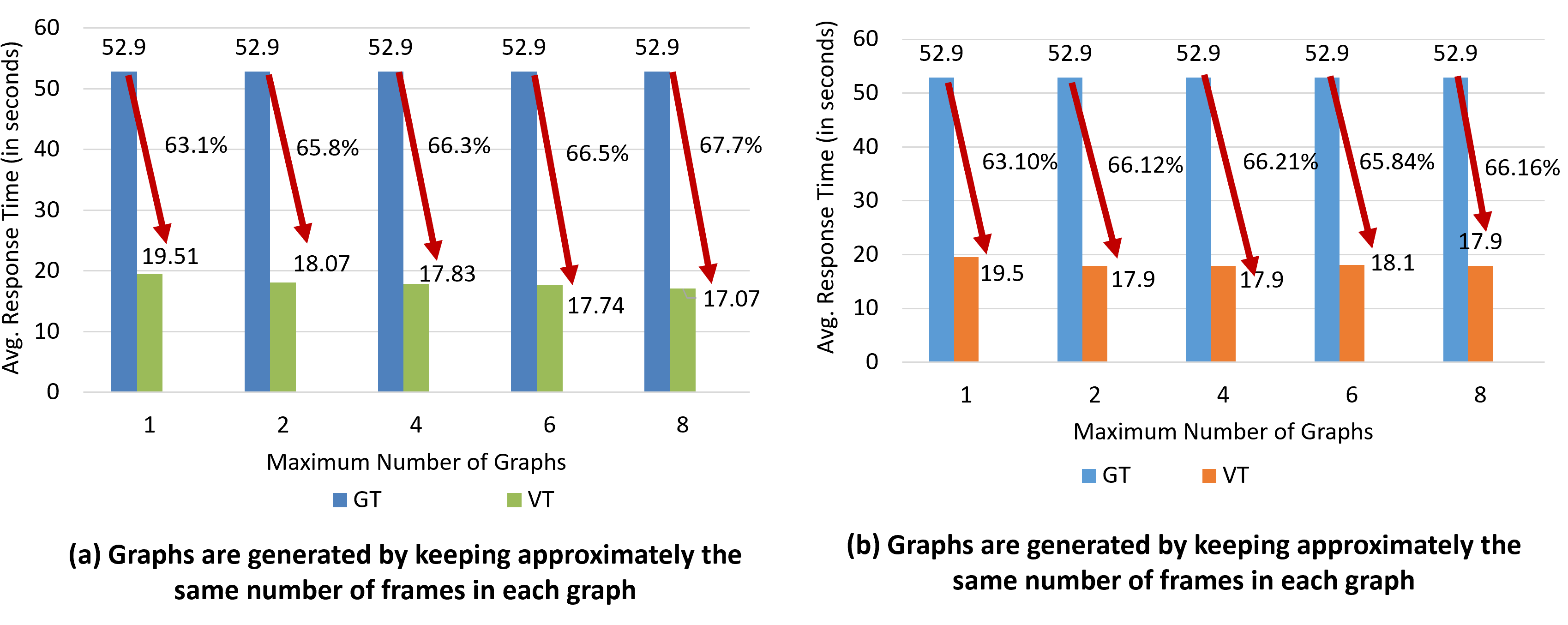}
    \caption{Performance of VT (using model MGV) for different numbers of graphs. The average response time was calculated by taking the average of the maximum response times for processing a graph of each video in the large category.}
    \label{fig:MGV-performance-Group-computation}
\end{figure}

\noindent \underline{\textit{Vertex Traversal Algorithm using Model MGV}:} As described earlier, algorithm VT can process all the graphs generated by model MGV in parallel. To stress test how much efficiency can be achieved by processing different numbers of graphs, we have set the value of maximum number of graphs that can be generated for model MGV ($maxGraph$ as 2,4,6, and 8) and generated graphs by having on average i) the same number of frames and ii) the same number of objects in each graph generated by model MGV. 
The response time is compared with the $baseline$ generated by model SGV (which has only 1 graph). Among all the graphs, the graph that takes the maximum amount of time is compared with the $baseline$ (generated by model SGV) response time. The performance of different values of $maxGraph$ and generating graphs by balancing the number of frames and the number of nodes is shown in Figure~\ref{fig:MGV-performance-Group-computation}. These experiments are conducted on large videos only.

\begin{figure}
    \centering
    \includegraphics[width=1\linewidth, keepaspectratio=True]{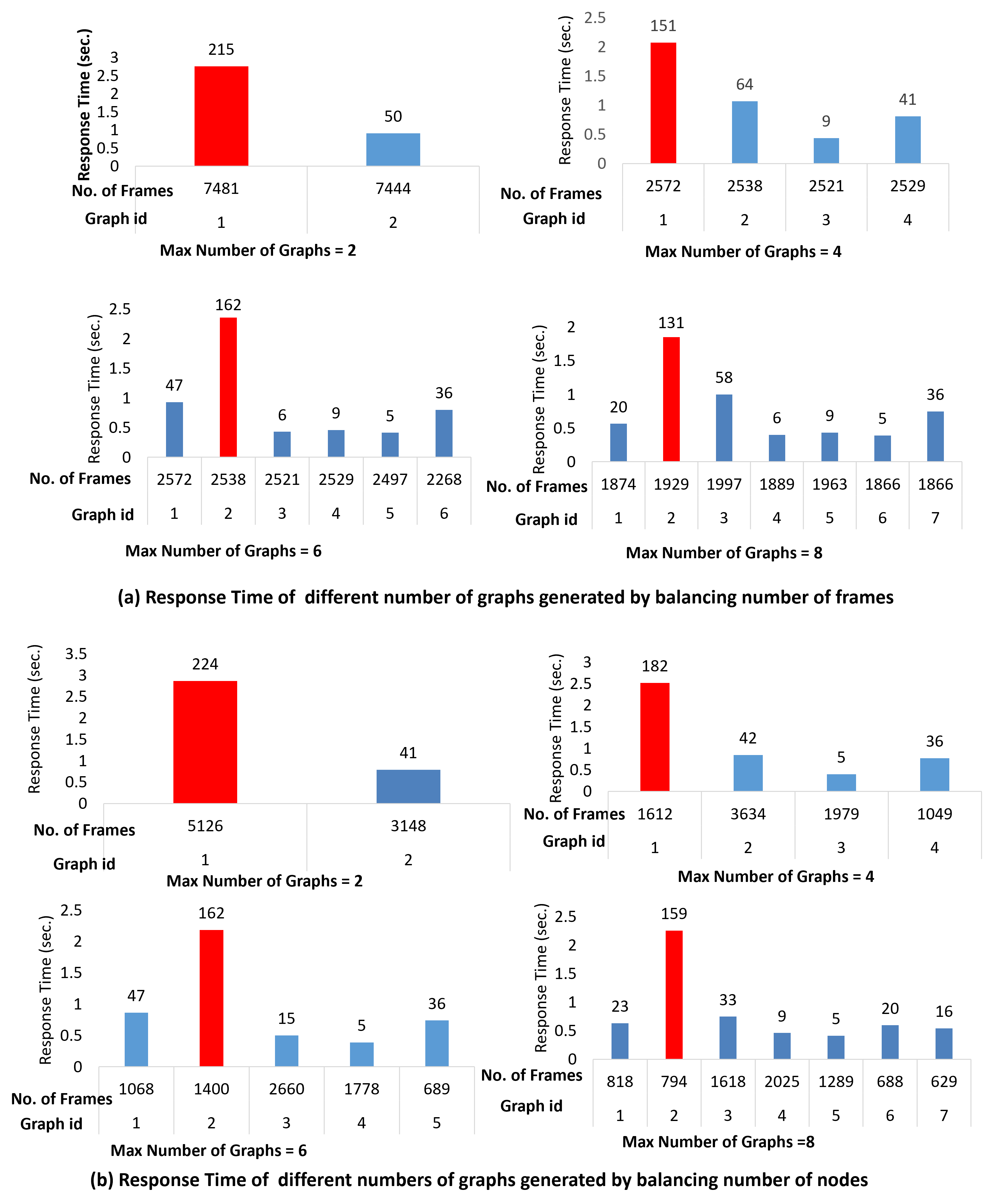}
    \caption{Performance of Algorithm VT for different numbers of graphs on video L9. The number of frames K-means and Elbow performed is displayed at the top of the bar. The graph with the maximum response time is highlighted using red color.}
    \label{fig:MGV-GC-individual-video}
\end{figure}

In this figure, when we generate the graphs by balancing the number of nodes, increasing the value of the maximum number of graphs (or $maxGraph$) does not improve performance. For $maxGraph$= 2,4, 6, and 8, the performance is the same, as the majority of the frames having max clusters are found in one graph. On the other hand, when the graphs are generated by balancing the number of frames, performance improves consistently as the value of $maxGraph$ increases. Note that the performance does not improve drastically after the value of $maxGraph$ is increased to 6 and 8. Because we are keeping the graphs disjoint, we cannot evenly distribute the frames across all the graphs. As a result, one graph can still have lots of frames in one connected component (and all of them can contain a max cluster, or we may end up finding the correct maximum cluster size very late). Efficiency optimization after this point is not possible.

To better showcase this, the maximum response time (and the number of frames for which K-means and Elbow are performed) for different values of $maxGraph$ (using two ways of generating MGV) for a video L9 is shown in Figure~\ref{fig:MGV-GC-individual-video}. Here, when the graphs are generated by balancing by frames, they have a better response time (as well as the number of frames K-means and Elbow is done), than when the graphs are generated by balancing the number of nodes. For example, when the maximum number of graphs is two and graphs are generated by balancing the number of frames, the graph that takes the maximum amount of time to be processed performs K-means and Elbow in 215 frames (shown in red bar in Figure~\ref{fig:MGV-GC-individual-video}(a)). When the graphs are generated by balancing the number of nodes, K-means and Elbow are performed in 224 frames (shown in red bar in Figure~\ref{fig:MGV-GC-individual-video}(b)). Similarly, when the maximum number of graphs is 8, balancing the frames allows K-means to be performed on 131 frames, whereas balancing the nodes takes the maximum amount of time, but it still performs K-means on 159 frames. Note, here, when we balance by nodes, each graph has an arbitrary number of frames, whereas when we balance by frames, each graph has approximately the same number of frames. Even if we end up performing K-means and Elbow on all the video frames of a Graph by balancing frames, we can get the best efficiency for this analysis. Furthermore, when we generate graphs by balancing on frames (max graph =8), we cannot reduce the number of frames, as K-means and Elbow are done beyond 131 frames, as all of them come from the same component. By balancing frames, we cannot further optimize performance by keeping the graphs disjoint.


\noindent \underline{\textit{Overall Comparison of All the Algorithms for Large Video}:} The summary of all the algorithms' accuracy and performance across models is summarized in Table~\ref{tab:summary-of-models-gc} for large videos. For algorithm VT, only the performance of model MGV (with a maximum number of graphs of 8) is shown, as it has the best performance on average among all different numbers of graphs, as explained above. In this table, if all the algorithms with 100\% accuracy are compared, then VT (using model MGV with the maximum number of graphs, 8) is the best for performance. 

Across all the algorithms GC\_Heuristic developed for model SGF is the most efficient among all the algorithms. Hence, if accuracy and performance are the priority, then VT (with maximum no. of graphs =8) should be chosen, and if performance is the priority with a small amount of loss of accuracy, then GC\_Heuristic algorithm (CS Heuristic, $th_{drop} \geq 1 fps$) should be chosen.

\begin{table}[H]
\scriptsize
\centering

\caption{\small{Summary of Accuracy and Performance of all Algorithms for Large Videos}}
\label{tab:summary-of-models-gc}
\begin{tabular}{|c|l|c|c|}
\hline
\textbf{Model} & \textbf{Algorithm} & \textbf{Accuracy} & \textbf{Response Time (seconds)} \\ \hline
SGF & HO &  \textbf{100\%} & 19.7s \\ \hline
MGV &  VT (maximum no. of graphs = 8) & \textbf{100\%} & 17.07s \\ \hline
SGF & GC\_Heuristic (CS Heuristic, $th_{drop} \geq 1 fps$) & 96\% & \textbf{14.6s}  \\ \hline
\end{tabular}

\end{table}

\subsection{Finding Clusters of Different Size} 
\begin{figure}
    \centering
    \includegraphics[width=1\linewidth, keepaspectratio=true]{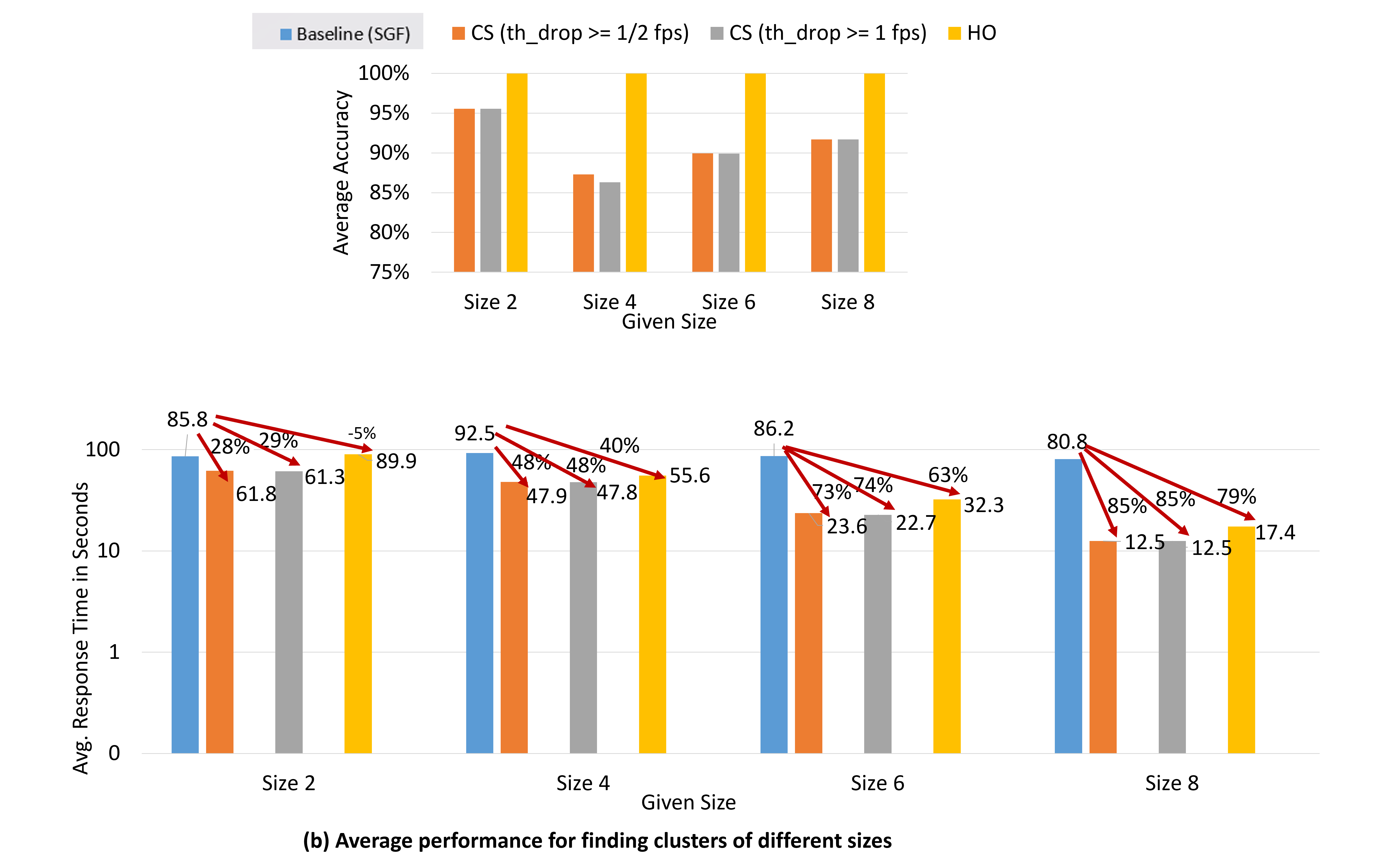}
    \caption{Accuracy and Performance of GC\_Heuristic and HO algorithm (model SGF) for finding clusters of given size s=2,4,8 on 5 medium videos. Response times are plotted on a log scale.}
    \label{fig:model-SGF-algo-clusters-of-diff-size}
\end{figure}

To showcase that the proposed algorithms can handle clusters of different sizes, we have experimented with the algorithms for finding clusters of given size, s = 2,4,6, and 8. For this set of experiments, 5 (out of 13) medium videos are chosen, which have cluster sizes ranging from 2-18. Among the five videos, the maximum cluster size is 8 in two videos, 9 in two videos, and 18 in one video. The experiment results for model SGF algorithms are shown in Figure~\ref{fig:model-SGF-algo-clusters-of-diff-size}.

The GC\_Heuristic algorithm, using the CS Heuristic, achieves the same accuracy and performance for both 1/2 fps and 1 fps thresholds across all different sizes. The accuracy varies for different values of the given size, as different numbers of frames can contain clusters of a given size. However, for all different sizes, the accuracy is above 87\% for this algorithm, which supports our initial claim that the heuristic is applicable for finding clusters of different sizes. As we increase the size, the number of dropped frames (having $<$ given size s objects) increases, and hence, performance also improves.

The HO algorithm shown in Figure~\ref{fig:model-SGF-algo-clusters-of-diff-size} always achieves 100\% accuracy for different sizes. The performance reduction percentage is -5\% when the given size is 2, because this algorithm performs K-means and Elbow on all the video frames. Additionally, it spends extra time building the Histogram for all the frames of the video. As the given size increases, this algorithm spends less time building histograms (i.e., builds a histogram for frames with $\geq s$ objects), and the number of dropped frames also increases. As a result, the performance improves. 

Among the two algorithms, if 100\% accuracy is not the goal, then the heuristic algorithm is suitable, even for finding smaller clusters, as it achieves reasonable accuracy and performance improvement. 
HO is good only if we want to find the largest cluster or clusters of large size with 100\% accuracy.

\begin{figure}
    \centering
    \includegraphics[width=0.7\linewidth, keepaspectratio=true]{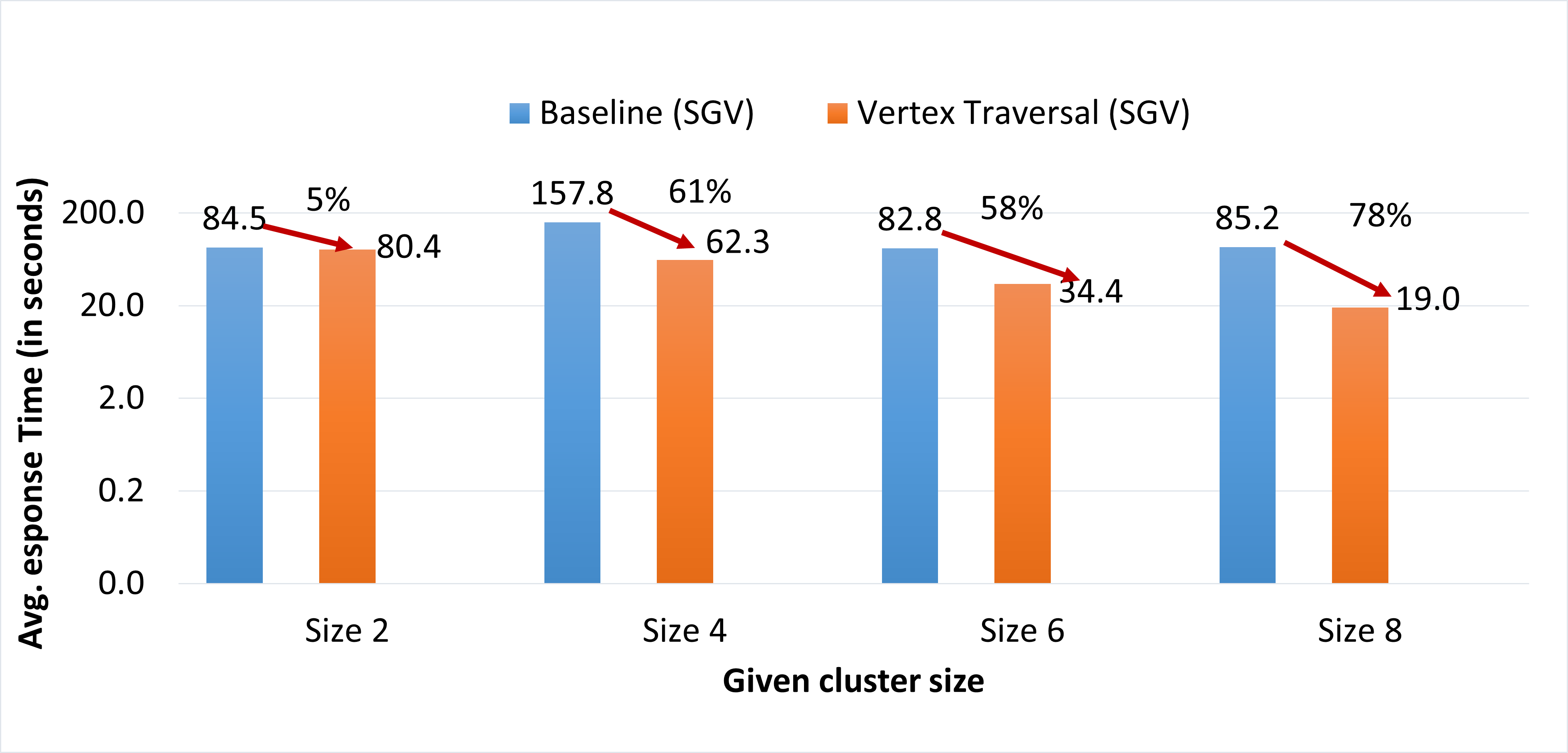}
    \caption{Accuracy and Performance of algorithm VT (for model SGV) for finding clusters of given size s=2,4,6,8. }
    \label{fig:model-SGV-algo-clusters-of-diff-size}
\end{figure}

We have also experimented with algorithm VT (max. number of graphs = 1 or model SGV) for given sizes s = 2, 4, 6, and 8. The experiment results are shown in Figure~\ref{fig:model-SGV-algo-clusters-of-diff-size}. For all different sizes, this algorithm achieves 100\% accuracy. The performance also improves when the size is increased. If both algorithms (VT and HO) with 100\% accuracy are compared, HO performs better than VT for all different sizes, even though both algorithms perform K-means and Elbow on the same number of frames for all the videos. The reason is that in all those videos, VT spends more time traversing vertices, and this time is more than the histogram building time of model HO on average. 


\section{Two objects moving closer/away from each other}
In videos, when two objects are moving closer or away from each other, the distances between these two objects monotonically increase or decrease for a certain amount of time (or number of frames). In this work, we do not include depth information and hence cannot separate which objects are in front or behind. The notion of moving closer or far apart is defined based on their spatial proximity in the image.

\begin{figure}[H]
    \centering
    \includegraphics[width=1\linewidth]{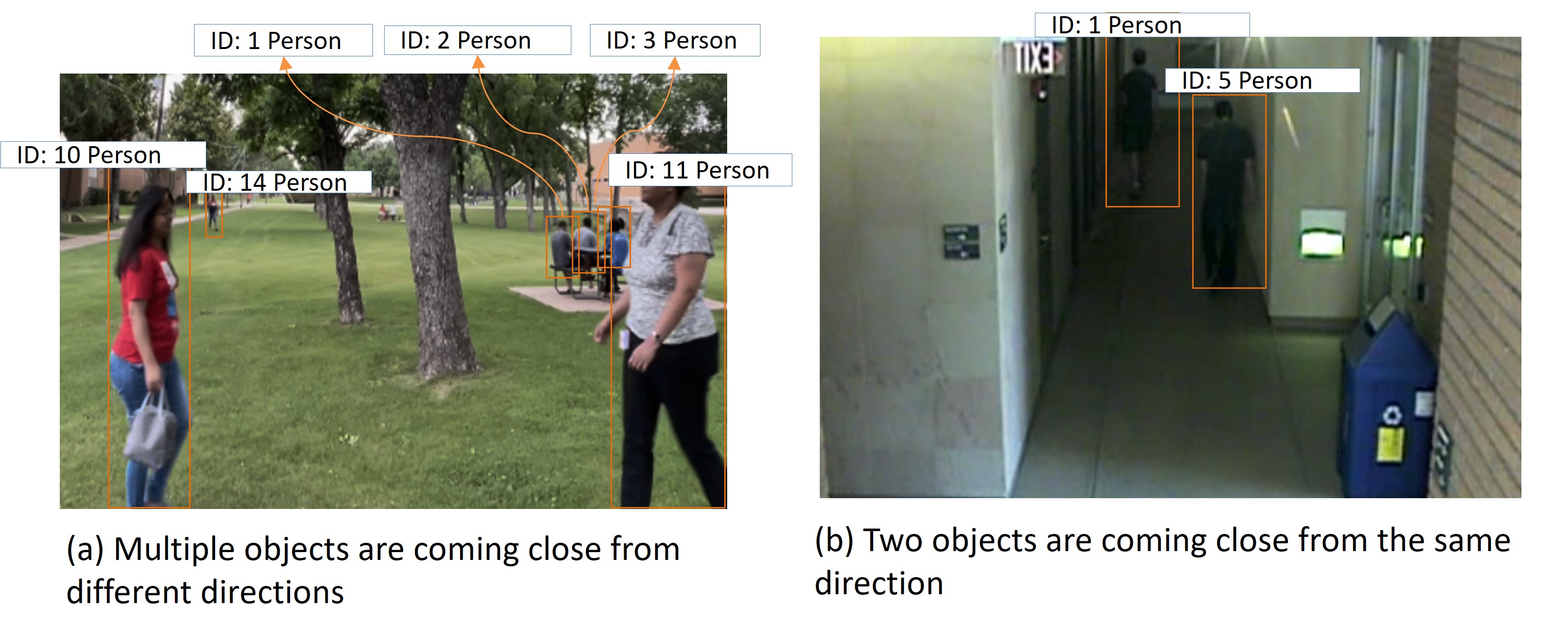}
    \caption{Example video frames from different videos containing the situation of two objects moving closer. }
    \label{fig:object-moving-closer-example}
\end{figure}

There can be several variations of this situation (shown in Figure~\ref{fig:object-moving-closer-example}). We will explain these variations in terms of objects moving closer for simplicity. The same variations hold when objects are moving far apart. The most straightforward case is when two objects are moving towards each other from opposite directions (shown in Figure~\ref{fig:object-moving-closer-example}(a), where objects 10 and 11 are coming closer to each other). Another possible example of moving closer is when one or more objects are not moving (e.g., sitting, standing, etc.) or are background objects (e.g., a check post or door).  One such example is shown in Figure~\ref{fig:object-moving-closer-example}(a), where object id 14 is coming closer to the person sitting on the bench. Note that here, object id's 10 and 11 are also coming closer to the objects sitting on the bench. Finally, it is possible that an object follows or moves behind another object, and at some point, they meet, depending upon the motion of the object that is following. This will happen when the second object (the object that is following) moves faster than the object is being followed. This is also considered an instance of two objects coming closer. In Figure~\ref{fig:object-moving-closer-example}(b), one such example is shown, where object id 5 is behind object id 1 and they are considered coming close. This phenomenon will not occur when objects are moving far apart.

The goal for this analysis is to find all the above cases of objects moving closer/farther apart. We can formulate the following problem definition for identifying the situation.

``\textit{Given a set of graphs $SGF^{\mathcal{V}},MGV^{\mathcal{V}}$ or a graph $SGV^{\mathcal{V}}$ frames, find out the situation instance (range of frames) where one/more pairs of objects of given class label $CL$ are coming closer/going away}".



\subsection{Challenges}   
To identify this situation, we need to analyze how the distance between object bounding box centroids changes over a sequence of frames. However, there are certain challenges in analyzing this situation.  

The bounding box locations estimated by the VCE algorithms are not exact across frames (even if the object pairs are not moving). Hence, for the same object pair, the distance between bounding box centroids will fluctuate across frames. This fluctuation will become larger when objects are moving at different rates and in different directions. As humans, when we look at these videos, we ignore these fluctuations. However, computing this situation in the presence of such fluctuations from the extracted video contents (without knowing what the video contains) is extremely difficult. 

One exhaustive solution to the above approach is to compare the bounding box centroids' distances in each frame they appear and check whether the distances decrease (for detecting if they are \textit{moving closer}) in consecutive frames or increase (for detecting if they are \textit{ moving away}). The consecutive frame sequence in which the two objects move closer or farther apart can be considered a situation instance. However, there are several problems with this approach. Because of fluctuating distance values across consecutive frames, a single situation instance can be detected as multiple instances, even though it is a single situation in the original video. Second, in videos, there may be lots of static objects or objects not moving but moving their body parts (e.g., a person waving their hands, sitting on a bench, etc.) This approach will treat all of these as situations, whereas they are actually minor object movements across frames. We term these minor movements as \textit{perturbations}.

To detect this situation, a supervised learning based approach is proposed by Samaras \textit{et al.} ~\cite{twoObjectClose/Samaras2012}, which identifies if two people are approaching each other. This approach does not identify multiple instances of the same situation (e.g., frames where an object pair is coming closer or when multiple object pairs are coming closer in the video at the same time). There are several other approaches that try to predict if two objects are going to collide from the acceleration and speed information of objects gathered from sensors~\cite{twoObjectClose/zeiller2025collision}. These works try to address a totally different type of problem (e.g., avoiding collisions between objects in real-time) than what we are addressing here.

In this thesis, we have proposed an algorithm for addressing the above issues of the exhaustive algorithm. There are certain other things that need to be mentioned from the VCE point of view. As mentioned several times in this thesis, the VCE output is an estimation and is not always correct. There are certain VCE quirks that are being addressed by the Image and Video Analysis (IVA) community. 

First, it is possible that a VCE algorithm may never detect a pair of objects together in a video frame. If those objects are not detected even if they come closer/go farther apart, it will not be possible to compute this situation from the VCE output. Moreover, even when the \textit{tracking} threshold is set equal to the video length, an object may still be assigned multiple object id's due to the inconsistency of the tracker. Consider a situation instance $S$ where two objects are coming closer for $f_i,\ldots,f_j$ consecutive frame sequences in the video. In this case, if an object is assigned two different object id's in frames $f_i,\ldots,f_j$, then an algorithm developed for processing only VCE output (without any other additional information) may detect multiple different instances. However, as VCE progresses, these inconsistencies will be minimized. Our goal is to develop a solution that can detect these situations, align as closely as possible with the manual ground truth, and filter out perturbations as much as possible. 



\subsection{How the situation can be addressed?}
Before explaining the proposed algorithm, we will give an overview of how each problem mentioned above can be solved, and then we will explain our algorithm. Let's assume an object id pair $o_i,o_j$ appears in the $\mathcal{F}_{o_i,o_j}$ sequence of frames together. The bounding box centroid distance between a pair of objects in a frame $f$ is denoted as $d_{o_i,o_j}^f$. The fluctuations in distance values across frames for an object pair can be reduced by smoothing. For example, we can smooth out the distance values by comparing the distances between $o_i,o_j$ in every $k^{th}$ frame that is $f, f+k,\ldots,f+nk, \ldots, $. If two objects' distance values are monotonically decreasing (for coming closer) or increasing (for coming far apart) from frame $f,\ldots,f+nk$, we will say the situation exists from frame  $f,\ldots,f+nk$ for these objects. This helps to suppress small fluctuations between consecutive frames and capture the overall direction of relative motion

Now comes the question, \textit{what should be the value of $k$, or how many frames should be skipped for an object pair to compare the distance values?} 
Intuitively, $k$ should be chosen to reflect the length of a situation instance in a video. However, in videos, objects may move at different rates, and it may not be possible to estimate a $k$ value applicable to all different motions in the video. However, our goal is to choose $k$ in such a manner that it can be used to identify, in general, most of the situation instances in the video. Care should also be taken so that $k$ is not too small, which may result in detecting perturbations as situation instances. Conversely, if $k$ is too large, it may overlook meaningful but short-lived instances. There are several alternatives for choosing $k$. They are described below with their advantages and disadvantages.

\noindent \underline{\textit{(i) Frame per second (fps) rate:}} A straightforward approach is to set $k$ equal to the $fps$ rate. Since $fps$ is a fixed value across different types of videos, it may not capture the diverse motion present in videos in general. Depending upon the object speed or movement rate, the situation in question may occur within fewer than $fps$ frames. If $fps$ is used, those instances will be missed.

\noindent \underline{\textit{(ii) Average number of frames an object appears in $avgDuration$ :}} This value is computed as $\frac{\mathcal{F}}{UO^{V}}$, where $UO^{\mathcal{V}}$ is the total number of unique object id instances in the video. This provides an approximation of how long, on average, an object remains in the field of view.  It serves as a \textit{useful approximation of the average length of a situation instance}, as the value changes depending on the video type and how objects are moving in the video. However, depending on the number of unique objects detected by VCE, this can be very small for some videos, introducing the problems of the exhaustive approach discussed above. This value can also become large, and in this case, many situation instances can be missed, specifically the object pairs that have come closer over a very short span of time. Since $avgDuration$ is an approximation from VCE output, we can use $k = \frac{avgDuration}{2}$ as well to identify the smaller situation instances. It may be possible that many situation instances occur in fewer $\frac{avgDuration}{2}$ number of frames, when $avgDuration$ becomes very large.

\noindent \underline{(iii)\textit{Minimum number of frames an object appears in the video or $minDuration$}}. This represents the smallest observed lifespan of an object and is accurately computed from VCE output as meta information. This value provides a fine-grained estimate of the shortest possible situation instance. Since $minDuration$ is directly computed from the VCE output, it offers a precise lower bound for situation length. The $minDuration$ can also become very low, specifically if VCE detects an object for a very small number of frames (e.g., 1/2 frames). $minDuration$ can also become very large if all the objects are present in a large number of frames. Note that all of the above information for computing $minDuration, avgDuration$, and $fps$ is already available as post-processed video meta information.

\noindent \underline{\textit{(iii) Automatically Choosing $k$}:}
To avoid the above extremes of $minDuration$ or $avgDuration$, an adaptive selection strategy can be used. We need to identify when $minDuration$ or $avgDuration$ becomes very small. It is possible that either of them or both of them become very small (e.g., 1 or 2 frames). In this case, if either one of the values is less than $\frac{1}{2} fps$, we can infer that these values are very small, and the chosen $k$ is $MAX (avgDuration,~minDuration)$. In this way, if either or both values are very small, the larger of the two will be chosen. This will also reduce the perturbations to some extent. To prevent $k$ from being excessively large, we can choose the minimum of $minDuration$ or $avgDuration$, when both of their values are more than $\frac{1}{2} fps$. This adaptive rule ensures that $k$ remains within a balanced range—neither too small to amplify fluctuations, nor too large to overlook short-lived but meaningful situation instances.

Finally, even after analyzing every $k$ frames for an object pair, some situations may arise due to fluctuations in distance values across frames, depending on which frames are chosen for comparison. For example, when objects are not moving (e.g., objects sitting on a bench shown in Figure~\ref{fig:object-moving-closer-example}(a)), the algorithm may identify that those object pairs are coming closer. Another important thing is that, in these cases, the distance values across frames will have very small differences and will be very close to each other, unlike when objects are really coming closer or farther apart. Let us assume the distance between a pair of object $o_i, o_j$ in frames $f$ and $f-k$ is $d_{o_i,o_i}^{f}$, $d_{o_i,o_i}^{f-k}$, and the absolute differences in distances is $\Delta_{o_i,o_j}^{f,f-k} = |d_{o_i,o_i}^{f}-d_{o_i,o_i}^{f-k}|$. If the $\Delta_{o_i,o_j}^{f_i,f_i+k}$ value is very small (less than a threshold), then we do not consider those frames for an object pair. In this way, the perturbations can be filtered out as much as possible.

Given all the aspects of addressing this situation, we have proposed an algorithm below. We have showcased how the same algorithm can be applied to models SGV and SGF individually below.


\begin{table}[!htb]
\centering
\scriptsize
\caption{\small Table of notations for the proposed of algorithms for identifying ``Two objects coming closer or going away''}
\label{tab:notations-comingclose-farapart}
\begin{tabular}{|p{0.18\textwidth}|p{0.77\textwidth}|}
\hline
\textbf{Notation} & \textbf{Description} \\
\hline
\textbf{Notation} & \textbf{Description and Example} \\ \hline
$SGF^{\mathcal{V}}$ & Set of input graphs generated by model SGF. \\ \hline
$SGV^{\mathcal{V}}$ & Input graphs generated by model SGV. \\ \hline
$MGV^{\mathcal{V}}$ & Set of input graphs generated by model MGV. \\ \hline
$BGF^{\mathcal{V}}$ & Base graph file(s) generated for model SGF/SGV/MGV \\ \hline
$IDF^{\mathcal{V}}$ & A data file indexed on frame id. Used to fetch bounding boxes for each object id in a frame. \\ \hline
$M^{\mathcal{V}}$ & Meta information associated with video $\mathcal{V}$ (stored in the graph file header). \\ \hline
$IDF^{\mathcal{V}}$ & A data file indexed on frame id. Used to fetch bounding boxes for each object id in a frame. \\ \hline
$CL$ & Given class label to filter objects of specific class (e.g., person). An optional input parameter. \\ \hline
$\epsilon$ & A threshold to filter out perturbation \\ \hline
$k$ & Number of frames to skip for comparison. \\
\hline
$d_{o_i,o_j}^f$ & The bounding box centroid distances between object id's $o_i,o_j$ in frame $f$ \\ \hline
$avgDuration$ & The number of frames an object appears on average. Computed as $F/UO^{\mathcal{V}}$ \\ \hline
$minDuration$ & The smallest number of frames an object appears in a video.   \\ \hline
$VD$ & A Vertex Dictionary containing vertex information of a graph, where key = vertex id or object id and value is an ordered list of frame ids in which the vertex appears. Used by the algorithm when MGV/SGV is used\\
\hline
$ED$ & An Edge Dictionary containing edge information of all the graphs generated by SGF. The key is an object id pair $(o_i, o_j)$ and value is a list containing a frame id $f$ and the bounding box centroid distances between $o_i, o_j$ in frame $f$ \\ \hline 
$Output$ & A dictionary containing the pairs of objects and the range of frames they are coming close or going far apart. Here, the key is an object id pair $(o_i,o_j)$, and the value is a list of frame intervals where the situation is detected. \\
\hline
\end{tabular}
\end{table}

\subsection{Model SGV and MGV algorithm}
As mentioned earlier, model SGV generates a graph $SGV^{\mathcal{V}}$, and the vertex have frame id information stored as part of vertex labels (stored in descending order of number of objects). In this model, there is an edge between a pair of objects $o_i, o_j$ if they were detected in one or more frames together. The edge labels store the minimum distance between object id pair $o_i, o_j$, and the sequence of frames having minimum distance, maximum distance between object id pair $o_i, o_j$, and the sequence of frames having maximum distance across all the frames $o_i, o_j$ appear. Note that the distances in each frame, $o_i, o_j$, appear not to be preserved in this model, and need to be computed when required.

To apply the above approach to the graph(s) of model SGV/MGV, we need to know the range of frames in which an object id appears in the whole video. This can be obtained by traversing the vertices and their labels. Once that information is available, the edges can be traversed. Now, there are several ways we can use the edges. One alternative is to use the information present in the edge label. We can take the frame id's stored as part of the edge label and then say a situation exists if the frame id's with the minimum distance appear later than the frame id's with the maximum distance. The intuition is that for an object pair that are coming closer, they will be the farthest, and when they are coming closer, they will be closest and have the minimum distance. However, in videos, it is possible that two objects start moving from two farthest points, come closer for some number of frames, and then go far apart and then come closest. The above approach cannot capture this case.

Another alternative is to take all the frame id's appearing in the edge label (for both minimum and maximum distance) and find out the smallest $f_s$ and largest frame id $f_i$ among them. Then analyze each $k^{th}$ frame in $f_s,\ldots,f_j$ using the above approach. Now, the main problem with this alternative is that the objects can come closer or far apart in other frames; they appear outside the range of frames $f_s, f_e$. Therefore, it is important to analyze the entire range of frames $o_i, o_k$ that appear, not a subset of them.

\begin{algorithm}[H]
\scriptsize
\caption{\small{Identifying if two objects are moving closer/far apart for model SGV/MGV}}
\label{alg:SGV-comingclose-farapart}
\begin{algorithmic}[1]
\Require A graph $G$ ($SGV^{\mathcal{V}}$ or $MGV_i^{\mathcal{V}}$), ${M^\mathcal{V}}$, $IDF^{\mathcal{V}}$,\underline{$CL$}, $\epsilon$, $direction$
\Ensure $Output$
\State $VD \gets \emptyset$, $Output \gets \emptyset$
\State Extract $fps$, $\mathcal{F}$, $minDuration$, and $UO^{\mathcal{V}}$ from $M^{\mathcal{V}}$
\State $avgDuration \gets \mathcal{F} / UO^{\mathcal{V}}$
\If{$minDuration < \frac{fps}{2}$ or $avgDuration < \frac{fps}{2}$ }
    \State $k \gets MAX (minDuration, avgDuration)$
\Else 
    \State $k \gets MIN (minDuration, avgDuration)$
\EndIf
\For{each vertex for object id $o_i$ in $G$}
    \If{$o_i$ has class label $CL$ in node label}
        \State Extract list of frame id's where $o_i$ appears (in sequential order) and put in $VD[o_i]$
    \EndIf
\EndFor

\For{each edge for object id pair $(o_i,o_j)$ in $G$}
    \If{$o_i, o_j$ in $VD$}
        \State $f_s, f_e \gets$ Extract first and last frames where $(o_i,o_j)$ appear from $VD$
        \State $f \gets f_s$
        \While{$f \le f_e$}
            \State Fetch the bounding boxes $BB_{o_i}^f$, and $BB_{o_j}^f$ from $IDF^{\mathcal{V}}$  
            \State$d_{o_i,o_j}^f \gets$ Distance between $BB_{o_i}^f$ and $BB_{o_j}^f$ centroid  
            \If{$|d_{o_i,o_j}^f -d_{o_i,o_j}^{f-k}| > \epsilon$} 
                \If {$(direction~=~''moving~closer''$ and $d_{o_i,o_j}^f  < d_{o_i,o_j}^{f-k})$ or $(direction~=~''far~apart''$ and $d_{o_i,o_j}^f  > d_{o_i,o_j}^{f-k})$}
                    \If{$Output[(o_i,o_j)]$ already contains a situation from frames $f-nk, f-k$}
                        \State  Update the frame interval to $f-nk, f$ 
                    \Else
                        \State Add new interval $f-k, f$ to $Output[(o_i,o_j)]$
                    \EndIf
                \EndIf
            \EndIf
            \State $f \gets f + k$
        \EndWhile
    \EndIf
\EndFor
\State \Return $Output$
\end{algorithmic}
\end{algorithm}

Finally, as mentioned earlier, there are notions of non-empty and empty frames in the VCE output. Now it is possible to analyze every $k$ frames from all the non-empty frames $o_i,o_k$ appear. However, given the quirks of VCE, it may skip a large number of non-empty frames, which could have been used for detecting the situation. For this, we must consider all frames, even if we end up analyzing consecutive non-empty frames for an object pair. 

By taking into account all the aspects of analysis, an algorithm is proposed in Algorithm~\ref{alg:SGV-comingclose-farapart} for model SGV/MGV. This algorithm takes as input a graph $G$ (stored as a graph file), an indexed (on frame id) data file $IDF^{\mathcal{V}}$, video meta information ${\mathcal{M^V}}$ (stored in graph file header), an optional parameter $CL$ to filter objects of given class, a threshold $\epsilon$ to filter perturbations, and a parameter $direction$. The threshold parameter $\epsilon$ is set (in terms of number of pixels). The $direction$ parameter accepts two values ``coming close'' or ``far apart'' to detect the two situations if objects are coming closer or going far apart.

Algorithm~\ref{alg:SGV-comingclose-farapart}, first extracts the $fps$, $minDuration$, total number of frames $\mathcal{F}$ in the video, total number of unique object id instances $UO^{\mathcal{V}}$ from the meta-information for video $\mathcal{V}$. It then computes the $avgDuration$ in line 3. In lines 4-8, the algorithm determines the value of $k$ using the formula discussed above for choosing $k$ adaptively. After it chooses a $k$, it starts iterating over the vertices input graph $G$, filters out vertices with given class label, extracts the list of frame id's a vertex or object id $o_i$ appears and stores it in the dictionary $VD$ (line 9-13). After that, it starts traversing the edges of the graph  $G$. For an edge for object id pair $(o_i, o_j)$ it extracts the first and last frame where $o_i$ and $o_j$ appear together in line 16. It then starts iterating over the frame id's $<f_s, f_s+k, \ldots, f_s+nk, \ldots, f_e>$. For each frame id $f$ in this sequence it fetches the bounding boxes of object id $o_i$ and $o_j$ for frame $f$ from $IDF^{\mathcal{V}}$, and computes the bounding box centroid distance of $o_i$ and $o_j$ in frame $f$. It also keeps track of the distance computed at frame $f-k$. If the absolute difference in distances in frame $f$ and $f-k$ is above the given threshold $\epsilon$, it then checks if the situation exists (line 21). If the given direction is ``moving closer'' then the algorithm checks if distances in frame $f-k$ and $f$ are decreasing, and when the direction is ``far apart'' it checks if the distances are increasing (line 22). Once one of the above conditions is satisfied, the algorithm checks if a situation is detected in previous frames from frame $f-nk$ to $f-k$ (previous frames). In this case, it updates the interval to $f-nk,f$ (line 24), which essentially means the distances are monotonically increasing or decreasing from frames $f-nk$ to $f-k$. If the situation has not been detected in the previous frame $f-k$, then it adds a new interval to $Output$, which means a new situation instance has been detected. The above is done for all pairs of objects by traversing the edges. 

This algorithm can be applied to the $\{MGV_1^{\mathcal{V}},\ldots,MGV_N^{\mathcal{V}}\}$ set of graphs generated by MGV in parallel. From this, we will obtain a list of partial output dictionaries, $<Ouput_1,\ldots,Ouput_N>$, where $Output_i$ is obtained from graph $MGV_i^{\mathcal{V}}$. After these final outputs are obtained, they are composed together by merging all the output dictionaries into a final $Output$ dictionary. Note, the composition phase is very simple here because the graphs do not have any object id's in common. Furthermore, the chosen $k$ is same for all the graphs of MGV as it is derived from the meta information for the whole video.


\noindent \underline{\textit{CPU Time Complexity of Algorithm~\ref{alg:SGV-comingclose-farapart}:}} For \textit{model SGV} for this algorithm, the main computation cost comes from traversing the node labels and extracting the frame id's in sequential order (building the dictionary $VD$), and from traversing the edges and computing distances. As mentioned earlier, for this model, there are $UO^{\mathcal{V}}$ nodes and $(UO^{\mathcal{V}})^2/2$ edges. In the worst case, each object id $o_i$ in $UO^{\mathcal{V}}$ will appear in all the $\mathcal{F}$ frames, and it will take $\mathcal{O}(UO^{\mathcal{V}} \times \mathcal{F} \times log \mathcal{F})$ to generate $VD$ from vertex labels in sequential order of frame id using a priority queue. Let's assume in the worst case, each object id pair will also appear in $\mathcal{F}$ frames and $k=1$. It will compute distance in every frame, and the worst complexity for traversing all the edges is $\mathcal{O}( \frac{(UO^{\mathcal{V}})^2}{2} \times \mathcal{F} \times d)$, where $d$ is the time required for computing distance between two bounding box centroids. The worst case CPU time complexity for this algorithm is $\mathcal{O} ( UO^{\mathcal{V}} \times \mathcal{F} \times log \mathcal{F} + \frac{(UO^{\mathcal{V}})^2}{2} \times \mathcal{F} \times d)$. This can be rewritten in simplified form as $\mathcal{O}((UO^{\mathcal{V}})^2 \times \mathcal{F})$. On average case, each object id can appear in $\frac{F}{UO^{\mathcal{V}}}$ frames on average, and the time complexity will be $\mathcal{O}((UO^{\mathcal{V}})^2 \times \frac{\mathcal{F}}{UO^{\mathcal{V}}})$, which is $\mathcal{O} (UO^{\mathcal{V}} \times \mathcal{F})$ in simplified form.

For \underline{\textit{Model MGV}}, the maximum time taken among the $N$ graphs is considered as the response time for this model. In the worst case, this model will generate a single graph; the time complexity will be the same as that of model SGV. Let us assume, each graph has $\mathcal{F}/N$ frames, $UO^{\mathcal{V}}/N$ nodes, $(UO^{\mathcal{V}})^2/2N$ on average. Each edge will appear in $\mathcal{F}/UO^{\mathcal{V}}$ frames for this model as well. The average case CPU time complexity will be $\mathcal{O}(\frac{(UO^{\mathcal{V}})^2}{N} \times \frac{\mathcal{F}}{UO^{\mathcal{V}}})$, which is $\mathcal{O} ( \frac{UO^{\mathcal{V}}}{N} \times \mathcal{F})$ in simplified form . The result composition time is nominal for this algorithm, and hence ignored. The average case CPU time complexity of model MGV will be $N$ times that of when model SGV is used, if every graph has approximately a similar number of frames, which may not always be possible.

\noindent \underline{\textit{I/O Time Complexity of Algorithm~\ref{alg:SGV-comingclose-farapart}:}} The I/O complexity for this algorithm depends on how many times a bounding box is fetched from file $IDF^{\mathcal{V}}$ and how many lines for an input base graph file are read. For model SGV, in the worst case, it will read a total of $UO^{\mathcal{V}}+\frac{(UO^{\mathcal{V}})^2}{2}$ lines (all the vertex and edge lines) from the graph file. In the worst case, each pair of object id $o_i, o_j$ will appear in $\mathcal{F}$ frames, and $k=1$. $IDF^{\mathcal{V}}$ will be accessed $\mathcal{F} \times \frac{(UO^{\mathcal{V}})^2}{2}$. The worst case I/O complexity will be $\mathcal{O}(\mathcal{F} \times \frac{(UO^{\mathcal{V}})^2}{2}+UO^{\mathcal{V}}+\frac{(UO^{\mathcal{V}})^2}{2})$, which is $\mathcal{O}(\mathcal{F} \times (UO^{\mathcal{V}})^2$ in simplified form. In the worst case, if an object pair appears in large number of times frames, it will dominate the overall response time of this algorithm.

In average case, each object pair will appear in $\frac{2\mathcal{F}}{(UO^{\mathcal{V}})^2}$ frames. $IDF^{\mathcal{V}}$ will be accessed $\frac{2\mathcal{F}}{(UO^{\mathcal{V}})^2} \times \frac{(UO^{\mathcal{V}})^2}{2}$ times, for all the edges. The base graph file is read the same number of times in both the worst case and the average case. The average case I/O complexity will be $\mathcal{O}(\mathcal{F} +UO^{\mathcal{V}}+\frac{(UO^{\mathcal{V}})^2}{2})$, which is $\mathcal{O}(\mathcal{F} \times (UO^{\mathcal{V}})^2$, which is $\mathcal{O}(\mathcal{F}+(UO^{\mathcal{V}})^2$.

For model MGV, the worst case I/O complexity will be the same as SGV, as it will generate one large connected component. In average case, it will read $\frac{UO^{\mathcal{V}}}{N}+\frac{(UO^{\mathcal{V}})^2}{2F}$ lines from the base graph file. The average case I/O complexity will be $\mathcal{O}(\frac{UO^{\mathcal{V}}}{N}+\frac{(UO^{\mathcal{V}})^2}{2F}+\frac{2 \mathcal{ F}}{(UO^{\mathcal{V}})^2} \times \frac{(UO^{\mathcal{V}})^2}{2N})$. This can be simplified as $\mathcal{O}(\frac{\mathcal{F}}{N}+(UO^{\mathcal{V}})^2$. For model MGV, the I/O time will be $N$ times faster than $SGV$ in the average case, if every graph has approximately a similar number of frames, which may not always be possible.


\subsection{Model SGF algorithm}
The same approach discussed for detecting the situation in question can be applied to the graphs generated by the model SGF. One advantage of the model SGF is that the distance values are stored as part of the edge label for each graph. However, unlike model SGV/MGV, this model can have different edges across graphs representing the relationship between object pair $o_i, o_k$. In other words, the object ID pairs $o_i, o_j$ and their corresponding frame id's are not grouped together in this model. Another quirk of this model is that an object id can appear in any frame of the video. Hence, each graph and its edges need to be traversed, and the distances extracted from edge labels need to be stored when this model is used. However, not all the distances across all the frames need to be stored for each object id pair $(o_i, o_j)$. We just need to store the distance values in frame $f-k$ for an object id pair and compare with the distance values extracted from frame $f$. For this, an edge dictionary ($ED$) is built while traversing the graphs of this model. Since the edge labels store the bounding box centroid distances, the indexed data file $IDF^{\mathcal{V}}$ is not required when this model is used. Given all of these aspects of model SGF, an algorithm is proposed in Algorithm~\ref{alg:SGF-comingclose-farapart}. 

Algorithm~\ref{alg:SGF-comingclose-farapart} takes as input the set of graphs $SGF^{\mathcal{V}}$ generated by model SGF (stored as a graph file), an indexed (on frame id) data file $IDF^{\mathcal{V}}$, video meta information ${\mathcal{M^V}}$ (stored in graph file header), an optional parameter $CL$ to filter objects of given type, a threshold $\epsilon$ to filter perturbations, and a parameter $direction$. This algorithm determines the $k$ value from the video meta information in the same manner described above for Algorithm~\ref{alg:SGF-comingclose-farapart}. Once $k$ is determined, this algorithm traverses each graph $SGF_{f}^{\mathcal{V}}$ in $SGF^{\mathcal{V}}$ at a time. For each edge for an object id pair $(o_i, o_j)$ in the graph $SGF_{f}^{\mathcal{V}}$, this algorithm extracts the distance $d_{o_i,o_j}^f$ from edge label of $(o_i, o_j)$ . Using the edge label information, it builds the dictionary $ED$. If object id pair $(o_i, o_j)$ have never appeared before, the frame id $f$ and the distance $d_{o_i,o_j}^f$ is stored in $ED$. If the object id pair $(o_i, o_j)$ has appeared before (a key for this pair exists in $ED$, then the algorithm first extracts the last frame id (denoted as $f_{prev}$) and the corresponding distance value in that frame ($d_{o_i,o_j}^{f_{prev}}$), where the existence of a situation was evaluated is extracted from $ED$. If the gap between the current frame id or graph id being analyzed and $f_{prev}$ is $k$, we check if the distances in frame $f_{prev}$ and $f$ are decreasing or increasing. The Outputs are updated in the same manner as Algorithm~\ref{alg:SGV-comingclose-farapart}, and perturbations are also filtered in the same manner. After all the comparisons are done, the value of the dictionary $ED$ is updated with frame id $f$ and $d_{o_i,o_j}^f$.

Note, this algorithm is the same as Algorithm~\ref{alg:SGV-comingclose-farapart}. Both the algorithms will generate the same output. The only difference is in how the graphs or vertices are traversed.

\begin{algorithm}[!htb]
\scriptsize
\caption{\small{Identifying if two objects are moving closer/going far apart (model SGF)}}
\label{alg:SGF-comingclose-farapart}
\begin{algorithmic}[1]
\Require Set of graphs $SGF^{\mathcal{V}}$, $M^{\mathcal{V}}$,  \underline{$CL$}, $\epsilon$, $direction$
\Ensure $Output$
\State $ED \gets \emptyset$, $Output \gets \emptyset$
\State Extract $fps$, $\mathcal{F}$, $minDuration$ and $UO^{\mathcal{V}}$ from $M^{\mathcal{V}}$
\State $avgDuration \gets \mathcal{F} / UO^{\mathcal{V}}$
\If{$minDuration < \frac{fps}{2}$ or $avgDuration < \frac{fps}{2}$}
    \State $k \gets \text{MAX}(minDuration, avgDuration)$
\Else
    \State $k \gets \text{MIN}(minDuration, avgDuration)$
\EndIf

\For{each graph $SGF_{f}^{\mathcal{V}}$ in $SGF^{\mathcal{V}}$}
    \For{each edge for object pair $(o_i,o_j)$ in graph $SGF_{f}^{\mathcal{V}}$}
        \If{$CL$ in vertex label of $o_i$ and $o_j$}
            \State $d_{o_i,o_j}^f \gets$ distance between $o_i$ and $o_j$ extracted from edge label
            \If{$(o_i, o_j)$ not in $ED$}
                \State $ED[(o_i,o_j)] \gets [f, d_{o_i,o_j}^f]$ 
            \Else
                \State $f_{prev}, d_{o_i,o_j}^{f_{prev}} \gets$ last frame id and distance value stored in $ED[(o_i,o_j)]$ 
                \If{$(f - f_{prev} = k)$ and $(|d_{o_i,o_j}^f - d_{o_i,o_j}^{f_{prev}}| > \epsilon)$} \Comment{$k$ frames have been skipped for this edge pair}
                    \If{$(direction = \texttt{"moving closer"}$ and $d_{o_i,o_j}^f < d_{o_i,o_j}^{f_{prev}})$ or $(direction = \texttt{"far apart"}$ and $d_{o_i,o_j}^f > d_{o_i,o_j}^{f_{prev}})$}
                        \If{$Output[(o_i,o_j)]$ already contains a situation from frames $f - nk, f - k$} 
                            \State  Update the frame interval to $f-nk, f$ in $Output[(o_i,o_j)]$
                        \Else
                            \State Add new interval $f_{prev}, f$ to $Output[(o_i,o_j)]$
                        \EndIf
                    \EndIf
                    \State $ED[(o_i,o_j)] \gets [f, d_{o_i,o_j}^f]$ 
                \EndIf
            \EndIf
        \EndIf
    \EndFor
\EndFor
\State \Return $Output$
\end{algorithmic}
\end{algorithm}

\underline{\textit{CPU Time Complexity of Algorithm~\ref{alg:SGF-comingclose-farapart}:}} For this algorithm, the main time complexity comes from traversing the edges and building the dictionary $ED$. Let's assume, there $OI^{\mathcal{V}}$ object id instances in a video $\mathcal{V}$ with $\mathcal{F}$ frames. There can be $OI^{\mathcal{V}}/F$ nodes and $\frac{(OI^{\mathcal{V}})^2}{2F^2}$ edges on average in each graph. Then the average case complexity for this algorithm will be $\mathcal{O}(F \times \frac{(OI^{\mathcal{V}})^2}{2F^2})$ which is $\mathcal{O}(OI^{\mathcal{V}^2})$ in simplified form. The CPU time complexity for this algorithm is worse than Algorithm~\ref{alg:SGV-comingclose-farapart} for model SGV and MGV both, as the number of object id instances $OI^{\mathcal{V}} << UI^{\mathcal{V}}$. Even though this algorithm does not perform any distance computation, the sheer amount of graphs that need to be traversed contributes to the CPU time complexity.

\underline{\textit{I/O Time Complexity of Algorithm~\ref{alg:SGF-comingclose-farapart}:}} 
This algorithm iterates over all the $\mathcal{F}$ graphs generated by model SGF, which means it reads all the lines in the base graph file $BGF^{\mathcal{V}}$. A graph file for model SGF will contain $\sum_{f \in \mathcal{F^V}} (|O_f|+\frac{|O_f|^2}{2})+\mathcal{F}+h$ rows in the graph file, where $h$ is the number of header lines, $\sum_{f \in \mathcal{F^V}} (|O_f|+\frac{|O_f|^2}{2})$ is number of edge and vertex lines in the graph file, and $\mathcal{F}$ is the number of lines to store graph id, the number of nodes, and the number of edges lines. The $IDF^{\mathcal{V}}$ file is never used by this algorithm. The I/O complexity for this algorithm will be $\mathcal{O}(\sum_{f\in \mathcal{F^V}} (|O_f|+|O_f|^2)+\mathcal{F})$. This is $\mathcal{O}(OI^{\mathcal{V}}+ (OI^{\mathcal{V}})^2/\mathcal{F}+\mathcal{F})$ for $OI^{\mathcal{V}}/\mathcal{F}$ average number of nodes per graph.

When comparing the CPU time complexities of the proposed approach across all three models, model MGV has the best CPU time complexity, followed by model SGV, and then model SGF. The I/O time complexity for model SGF is the least, followed my model MGV, and then SGV

\subsection{Experiment Results}
As mentioned earlier, what we perceive humans as a ground truth (GT) for a situation and what can be computed from VCE output are significantly different. For this particular situation, computing the ground truth is extremely difficult due to the various quirks of VCE. It is also extremely difficult to analyze the VCE outputs manually and evaluate the accuracy for this situation using the standard accuracy formula, as done for the other situation discussed in this chapter. It is also extremely difficult to produce GT containing the range of frames in a situation that exists from the VCE output.

Even though there exist some datasets in the literature with a handful of videos and labeled situation instances for these videos, they are currently unavailable. Furthermore, the existing algorithms for this situation in the literature are not open source, making it very difficult to compare our results with a baseline approach.

Hence, for this situation, we have counted the number of situation instances by watching raw videos manually for 6 small videos, 1 medium video, and 2 large videos from our dataset described in table~\ref{tab:dataset}. We compared the number of situation instances generated by our algorithm with the GT manually generated. We consider the result to be good if the number of instances brought out by our algorithm is closer to the manual GT. It is also worth mentioning, for evaluating the algorithm, choosing any arbitrary video (e.g., sports video with audiences in the background) is not a good choice, as generating the manual GT becomes very hard. Hence, the videos that contain smaller number of objects (e.g., CCTV footage of a corridor or captured in a controlled environment) are chosen for analyzing the accuracy of the algorithm.


The algorithms are implemented in Python, and the experiments were conducted on a machine equipped with 2 Intel Xeon processors (48 cores, 756 GB of main memory). For comparison purposes, we have used $k =1/2 fps, 1 fps$, and the adaptive $k$ was selected using $avgDuration/2$ and $minDuration$ or $avgDuration$ and $minDuration$. The threshold value $\epsilon$ for filtering perturbation is fixed to 5 pixels. Since both Algorithm~\ref{alg:SGV-comingclose-farapart} and Algorithm~\ref{alg:SGF-comingclose-farapart} generate the same output, we will not showcase results for individual models. For performance, we will compare the results for the three models individually.

\underline{\textit{Experiment results for two objects coming closer:}} In table~\ref{tab:results-object-coming-closer} are shown for the situation when two objects are coming closer. In this table, the experiment results for 6 small videos (rows 1-7), 1 medium video (row 8), and two large videos (rows 9-10) are showcased. For each video, the total number of frames $\mathcal{F}$, the number of unique object id instances detected by VCE $UO^{\mathcal{V}}$, the computed average duration of objects using the formula introduced earlier, and the minimum duration an object appears in the videos are shown. In this table for video S13, there are 3 instances where two objects were coming closer in the original video. It detected a total of 7 unique object id instances, and the average duration is 215. Note that the minimum duration for this video is 1276 frames, and the total number of frames is 1509. Since the minimum duration is very large, the proposed adaptive K selection criteria chooses the average duration. Among all four different values of K, when the average duration is chosen, it obtains the best results in terms of the number of situation instances. 

On the other hand, when fps is used for this video, the number of situation instances is 31 and 25, which is way more than the number of situation instances in the original GT. Most of these extra detected instances are where objects are moving hands (not moving) and perturbations, when fps is used, even after filtering using $\epsilon$. On the other hand, when $avgDuration$ or $avgDuration/2$ is chosen for this video, these perturbations are significantly reduced. There are still some extra instances detected for this video, because the object bounding boxes are very large across all the frames of the video. The absolute difference in distances in two frames $f-k$ and $f$ is around 20-40 pixels. Even with that, the chosen $k$ and the algorithm are getting better results.

\begin{table*}[!htb]
\centering
\scriptsize
\caption{\small{Summary of experiment results on different video types. $\mathcal{F}$: No. of frames in the video, $UO^{\mathcal{V}}$: No. of unique object id instances in a video. }}
\label{tab:results-object-coming-closer}
\begin{tabular}{|p{.045\linewidth}|p{.05\linewidth}|p{.06\linewidth}|p{.12\linewidth}|p{.12\linewidth}|p{.04\linewidth}|p{.04\linewidth}|p{.12\linewidth}|p{.04\linewidth}|p{.12\linewidth}|}
\hline
\multirow{2}{.04\linewidth}{\textbf{Video id}} & 
\multirow{2}{*}{$\mathcal{F}$} & 
\multirow{2}{*}{\textbf{$UO^{\mathcal{V}}$}} & 
\multirow{3}{.15\linewidth}{$minDuration$} & 
\multirow{3}{.15\linewidth}{$avgDuration$} &
\multicolumn{5}{c|}{\textbf{No. of situation instances for different values of $k$}} \\ 
\cline{6-10}
~ & ~ & ~ & ~ & ~ & 
\textbf{GT} &
\textbf{$\frac{1}{2} fps$} &
\textbf{$\frac{avgDuration}{2}$ or $minDuration$} &
\textbf{$1fps$} & 
\textbf{$avgDuration$ or $minDuration$} \\ 
\hline
S13 & 1509 & 7 & 1276 & 215 & 3 & 31 & 11 & 25 & \textbf{5} \\ \hline
S21 & 2575 & 113 & 2 & 27 & 43 & 48 & \textbf{47} & 53 & 53 \\ \hline 
S62 & 2031 & 7 & 3 & 139 & 10 & \textbf{8} & 5 & \textbf{8} & 4 \\ \hline 
S63 & 2096 & 29 & 2 & 72 & 17 & \textbf{13} & 10 & 11 & 9 \\ \hline  
S64 & 1242 & 14 & 128 & 88 & 7 & \textbf{8} & \textbf{8} & 9 & 4 \\ \hline
S66 & 1458 & 19 & 242 & 76 & 26 & 29 & \textbf{24} & 25 & 17 \\ \hline
S73 & 2229 & 8 & 24 & 278 & 10 & \textbf{8} & 6 & 5 & 6 \\ \hline
\hline
M9 & 9075 & 221 & 2 & 41 & 29 & \textbf{213} & 324 & 273 & 252 \\ \hline \hline
L5 & 22393 & 233 & 2 & 96 & 4 & 34 & 16 & 16 & \textbf{15} \\ \hline
L6 & 30486 & 483 & 3 & 43 & 9 & 65 & \textbf{5} & 6 & \textbf{5} \\ \hline

\end{tabular}
\end{table*}


The results obtained for video S21 are more interesting as this is a complex video (a snapshot of a frame for this video is shown in Figure~\ref{fig:object-moving-closer-example}(a)). Now, for these videos, there were a lot of perturbations (coming from objects sitting on the bench). The GT had 43 instances of objects coming closer. For this video, we have also experimented by setting the value of $k$ frame, which yielded 138 instances. This establishes our earlier claim that using a very small $k$ is not a good choice when the distances of objects fluctuate across frames. Now, for this video, the adaptive $k$ selection strategy of the proposed algorithm chooses $avgDuration$ as $minDuration$ is very small (2 frames). Now, among all four, using $avgDuration/2$ achieves the number of instances closer to the GT. It also filters out lots of perturbations coming from objects sitting on the bench and does not filter out the actual situation instances, whereas when we use fps, some of the actual situation instances are not brought out because the situations were ranging from $< 15$ frames (fps for this video is 30). 

For the other small videos in this table, in terms of the number of situation instances, choosing $k$ from $ \frac{avgDuration}{2}$ or $minDuration$ consistently obtains results closer to GT for different video types. Even though using $1/2 fps$ or $1 fps$ yields better results for two videos, in general, they do not consistently show good results for all the different videos. Even for the videos where $fps$ shows good results, it contains many instances that are perturbations. For example, for the video S62, using 1/2 fps yields the best result. However, when we analyzed the VCE output, we found that there are 6 situation instances where two people were coming closer. In the other 4 instances in the manual GT, VCE did not detect any object pairs in a frame together. Now, if the GT obtained from the VCE output is compared with the results of our algorithm, it is evident that choosing $k$ from $\frac{avgDuration}{2}$ or $minDuration$ yields better results overall for small videos.

For the medium and large videos (rows 8-10), it is also evident that choosing $k$ from $ \frac{avgDuration}{2}$  or $minDuration$, consistently yields results closer to $GT$. For video M9, the algorithm identifies a lot more situation instances than GT instances for all the different values of $k$. As mentioned earlier, due to the VCE quirk, the number of unique object id instances in a video can become very large. In that case, the algorithm can detect a single situation instance present in the video as multiple instances. For this video, in the original video there were 15-20 unique objects (persons), whereas VCE detected 215 person objects among the 221 detected objects. This also showcases the challenges of identifying this situation. In summary, it can be concluded that, given all the problems introduced by VCE, the algorithm's accuracy is reasonable. 

\begin{figure}
    \centering
    
    \includegraphics[width=1\linewidth, keepaspectratio=True]{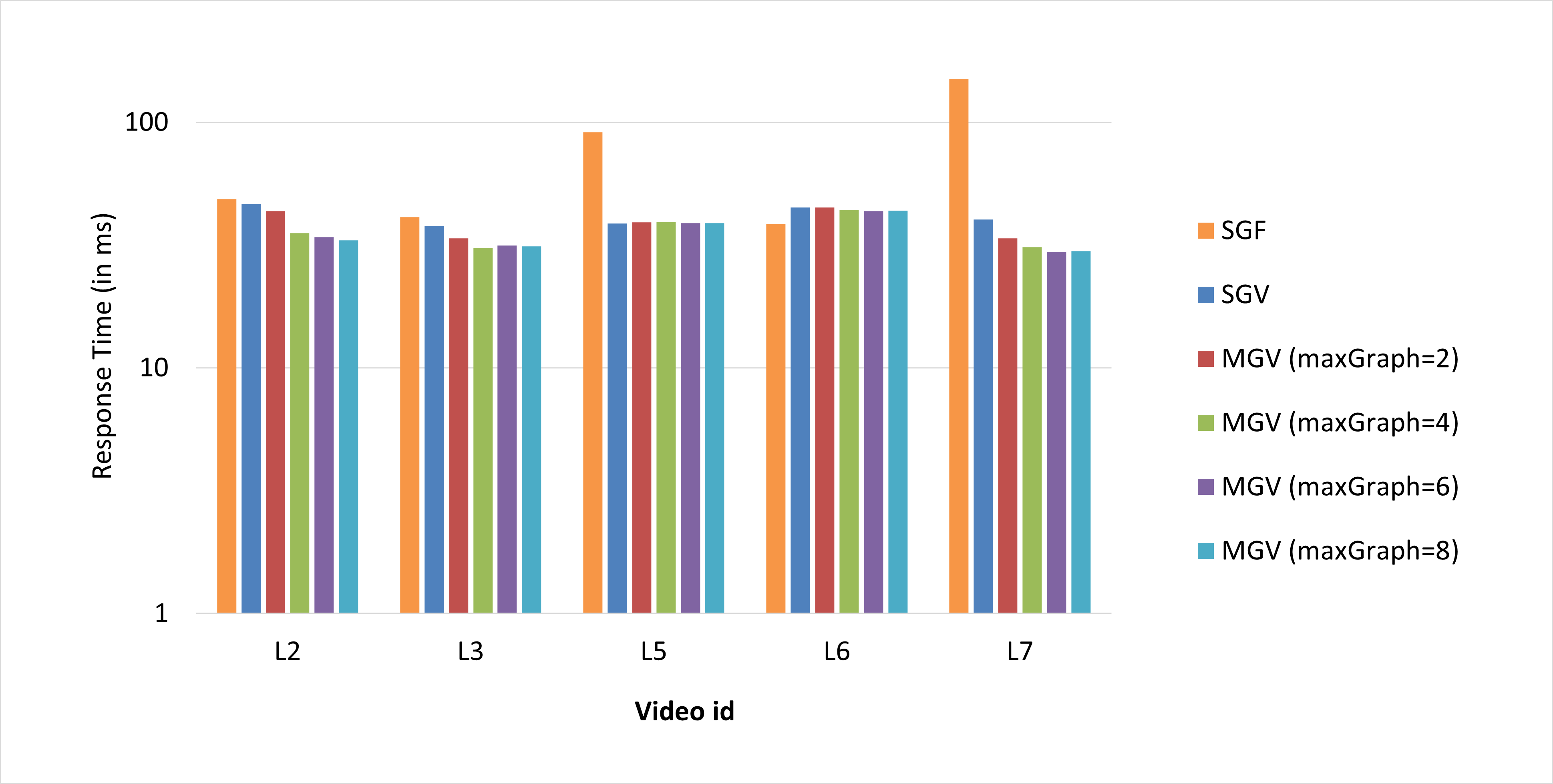}
    \caption{Response time for the proposed algorithm for different graph models. Response time is plotted in logarithmic scale.}
    \label{fig:object-coming-close-RT}
\end{figure}

For this algorithm, the response time for 5 large videos for models SGF, SGV, and MGV is shown in Figure~\ref{fig:object-coming-close-RT}. For MGV, the graphs were generated by balancing the frames. We have generated $2,4,6,8$ maximum number of graphs ($maxGraph$) for this model, and the maximum response time is taken among all the graphs generated for a given $maxGraph$. As it is evident from this figure, model SGF takes the highest amount of time for all the videos except for video L6. The response time for model SGV and SGF is similar (38 ms vs 44 ms) for this video. The reason is that this video generates a single large connected component, and the SGV model spends extra time traversing the vertex labels (4 ms in total). Because a large connected component is generated for this video, partitions of $2,4,6,8$ graphs cannot be created for this video using model MGV. For the other videos, it is evident that model MGV (with the $maxGRaph$= 8) achieves a better performance. Note that, for video L5 and L7, the response time is almost when the $maxGraph$ is set as 6 or 8. The reason is that for videos L5 and L7, there are some nodes with a huge number of frames. As a result, even after analyzing a small number of frames when the $maxGraph$ is set to 8, it achieves the same response time when we set the maximum no. of graphs = 6. 

\begin{figure}
    \centering
    \includegraphics[width=1\linewidth, keepaspectratio=True]{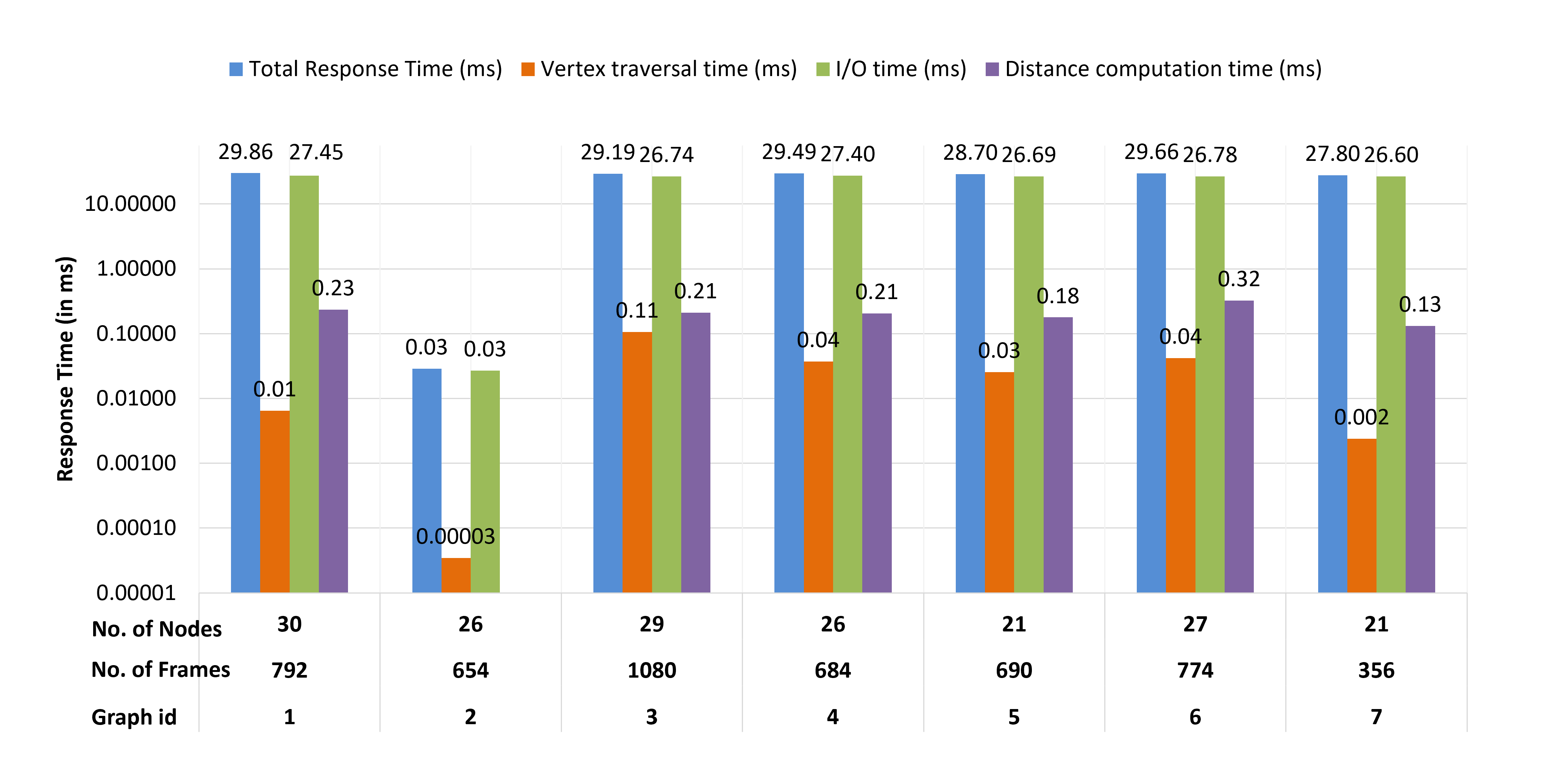}
    \caption{Response time (broken) down for model MGV ($maxGraph = 8$) for video $L7$. Response time is plotted in logarithmic scale.}
    \label{fig:object-coming-close-L7}
\end{figure}

For video L7, the response time of individual graphs when the maximum no. of graphs is set as  8 is shown in Figure~\ref{fig:object-coming-close-L7}. Here, the total Response Time (in milliseconds) is broken down for all the graphs generated for video $L7$. Now the response time comes from several components of the algorithm, such as I/O time for fetching the bounding box information, time taken for vertex traversal, and computing the distance between the bounding box centroids. For all the graphs, I/O time is the highest and similar across graphs. In this video, Graph 1 takes the maximum amount of time overall, even though it has fewer frames than Graph 3. The response time of graphs 1 and 3 is very similar, even though the number of frames is very different. There are two reasons for this: first, the time taken for traversing the vertices, and second, the time taken for computing distances. For graph 1, the vertex traversal time is higher than that of graph 3, whereas the distance computation time is higher for graph 3. Together, these 3 components contribute to the response time, and as a result, the graphs that compute distances in a larger number of frames and have a vertex with a large number of frames, take the highest amount of time to be processed. For L7, on average, it takes. This is true for the other largest videos as well. Given all the different aspects of response times, it is hard to choose a way to generate model MGV for this situation. Since situations are detected by analyzing the frames, and model MGV processes a certain number of frames, we have chosen this way of generating MGV for this analysis. 
Finally, even though the experiment results are shown for the situation ``two objects coming closer'', the same algorithm can be used to identify ``two objects going far apart''. Since generating GT for this situation is extremely difficult, experiments for that situation are not conducted in this thesis.

\section{Summary}
In this section, we have showcased that different analysis algorithms can be developed for a given situation. We have also showcased that, depending upon the situation in question, one modeling alternative is useful. We have established that model MGV (generated by a different way of balancing), in general, is efficient because of parallel processing graphs, even though efficiency can be achieved up to a certain point. Finally, this section established the applicability of graph models for different analyses without depending upon the video type in a domain-independent manner.



 \chapter{Continuous Query Language for Video Analysis (CQL-VA)}
\label{chap:CQL-VA}
\noindent Operators introduced in CQL-VA are based on the requirements discussed in Chapter~\ref{chap:problem-statement} and can play a role in CQL-VA query optimization in the future. These operators process tuples of an R++ relation like any other relational operator but have some constraints due to their applicability to specific attributes or data types. For example, to filter objects from a video using the relational algebra SELECT ($\sigma$) operator, one needs to provide a feature vector or an image from which the feature vector can be extracted. On the other hand, processing a bounding box attribute is relatively simpler using integer values and available wild cards. Joining video streams for object matching also requires dealing with feature vectors. These are not semantically valid for other attributes. This is the same as using operators, such as average, applicable to numeric values. Care has been taken to introduce a minimum number of primitive operators, and composition is used (using the closure property) to express larger computations. The CQL-VA operators and conditions, syntax, and complexity are shown in Table~\ref{tab:operator-syntax}.

\begin{table}[!h]
\footnotesize
\vspace{-15pt}
\centering
\caption{\small{\textbf{CQL-VA Operators and Conditions}. Here R$_i$: R++ relation; AR$_i$: R++ relation with extended arrables; gba, aoa: group by and assuming order attributes; N: number of tuples in $R_i$; M: unique number of objects; $th$: threshold; $S$: the complexity of sMatch; $a_i$: scalar attributes; G: number of rows in $AR_i$; Nl$_{g}$, Nr$_{g}$: average number of tuples in left and right groups in join.}}
\vspace{3pt}
\begin{tabular}{|p{.25\textwidth}|p{.45\textwidth}|p{.25\textwidth}|}
\hline
\textbf{CQL-VA Operators} & \textbf{Syntax} & \textbf{Complexity} \\ \hline
Similarity Matching Condition & \texttt{sMatch (R$_1$.[FV], R$_2$.[FV]) <comparison operator> $th$ } & $\mathcal{O}(S)$ \\ \hline
R++ to Arrable & \texttt{R2A (R$_1$, gba=R$_1$.a$_1$, aoa=R$_1$.a$_2$)} & $\mathcal{O}(N * \log N)$ \\ \hline
Compress Consecutive Tuples & \texttt{CCT (AR$_1$, \{\underline{first}|last|both\})} & $\mathcal{O}(G)$ \\ \hline
Consecutive Join & \texttt{AR$_1$ cJoin AR$_2$ (condition)} & $\mathcal{O}(G * Nl_{g} * Nr_{g} * S)$ \\ \hline
CCT Join & \texttt{AR$_1$ cctJoin AR$_2$ (condition)} & $\mathcal{O}(G^2 * S)$ \\ \hline
Direction & \texttt{Direction (AR$_1$.[[BB]])} & $\mathcal{O}(M)$ \\ \hline 
\end{tabular}
\vspace{-10pt}
\label{tab:operator-syntax}
\end{table}

\noindent\textbf{1. Similarity Matching Condition:}
The similarity matching condition (see the syntax in row 1, Table~\ref{tab:operator-syntax}) in CQL-VA compares the similarity of two feature vector attributes of an R++ relation (with or without extended arrables) using the sMatch operator. The sMatch operator computes the similarity score (a numerical distance) between the two feature vector attributes. In the similarity matching condition, the output of sMatch is compared (using comparison operators of the relational model) with a given threshold value ($th$), which controls to what extent two vectors can be considered similar. Different distance metrics are needed to compute the similarity score since feature vector semantics and size differ based on the VCE algorithm. Two standard distance metrics (values ranging from 0-1), cosine distance (for deep sort) and Euclidean distance (for SIFT and histograms), are supported by CQL-VA. Complexity is shown as $\mathcal{O}(S)$ for this condition in Table~\ref{tab:operator-syntax} as different distance metrics can have different complexity.

\noindent\textbf{2. R++ to Arrable (\texttt{R2A}):}
This operator converts an R++ relation into an R++ relation with extended arrables. The group by (gba) and assuming order (aoa) parameters need to be numeric or categorical attributes. They perform grouping and ordering, respectively, according to AQuery semantics. This operator can apply CCT Join (discussed later in this section) to improve efficiency. It can also be used to perform the running aggregate operations of AQuery on the extended arrables, again to improve efficiency and avoid self-join. The group by operation using hashing takes $\mathcal{O}(N)$ time, and sorting $G$ groups can be approximated by $\mathcal{O}(\log N)$ (upper bound), making the overall complexity $\mathcal{O}(N * \log N)$ for $N$ tuples in an R++ relation or window.

\noindent \textbf{3. Compress Consecutive Tuples (\texttt{CCT}):} 
In videos, the same object appears consecutively in multiple frames, and there can be more than one tuple for each object. Computations such as select or join will compare the same object multiple times (quadratically for join). CCT operator is introduced in CQL-VA to reduce the consecutive occurrences of the same object to one or two tuples. This can eventually improve the performance of select and join. CCT takes an R++ relation with extended arrables as input and an option argument, which can be first, last, or both (syntax shown in row 3, Table~\ref{tab:operator-syntax}). CCT compresses each extended arrable attribute by keeping the first or last element or the first and last both elements. If the option is first or last, all the extended arrable attributes are converted to their original attribute type (basic/vector). If the option is both, the attribute types remain the same as the input relation. The computational complexity of CCT is $\mathcal{O}(G)$, where $G$ is the number of tuples in an R++ relation with extended arrables.


\begin{figure}
\begin{center}
\includegraphics[width=1\textwidth, keepaspectratio=true]{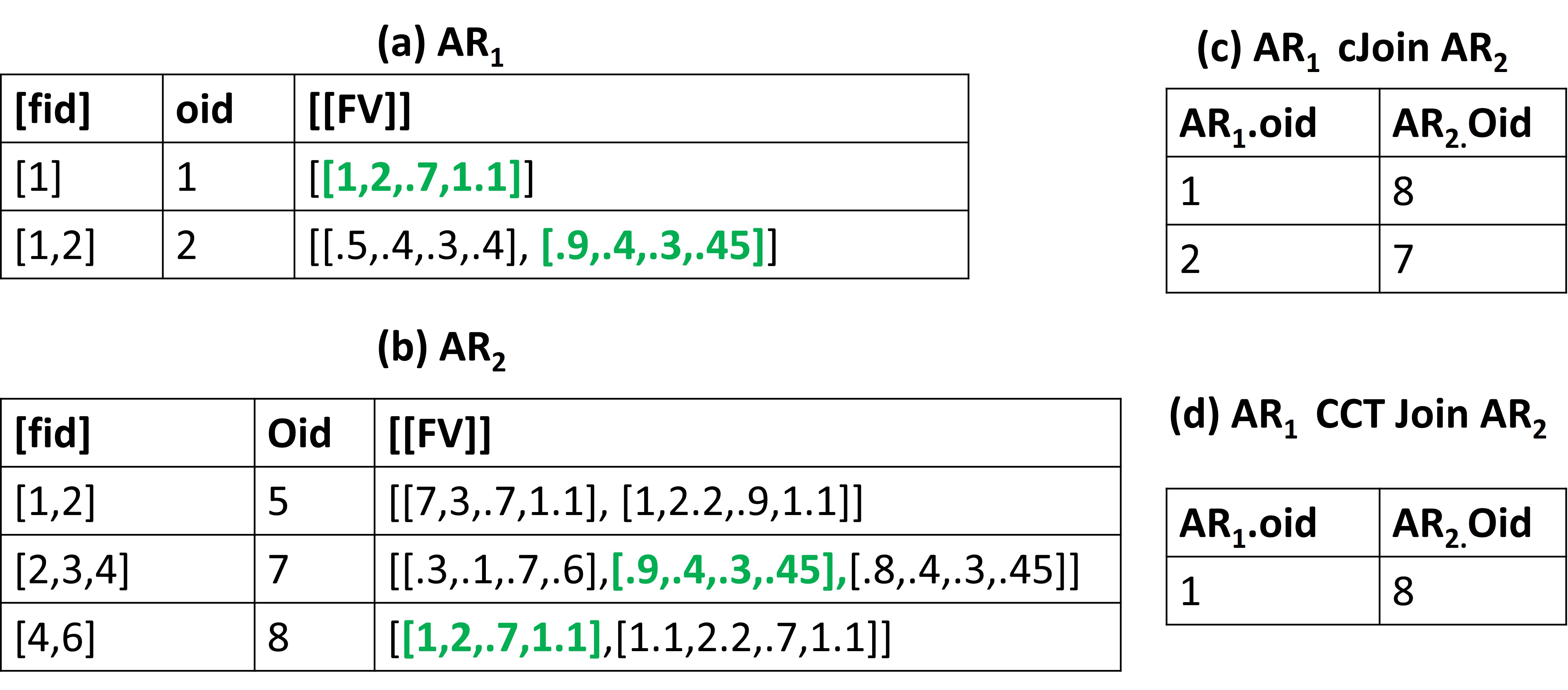}
\caption{\small{(a) AR$_1$, (b) AR$_2$ are two R++ tables with extended arrables from two different videos, generated by grouping on oid and ordering on fid, (c) cJoin on AR$_1$ and AR$_2$, (d) CCT Join on AR$_1$ and AR$_2$. [[BB]] and [ts] columns are not shown due to space constraints.}}
\label{fig:cJoinExample}
\end{center}
\end{figure}
\noindent \textbf{4. Consecutive Join (\texttt{cJoin}):}
Although regular join works on extended arrables using the CQL-VA similarity matching condition and existing relational comparison operator, it is still a join that compares each pair from two columns. Situation ST7 in Table~\ref{tab:PrimitiveSituations} can be answered efficiently and accurately without comparing all the pairs of tuples. cJoin is introduced as an alternative to improve accuracy and reduce processing time by performing less number of matches than regular join. cJoin is applied to two R++ relations with extended arrables, and the tuples in each group from both relations are compared until a match is found for a condition. Once a match is found, the rest of the tuples are not compared in that pair of partitions. Although cJoin can be applied to any R++ relations, it provides the best efficiency for feature vector similarity matches from two relations grouped on oid. In general, this will improve efficiency, although in the worst case, similarity matching may succeed only on the last tuple from both sides. Another alternative of cJoin is to drop tuples using CCT and perform a regular join with the similarity matching condition. This may result in incorrect answers. cJoin will provide better accuracy than CCT Join for the same computation, as only 2 tuples can be retained per partition in CCT, which may not match.

An example of the cJoin operation on two R++ relations with extended arrables is shown in Fig.~\ref{fig:cJoinExample}(a-c). Here, $AR_1$ and $AR_2$ are generated from two different videos. They were grouped on oid and ordered on fid. For simplicity, the [[BB]] and [ts] attributes are not shown here. In $AR_1$, there are two objects with oid 1 and 2. In $AR_2$ there are three objects with oid 5,7, and 8. The result of performing cJoin using the similarity matching condition with a threshold value of 1 is shown in Fig.~\ref{fig:cJoinExample}(c).
The elements of [[FV]] column for oid 1 in $AR_1$ is matched with all the elements of 
[[FV]] column of $AR_2$. There is no matching similar feature vectors for oid pairs (1,5) and (1,7). However, for the tuples of oid pairs (1,8) from $AR_1$ and $AR_2$, the first element of each feature vector array matches (colored green). In this case, only one comparison is made between the feature vectors in [[FV]] attribute for the rows corresponding to oid 1 and 8. Similarly, for oid 2 and 7 from $AR_1$ and $AR_2$, the second element of each feature vector from [[FV]] column was matched. The rest of the elements are not compared in this case. cJoin performs nested loop join instead of hash join as the comparison is not "equality". However, its performance should be much better than a regular nested loop join. The worst case complexity of cJoin is $\sum_{g=1}^{G}\mathcal{O}(Nl_{g} * Nr_{g} * S)$, assuming G groups in both (left: $l$, right: $r$) relations, each group $g$ with $Nl_{g}$ and $Nr_{g}$ number of tuples. $S$ is the complexity of similarity matching condition. This can be simplified to  $\mathcal{O}(G * Nl_{g} * Nr_{g} * S)$. 

\noindent \textbf{5. CCT Join:} 
\label{sec:CCT_Join_Operator}
This operation first applies the CCT operator on an R++ relation with extended arrables and then a regular join. Since CCT reduces the number of elements in each row of the extended arrable attributes to one or two, there are fewer comparisons than cJoin. An example of the CCT Join is shown in Fig.~\ref{fig:cJoinExample}(d). Here, only the first and last elements of the arrays in [[FV]] column of $AR_1$ and $AR_2$ relation are kept. Hence, for oid 7 the middle element of the feature vector array is dropped and oid 2 and 7 do not come out in the result with threshold 1. The worst case complexity of CCT Join is $\mathcal{O}((G*2) * (G*2) * S)$, assuming $G$ groups in both R++ relation, each group $g \in G$ with $Nl_{g}$ and $Nr_{g}$ tuples. If first or last is chosen, complexity will be $\mathcal{O}(G * G * S)$. This can be simplified to  $\mathcal{O}(G^2 * S)$. 


\noindent \textbf{6. Direction (\texttt{Direction}):} This primitive operator has been introduced to answer queries that require direction (e.g., ST5 in Table~\ref{tab:PrimitiveSituations}) and can only be applied to the bounding box attribute of an R++ relation with extended arrables (grouped on object id). Direction outputs one of the 8 directions as an enumerated type in a new column (used like an aggregate function). It is computed by default using the first and last bounding box elements of each extended arrable. It is also possible to compute the direction between the $i^{th}$ and $j^{th}$ element of the extended arrables of bounding boxes, where $i < j$ if specified in the operator explicitly. The complexity of the direction operator is $\mathcal{O}(M)$, where $M$ is the number of unique objects in that particular relation or window.

 \chapter{Expressing Situations as Queries}
\label{chap:situation-as-query}
This will be addressed after proposal.
\chapter{CONCLUSION}
\label{chap:conclusion}
In this thesis, we identified the various components of a domain-independent video situation analysis framework. We made a case for the applicability of various graph models for analyzing different situations. The accuracy and efficiency of the proposed algorithms are evaluated with extensive experimentation. We also showcased that even in the presence of various errors induced by video content extractors, the proposed algorithms will not change. In other words, as video content extractors evolve, the proposed algorithm's accuracy will increase. In our view, this thesis lays the groundwork for a domain-independent video analysis framework based on graph models. We have also introduced the relational model in this thesis. Our initial work on designing primitive operators for the relational model is also included. A complete generalized situation analysis system supporting more primitive operators and more complex situation analysis will be addressed after the proposal.


\bibliographystyle{./bibliography/IEEEtran}



\bibliography{bibliography/hafsaResearch,bibliography/itlabPublications,bibliography/itlabTheses,bibliography/itlabCollectionPart1,bibliography/itlabCollectionPart2}

@inproceedings{Chellappa-NSF-report-2014,
  author    = {Rama Chellappa},
  title     = {Frontiers in Image and Video Analysis NSF/FBI/DARPA Workshop Report},
  booktitle = {Workshop},
  year      = {2014},
  pages     = {120},
  url		= {www.umiacs.umd.edu/~rama/NSF_report.pdf}
}

@article{SP/abadi2003aurora,
  title={Aurora: a new model and architecture for data stream management},
  author={Abadi, Daniel J and Carney, Don and Cetintemel, Ugur and Cherniack, Mitch and Convey, Christian and Lee, Sangdon and Stonebraker, Michael and Tatbul, Nesime and Zdonik, Stan},
  journal={the VLDB Journal},
  volume={12},
  pages={120--139},
  year={2003},
  publisher={Springer}
}

@inproceedings{SP/abadi2005designBorealis,
  title={The design of the borealis stream processing engine.},
  author={Abadi, Daniel J and Ahmad, Yanif and Balazinska, Magdalena and Cetintemel, Ugur and Cherniack, Mitch and Hwang, Jeong-Hyon and Lindner, Wolfgang and Maskey, Anurag and Rasin, Alex and Ryvkina, Esther and others},
  booktitle={Cidr},
  volume={5},
  number={2005},
  pages={277--289},
  year={2005}
}

@inproceedings{SP/arasu2003stream,
  title={STREAM: the stanford stream data manager (demonstration description)},
  author={Arasu, Arvind and Babcock, Brian and Babu, Shivnath and Datar, Mayur and Ito, Keith and Nishizawa, Itaru and Rosenstein, Justin and Widom, Jennifer},
  booktitle={Proceedings of the 2003 ACM SIGMOD international conference on Management of data},
  pages={665--665},
  year={2003}
}

@inproceedings{SP/chandrasekaran2003telegraphcq,
  title={TelegraphCQ: continuous dataflow processing},
  author={Chandrasekaran, Sirish and Cooper, Owen and Deshpande, Amol and Franklin, Michael J and Hellerstein, Joseph M and Hong, Wei and Krishnamurthy, Sailesh and Madden, Samuel R and Reiss, Fred and Shah, Mehul A},
  booktitle={Proceedings of the 2003 ACM SIGMOD international conference on Management of data},
  pages={668--668},
  year={2003}
}

@article{flinkarticle,
author       = {Paris Carbone and
                  Asterios Katsifodimos and
                  Stephan Ewen and
                  Volker Markl and
                  Seif Haridi and
                  Kostas Tzoumas},
  title        = {Apache Flink{\texttrademark}: Stream and Batch Processing in a Single
                  Engine},
  journal      = {{IEEE} Data Eng. Bull.},
  volume       = {38},
  number       = {4},
  pages        = {28--38},
  year         = {2015},
  url          = {http://sites.computer.org/debull/A15dec/p28.pdf},
  bibsource    = {dblp computer science bibliography, https://dblp.org}
}

@article{thein2014apache,
  title="{Apache kafka: Next generation distributed messaging system}",
  author={Thein, Khin Me Me},
  journal={International Journal of Scientific Engineering and Technology Research},
  volume={3},
  number={47},
  pages={9478--9483},
  year={2014}
}

@inproceedings{2015TwitterHS,
  title="{Twitter Heron: Stream Processing at Scale}",
  author={Kulkarni, Sanjeev and Bhagat, Nikunj and Fu, Maosong and Kedigehalli, Vikas and Kellogg, Christopher and Mittal, Sailesh and Patel, Jignesh M and Ramasamy, Karthik and Taneja, Siddarth},
  booktitle={Proceedings of the 2015 ACM SIGMOD international conference on Management of data},
  pages={239--250},
  year={2015}
}

@article{survey/ibrahim2021survey,
  title="{A Survey of Performance Optimization in Neural Network-Based Video Analytics Systems}",
  author={Ibrahim, Nada and Maurya, Preeti and Jafari, Omid and Nagarkar, Parth},
  journal={arXiv preprint arXiv:2105.14195},
  year={2021}
}

@article{lowe2004sift,
  title={Sift-the scale invariant feature transform},
  author={Lowe, G},
  journal={Int. J},
  volume={2},
  number={91-110},
  pages={2},
  year={2004}
}

@article{SURF/bay2008speeded,
  title={Speeded-up robust features (SURF)},
  author={Bay, Herbert and Ess, Andreas and Tuytelaars, Tinne and Van Gool, Luc},
  journal={Computer vision and image understanding},
  volume={110},
  number={3},
  pages={346--359},
  year={2008},
  publisher={Elsevier}
}

@ARTICLE{Survey/ObjectDetection,
  author={Zou, Zhengxia and Chen, Keyan and Shi, Zhenwei and Guo, Yuhong and Ye, Jieping},
  journal={Proceedings of the IEEE}, 
  title={Object Detection in 20 Years: A Survey}, 
  year={2023},
  volume={111},
  number={3},
  pages={257-276},
  doi={10.1109/JPROC.2023.3238524}}

@inproceedings{VideoQuerying/hjelsvold1994modelling,
  title={Modelling and querying video data},
  author={Hjelsvold, Rune and Midtstraum, Roger},
  booktitle={VLDB},
  volume={94},
  pages={686--694},
  year={1994},
  organization={Citeseer}
}

@inproceedings{VideoQuerying/goswami2023active,
  title={Active learning for video classification with frame level queries},
  author={Goswami, Debanjan and Chakraborty, Shayok},
  booktitle={2023 International Joint Conference on Neural Networks (IJCNN)},
  pages={1--9},
  year={2023},
  organization={IEEE}
}

@article{VideoQuerying/QBIC,
  title={Query by image and video content: The QBIC system},
  author={Flickner, Myron and Sawhney, Harpreet and Niblack, Wayne and Ashley, Jonathan and Huang, Qian and Dom, Byron and Gorkani, Monika and Hafner, Jim and Lee, Denis and Petkovic, Dragutin and others},
  journal={computer},
  volume={28},
  number={9},
  pages={23--32},
  year={1995},
  publisher={IEEE}
}

@article{VideoQuerying/kuo2000content,
  title={Content-based query processing for video databases},
  author={Kuo, Tony CT and Chen, Arbee LP},
  journal={IEEE Transactions on Multimedia},
  volume={2},
  number={1},
  pages={1--13},
  year={2000},
  publisher={IEEE}
}

@inproceedings{videoQuerying/aref2003video,
  title="{Video query processing in the VDBMS testbed for video database research}",
  author={Aref, Walid and Hammad, Moustafa and Catlin, Ann Christine and Ilyas, Ihab and Ghanem, Thanaa and Elmagarmid, Ahmed and Marzouk, Mirette},
  booktitle={Proceedings of the 1st ACM international workshop on Multimedia databases},
  pages={25--32},
  year={2003}
  }

@inproceedings{videoQuerying/zhang2017live,
  title="{Live video analytics at scale with approximation and delay-tolerance}",
  author={Zhang, Haoyu and Ananthanarayanan, Ganesh and Bodik, Peter and Philipose, Matthai and Bahl, Paramvir and Freedman, Michael J},
  booktitle={14th $\{$USENIX$\}$ Symposium on Networked Systems Design and Implementation ($\{$NSDI$\}$ 17)},
  pages={377--392},
  year={2017}
}

@article{videoQuerying/cao2021thia,
  title="{THIA: Accelerating Video Analytics using Early Inference and Fine-Grained Query Planning}",
  author={Cao, Jiashen and Hadidi, Ramyad and Arulraj, Joy and Kim, Hyesoon},
  journal={arXiv preprint arXiv:2102.08481},
  year={2021}
}

@inproceedings{videoQuerying/zhang2017vsql,
  title="{vSQL: Verifying arbitrary SQL queries over dynamic outsourced databases}",
  author={Zhang, Yupeng and Genkin, Daniel and Katz, Jonathan and Papadopoulos, Dimitrios and Papamanthou, Charalampos},
  booktitle={2017 IEEE Symposium on Security and Privacy (SP)},
  pages={863--880},
  year={2017},
  organization={IEEE}
}

@inproceedings{videoQuerying/csaykol2005database,
  title="{A database model for querying visual surveillance videos by integrating semantic and low-level features}",
  author={{\c{S}}aykol, Ediz and G{\"u}d{\"u}kbay, U{\u{g}}ur and Ulusoy, {\"O}zg{\"u}r},
  booktitle={International Workshop on Multimedia Information Systems},
  pages={163--176},
  year={2005},
  organization={Springer}
}

@inproceedings{videoQuerying/nguyen2021traffic,
  title="{Traffic video event retrieval via text query using vehicle appearance and motion attributes}",
  author={Nguyen, Tien-Phat and Tran-Le, Ba-Thinh and Thai, Xuan-Dang and Nguyen, Tam V and Do, Minh N and Tran, Minh-Triet},
  booktitle={IEE/CVF CVPR},
  pages={4165--4172},
  year={2021}
}

@inproceedings{videoQuerying/zhang2021temporal,
  title="{Temporal Query Networks for Fine-grained Video Understanding}",
  author={Zhang, Chuhan and Gupta, Ankush and Zisserman, Andrew},
  booktitle={IEEE/CVF CVPR},
  pages={4486--4496},
  year={2021}
}

@inproceedings{videoQuerying/bastani2020miris,
  title="{MIRIS: Fast Object Track Queries in Video}",
  author={Bastani, Favyen and He, Songtao and Balasingam, Arjun and Gopalakrishnan, Karthik and Alizadeh, Mohammad and Balakrishnan, Hari and Cafarella, Michael and Kraska, Tim and Madden, Sam},
  booktitle={ACM SIGMOD},
  pages={1907--1921},
  year={2020}
}

@article{ObjectDetection/ren2015faster,
  title="{Faster r-cnn: Towards real-time object detection with region proposal networks}",
  author={Ren, Shaoqing and He, Kaiming and Girshick, Ross and Sun, Jian},
  journal={NeurIPS},
  volume={28},
  year={2015}
}

@article{videoQuerying/kang2017noscope,
  author       = {Daniel Kang and
                  John Emmons and
                  Firas Abuzaid and
                  Peter Bailis and
                  Matei Zaharia},
  title        = {NoScope: Optimizing Deep CNN-Based Queries over Video Streams at Scale},
  journal      = {PVLDB},
  volume       = {10},
  number       = {11},
  pages        = {1586--1597},
  year         = {2017}
}

@inproceedings{videoQuerying/yadav2019vidcep,
  title="{Vidcep: Complex event processing framework to detect spatiotemporal patterns in video streams}",
  author={Yadav, Piyush and Curry, Edward},
  booktitle={IEEE BigData},
  pages={2513--2522},
  year={2019},
  organization={IEEE}
}

@article{videoQuerying/yadav2021vidwin,
  title="{Vid-win: Fast video event matching with query-aware windowing at the edge for the internet of multimedia things}",
  author={Yadav, Piyush and Salwala, Dhaval and Curry, Edward},
  journal={IEEE IoT Journal},
  volume={8},
  number={13},
  pages={10367--10389},
  year={2021}}

@article{videoQuerying/EQUI-VOCAL,
  title={EQUI-VOCAL: Synthesizing Queries for Compositional Video Events from Limited User Interactions},
  author={Zhang, Enhao and Daum, Maureen and He, Dong and Haynes, Brandon and Krishna, Ranjay and Balazinska, Magdalena},
  journal={VLDB},
  volume={16},
  number={11},
  pages={2714--2727},
  year={2023},
summary= {This paper uses whole video as a graph model and the queries are expressed as SQL like queries. They assume the queries submitted by user is unknown, and iteratively asks the user for handful of videos (or labels) that relates to the query they the user want to ask. After some iteration (in supervised learning) they select top k queries, and give them to user. User can either select the query manually, or the system can automatically execute the suggested query. This is something similar to Query By Example or QBIC. This system can answer queries based on spatial bounding box relationship of objects only. They do not even answer queries involving more than two objects.}
}

@inproceedings{videoQuerying/chao2020svq,
  title="{SVQ++: Querying for Object Interactions in Video Streams}",
  author={Chao, Daren and Koudas, Nick and Xarchakos, Ioannis},
  booktitle={2020 ACM SIGMOD},
  pages={2769--2772},
  year={2020}
}

@article{VideoRepresentation/yadav2020knowledge,
  title="{Knowledge Graph Driven Approach to Represent Video Streams for Spatiotemporal Event Pattern Matching in Complex Event Processing}",
  author={Yadav, Piyush and Salwala, Dhaval and Das, Dibya Prakash and Curry, Edward},
  journal={International Journal of Semantic Computing},
  volume={14},
  number={03},
  pages={423--455},
  year={2020},
  publisher={World Scientific}
}

@article{videoQuerying/donderler2005bilvideo,
  title="{BilVideo: Design and implementation of a video database management system}",
  author={D{\"o}nderler, Mehmet Emin and {\c{S}}aykol, Ediz and Arslan, Umut and Ulusoy, {\"O}zg{\"u}r and G{\"u}d{\"u}kbay, U{\u{g}}ur},
  journal={Multimedia Tools and Applications},
  volume={27},
  number={1},
  pages={79--104},
  year={2005},
  publisher={Springer}
}

@article{videoQuerying/kang2018blazeit,
  title="{BlazeIt: optimizing declarative aggregation and limit queries for neural network-based video analytics}",
  author={Kang, Daniel and Bailis, Peter and Zaharia, Matei},
  journal={arXiv preprint arXiv:1805.01046},
  year={2018}
}

@article{objectRecognintion/zhu2020deformable,
  title={Deformable detr: Deformable transformers for end-to-end object detection},
  author={Zhu, Xizhou and Su, Weijie and Lu, Lewei and Li, Bin and Wang, Xiaogang and Dai, Jifeng},
  journal={arXiv preprint arXiv:2010.04159},
  year={2020}
}

@inproceedings{objectRecognition/dalal2005histograms,
  title={Histograms of oriented gradients for human detection},
  author={Dalal, Navneet and Triggs, Bill},
  booktitle={2005 IEEE computer society conference on computer vision and pattern recognition (CVPR'05)},
  volume={1},
  pages={886--893},
  year={2005},
  organization={Ieee}
}

@inproceedings{ObjectTracking/li2010multiple,
  title={A multiple object tracking method using Kalman filter},
  author={Li, Xin and Wang, Kejun and Wang, Wei and Li, Yang},
  booktitle={The 2010 IEEE international conference on information and automation},
  pages={1862--1866},
  year={2010},
  organization={IEEE}
}

@inproceedings{ObjectTracking/sundaram2010dense,
  title={Dense point trajectories by gpu-accelerated large displacement optical flow},
  author={Sundaram, Narayanan and Brox, Thomas and Keutzer, Kurt},
  booktitle={European conference on computer vision},
  pages={438--451},
  year={2010},
  organization={Springer}
}

@inproceedings{ObjectTracking/Wojke2018deep,
  title="{Deep Cosine Metric Learning for Person Re-identification}",
  author={Wojke, Nicolai and Bewley, Alex},
  booktitle={WACV},
  year={2018},
  pages={748--756},
  organization={IEEE},
  doi={10.1109/WACV.2018.00087}
}

@article{ObjectTracking/zhu2023cross,
title={Cross-modal Orthogonal High-rank Augmentation for RGB-Event Transformer-trackers},
author={Zhu, Zhiyu and Hou, Junhui and Wu, Dapeng Oliver},
journal={International Conference on Computer Vision},
year={2023}
}

@inproceedings{objectRecognintion/MRCNN,
  title="{Mask r-cnn}",
  author={He, Kaiming and Gkioxari, Georgia and Doll{\'a}r, Piotr and Girshick, Ross},
  booktitle={ICCV},
  pages={2961--2969},
  year={2017}
}

@article{objectRecognintion/wang2022yolov7,
  title="{YOLOv7: Trainable bag-of-freebies sets new state-of-the-art for real-time object detectors}",
  author={Wang, Chien-Yao and Bochkovskiy, Alexey and Liao, Hong-Yuan Mark},
  journal={arXiv preprint arXiv:2207.02696},
  year={2022}
}

@article{objectRecognintion/yolov8,
  title="{YOLOv8}",
  author={Ultralytics},
  journal={},
  url={https://github.com/ultralytics/ultralytics/tree/main},
  year={2023},
   note = {Accessd: June 30, 2023}
}

@article{objectRecognintion/yolov10,
  author       = {Ao Wang and
                  Hui Chen and
                  Lihao Liu and
                  Kai Chen and
                  Zijia Lin and
                  Jungong Han and
                  Guiguang Ding},
  title        = {YOLOv10: Real-Time End-to-End Object Detection},
  journal      = {CoRR},
  volume       = {abs/2405.14458},
  year         = {2024},
  url          = {https://doi.org/10.48550/arXiv.2405.14458},
  doi          = {10.48550/ARXIV.2405.14458},
  eprinttype    = {arXiv},
  eprint       = {2405.14458},
  bibsource    = {dblp computer science bibliography, https://dblp.org}
}

@inproceedings{ActivityRecognitionPose/duan2022revisiting,
  title={Revisiting skeleton-based action recognition},
  author={Duan, Haodong and Zhao, Yue and Chen, Kai and Lin, Dahua and Dai, Bo},
  booktitle={Proceedings of the IEEE/CVF conference on computer vision and pattern recognition},
  pages={2969--2978},
  year={2022}
}

@article{ActivityRecognitionPose/reilly2023just,
  title={Just Add $\pi$! Pose Induced Video Transformers for Understanding Activities of Daily Living},
  author={Reilly, Dominick and Das, Srijan},
  journal={arXiv preprint arXiv:2311.18840},
  year={2023}
}

@misc{ActivityRecognitionPose/2020mmaction2,
    title={OpenMMLab's Next Generation Video Understanding Toolbox and Benchmark},
    author={MMAction2 Contributors},
    howpublished = {\url{https://github.com/open-mmlab/mmaction2}},
    year={2020}
}

@ARTICLE{ActivityRecognitionPose/MMNET,
  author={Yu, Bruce X.B. and Liu, Yan and Zhang, Xiang and Zhong, Sheng-hua and Chan, Keith C.C.},
  journal={IEEE Transactions on Pattern Analysis and Machine Intelligence}, 
  title={MMNet: A Model-based Multimodal Network for Human Action Recognition in RGB-D Videos}, 
  year={2022},
  volume={},
  number={},
  pages={1-1},
  doi={10.1109/TPAMI.2022.3177813}
  }

@inproceedings{ActivityRecognition/ryali2023hiera,
  title={Hiera: A hierarchical vision transformer without the bells-and-whistles},
  author={Ryali, Chaitanya and Hu, Yuan-Ting and Bolya, Daniel and Wei, Chen and Fan, Haoqi and Huang, Po-Yao and Aggarwal, Vaibhav and Chowdhury, Arkabandhu and Poursaeed, Omid and Hoffman, Judy and others},
  booktitle={International Conference on Machine Learning},
  pages={29441--29454},
  year={2023},
  organization={PMLR}
}

@article{ActivityRecognition/simonyan2014two,
  title="{Two-stream convolutional networks for action recognition in videos}",
  author={Simonyan, Karen and Zisserman, Andrew},
  journal={NeurIPS},
  volume={27},
  year={2014}
}

@inproceedings{ActivityRecognition/feichtenhofer2017spatiotemporal,
  title="{Spatiotemporal multiplier networks for video action recognition}",
  author={Feichtenhofer, Christoph and Pinz, Axel and Wildes, Richard P},
  booktitle={IEEE CVPR},
  pages={4768--4777},
  year={2017}
}

@inproceedings{DepthMonocular/fu2018deep,
  title="{Deep ordinal regression network for monocular depth estimation}",
  author={Fu, Huan and Gong, Mingming and Wang, Chaohui and Batmanghelich, Kayhan and Tao, Dacheng},
  booktitle={Proceedings of the IEEE conference on computer vision and pattern recognition},
  pages={2002--2011},
  year={2018}
}

@article{DepthMonocular/saxena2024surprising,
  title={The surprising effectiveness of diffusion models for optical flow and monocular depth estimation},
  author={Saxena, Saurabh and Herrmann, Charles and Hur, Junhwa and Kar, Abhishek and Norouzi, Mohammad and Sun, Deqing and Fleet, David J},
  journal={Advances in Neural Information Processing Systems},
  volume={36},
  year={2024}
}

@inproceedings{DepthMonocular/wang2024sqldepth,
  title={SQLdepth: Generalizable self-supervised fine-structured monocular depth estimation},
  author={Wang, Youhong and Liang, Yunji and Xu, Hao and Jiao, Shaohui and Yu, Hongkai},
  booktitle={Proceedings of the AAAI Conference on Artificial Intelligence},
  volume={38},
  number={6},
  pages={5713--5721},
  year={2024}
}

@inproceedings{DepthMonocular/junayed2022himode,
  title={HiMODE: A hybrid monocular omnidirectional depth estimation model},
  author={Junayed, Masum Shah and Sadeghzadeh, Arezoo and Islam, Md Baharul and Wong, Lai-Kuan and Ayd{\i}n, Tarkan},
  booktitle={Proceedings of the IEEE/CVF Conference on Computer Vision and Pattern Recognition},
  pages={5212--5221},
  year={2022}
}

@inproceedings{DepthMonocular/yun2022improving,
  title={Improving 360 monocular depth estimation via non-local dense prediction transformer and joint supervised and self-supervised learning},
  author={Yun, Ilwi and Lee, Hyuk-Jae and Rhee, Chae Eun},
  booktitle={Proceedings of the AAAI Conference on Artificial Intelligence},
  volume={36},
  number={3},
  pages={3224--3233},
  year={2022}
}

@inproceedings{he2023fastreid,
  title={Fastreid: A pytorch toolbox for general instance re-identification},
  author={He, Lingxiao and Liao, Xingyu and Liu, Wu and Liu, Xinchen and Cheng, Peng and Mei, Tao},
  booktitle={Proceedings of the 31st ACM International Conference on Multimedia},
  pages={9664--9667},
  year={2023}
}

@inproceedings{
  PoseEstimation/xu2022vitpose,
  title={Vi{TP}ose: Simple Vision Transformer Baselines for Human Pose Estimation},
  author={Yufei Xu and Jing Zhang and Qiming Zhang and Dacheng Tao},
  booktitle={Advances in Neural Information Processing Systems},
  year={2022},
}

@article{PoseEstimationGeneric/xu2022vitpose+,
  title={ViTPose+: Vision Transformer Foundation Model for Generic Body Pose Estimation},
  author={Xu, Yufei and Zhang, Jing and Zhang, Qiming and Tao, Dacheng},
  journal={arXiv preprint arXiv:2212.04246},
  year={2022}
}

@inproceedings{PoseEstimation/OpenPose,
  title="{Realtime multi-person 2d pose estimation using part affinity fields}",
  author={Cao, Zhe and Simon, Tomas and Wei, Shih-En and Sheikh, Yaser},
  booktitle={IEEE CVPR},
  pages={7291--7299},
  year={2017}
}

@article{PoseEstimation/LCRNet++,
  title="{Lcr-net++: Multi-person 2d and 3d pose detection in natural images}",
  author={Rogez, Gregory and Weinzaepfel, Philippe and Schmid, Cordelia},
  journal={IEEE transactions on pattern analysis and machine intelligence},
  volume={42},
  number={5},
  pages={1146--1161},
  year={2019},
  publisher={IEEE}
}

@inproceedings{PoseEstimation/rogez2017lcr,
  title={Lcr-net: Localization-classification-regression for human pose},
  author={Rogez, Gregory and Weinzaepfel, Philippe and Schmid, Cordelia},
  booktitle={Proceedings of the IEEE conference on computer vision and pattern recognition},
  pages={3433--3441},
  year={2017}
}

@article{PoseEstimation/zhang20243d,
  title={3D Graph Convolutional Feature Selection and Dense Pre-Estimation for Skeleton Action Recognition},
  author={Zhang, Junxian and Yang, Aiping and Miao, Changwu and Li, Xiang and Zhang, Rui and Thanh, Dang NH},
  journal={IEEE Access},
  year={2024},
  publisher={IEEE}
}

@article{PoseEstimation/HRNeT,
  title="{Deep high-resolution representation learning for visual recognition}",
  author={Wang, Jingdong and Sun, Ke and Cheng, Tianheng and Jiang, Borui and Deng, Chaorui and Zhao, Yang and Liu, Dong and Mu, Yadong and Tan, Mingkui and Wang, Xinggang and others},
  journal={IEEE PAMI},
  volume={43},
  number={10},
  pages={3349--3364},
  year={2020},
  publisher={IEEE},
 note={Implementation in https://github.com/HRNet/HRNet-Human-Pose-Estimation, last accessed=2025-04-29}
}

@inproceedings{Scenegraph/kim2024groupwise,
      title={Groupwise Query Specialization and Quality-Aware Multi-Assignment for Transformer-based Visual Relationship Detection}, 
      author={Kim, Jongha and Park, Jihwan and Park, Jinyoung and Kim, Jinyoung and Kim, Sehyung and Kim, Hyunwoo J},
      booktitle={CVPR},
      year={2024},
}

@inproceedings{Scenegraph/li2024panoptic,
  title={Panoptic scene graph generation with semantics-prototype learning},
  author={Li, Li and Ji, Wei and Wu, Yiming and Li, Mengze and Qin, You and Wei, Lina and Zimmermann, Roger},
  booktitle={Proceedings of the AAAI Conference on Artificial Intelligence},
  volume={38},
  number={4},
  pages={3145--3153},
  year={2024}
}

@inproceedings{Scenegraph/xu2017scene,
  title="{Scene graph generation by iterative message passing}",
  author={Xu, Danfei and Zhu, Yuke and Choy, Christopher B and Fei-Fei, Li},
  booktitle={IEEE CVPR},
  pages={5410--5419},
  year={2017}
}

@inproceedings{Scenegraph/ActionGenome,
  title="{Action genome: Actions as compositions of spatio-temporal scene graphs}",
  author={Ji, Jingwei and Krishna, Ranjay and Fei-Fei, Li and Niebles, Juan Carlos},
  booktitle={CVPR},
  pages={10236--10247},
  year={2020}
}

@inproceedings{Scenegraph/VidSGG,
  title="{Target adaptive context aggregation for video scene graph generation}",
  author={Teng, Yao and Wang, Limin and Li, Zhifeng and Wu, Gangshan},
  booktitle={CVPR},
  pages={13688--13697},
  year={2021}
}

@inproceedings{vqa/xiong2019visual,
  title="{Visual query answering by entity-attribute graph matching and reasoning}",
  author={Xiong, Peixi and Zhan, Huayi and Wang, Xin and Sinha, Baivab and Wu, Ying},
  booktitle={CVPR},
  pages={8357--8366},
  year={2019}
}

@article{objectReidentification/BotSort,
  author       = {Nir Aharon and
                  Roy Orfaig and
                  Ben{-}Zion Bobrovsky},
  title        = {BoT-SORT: Robust Associations Multi-Pedestrian Tracking},
  journal      = {CoRR},
  volume       = {abs/2206.14651},
  year         = {2022},
  url          = {https://doi.org/10.48550/arXiv.2206.14651},
  doi          = {10.48550/ARXIV.2206.14651},
  eprinttype    = {arXiv},
  eprint       = {2206.14651},
  bibsource    = {dblp computer science bibliography, https://dblp.org},
note={Implementation in https://github.com/NirAharon/BoT-SORT.git, last accessed=2025-04-29}
}

@article{SP/iqbal2015big,
  title={Big data analysis: Apache storm perspective},
  author={Iqbal, Muhammad Hussain and Soomro, Tariq Rahim and others},
  journal={International journal of computer trends and technology},
  volume={19},
  number={1},
  pages={9--14},
  year={2015}
}

@article{VSP/AmazonKinesis,
  title="{Overview of amazon web services}",
  author={Mathew, Sajee and Varia, J},
  journal={Amazon Whitepapers},
  volume={105},
  pages={1--22},
  year={2014}
}

@inproceedings{VSP/salehe2019videopipe,
  title={Videopipe: Building video stream processing pipelines at the edge},
  author={Salehe, Mohammad and Hu, Zhiming and Mortazavi, Seyed Hossein and Mohomed, Iqbal and Capes, Tim},
  booktitle={Proceedings of the 20th international middleware conference industrial track},
  pages={43--49},
  year={2019}
}

@article{VSP/zhang2019streaming,
  title={A streaming cloud platform for real-time video processing on embedded devices},
  author={Zhang, Weishan and Sun, Haoyun and Zhao, Dehai and Xu, Liang and Liu, Xin and Ning, Huansheng and Zhou, Jiehan and Guo, Yi and Yang, Su},
  journal={IEEE Transactions on Cloud Computing},
  volume={9},
  number={3},
  pages={868--880},
  year={2019},
  publisher={IEEE}
}

@article{arrayDB/baumann2021array,
  title="{Array databases: concepts, standards, implementations}",
  author={Baumann, Peter and Misev, Dimitar and Merticariu, Vlad and Huu, Bang Pham},
  journal={Journal of Big Data},
  volume={8},
  number={1},
  pages={1--61},
  year={2021},
  publisher={SpringerOpen}
}

@inproceedings{CustomAlgoSurveillance/ivanov1999video,
  title={Video surveillance of interactions},
  author={Ivanov, Yuri and Stauffer, Chris and Bobick, Aaron and Grimson, WEL},
  booktitle={Proceedings Second IEEE Workshop on Visual Surveillance (VS'99)(Cat. No. 98-89223)},
  pages={82--89},
  year={1999},
  organization={IEEE}
}

@article{CustomAlgoSurveillance/collins2000system,
  title={A system for video surveillance and monitoring},
  author={Collins, Robert T and Lipton, Alan J and Kanade, Takeo and Fujiyoshi, Hironobu and Duggins, David and Tsin, Yanghai and Tolliver, David and Enomoto, Nobuyoshi and Hasegawa, Osamu and Burt, Peter and others},
  journal={VSAM final report},
  volume={2000},
  number={1-68},
  pages={1},
  year={2000},
  publisher={Citeseer}
}

@inproceedings{CustomAlgoSurveillance/freer1996automatic,
  title={Automatic video surveillance with intelligent scene monitoring and intruder detection},
  author={Freer, JA and Beggs, BJ and Fernandez-Canque, HL and Chevrier, F and Goryashko, A},
  booktitle={1996 30th Annual International Carnahan Conference on Security Technology},
  pages={89--94},
  year={1996},
  organization={IEEE}
}

@inproceedings{VIRAT-2013,
  title={Video-based activity analysis using the L1 tracker on VIRAT data},
  author={Blasch, Erik and Wang, Zhonghai and Ling, Haibin and Palaniappan, Kannappan and Chen, Genshe and Shen, Dan and Aved, Alex and Seetharaman, Guna},
  booktitle={2013 IEEE Applied Imagery Pattern Recognition Workshop (AIPR)},
  pages={1--8},
  year={2013},
  organization={IEEE}
}

@misc{VIRATDataSet,
title={Introduction to VIRAT Video Data-set},
howpublished = {\url{https://data.kitware.com/api/v1/file/56f581c88d777f753209c9ce/download}}
}

@inproceedings{DataSet/COCO,
  title="{Microsoft coco: Common objects in context}",
  author={Lin, Tsung-Yi and Maire, Michael and Belongie, Serge and Hays, James and Perona, Pietro and Ramanan, Deva and Doll{\'a}r, Piotr and Zitnick, C Lawrence},
  booktitle={ECCV},
  pages={740--755},
  year={2014},
  organization={Springer}
}

@misc{ORACLE-Multimedia,
title= {Oracle Multimedia User's Guide},
howpublished = {\url{docs.oracle.com/database/121/IMURG/ch_intr.htm#IMURG1000}{Oracle Multimedia}
},
year = {2014}
}

@misc{ArrayDBMS/ArcMap,
title= "{ArcMap}",
url = {https://desktop.arcgis.com/en/arcmap/latest/map/main/what-is-arcmap},
year ={last accessed 07/25/2023}
}

@misc{DataSet/MavVid,
title= "{MavVid dataset prepared in ITLab}",
howpublished = {\url{https://itlab.uta.edu/downloads/datasets/MavVid.zip}
}
}

@misc{DataSet/MavVid_merged,
title= "{MavVid dataset}",
howpublished = {\url{https://itlab.uta.edu//downloads/mavVid-datasets/MavVid_Merged_v1.zip}
},
year      = {2023}
}

@article{Dataset/CVIU_UA-DETRAC,
             author    = {Longyin Wen and Dawei Du and Zhaowei Cai and Zhen Lei and Ming{-}Ching Chang and
               Honggang Qi and Jongwoo Lim and Ming{-}Hsuan Yang and Siwei Lyu},
             title     = "{ {UA-DETRAC:} {A} New Benchmark and Protocol for Multi-Object Detection and Tracking}",
             journal   = {Computer Vision and Image Understanding},
             year      = {2020}
             }

@misc{ArrayDBMS/maier2013arrayql,
  title="{ArrayQL algebra: version 3}",
  author={Maier, David and Baumann, Peter and Kersten, Martin and Lim, Kian-Tat and Stonebraker, Michael},
  year={2013}
}

@inproceedings{Sequence/SRQL,
  author       = {Raghu Ramakrishnan and
                  Donko Donjerkovic and
                  Arvind Ranganathan and
                  Kevin S. Beyer and
                  Muralidhar Krishnaprasad},
  editor       = {Maurizio Rafanelli and
                  Matthias Jarke},
  title        = {{SRQL:} Sorted Relational Query Language},
  booktitle    = {10th International Conference on Scientific and Statistical Database
                  Management, Proceedings, Capri, Italy, July 1-3, 1998},
  pages        = {84--95},
  publisher    = {{IEEE} Computer Society},
  year         = {1998},
  url          = {https://doi.org/10.1109/SSDM.1998.688114},
  doi          = {10.1109/SSDM.1998.688114},
  bibsource    = {dblp computer science bibliography, https://dblp.org}
}

@inproceedings{DataSet/Camnet,
  title="{A Camera Network Tracking (CamNeT) Dataset and Performance Baseline}",
  author={Zhang, Shu and Staudt, Elliot and Faltemier, Tim and Roy-Chowdhury, Amit K},
  booktitle={IEEE WACV},
  pages={365--372},
  year={2015}
}

@ARTICLE{DataSet/Toyota_Smarthome,
  author={Dai, Rui and Das, Srijan and Sharma, Saurav and Minciullo, Luca and Garattoni, Lorenzo and Bremond, Francois and Francesca, Gianpiero},
  journal={IEEE Transactions on Pattern Analysis and Machine Intelligence}, 
  title="{Toyota Smarthome Untrimmed: Real-World Untrimmed Videos for Activity Detection}", 
  year={2022},
  volume={},
  number={},
  pages={1-1},
  doi={10.1109/TPAMI.2022.3169976}}

@article{DataSet/ucf101,
  title="{UCF101: A dataset of 101 human actions classes from videos in the wild}",
  author={Soomro, Khurram and Zamir, Amir Roshan and Shah, Mubarak},
  journal={arXiv preprint arXiv:1212.0402},
  year={2012}
}

@inproceedings{DataSet/HMDB,
  title={HMDB: a large video database for human motion recognition},
  author={Kuehne, Hildegard and Jhuang, Hueihan and Garrote, Est{\'\i}baliz and Poggio, Tomaso and Serre, Thomas},
  booktitle={2011 International conference on computer vision},
  pages={2556--2563},
  year={2011},
  organization={IEEE}
}

@inproceedings{DataSet/jodoin2014urban,
  title={Urban tracker: Multiple object tracking in urban mixed traffic},
  author={Jodoin, Jean-Philippe and Bilodeau, Guillaume-Alexandre and Saunier, Nicolas},
  booktitle={IEEE Winter Conference on Applications of Computer Vision},
  pages={885--892},
  year={2014},
  organization={IEEE}
}

@misc{DataSet/UT-Interaction-Data,
      author = "Ryoo, M. S. and Aggarwal, J. K.",
      title = "{UT}-{I}nteraction {D}ataset, {ICPR} contest on {S}emantic {D}escription of {H}uman {A}ctivities ({SDHA})",
      year = "2010",
      howpublished = "http://cvrc.ece.utexas.edu/SDHA2010/Human\_Interaction.html"
}

@inproceedings{DataSet/Activitynet,
  title="{ActivityNet: A Large-Scale Video Benchmark for Human Activity Understanding}",
  author={Fabian Caba Heilbron, Victor Escorcia, Bernard Ghanem and Juan Carlos Niebles},
  booktitle={Proceedings of the IEEE Conference on Computer Vision and Pattern Recognition},
  pages={961--970},
  year={2015}
}

@inproceedings{GaitMonitoring/ajay2018pervasive,
  title="{A pervasive and sensor-free deep learning system for Parkinsonian gait analysis}",
  author={Ajay, Jerry and Song, Chen and Wang, Aosen and Langan, Jeanne and Li, Zhinan and Xu, Wenyao},
  booktitle={2018 IEEE EMBS International Conference on Biomedical \& Health Informatics (BHI)},
  pages={108--111},
  year={2018},
  organization={IEEE}
}

@article{GaitMonitoring/romeo2023video,
  title="{Video Based Mobility Monitoring of Elderly People Using Deep Learning Models}",
  author={Romeo, Laura and Marani, Roberto and D’Orazio, Tiziana and Cicirelli, Grazia},
  journal={IEEE Access},
  volume={11},
  pages={2804--2819},
  year={2023},
  publisher={IEEE}
}

@article{FallDetection/amsaprabhaa2023multimodal,
  title="{Multimodal spatiotemporal skeletal kinematic gait feature fusion for vision-based fall detection}",
  author={Amsaprabhaa, M and others},
  journal={Expert Systems with Applications},
  volume={212},
  pages={118681},
  year={2023},
  publisher={Elsevier}
}

@inproceedings{FallDetection/tohidypour2022deep,
  title="{A Deep Learning Based Human Fall Detection Solution}",
  author={Tohidypour, Hamid Reza and Shojaei-Hashemi, Anahita and Nasiopoulos, Panos and Pourazad, Mahsa T},
  booktitle={Proceedings of the 15th International Conference on PErvasive Technologies Related to Assistive Environments},
  pages={89--92},
  year={2022}
}

@article{SocialIsolation/prenkaj2023self,
  title="{A self-supervised algorithm to detect signs of social isolation in the elderly from daily activity sequences}",
  author={Prenkaj, Bardh and Aragona, Dario and Flaborea, Alessandro and Galasso, Fabio and Gravina, Saverio and Podo, Luca and Reda, Emilia and Velardi, Paola},
  journal={AI in Medicine},
  volume={135},
  pages={102454},
  year={2023},
  publisher={Elsevier}
}

@article{ActionRecognition/Skeletal-Francisco,
  title="{Improved action recognition with separable spatio-temporal attention using alternative skeletal and video pre-processing}",
  author={Climent-P{\'e}rez, Pau and Florez-Revuelta, Francisco},
  journal={Sensors},
  volume={21},
  number={3},
  pages={1005},
  year={2021},
  publisher={MDPI}
}

@inproceedings{GroupExtraction/chen2017anchor,
  title="{Anchor-based group detection in crowd scenes}",
  author={Chen, Mulin and Wang, Qi and Li, Xuelong},
  booktitle={ICASSP},
  pages={1378--1382},
  year={2017},
  organization={IEEE}
}

@inproceedings{GroupExtraction/li2022evolutionary,
  title={Evolutionary clustering of moving objects},
  author={Li, Tianyi and Chen, Lu and Jensen, Christian S and Pedersen, Torben Bach and Gao, Yunjun and Hu, Jilin},
  booktitle={2022 IEEE 38th International Conference on Data Engineering (ICDE)},
  pages={2399--2411},
  year={2022},
  organization={IEEE}
}

@article{GroupExtraction/dynamic-network,
  title="{A particle-and-density based evolutionary clustering method for dynamic networks}",
  author={Kim, Min-Soo and Han, Jiawei},
  journal={VLDB},
  volume={2},
  number={1},
  pages={622--633},
  year={2009},
  publisher={}
}

@article{LLMSituationDetection/yuan2025empowering,
  title={Empowering Large Language Models with 3D Situation Awareness},
  author={Yuan, Zhihao and Peng, Yibo and Ren, Jinke and Liao, Yinghong and Han, Yatong and Feng, Chun-Mei and Zhao, Hengshuang and Li, Guanbin and Cui, Shuguang and Li, Zhen},
  journal={arXiv preprint arXiv:2503.23024},
  year={2025}
}

@article{LLMVideoUnderstanding/tang2025video,
  title={Video understanding with large language models: A survey},
  author={Tang, Yunlong and Bi, Jing and Xu, Siting and Song, Luchuan and Liang, Susan and Wang, Teng and Zhang, Daoan and An, Jie and Lin, Jingyang and Zhu, Rongyi and others},
  journal={IEEE Transactions on Circuits and Systems for Video Technology},
  year={2025},
  publisher={IEEE}
}

@inproceedings{twoObjectClose/Samaras2012,
  author={Yun, Kiwon and Honorio, Jean and Chattopadhyay, Debaleena and Berg, Tamara L. and Samaras, Dimitris},
  booktitle={2012 IEEE Computer Society Conference on Computer Vision and Pattern Recognition Workshops}, 
  title={Two-person interaction detection using body-pose features and multiple instance learning}, 
  year={2012},
  volume={},
  number={},
  pages={28-35},
  doi={10.1109/CVPRW.2012.6239234}}

@article{twoObjectClose/zeiller2025collision,
  title={Collision detection for objects modelled by csg},
  author={Zeiller, Michael},
  journal={WIT Transactions on Information and Communication Technologies},
  volume={5},
  year={2025},
  publisher={WIT Press}
}

@article{LLM/zhang2023video,
  title={Video-llama: An instruction-tuned audio-visual language model for video understanding},
  author={Zhang, Hang and Li, Xin and Bing, Lidong},
  journal={arXiv preprint arXiv:2306.02858},
  year={2023}
}

@inproceedings{LLM/munasinghe2025videoglamm,
  title={Videoglamm: A large multimodal model for pixel-level visual grounding in videos},
  author={Munasinghe, Shehan and Gani, Hanan and Zhu, Wenqi and Cao, Jiale and Xing, Eric and Khan, Fahad Shahbaz and Khan, Salman},
  booktitle={Proceedings of the Computer Vision and Pattern Recognition Conference},
  pages={19036--19046},
  year={2025}
}

@inproceedings{objectRecognintion/wang2025yolov9,
  title={Yolov9: Learning what you want to learn using programmable gradient information},
  author={Wang, Chien-Yao and Yeh, I-Hau and Mark Liao, Hong-Yuan},
  booktitle={European conference on computer vision},
  pages={1--21},
  year={2025},
  organization={Springer}
}

@TechReport{ChaMis91,
  author      = {S. Chakravathy and D. Mishra},
  date        = {1991-09},
  institution = {Database Systems R\&D Center, CIS Department, University of Florida},
  title       = {An Event Specification Language (Snoop) for Active Databases and its Detection},
  location    = {E470-CSE, Gainesville, FL 32611},
  number      = {UF-CIS TR-91-23},
}

@TechReport{Vidya94,
  author      = {S. Chakravarthy and V. Krishnaprasad and Z. Tamizuddin and R.H. Badani},
  date        = {1994-05},
  institution = {University of Florida, Gainesville, FL},
  title       = {ECA Rule Integration into an OODBMS:Architecture and Implementation},
  number      = {Tech. report UF-CIS-TR-94-023},
}

@Article{sentinel::sharma,
  author       = {Chakravarthy, S. and others},
  date         = {1994},
  journal = {Information and Software Technology},
  title        = {Design of Sentinel: An Object-Oriented DBMS with Event-Based Rules},
  number       = {9},
  pages        = {559-568},
  volume       = {36},
}

@InProceedings{seshadri1995seq,
  author       = {Seshadri, Praveen and Livny, Miron and Ramakrishnan, Raghu},
  booktitle    = {Proc. of the IEEE Int'l Conference on Data Engineering},
  date         = {1995},
  title        = {SEQ: A model for sequence databases},
  location     = {Taiwan},
  organization = {IEEE},
  pages        = {232--239},
}

@InProceedings{lerner2003aquery,
  author       = {Lerner, Alberto and Shasha, Dennis},
  booktitle    = {Proceedings of the 29th international conference on Very large data bases-Volume 29},
  date         = {2003},
  year         = {2003},
  title        = {Aquery: Query language for ordered data, optimization techniques, and experiments},
  organization = {VLDB Endowment},
  pages        = {345--356},
}

@InProceedings{snoopibcont::raman,
  author    = {Adaikkalavan, Raman and Sharma Chakravarthy},
  booktitle = {Proc. of ADBIS},
  date      = {2004},
  title     = {{{Formalization and Detection of Events Over a Sliding Window in Active Databases Using Interval-Based Semantics}}},
}

@PhdThesis{aved2013scene,
  author      = {Aved, Alexander J},
  date        = {2013},
  institution = {University of Central Florida Orlando, Florida},
  title       = {Scene Understanding for Real Time Processing of Queries Over Big Data Streaming Video},
}

@Article{aved2014informatics,
  author       = {Aved, Alexander J and Hua, Kien A},
  date         = {2014},
  journal = {Multimedia Tools and Applications},
  title        = {An informatics-based approach to object tracking for distributed live video computing},
  number       = {1},
  pages        = {111--133},
  volume       = {68},
  publisher    = {Springer},
}

@Misc{cougaar,
  title   = {{Cognitive Agent Architecture (Cougaar) Open Source Project site}},
  url     = {http://www.cougaar.org/},
  sortkey = {cougaar},
}

@TechReport{Cha+96:snoop-sem:tr,
  author      = {S. Chakravarthy and J. D. Yang and S. Yang},
  date        = {1998-11},
  institution = {University of Florida},
  title       = {A Formal Framework for Computing Composite Events over Histories and Logs},
  location    = {E470-CSE, Gainesville, FL 32611},
  number      = {UF-CIS TR-98-017},
}

@InProceedings{ICDM/YanH2002,
  author    = {Xifeng Yan and Jiawei Han},
  booktitle = {IEEE International Conference on Data Mining},
  date      = {2002},
  year      = {2002},
  title     = {{gSpan: Graph-Based Substructure Pattern Mining}},
  doi       = {10.1109/ICDM.2002.1184038},
  pages     = {721--724},
  masid     = {25710},
}

@Article{datamine/KuramochiK05,
  author       = {Michihiro Kuramochi and George Karypis},
  date         = {2005},
  year         = {2005},
  journal = {Data Min. Knowl. Discov.},
  title        = {Finding Frequent Patterns in a Large Sparse Graph\({}^{\mbox{*}}\)},
  number       = {3},
  pages        = {243--271},
  volume       = {11},
}

@Misc{StreamSQL:URL,
  author = {{StreamSQL}},
  date   = {2006},
  title  = {{StreamSQL Home page}},
  note   = {\url{http://blogs.streamsql.org}},
}

@Article{Balakrishnan:Aurora,
  author       = {H. Balakrishnan and others},
  date         = {2004},
  journal = {VLDB Journal: Special Issue on Data Stream Processing},
  title        = {Retrospective on Aurora},
  number       = {4},
  pages        = {370--383},
  volume       = {13},
}

@InProceedings{KDD/HolderCD1994,
  author    = {Lawrence B. Holder and Diane J. Cook and Surnjani Djoko},
  booktitle = {Knowledge Discovery and Data Mining},
  title     = {{Substucture Discovery in the SUBDUE System}},
  year      = {1994},
  pages     = {169--180},
  date      = {1994},
  masid     = {473658},
}

@InCollection{CommFortunatoC09,
  author    = {S. Fortunato and C. Castellano},
  booktitle = {Ency. of Complexity and Systems Science},
  title     = {Community Structure in Graphs},
  year      = {2009},
  pages     = {1141--1163},
  bibsource = {dblp computer science bibliography, http://dblp.org},
  date      = {2009},
  doi       = {10.1007/978-0-387-30440-3_76},
  url       = {http://dx.doi.org/10.1007/978-0-387-30440-3_76},
}

@Article{MultiLayerSurveyKivelaABGMP13,
  author    = {M. Kivel{\"{a}} and A. Arenas and M. Barthelemy and J. P. Gleeson and Y. Moreno and M. A. Porter},
  journal   = {CoRR},
  title     = {Multilayer Networks},
  year      = {2013},
  volume    = {abs/1309.7233},
  bibsource = {dblp computer science bibliography, http://dblp.org},
  date      = {2013},
  opturl    = {http://arxiv.org/abs/1309.7233},
}

@Article{aleta2019multilayer,
  author    = {Aleta, Alberto and Moreno, Yamir},
  journal   = {Annual Review of Condensed Matter Physics},
  title     = {Multilayer networks in a nutshell},
  year      = {2019},
  pages     = {45--62},
  volume    = {10},
  date      = {2019},
  publisher = {Annual Reviews},
}

@string{is    =  "Information Systems"}

@string{sigmod = "Proceedings, International Conference on Management of Data (SIGMOD)"}

@inproceedings{ICABCD2025/SantraC25,
  author={Santra, Abhishek and Chakravarthy, Sharma},
  booktitle={2025 International Conference on Artificial Intelligence, Big Data, Computing and Data Communication Systems (icABCD), Cape Town, South Africa, November 26-27, 2025}, 
  title={Contrasting Traditional and LLMs-Based Approaches For Holistic Analysis of Disparate Data Types}, 
  publisher={ACM},
  year={2025},
  volume={},
  number={},
  pages={1-12},
}

@inproceedings{DEXA/BillahC24,
  author       = {Hafsa Billah and
                  Sharma Chakravarthy},
  title        = {Video Situation Monitoring Using Continuous Queries},
  booktitle    = {Database and Expert Systems Applications - 35th International Conference,
                  {DEXA} 2024, Naples, Italy, August 26-28, 2024, Proceedings, Part
                  {II}},
  series       = {LNCS},
  volume       = {14911},
  pages        = {125--141},
  publisher    = {Springer},
  year         = {2024},
  url          = {https://doi.org/10.1007/978-3-031-68312-1\_10},
  doi          = {10.1007/978-3-031-68312-1\_10},
  bibsource    = {dblp computer science bibliography, https://dblp.org}
}

@inproceedings{PETRA2024/BillahSC24,
  author       = {Umme Hafsa Billah and
                  Abhishek Santra and
                  Sharma Chakravarthy},
  title        = {Leveraging Video Situation Monitoring in Assisted Living Environment},
  booktitle    = {Proceedings of the 17th International Conference on PErvasive Technologies
                  Related to Assistive Environments, {PETRA} 2024, Crete, Greece, June
                  26-28, 2024},
  publisher    = {{ACM}},
  year         = {2024},
  url          = {https://doi.org/10.1145/3652037.3652057},
  doi          = {10.1145/3652037.3652057},
  bibsource    = {dblp computer science bibliography, https://dblp.org}
}

@InProceedings{ADBIS2023/BillahSC,
  author    = {Hafsa Billah and Abhishek Santra and Sharma Chakravarthy},
  booktitle = {New Trends in Database and Information Systems - {ADBIS} 2023 Short Papers, Doctoral Consortium and Workshops: AIDMA, DOING, K-Gals, MADEISD, PeRS, Barcelona, Spain, September 4-7, 2023, Proceedings},
  title     = {Video Situation Monitoring to Improve Quality of Life},
  year      = {2023},
  editor    = {Alberto Abell{\'{o}} and Panos Vassiliadis and Oscar Romero and Robert Wrembel and Francesca Bugiotti and Johann Gamper and Genoveva Vargas{-}Solar and Ester Zumpano},
  pages     = {35--45},
  publisher = {Springer},
  series    = {Communications in Computer and Information Science},
  volume    = {1850},
  bibsource = {dblp computer science bibliography, https://dblp.org},
  doi       = {10.1007/978-3-031-42941-5},
  url       = {https://doi.org/10.1007/978-3-031-42941-5},
}

@InProceedings{IC3K2023/PavelRSC23,
  author    = {Hamza Reza Pavel and Anamitra Roy and Abhishek Santra and Sharma Chakravarthy},
  booktitle = {Proceedings of the 15th International Joint Conference on Knowledge Discovery, Knowledge Engineering and Knowledge Management, {IC3K} 2023, Volume 1: KDIR, Rome, Italy, November 13-15, 2023},
  title     = {Closeness Centrality Detection in Homogeneous Multilayer Networks},
  year      = {2023},
  editor    = {Ana L. N. Fred and Frans Coenen and Jorge Bernardino},
  pages     = {17--29},
  publisher = {{SCITEPRESS}},
  bibsource = {dblp computer science bibliography, https://dblp.org},
  doi       = {10.5220/0012159500003598},
  url       = {https://doi.org/10.5220/0012159500003598},
}

@Article{DBLP:journals/dke/SantraKBC22,
  author    = {Abhishek Santra and Kanthi Sannappa Komar and Sanjukta Bhowmick and Sharma Chakravarthy},
  journal   = {Data Knowl. Eng.},
  title     = {From base data to knowledge discovery - \emph{A life cycle approach} - Using multilayer networks},
  year      = {2022},
  pages     = {102058},
  volume    = {141},
  bibsource = {dblp computer science bibliography, https://dblp.org},
  doi       = {10.1016/j.datak.2022.102058},
  url       = {https://doi.org/10.1016/j.datak.2022.102058},
}

@InProceedings{CICLing:Vu2019,
  author    = {Xuan-Son Vu and Abhishek Santra and Sharma Chakravarthy and Lili Jiang},
  booktitle = {CICLing 2019, La Rochelle, France},
  title     = {Generic Multilayer Network Data Analysis with the Fusion of Content and Structure},
  year      = {2019},
}

@Article{Arxiv/SantraKBC20,
  author        = {Abhishek Santra and Kanthi Sannappa Komar and Sanjukta Bhowmick and Sharma Chakravarthy},
  journal       = {CoRR},
  title         = {A New Community Definition For MultiLayer Networks And {A} Novel Approach For Its Efficient Computation},
  year          = {2020},
  volume        = {abs/2004.09625},
  archiveprefix = {arXiv},
  bibsource     = {dblp computer science bibliography, https://dblp.org},
  eprint        = {2004.09625},
  url           = {https://arxiv.org/abs/2004.09625},
}

@InProceedings{BDA/SantraB17,
  author    = {Abhishek Santra and Sanjukta Bhowmick},
  booktitle = {Big Data Analytics - 5th International Conference, {BDA} 2017, Hyderabad, India, December 12-15, 2017, Proceedings},
  title     = {Holistic Analysis of Multi-source, Multi-feature Data: Modeling and Computation Challenges},
  year      = {2017},
  editor    = {P. Krishna Reddy and Ashish Sureka and Sharma Chakravarthy and Subhash Bhalla},
  pages     = {59--68},
  publisher = {Springer},
  series    = {Lecture Notes in Computer Science},
  volume    = {10721},
  bibsource = {dblp computer science bibliography, https://dblp.org},
  doi       = {10.1007/978-3-319-72413-3\_4},
  url       = {https://doi.org/10.1007/978-3-319-72413-3\_4},
}

@InProceedings{ICCS/SantraBC17,
  author    = {Abhishek Santra and Sanjukta Bhowmick and Sharma Chakravarthy},
  booktitle = {International Conference on Computational Science, {ICCS} 2017, 12-14 June 2017, Zurich, Switzerland},
  title     = {Efficient Community Re-creation in Multilayer Networks Using Boolean Operations},
  year      = {2017},
  editor    = {Petros Koumoutsakos and Michael Lees and Valeria V. Krzhizhanovskaya and Jack J. Dongarra and Peter M. A. Sloot},
  pages     = {58--67},
  publisher = {Elsevier},
  series    = {Procedia Computer Science},
  volume    = {108},
  bibsource = {dblp computer science bibliography, https://dblp.org},
  doi       = {10.1016/j.procs.2017.05.246},
  url       = {https://doi.org/10.1016/j.procs.2017.05.246},
}

@InProceedings{ICDMW/SantraBC17,
  author    = {Abhishek Santra and Sanjukta Bhowmick and Sharma Chakravarthy},
  booktitle = {2017 {IEEE} International Conference on Data Mining Workshops, {ICDM} Workshops 2017, New Orleans, LA, USA, November 18-21, 2017},
  title     = {HUBify: Efficient Estimation of Central Entities Across Multiplex Layer Compositions},
  year      = {2017},
  editor    = {Raju Gottumukkala and Xia Ning and Guozhu Dong and Vijay Raghavan and Srinivas Aluru and George Karypis and Lucio Miele and Xindong Wu},
  pages     = {142--149},
  publisher = {{IEEE} Computer Society},
  bibsource = {dblp computer science bibliography, https://dblp.org},
  doi       = {10.1109/ICDMW.2017.24},
  url       = {https://doi.org/10.1109/ICDMW.2017.24},
}

@inproceedings{book/PadmanabhanC2009,
author = {Srihari Padmanabhan and Sharma Chakravarthy},
title = {{HDB-Subdue: A Scalable Approach to Graph Mining}},
booktitle = {Data Warehousing and Knowledge Discovery},
year = {2009},
pages = {325--338},
doi = {10.1007/978-3-642-03730-6_26},
masid = {6032907}
}

@Article{DKE/AdaikkalavanC06,
  author  = {Raman Adaikkalavan and Sharma Chakravarthy},
  journal = {Transactions of Data Knowledge and Engineering},
  title   = {{SnoopIB: Interval-based event specification and detection for active databases}},
  year    = {2006},
  number  = {1},
  pages   = {139-165},
  volume  = {59},
}

@InProceedings{ICDE/ChakravarthyKTB95,
  author        = {Sharma Chakravarthy and V. Krishnaprasad and Z. Tamizuddin and R. H. Badani},
  booktitle     = {ICDE},
  title         = {{ECA Rule Integration into an OODBMS: Architecture and Implementation}},
  year          = {1995},
  pages         = {341-348},
}

@PHDTHESIS{phdThesis/Santra20,
  AUTHOR =       {Abhishek Santra},
  TITLE =        "{Analysis of Complex Data Sets Using Multilayer Networks: a Decoupling-Based Framework}",
  SCHOOL =       {The University of Texas at Arlington},
  YEAR =         {2020},
  month =        {August},
  url = 	     {https://itlab.uta.edu/students/alumni/PhD/Abhishek_Santra/ASantra_PhD2020.pdf}
}

@PHDTHESIS{phdThesis/Adaikkalavan06,
  AUTHOR =       {Raman Adaikkalavan},
  TITLE =        "{Generalization and Enforcement of Role-Based Access Control using a Novel Event-based Approach.}",
  SCHOOL =       {The University of Texas at Arlington},
  YEAR =         {2006},
  month =        {August},
  source = 	     {http://itlab.uta.edu/ITLABWEB/Students/sharma/theses/Ada06PHD.pdf}
}

@PHDTHESIS{phdThesis/Jiang05,
  AUTHOR =       {Jiang Qingchun},
  TITLE =        "{A Framework for Supporting Quality of Service Requirements in a Data Stream Management System.}",
  SCHOOL =       {The University of Texas at Arlington},
  YEAR =         {2005},
  month =        {August},
  source = 	   {http://itlab.uta.edu/ITLABWEB/Students/sharma/theses/Jia05PHD.pdf}
}

@MASTERSTHESIS{msThesis/Kendai06,
  AUTHOR =       {Balakumar Kendai},
  TITLE =        "{Runtime Optimization and Load Shedding in MavStream: Design and Implementation.}",
  SCHOOL =       {The University of Texas at Arlington},
  YEAR =         {2006},
  month =        {December},
  source =       {http://itlab.uta.edu/ITLABWEB/Students/sharma/theses/Ken06MS.pdf}
}

@MASTERSTHESIS{msThesis/Garg05,
  AUTHOR =       {Vihang Garg},
  TITLE =        "{Estream: An Integration and Event and Stream Processing}",
  SCHOOL =       {The University of Texas at Arlington},
  YEAR =         {2005},
  month =        {December},
  source = 		 {http://itlab.uta.edu/ITLABWEB/Students/sharma/theses/Gar05MS.pdf}
}

@MASTERSTHESIS{msThesis/Sonune03,
  AUTHOR =       {Satyajeet Sonune},
  TITLE =        "{Design and Implementation of Windowed Operators and Scheduler for Stream Data.}",
  SCHOOL =       {The University of Texas at Arlington},
  YEAR =         {2003},
  month =        {December},
  source = 		 {http://itlab.uta.edu/ITLABWEB/Students/sharma/theses/Son03MS.pdf}
}

@MASTERSTHESIS{msThesis/Gilani03,
  AUTHOR =       {Altaf Gilani},
  TITLE =        "{Design and Implementation of Stream Operators, Query Instantiator and Stream Buffer Manager.}",
  SCHOOL =       {The University of Texas at Arlington},
  YEAR =         {2003},
  month =        {December},
  source = 		 {http://itlab.uta.edu/ITLABWEB/Students/sharma/theses/Gil03MS.pdf}
}

@MASTERSTHESIS{msThesis/Krishnaprasad94,
  AUTHOR =       {V. Krishnaprasad},
  TITLE =        "{Event Detection For Supporting Active Capability in an Object-Oriented DBMSs: Semantics Architecture and Implementation.}",
  SCHOOL =       {The University of Florida at Gainesville},
  YEAR =         {1994},
  month =        {May},
  source = 		 {http://itlab.uta.edu/sharma/People/ThesisWeb/vk_thesis.pdf}
}



\end{document}